\documentclass{article}
\usepackage{amsmath}
\usepackage{amssymb}
\usepackage{times}
\usepackage{graphicx} 
\usepackage{natbib}
\usepackage{multirow}
\usepackage{algorithm}
\usepackage{algorithmic}
\usepackage{hyperref}
\usepackage{subcaption}
\usepackage{verbatim}
\usepackage{tikz}
\usepackage{enumitem}

\usepackage[accepted]{icml2016}

\newcommand\REMOVE[1]{\textcolor{red}{}}
\usepackage{comment}

\newcommand{\cutabstractup}{\vspace*{-0.3in}}
\newcommand{\cutabstractdown}{\vspace*{-0.25in}}

\newcommand{\cutsectionup}{\vspace*{-0.1in}}
\newcommand{\cutsectiondown}{\vspace*{-0.05in}}

\newcommand{\cutsubsectionup}{\vspace*{-0.1in}} 
\newcommand{\cutsubsectiondown}{\vspace*{-0.04in}} 

\newcommand{\cutparagraphup}{\vspace*{-0.15in}}
\newcommand{\cutparagraphdown}{\vspace*{-0.00in}}

\icmltitlerunning{Control of Memory, Active Perception, and Action in Minecraft}

\begin{document} 

\twocolumn[
\icmltitle{Control of Memory, Active Perception, and Action in Minecraft}

\icmlauthor{Junhyuk Oh}{junhyuk@umich.edu}
\icmlauthor{Valliappa Chockalingam}{valli@umich.edu}
\icmlauthor{Satinder Singh}{baveja@umich.edu}
\icmlauthor{Honglak Lee}{honglak@umich.edu}
\icmladdress{Computer Science \& Engineering, University of Michigan}
\icmlkeywords{reinforcement learning, deep learning}

\vskip 0.3in
]

\cutabstractup
\begin{abstract} 
\vspace*{-0.05in}

In this paper, we introduce a new set of reinforcement learning (RL) tasks in Minecraft (a flexible 3D world). We then use these tasks to systematically compare and contrast existing deep reinforcement learning (DRL) architectures with our new memory-based DRL architectures. These tasks are designed to emphasize, in a controllable manner, issues that pose challenges for RL methods including partial observability (due to first-person visual observations), delayed rewards, high-dimensional visual observations, and the need to use active perception in a correct manner so as to perform well in the tasks. While these tasks are conceptually simple to describe, by virtue of having all of these challenges simultaneously they are difficult for current DRL architectures. Additionally, we evaluate the generalization performance of the architectures on environments not used during training. The experimental results show that our new architectures generalize to unseen environments better than existing DRL architectures.

\end{abstract} 
\cutabstractdown
\cutsectionup
\section{Introduction} 
\cutsectiondown
Deep learning approaches (surveyed in ~\citealp{lecun2015deep,schmidhuber2015deep}) have made advances in many low-level perceptual supervised learning problems \cite{krizhevsky2012imagenet,girshick2014rich,simonyan2014very}. This success has been extended to reinforcement learning (RL) problems that involve visual perception. For example, the Deep Q-Network (DQN)~\cite{mnih2015human} architecture has been shown to successfully learn to play many Atari 2600 games in the Arcade Learning Environment (ALE) benchmark~\cite{bellemare13arcade} by learning visual features useful for control directly from raw pixels using Q-Learning~\cite{watkins1992q}.

\begin{figure}
    \small
    \centering
    \begin{subfigure}{0.04\linewidth}
        \raggedleft
        \hspace{-0.5cm}
        \rotatebox{90}{
        \hspace{0.1cm}
        \parbox{2cm}{\centering Top-Down\\ View} \hspace{-0.3cm} \parbox{2cm}{\centering First-Person\\ View}
        \hspace{-0.4cm}}
    \end{subfigure}
    \begin{subfigure}{0.21\linewidth}
	    \includegraphics[width=1\linewidth]{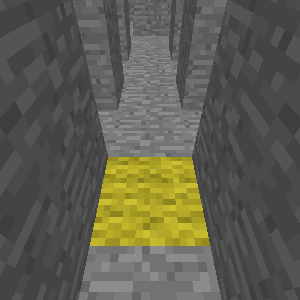} 
   		\includegraphics[width=1\linewidth]{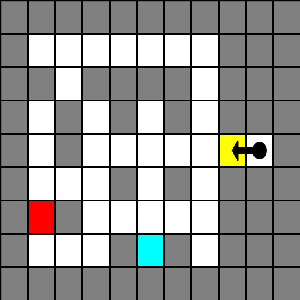} 
   	    \caption{t=3}
	\end{subfigure} 
    \begin{subfigure}{0.21\linewidth}
	    \includegraphics[width=1\linewidth]{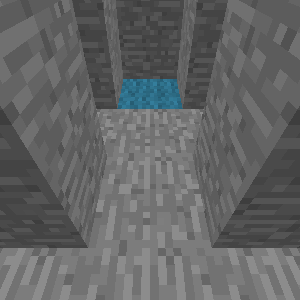} 
   		\includegraphics[width=1\linewidth]{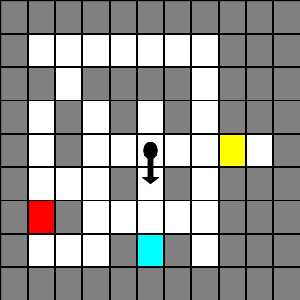} 
   	    \caption{t=10}
	\end{subfigure} 
    \begin{subfigure}{0.21\linewidth}
	    \includegraphics[width=1\linewidth]{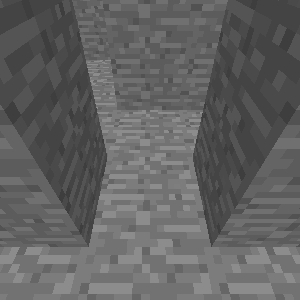} 
   		\includegraphics[width=1\linewidth]{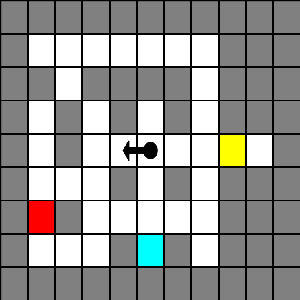} 
   	    \caption{t=11}
	\end{subfigure} 
    \begin{subfigure}{0.21\linewidth}
	    \includegraphics[width=1\linewidth]{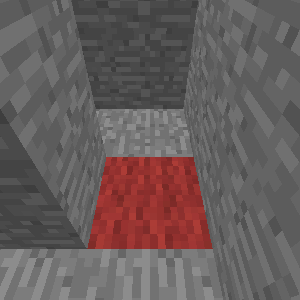} 
   		\includegraphics[width=1\linewidth]{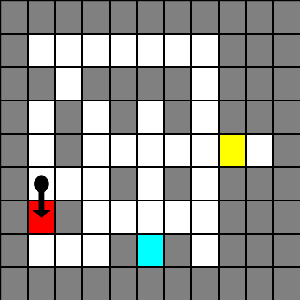} 
   	    \caption{t=19}
	\end{subfigure}
	\vspace{-5pt}
    \caption{Example task in Minecraft. In this task, the agent should visit the red block if the indicator (next to the start location) is yellow. Otherwise, if the indicator is green, it should visit the blue block. The top row shows the agent's first-person observation. The bottom row visualizes the map and the agent's location; this is not available to the agent. 
    (a) The agent observes the yellow indicator. 
    (b) The agent looks left and sees the blue block, (c) but it decides to keep going straight having previously seen the yellow indicator. (d) Finally, it visits the red block and receives a positive reward.}
\label{fig:intro}
\vspace{-15pt}
\end{figure}

Recently, researchers have explored problems that require faculties associated with higher-level cognition (e.g., inferring simple general purpose algorithms: \citealp{graves2014neural}, and, Q\&A: \citealp{weston2014memory}). Most of these advances, however, are restricted to the supervised learning setting, which provides clear error signals. In this paper, we are interested in extending this success to similarly cognition-inspired RL tasks. Specifically, this paper introduces a set of tasks in \textit{Minecraft}\footnote{\url{https://minecraft.net/}}, a flexible 3D world in which an agent can collect resources, build structures, and survive attacks from enemies. Our RL tasks (one example is illustrated in Figure~\ref{fig:intro}) not only have the usual RL challenges of partial observability, high-dimensional (visual) perception, and delayed reward, but also require an agent to develop movement policies by learning how to use its active perception to  observe useful information and collect reward. In addition, our RL tasks require an agent to learn to use any memory it possesses including its interaction with active perception which feeds observations into memory. We note that for simplicity we hereafter refer to these cognition-inspired tasks as cognitive tasks but acknowledge that they form at best a very limited exploration of the range of cognitive faculties in humans.

In this work, we aim to not only systematically evaluate the performance of different neural network architectures on our tasks, but also examine how well such architectures generalize to unseen or larger topologies (Minecraft maps). 
The empirical results show that existing DRL architectures~\cite{mnih2015human,hausknecht2015deep} perform worse on unseen or larger maps compared to training sets of maps, even though they perform reasonably well on the training maps. Motivated by the lack of generalization of existing architectures on our tasks, we also propose new memory-based DRL architectures. Our proposed architectures store recent observations into their memory and retrieve relevant memory based on the temporal context, whereas memory retrieval in existing architectures used in RL problems is not conditioned on the context. 
In summary, we show that our architectures outperform existing ones on most of the tasks as well as generalize better to unseen maps by exploiting their new memory mechanisms.


\cutsectionup
\section{Related Work} \label{sec:related-work}

\paragraph{Neural Networks with External Memory.}
\cutparagraphdown

\citet{graves2014neural} introduced a Neural Turing Machine (NTM), a differentiable external memory architecture, and showed that it can learn algorithms such as copy and reverse.
\citet{zaremba2015reinforcement} proposed RL-NTM that has a non-differentiable memory to scale up the addressing mechanism of NTM and applied policy gradient to train the architecture. 
\citet{NIPS2015_5857} implemented a stack using neural networks and demonstrated that it can infer several algorithmic patterns.
\citet{sukhbaatar2015end} proposed a Memory Network (MemNN) for Q\&A and language modeling tasks, which stores all inputs and retrieves relevant memory blocks depending on the question.

\cutparagraphup
\paragraph{Deep Reinforcement Learning.}
\cutparagraphdown
Neural networks have been used to learn features for RL tasks for a few decades (e.g., \citealp{tesauro1995temporal} and \citealp{lange2010deep}). Recently,  \citet{mnih2015human} proposed a Deep Q-Network (DQN) for training deep convolutional neural networks (CNNs) through Q-Learning in an end-to-end fashion; this achieved state-of-the-art performance on Atari games. \citet{guo2014deep} used slow Monte-Carlo Tree Search (MCTS)~\cite{kocsis2006bandit} to generate a relatively small amount of data to train fast-playing convolutional networks in Atari games. \citet{schulman2015trust}, \citet{levine2015end}, and \citet{lillicrap2015continuous} have successfully trained deep neural networks to directly learn  policies and applied their architectures to robotics problems. In addition, there are deep RL approaches to tasks other than Atari such as learning algorithms~\cite{zaremba2015learning} and text-based games~\cite{sukhbaatar2015mazebase,narasimhan2015language}. There have also been a few attempts to learn state-transition models using deep learning to improve exploration in RL~\cite{oh2015action,stadie2015incentivizing}. Most recently, \citet{mnih2016asynchronous} proposed asynchronous DQN and showed that it can learn to explore a 3D environment similar to Minecraft. Unlike their work, we focus on a systematic evaluation of the ability to deal with partial observability, active perception, and external memory in different neural network architectures as well as generalization across size and maps.  

\cutparagraphup
\paragraph{Model-free Deep RL for POMDPs.}
\cutparagraphdown
Building a model-free agent in partially observable Markov decision processes (POMDPs) is a challenging problem because the agent needs to learn how to summarize history for action-selection. 
To deal with such a challenge, \citet{bakker2003robot} used a Long Short-Term Memory (LSTM) network~\cite{hochreiter1997long} in an offline policy learning framework to show that a robot controlled by an LSTM network can solve \textit{T-Mazes} where the robot should go to the correct destination depending on the traffic signal at the beginning of the maze. 
\citet{wierstra2010recurrent} proposed a \textit{Recurrent Policy Gradient} method and showed that an LSTM network trained using this method outperforms other methods in several tasks including T-Mazes. More recently, \citet{zhang2015policy} introduced continuous memory states to augment the state and action space and showed it can memorize salient information through \textit{Guided Policy Search}~\cite{levine2013guided}. \citet{hausknecht2015deep} proposed Deep Recurrent Q-Network (DRQN) which consists of an LSTM on top of a CNN based on the DQN framework and demonstrated improved handling of partial observability in Atari games.

\cutparagraphup
\paragraph{Departure from Related Work.}
\cutparagraphdown
The architectures we introduce use memory mechanisms similar to MemNN, but our architectures have a layer that constructs a query for memory retrieval based on temporal context. Our architectures are also similar to NTM in that a recurrent controller interacts with an external memory, but ours have a simpler writing and addressing mechanism which makes them easier to train. Most importantly, our architectures are used in an RL setting and must learn from a delayed reward signal, whereas most previous work in exploring architectures with memory is in the supervised learning setting with its much more direct and undelayed error signals. We describe details of our architectures in Section~\ref{sec:method}.

The tasks we introduce are inspired by the T-maze experiments~\cite{bakker2003robot} as well as MazeBase~\cite{sukhbaatar2015mazebase}, which has natural language descriptions of mazes available to the agent. Unlike these previous tasks, our mazes have high-dimensional visual observations with deep partial observability due to the nature of the 3D worlds. In addition, the agent has to learn how best to control its active perception system to collect useful information at the right time in our tasks; this is not necessary in previous work.



\begin{figure}[t]
    \centering
        \begin{subfigure}{0.49\linewidth}
	        \centering
	        \includegraphics[width=\linewidth]{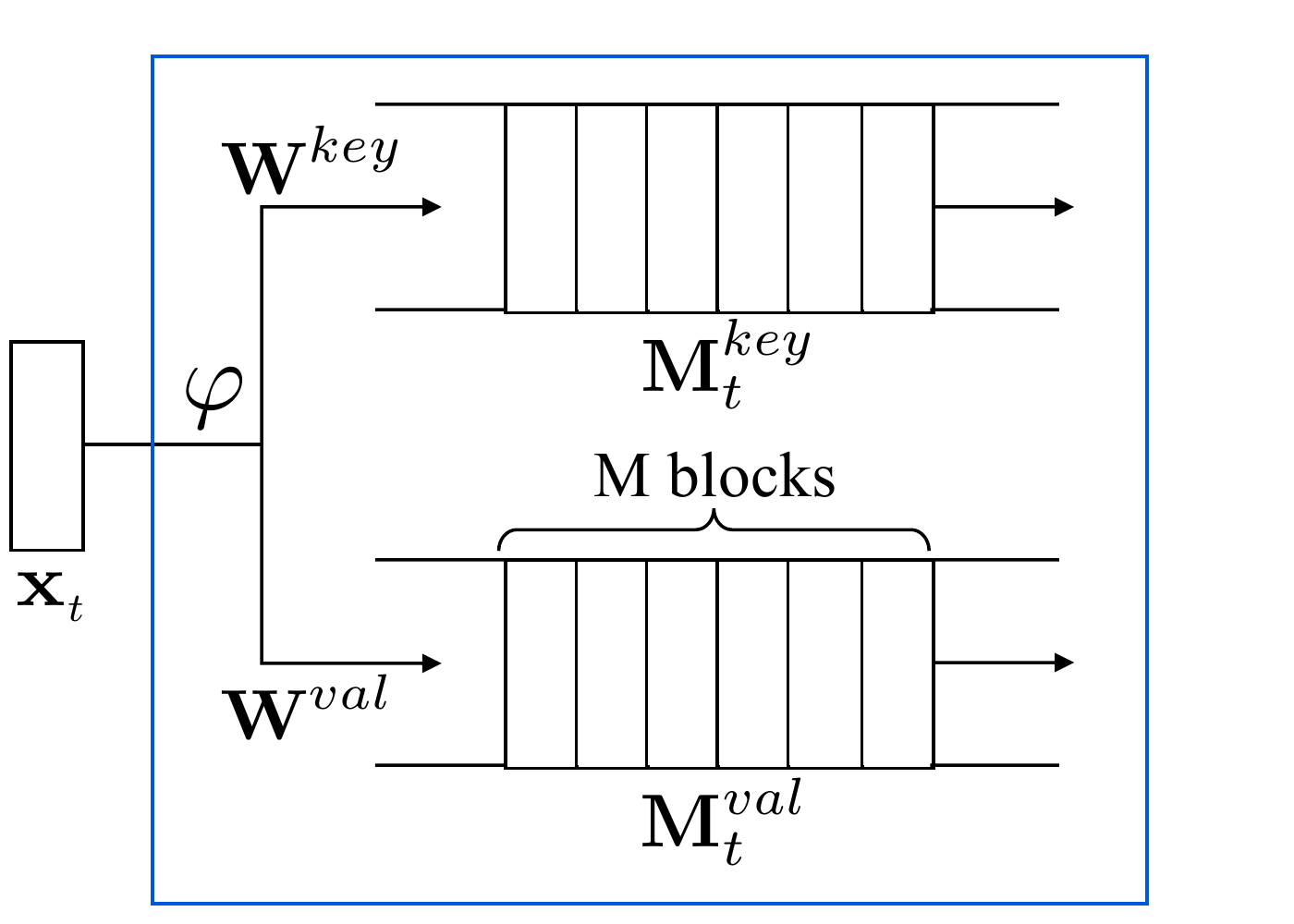}
	        \caption{Write}
	        \label{fig:mem-write}
        \end{subfigure}
        \begin{subfigure}{0.49\linewidth}
	        \centering	
	        \includegraphics[width=\linewidth]{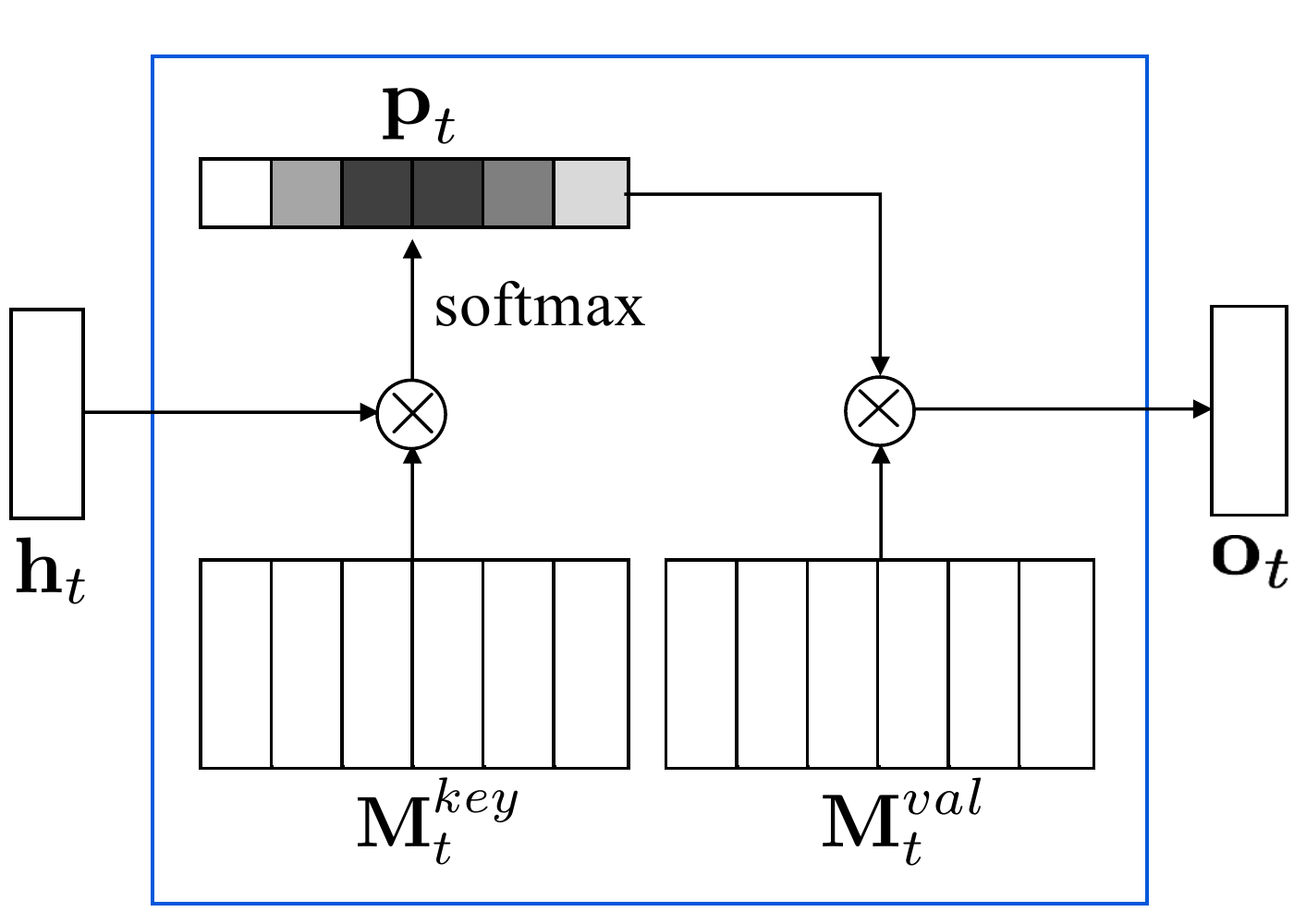}
	        \caption{Read}
	        \label{fig:mem-read}
        \end{subfigure}
        \vspace{-5pt}
        \caption{Illustration of memory operations. }
        \label{fig:memory}
        \vspace{-10pt}
\end{figure}
\cutsectionup
\section{Background: Deep Q-Learning}
\cutsectiondown
Denote the state, immediate reward, and action at time $t$ as $s_{t}, r_t, a_t$ respectively.
In the DQN framework, every transition $T_{t}=\left(s_{t},s_{t+1},a_{t},r_{t}\right)$ is stored in a \textit{replay memory}. 
For (each) iteration $i$, the deep neural network (with parameters $\theta$) is trained to approximate the action-value function from transitions $\left\{(s,s',a,r) \right \}$ by minimizing the loss functions $L_{i}\left(\theta_{i}\right)$ as follows:
\begin{eqnarray*}
\mathcal{L}_{i}\left(\theta\right) & = &  \mathbb{E}_{s,a\sim\pi_{\theta}}\left[\left(y_{i}-Q\left(s,a;\theta\right)\right)^{2}\right] \\
\nabla_{\theta}\mathcal{L}_{i}\left(\theta\right) & = &  \mathbb{E}_{s,a\sim\pi_{\theta}}\left[\left(y_i - Q\left(s,a;\theta\right)\right)\nabla_{\theta}Q\left(s,a;\theta\right)\right] \label{eq:update-rule}
\end{eqnarray*}	
where $y_{i}=\mathbb{E}_{s'\sim\pi_{\theta}}\left[r+\gamma\max_{a'}Q\left(s',a';\theta'\right)\right]$
is the target Q-value estimated by a \textit{target Q-network} ($\theta'$). 
In practice, the expectation terms are approximated by sampling a mini-batch of transitions from the replay memory. The parameter of target Q-network ($\theta'$) is synchronized with the learned network ($\theta$) after a fixed number of iterations.

\cutsectionup
\section{Architectures} \label{sec:method}
\cutsectiondown
The importance of retrieving a prior observation from memory depends on the current context. For example, in the maze of Figure~\ref{fig:intro} where the color of the indicator block determines the desired target color, the indicator information is important only when the agent is seeing a potential target and has to decide whether to approach it or find a different target. Motivated by the lack of ``context-dependent memory retrieval'' in existing DRL architectures, we present three new memory-based architectures in this section.

\begin{figure}[t]
    \centering
    \begin{subfigure}{0.19\linewidth}
    	    \centering
	    \includegraphics[width=\linewidth]{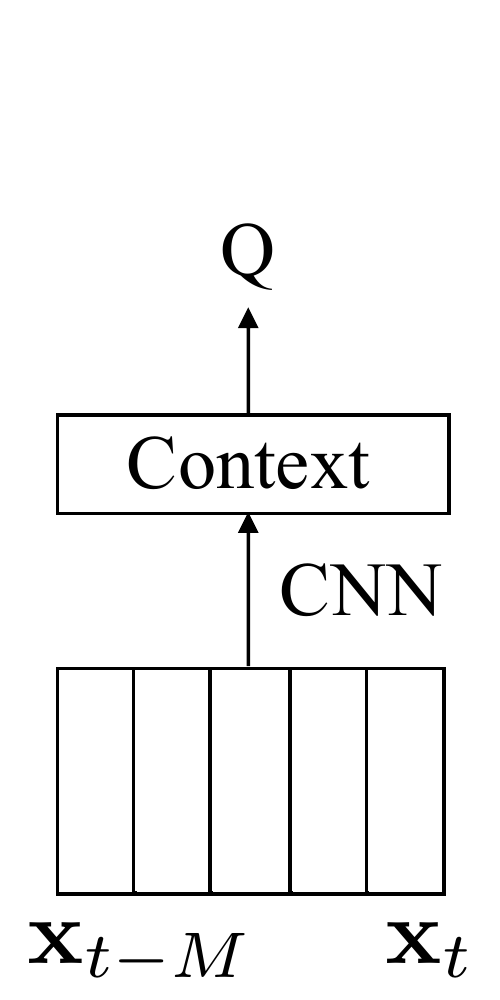} 
  	    \caption{DQN}
  	    \label{fig:model-dqn}
  	\end{subfigure}
  	\hfill
	\begin{subfigure}{0.19\linewidth}
	    \centering
	    \includegraphics[width=\linewidth]{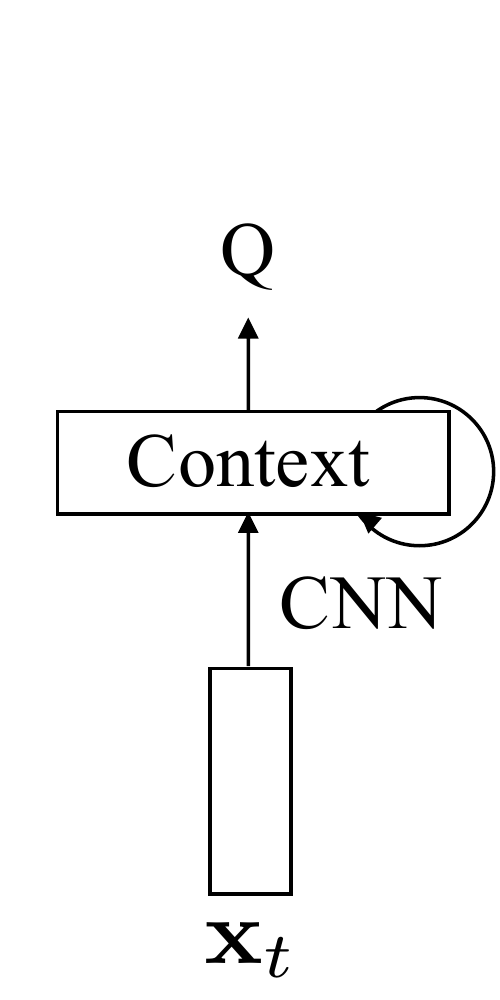} 
  	    \caption{DRQN}
  	    \label{fig:model-drqn}
	\end{subfigure}
	\hfill
	\begin{subfigure}{0.19\linewidth}
	    \centering
	    \includegraphics[width=\linewidth]{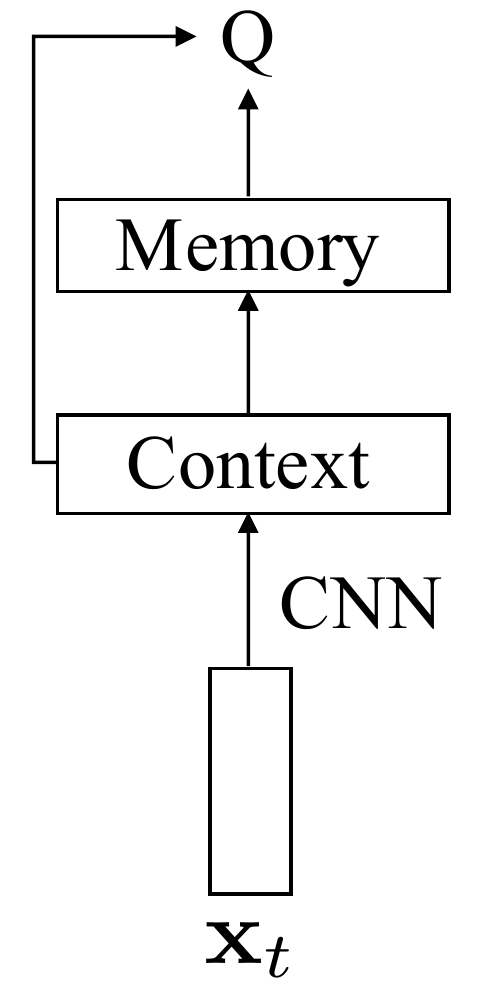} 
  	    \caption{MQN}
  	    \label{fig:model-mqn}
	\end{subfigure}
	\hfill
	\begin{subfigure}{0.19\linewidth}
	    \centering
	    \includegraphics[width=\linewidth]{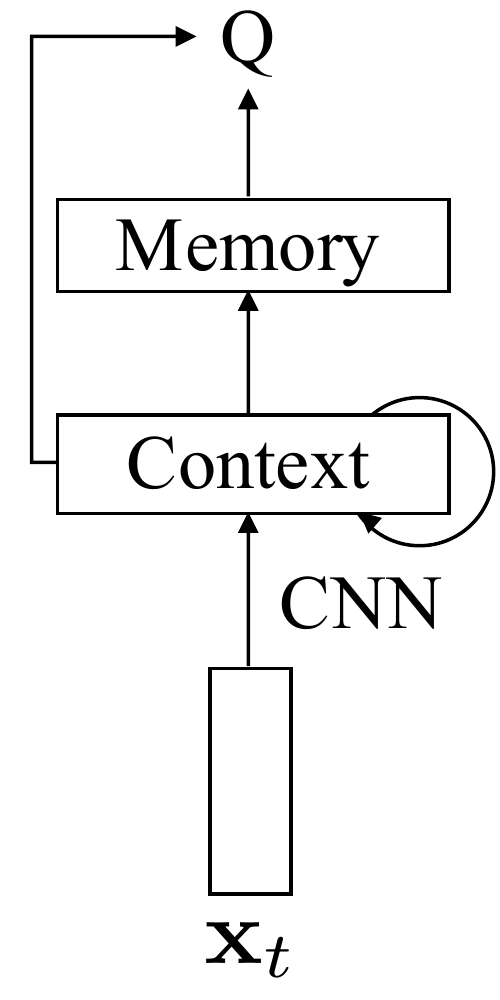} 
  	    \caption{RMQN}
  	    \label{fig:model-lmqn}
	\end{subfigure}
	\hfill
	\begin{subfigure}{0.19\linewidth}
	    \centering
	    \includegraphics[width=\linewidth]{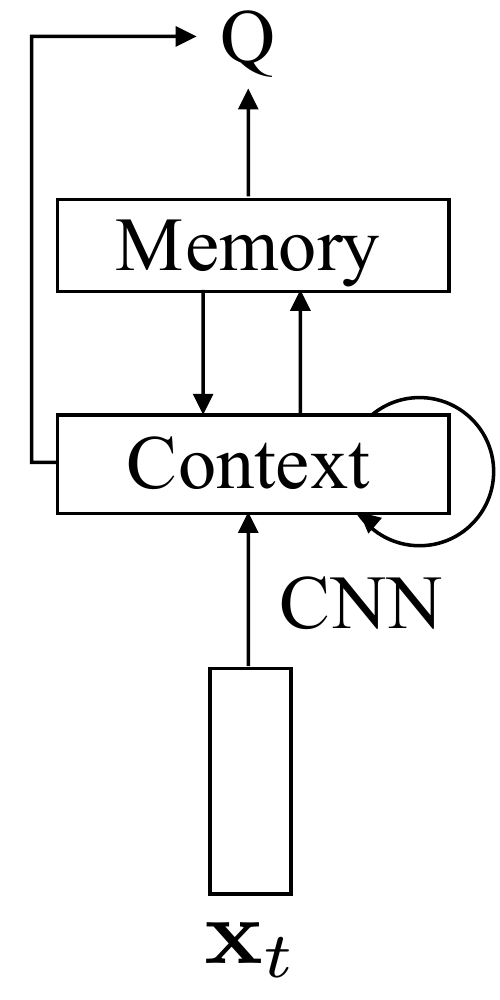} 
  	    \caption{FRMQN}
  	    \label{fig:model-frmqn}
	\end{subfigure}		
	\vspace{-5pt}
\caption{Illustration of different architectures}  
\label{fig:architectures}
    \vspace{-10pt}
\end{figure}

Our proposed architectures (Figure~\ref{fig:architectures}c-e) consist of convolutional networks for extracting high-level features from images (\S\ref{sec-encoding}), a memory that retains a recent history of observations (\S\ref{sec-memory}), and a context vector used both for memory retrieval and (in part for) action-value estimation (\S\ref{sec-controller}). Depending on how the context vector is constructed, we obtain three new architectures: Memory Q-Network (MQN), Recurrent Memory Q-Network (RMQN), and Feedback Recurrent Memory Q-Network (FRMQN).

\subsection{Encoding} \label{sec-encoding}
\cutsubsectiondown

For each time-step, a raw observation (pixels) is encoded to a fixed-length vector as follows:
\begin{eqnarray}
\textbf{e}_t = \varphi^{enc} \left(\textbf{x}_t\right)
\end{eqnarray}
where $\textbf{x}_t \in \mathbb{R}^{c\times h \times w}$ is $h\times w$ image with $c$ channels, and $\textbf{e}_t \in \mathbb{R}^{e}$ is the encoded feature at time $t$. In this work, we use a CNN to encode the observation.

\cutsubsectionup
\subsection{Memory} \label{sec-memory}
\cutsubsectiondown
The memory operations in the proposed architectures are similar to those proposed in MemNN. 

\cutparagraphup
\paragraph{Write.} \label{sec-memory-write}
\cutparagraphdown

The encoded features of last $M$ observations are linearly transformed and stored into the memory as \textit{key} and \textit{value} memory blocks as illustrated in Figure~\ref{fig:mem-write}. More formally, two types of memory blocks are defined as follows: 
\begin{eqnarray}
& \textbf{M}^{key}_t = \textbf{W}^{key} \textbf{E}_{t} \\
& \textbf{M}^{val}_t = \textbf{W}^{val} \textbf{E}_{t}
\end{eqnarray}
where $\textbf{M}^{key}_t,\textbf{M}^{val}_t \in \mathbb{R}^{m\times M}$ are memory blocks with $m$-dimensional embeddings, and $\textbf{W}^{key},\textbf{W}^{val} \in \mathbb{R}^{m\times e}$ are parameters of the linear transformations for keys and values respectively. $\textbf{E}_{t}=\left[\textbf{e}_{t-1}, \textbf{e}_{t-2}, ... , \textbf{e}_{t-M} \right] \in \mathbb{R}^{e\times M}$ is the concatenation of features of the last $M$ observations. 

\cutparagraphup
\paragraph{Read.} \label{sec-memory-read}
\cutparagraphdown

The reading mechanism of the memory is based on soft attention~\cite{graves2013generating,bahdanau2014neural} as illustrated in Figure~\ref{fig:mem-read}. Given a context vector $\textbf{h}_t \in \mathbb{R}^{m}$ (\S\ref{sec-controller}), the memory module draws soft attention over memory locations (and implicitly time) by computing the inner-product between the context and all key memory blocks as follows: 
\begin{eqnarray}
p_{t,i} = \frac{\exp{\left( \mathbf{h}^{\top}_t\mathbf{M}^{key}_{t}[i] \right )}}{\sum_{j=1}^{M} \exp{\left( \mathbf{h}^{\top}_t\mathbf{M}^{key}_{t}[j] \right )}}
\end{eqnarray}
where $p_{t,i} \in \mathbb{R}$ is an attention weight for i-th memory block ($t-i$ time-step).
The output of the read operation is the linear sum of the value memory blocks based on the attention weights as follows:
\begin{eqnarray}
\textbf{o}_t = \textbf{M}^{val}_{t}\textbf{p}_{t} 
\end{eqnarray}
where $\textbf{o}_t \in \mathbb{R}^{m}$ and $\textbf{p}_t \in \mathbb{R}^{M}$ are the retrieved memory and the attention weights respectively.

\begin{figure}[t]
	\centering
        \includegraphics[width=0.60\linewidth]{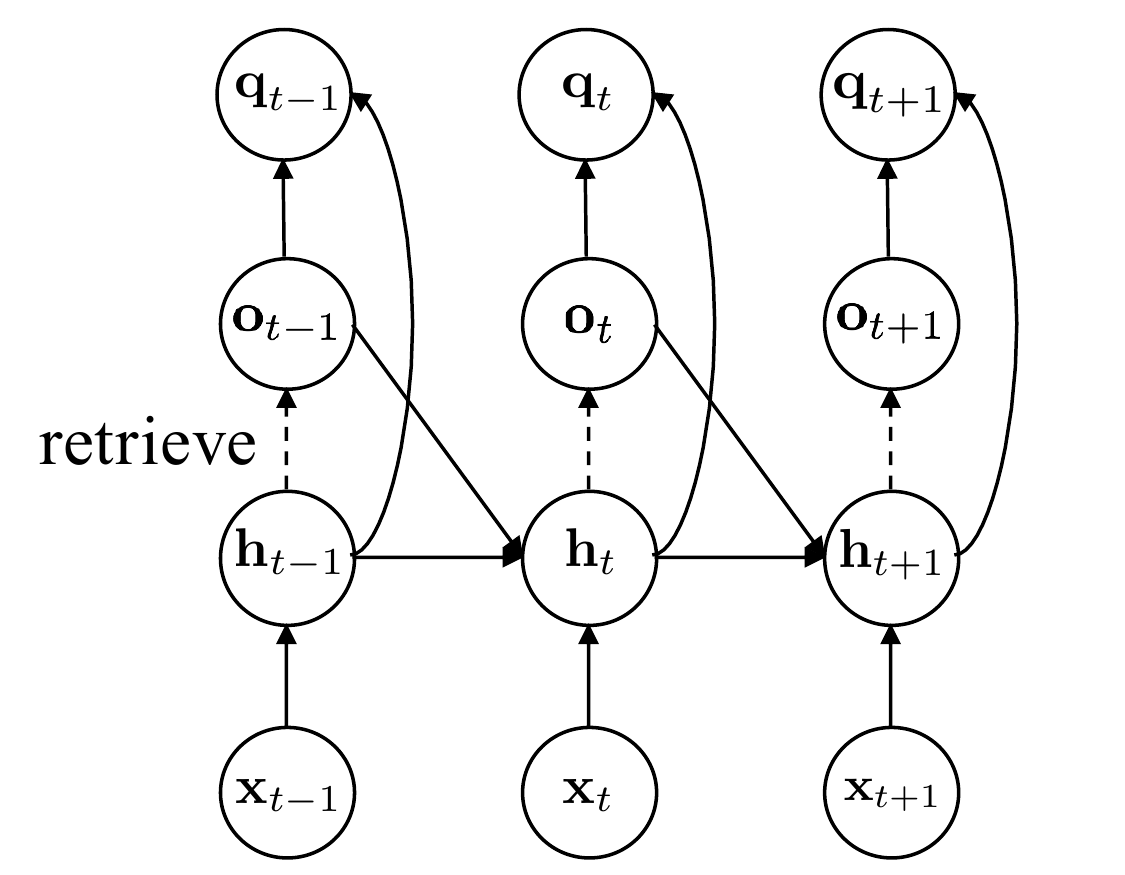}
      \vspace{-10pt}
      \caption{Unrolled illustration of FRMQN. }
      \label{fig:controller}
      \vspace{-10pt}
\end{figure}

\subsection{Context} \label{sec-controller}
\cutsubsectiondown

To retrieve useful information from memory, the context vector should capture relevant spatio-temporal information from the observations. To this end, we present three different architectures for constructing the context vector:
\begin{flalign}
\mbox{MQN: }& \textbf{h}_t = \textbf{W}^{c}\textbf{e}_t \\
\mbox{RMQN: }& \left[\textbf{h}_t, \textbf{c}_t \right] = \mbox{LSTM}\left(\textbf{e}_t, \textbf{h}_{t-1}, \textbf{c}_{t-1}\right) \\
\mbox{FRMQN: }& \left[\textbf{h}_t, \textbf{c}_t \right] = \mbox{LSTM}\left(\left[\textbf{e}_t, \textbf{o}_{t-1} \right], \textbf{h}_{t-1}, \textbf{c}_{t-1}\right)
\end{flalign}
where $\textbf{h}_t,\textbf{c}_t \in \mathbb{R}^{m}$ are a context vector and a memory cell of LSTM respectively, and $\left[\textbf{e}_t, \textbf{o}_{t-1} \right]$ denotes concatenation of the two vectors as input for LSTM. \textbf{MQN} is a feedforward architecture that constructs the context based on only the current observation, which is very similar to MemNN except that the current input is used for memory retrieval in the temporal context of an RL problem. \textbf{RMQN} is a recurrent architecture that captures spatio-temporal information from the history of observations using LSTM. This architecture allows for retaining temporal information through LSTM as well as external memory. Finally, \textbf{FRMQN} has a feedback connection from the retrieved memory to the context vector as illustrated in Figure~\ref{fig:controller}. This allows the FRMQN architecture to refine its context based on the previously retrieved memory so that it can do more complex reasoning as time goes on. Note that feedback connections are analogous to the idea of \textit{multiple hops} in MemNN in the sense that the architecture retrieves memory blocks multiple times based on the previously retrieved memory. However, FRMQN retrieves memory blocks through time, while MemNN does not. 

Finally, the architectures estimate action-values by incorporating the retrieved memory and the context vector: 
\begin{eqnarray}
& & \textbf{q}_{t} = \varphi^{q}\left(\textbf{h}_t, \textbf{o}_t\right)
\end{eqnarray}
where $\textbf{q}_t \in \mathbb{R}^{a}$ is the estimated action-value, and $\varphi^{q}$ is a multi-layer perceptron (MLP) taking two inputs. In the results we report here, we used an MLP with one hidden layer as follows: $\textbf{g}_t = f\left(\textbf{W}^{h}\textbf{h}_t + \textbf{o}_{t}\right), \textbf{q}_{t} = \textbf{W}^{q}\textbf{g}_t$
where $f$ is a rectified linear function~\cite{nair2010rectified} applied only to half of the hidden units for easy optimization by following \citet{sukhbaatar2015end}.

\section{Experiments}
\cutsectiondown
\begin{figure}[t]
\setlength{\tabcolsep}{1pt}
\begin{tabular}{ll}
    \small
	\begin{subfigure}{0.44\linewidth}
	    \centering
	    \includegraphics[width=0.8\linewidth]{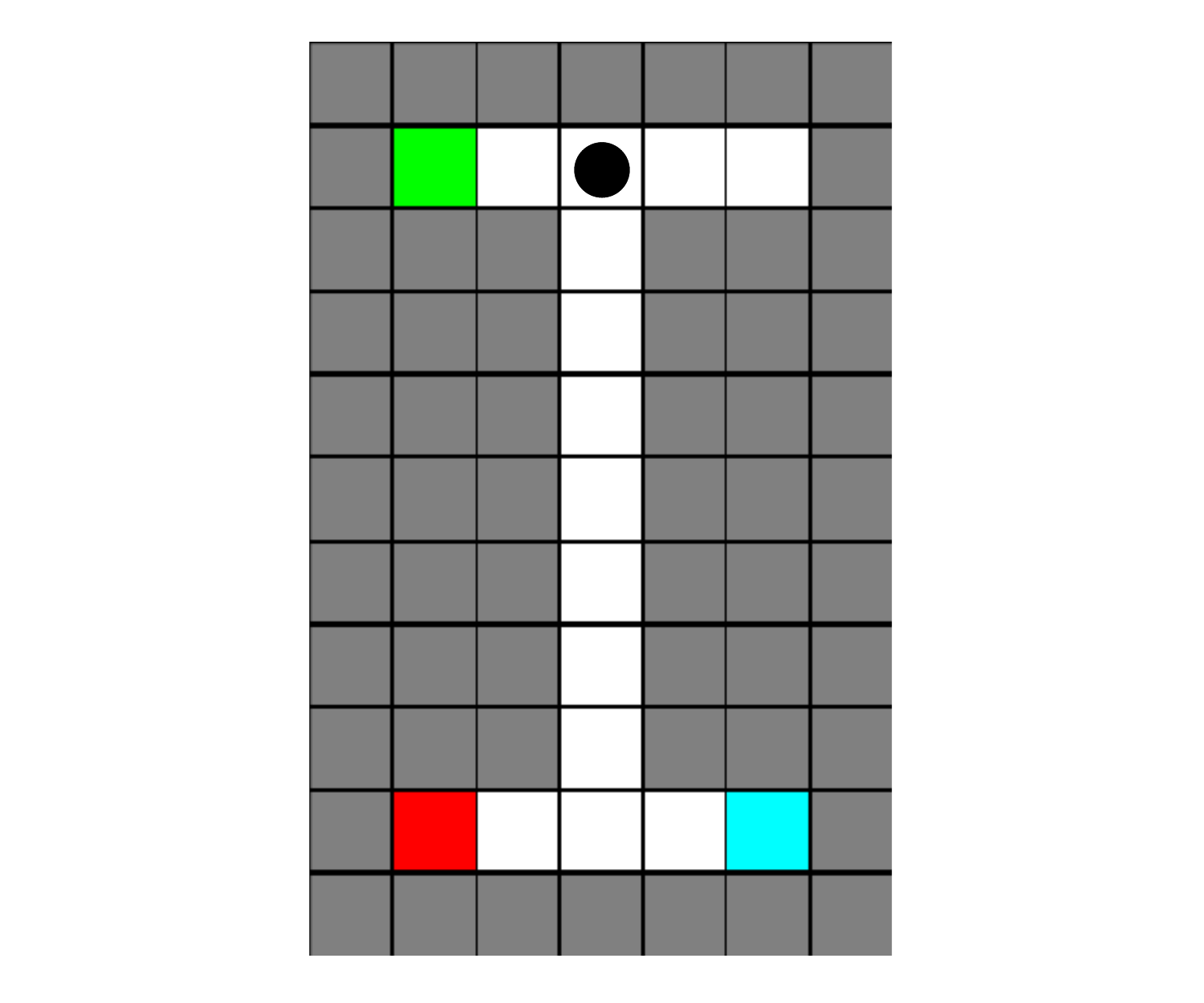} 
   	    \caption{I-Maze}
   	    \label{fig:map-i-maze}
	\end{subfigure}
	&
	\begin{subfigure}{0.44\linewidth}
	    \centering
	    \includegraphics[width=0.8\linewidth]{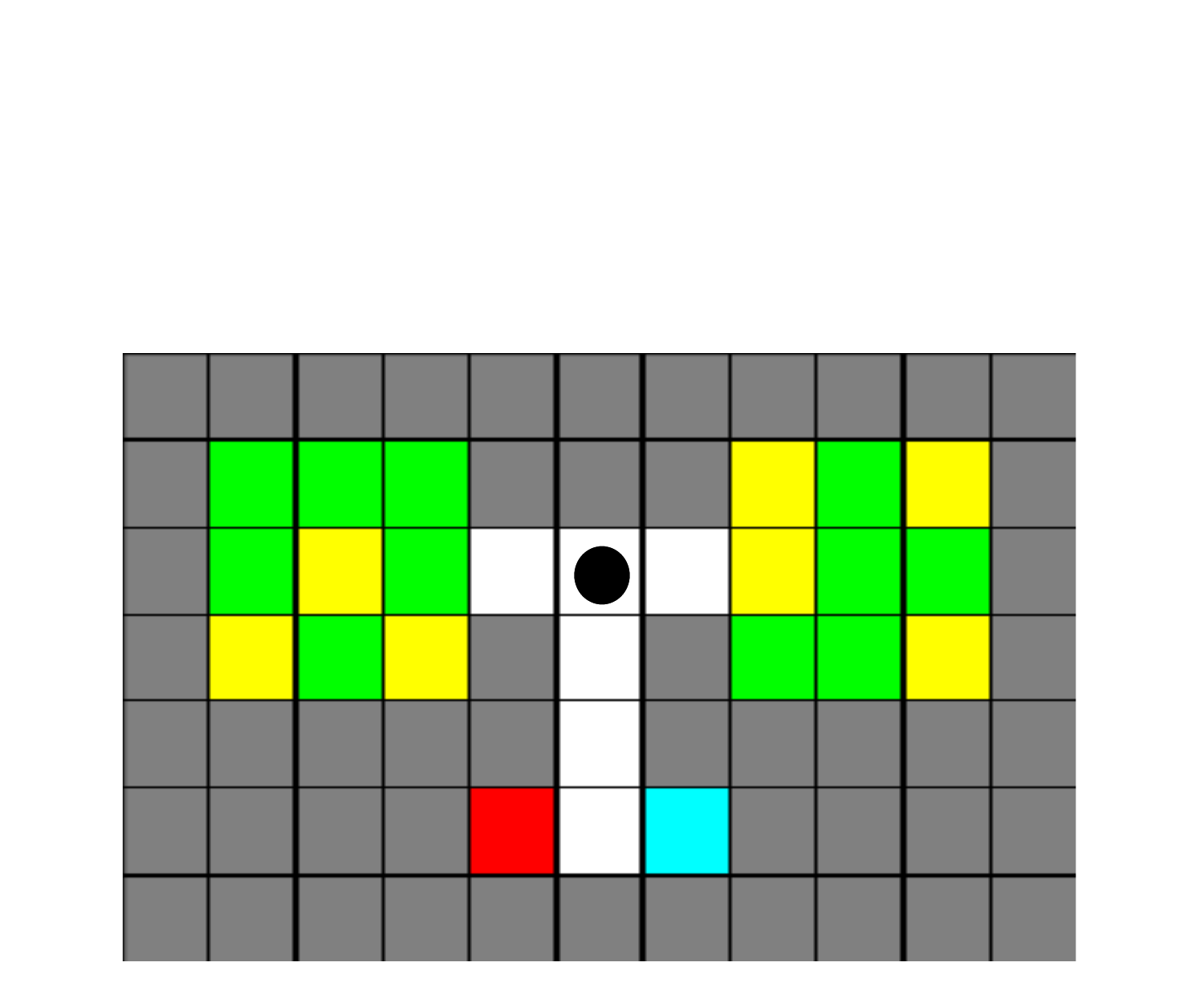} 
   	    \caption{Pattern Matching}
   	    \label{fig:map-pattern-matching}
	\end{subfigure}
	\\
    \begin{subfigure}{0.44\linewidth}
	    \includegraphics[width=\linewidth]{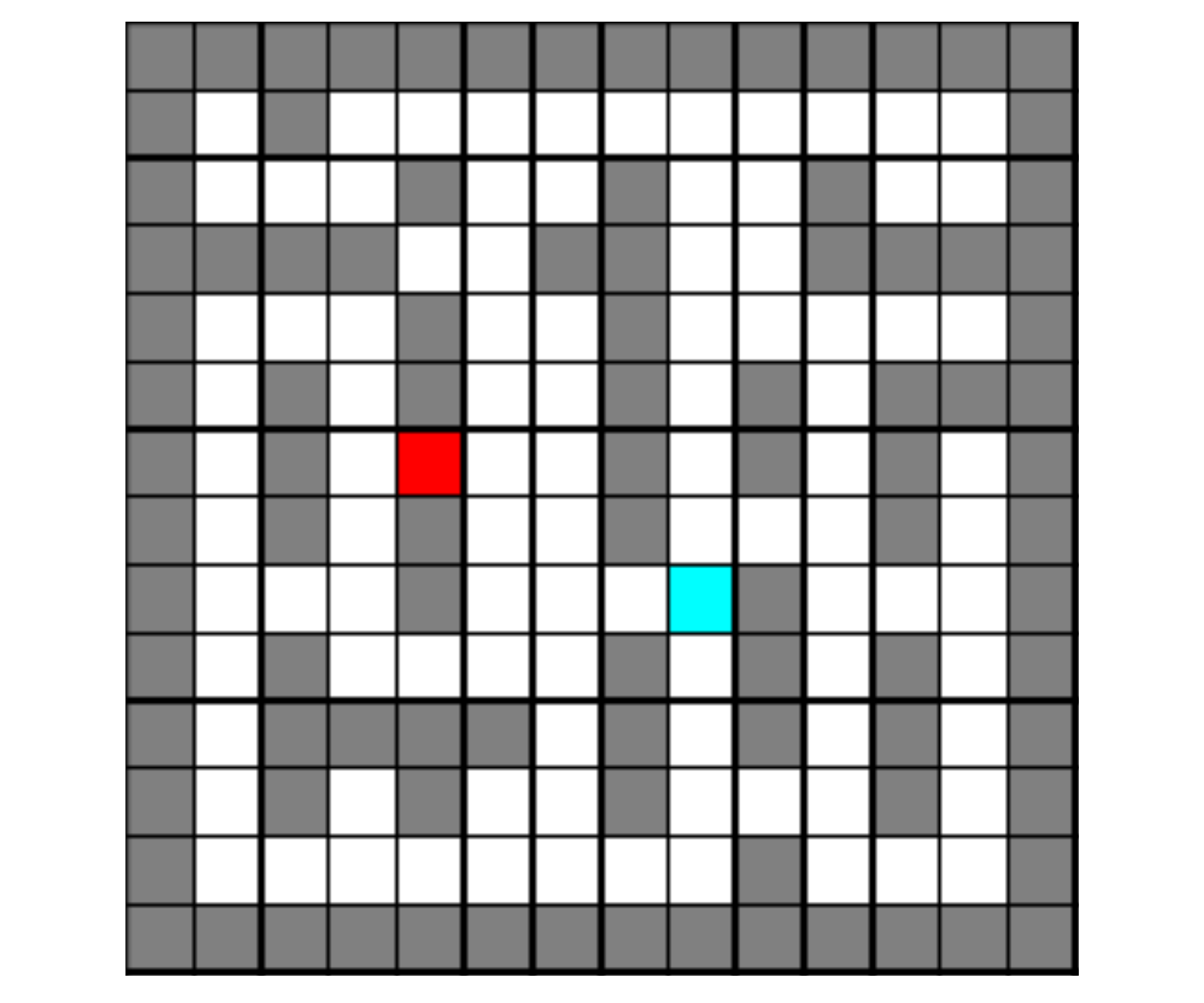} 
   	    \caption{Random Maze}
   	    \label{fig:map-rand}
   	\end{subfigure}
	&
	\begin{subfigure}{0.44\linewidth}
	    \includegraphics[width=\linewidth]{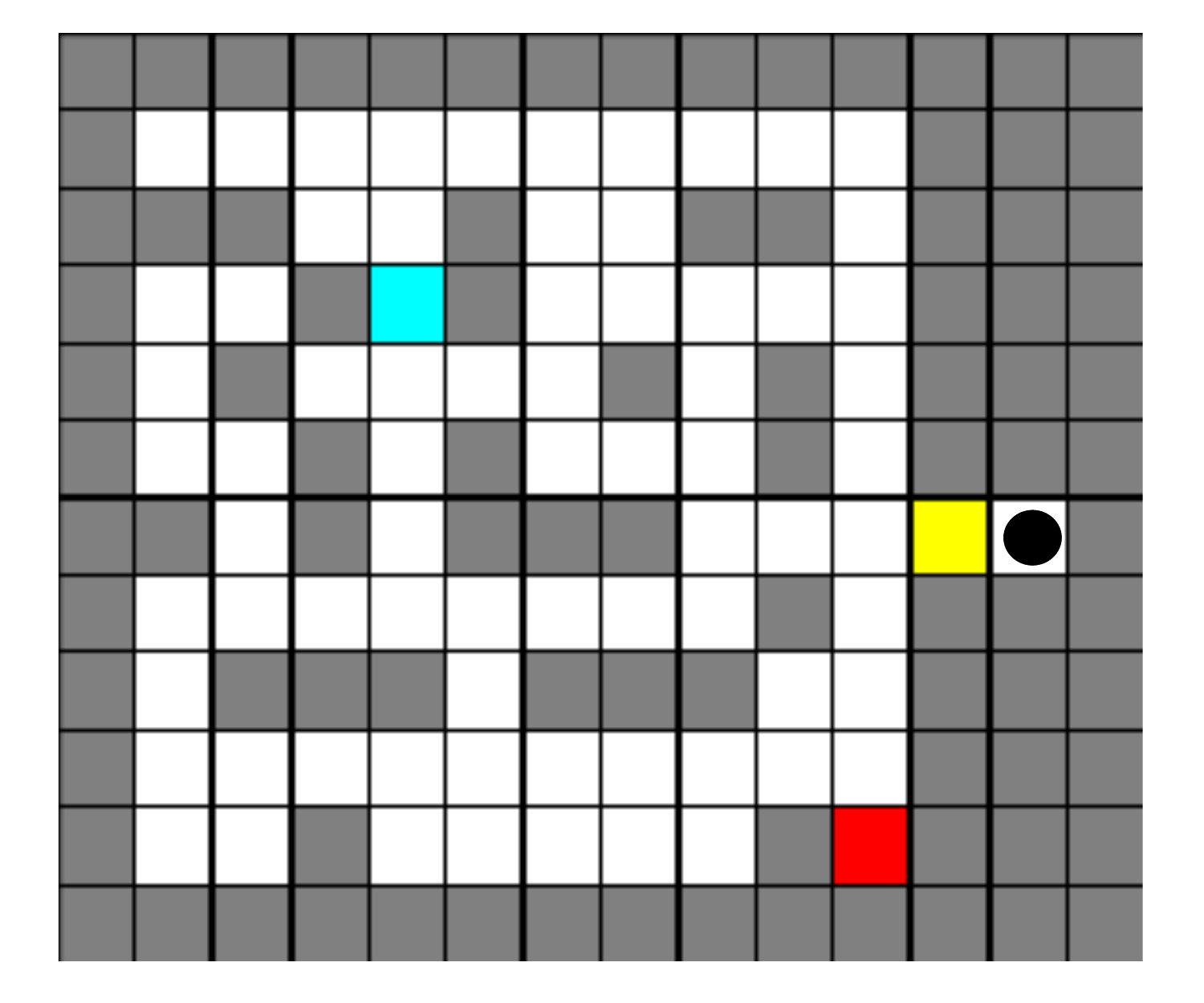} 
   	    \caption{Random Maze w/ Ind}
   	    \label{fig:map-rand-i}
	\end{subfigure} 
\end{tabular}
\caption{Examples of maps. (a) has an I-structured topology where the location of indicator (yellow/green), goals (red/blue), and spawn locations (black circle) are fixed across episodes. (b) has two goals and two rooms with color patterns. (c) consists of randomly generated walls and two goals. The agent can be spawned anywhere except for goal locations. (d) is similar to (c) except that it has an indicator at the fixed location (yellow/green) and a fixed spawn location.
}  
\vspace{-10pt}
\label{fig:maps}
\end{figure}

The experiments, baselines, and tasks are designed to investigate how useful context-dependent memory retrieval is for generalizing to unseen maps, and when memory feedback connections in FRMQN are helpful. Game play videos can be found in the supplementary material and at the following website: \url{https://sites.google.com/a/umich.edu/junhyuk-oh/icml2016-minecraft}.
Next, we describe aspects that are common to all tasks and our training methodology.


\cutparagraphup
\paragraph{Environment.}
In all the tasks, episodes terminate either when the agent finishes the task or after 50 steps. An agent receives -0.04 reward at every time step. The agent's initial looking direction is randomly selected among four directions: north, south, east, and west. For tasks where there is randomness (e.g., maps, spawn points), we randomly sampled an instance after every episode. 

\cutparagraphup
\paragraph{Actions.}
The following six actions are available: Look left/right ($\pm 90^\circ$ in yaw), Look up/down ($\pm 45^\circ$ in pitch), and Move forward/backward. Moving actions move the agent one block forward or backward in the direction it is facing. The pitch is limited to $\left[-45^\circ, 0^\circ \right]$.

\cutparagraphup
\paragraph{Baselines.}
We compare our three architectures with two baselines: \textbf{DQN}~\cite{mnih2015human} (see Figure~\ref{fig:architectures}a) and \textbf{DRQN}~\cite{hausknecht2015deep} (see Figure~\ref{fig:architectures}b). DQN is a CNN architecture that takes a fixed number of frames as input. DRQN is a recurrent architecture that has an LSTM layer on top of the CNN. Note that DQN cannot take more than the number of frames used during training because its first convolution layer takes a fixed number of observations. However, DRQN and our architectures can take arbitrary number of input frames using their recurrent layers. Additionally, our architectures can use an arbitrarily large size of memory during evaluation as well.

\cutparagraphup
\paragraph{Training details.}
Input frames from Minecraft are captured as $32 \times 32$ RGB images.
All the architectures use the same 2-layer CNN architecture as described in the supplementary material. In the DQN and DRQN architectures, the last convolutional layer is followed by a fully-connected layer with 256 hidden units. In our architectures, the last convolution layer is given as the encoded feature for memory blocks. In addition, 256 LSTM units are used in DRQN, RMQN, and FRMQN. More details including hyperparameters for Deep Q-Learning are described in the supplementary material.
Our implementation is based on Torch7~\cite{collobert2011torch7}, a public DQN implementation~\cite{mnih2015human}, and a Minecraft Forge Mod.\footnote{\url{http://files.minecraftforge.net/}} 

\begin{figure*}
    \centering
    \begin{subfigure}{0.19\linewidth}
	    \centering
	    \includegraphics[width=\linewidth]{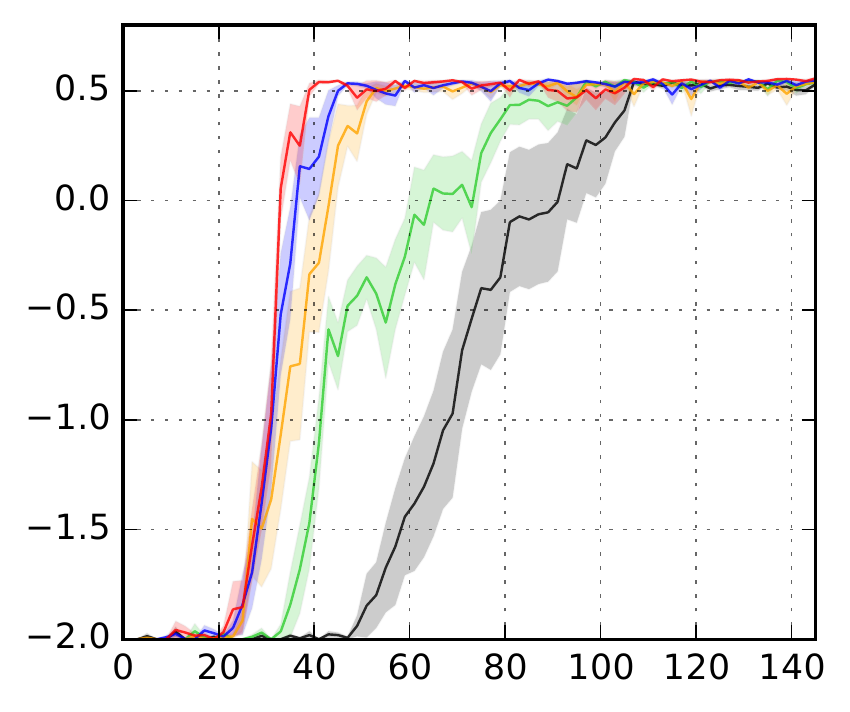} 
	    \caption{I-Maze (Train)}
	    \label{fig:plot-imaze}
	\end{subfigure}
    \begin{subfigure}{0.19\linewidth}
		\centering
	    \includegraphics[width=\linewidth]{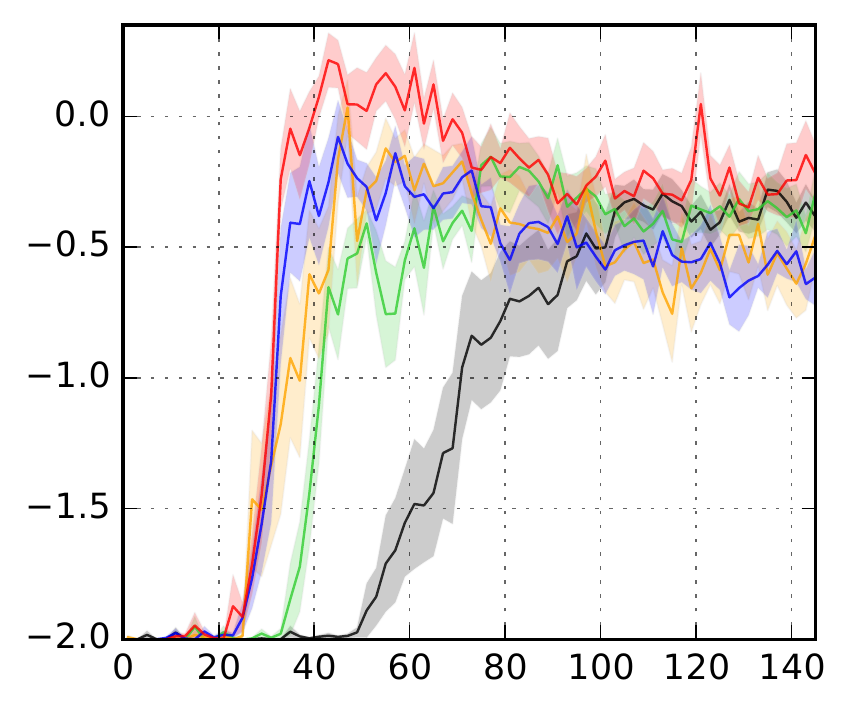} 	    
	    \caption{I-Maze (Unseen)}
	    \label{fig:plot-imaze-test}
	\end{subfigure}
     \begin{subfigure}{0.19\linewidth}
 		\centering
 	    \includegraphics[width=\linewidth]{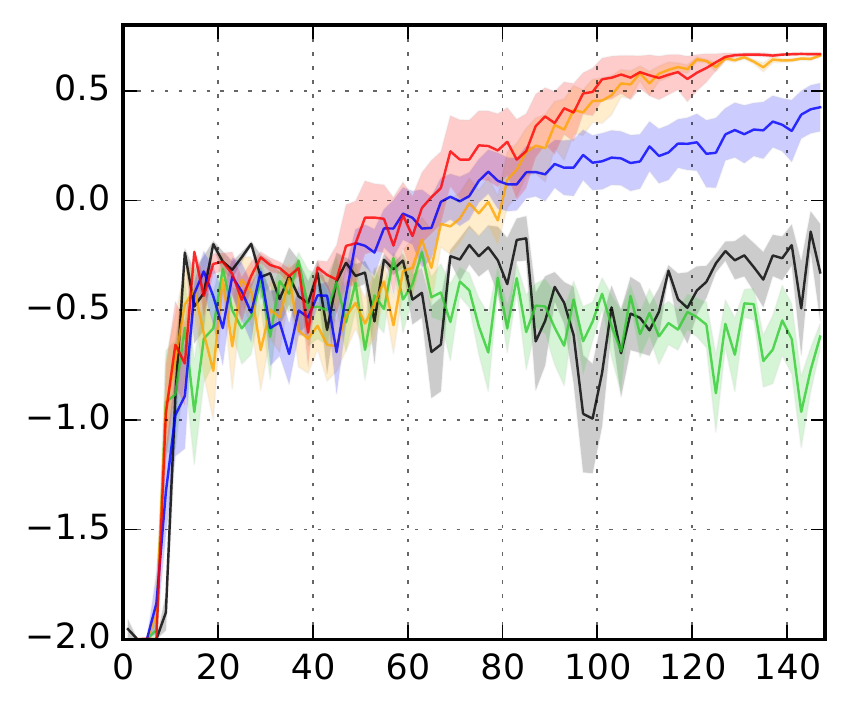} 
 	    \caption{Matching (Train)}
 	    \label{fig:plot-match}
 	\end{subfigure}
     \begin{subfigure}{0.19\linewidth}
 		\centering
 	    \includegraphics[width=\linewidth]{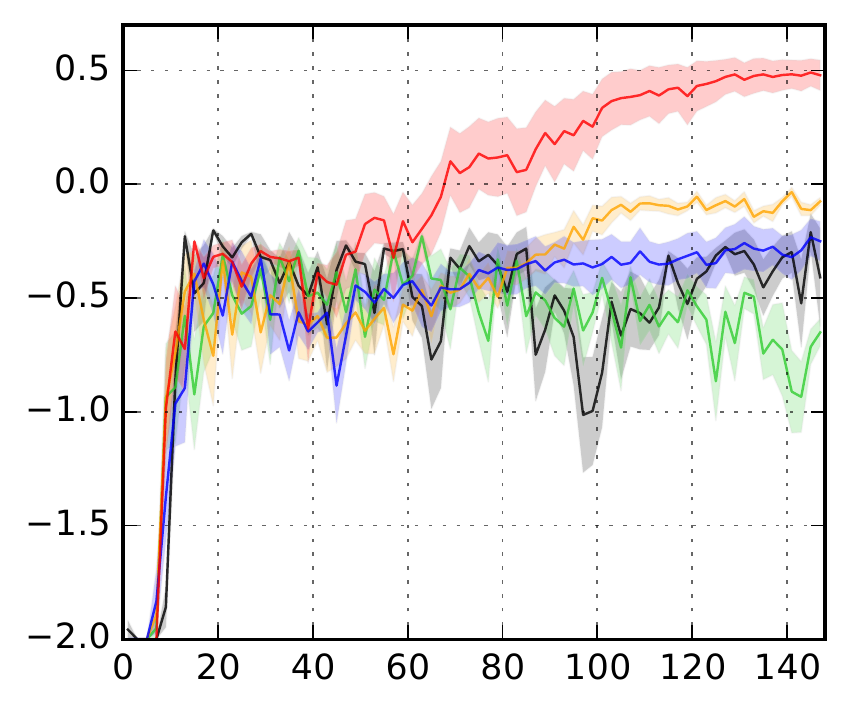} 
 	    \caption{Matching (Unseen)}
 	    \label{fig:plot-match-test}
	\end{subfigure} \\
    \begin{subfigure}{0.19\linewidth}
		\centering
	    \includegraphics[width=\linewidth]{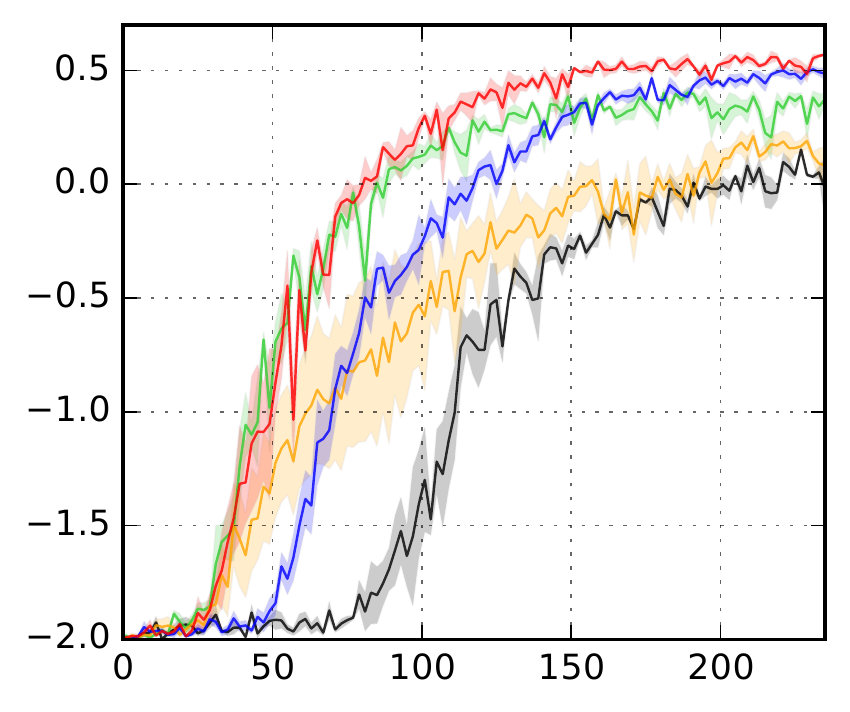} 
	    \caption{Seq+I (Train)}
 	    \label{fig:plot-rand}
 	\end{subfigure}
    \begin{subfigure}{0.19\linewidth}
 		\centering
 	     \includegraphics[width=\linewidth]{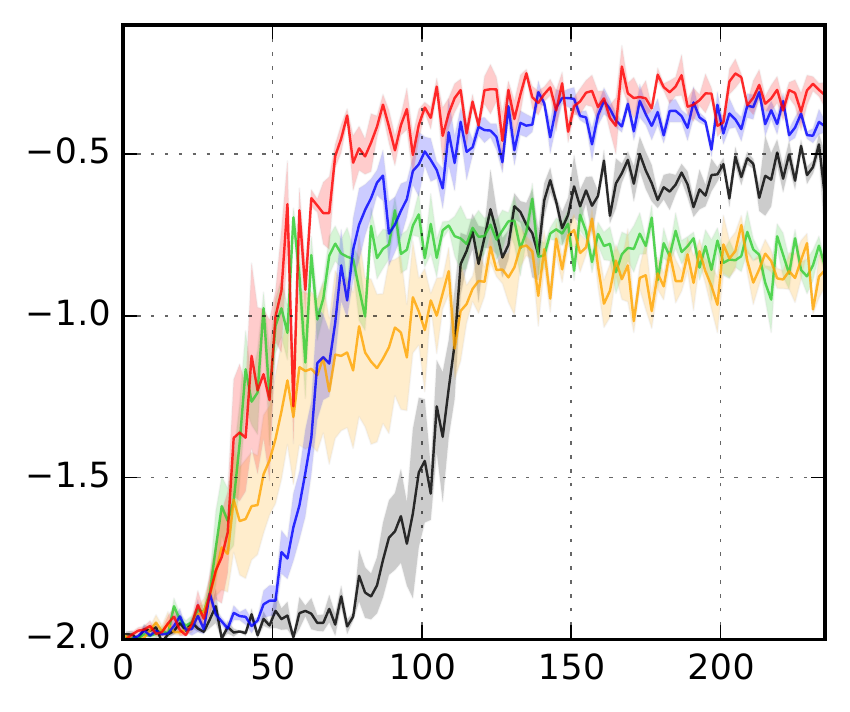} 
	    \caption{Seq+I (Unseen)}
 	    \label{fig:plot-rand-test}
    \end{subfigure}
    \begin{subfigure}{0.19\linewidth}
 		\centering
 	     \includegraphics[width=\linewidth]{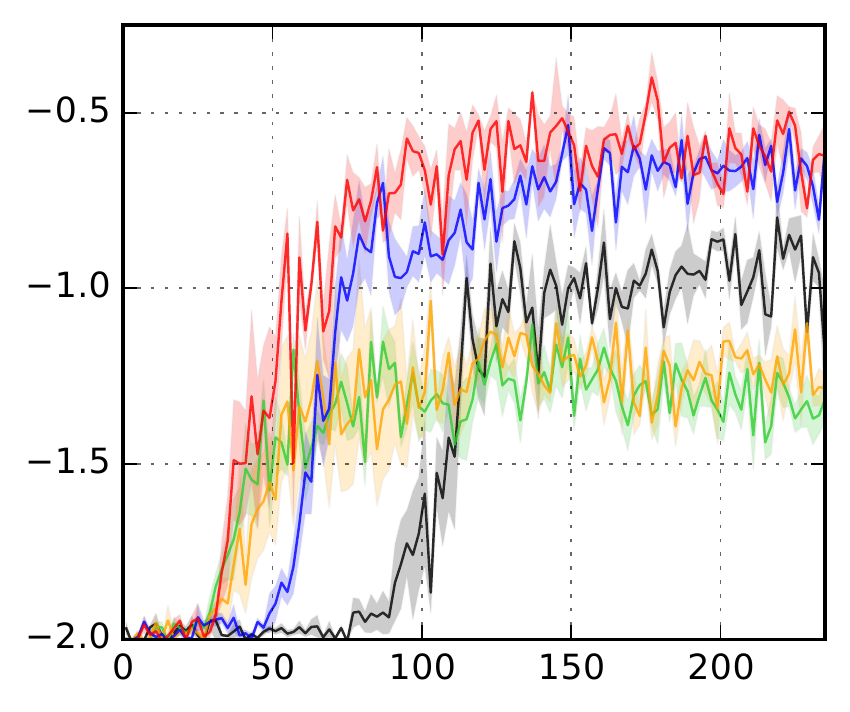} 
	    \caption{Seq+I (Unseen-L)}
 	    \label{fig:plot-rand-large}
    \end{subfigure}
    \begin{subfigure}{0.19\linewidth}
 		\centering
 	    \raisebox{15mm}{\includegraphics[width=0.7\linewidth]{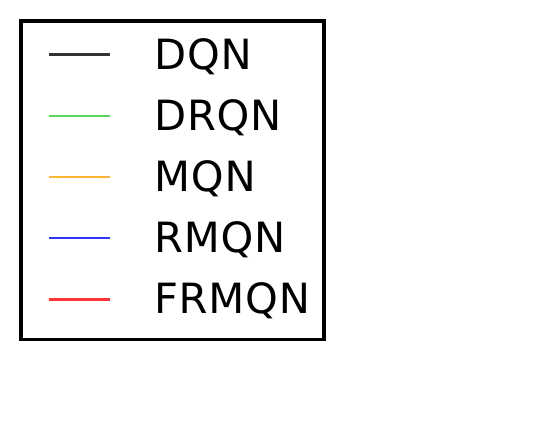}}
    \end{subfigure}
    \vspace{-5pt}
    \caption{Learning curves for different tasks: (a-b) I-maze (\S\ref{sec:exp-imaze}), (c-d) pattern matching (\S\ref{sec:exp-pattern-matching}), (e-g) random mazes (\S\ref{sec:exp-random-mazes}). X-axis and y-axis correspond to the number of training epochs (1 epoch = 10K steps) and the average reward. For (b) and (d), `Unseen' represents unseen maps with different sizes and different patterns respectively. For (f) and (g), `Unseen' and `Unseen-L' indicate unseen topologies with the same sizes and larger sizes of maps, respectively. The performance was measured from 4 runs for random mazes and 10 runs for I-Maze and Pattern Matching. For the random mazes, we only show the results on Sequential Goals with Indicator due to space constraints. More plots are provided in the supplementary material. }     
\label{fig:plot}
\vspace{-15pt}
\end{figure*}

\begin{table}
\vspace*{-0.08in}
\caption{Performance on I-Maze. Each entry shows the average success rate with standard error measured from 10 runs. For each run, we measured the average success rate of 10 best-performing parameters based on the performance on unseen set of maps.  The success rate is defined as the number of episodes that the agent reaches the correct goal within 100 steps divided by the total number of episodes. `Size' represents the number of blocks of the vertical corridor. `\checkmark' indicates that such sizes of I-Mazes belong to the training set of maps. }
\label{table:i-maze}
\begin{center}
\begin{scriptsize}
\begin{sc}
\setlength{\tabcolsep}{2.6pt}
\begin{tabular}{l|c|ccccc}
\hline
Size & Train & DQN & DRQN & MQN & RMQN & FRMQN  \\
\hline 
4 & & $ 92.1(1.5) $& $ 94.8(1.5) $ & $ 87.2(2.3) $ & $ 89.2(2.4) $ & $ \textbf{96.9}(1.0) $  \\
5 & \checkmark & $ \textbf{99.3}(0.5) $& $ 98.2(1.1) $ & $ 96.2(1.0) $ & $ 98.6(0.5) $ & $ \textbf{99.3}(0.7) $  \\
6 & & $ \textbf{99.4}(0.4) $& $ 98.2(1.0) $ & $ 96.0(1.0) $ & $ 99.0(0.4) $ & $ \textbf{99.7}(0.3) $  \\
7 & \checkmark & $ 99.6(0.3) $& $ 98.8(0.8) $ & $ 98.0(0.6) $ & $ 98.8(0.5) $ & $ \textbf{100.0}(0.0) $  \\
8 & & $ 99.3(0.4) $& $ 98.3(0.8) $ & $ 98.3(0.5) $ & $ 98.0(0.8) $ & $ \textbf{100.0}(0.0) $  \\
9 & \checkmark & $ 99.0(0.5) $& $ 98.4(0.6) $ & $ 98.0(0.7) $ & $ 94.6(1.8) $ & $ \textbf{100.0}(0.0) $  \\
10 & & $ 96.5(0.7) $& $ 97.4(1.1) $ & $ 98.2(0.7) $ & $ 87.5(2.6) $ & $ \textbf{99.6}(0.3) $  \\
15 & & $ 50.7(0.9) $& $ 83.3(3.2) $ & $ \textbf{96.7}(1.3) $ & $ 89.8(2.4) $ & $ \textbf{97.4}(1.1) $  \\
20 & & $ 48.3(1.0) $& $ 63.6(3.7) $ & $ 97.2(0.9) $ & $ 96.3(1.2) $ & $ \textbf{98.8}(0.5) $  \\
25 & & $ 48.1(1.0) $& $ 57.6(3.7) $ & $ \textbf{98.2}(0.7) $ & $ 90.3(2.5) $ & $ \textbf{98.4}(0.6) $  \\
30 & & $ 48.6(1.0) $& $ 60.5(3.6) $ & $ \textbf{97.9}(0.9) $ & $ 87.1(2.4) $ & $ \textbf{98.1}(0.6) $  \\
35 & & $ 49.5(1.2) $& $ 59.0(3.4) $ & $ \textbf{95.0}(1.1) $ & $ 84.0(3.2) $ & $ \textbf{94.8}(1.2) $  \\ 
40 & & $ 46.6(1.2) $& $ 59.2(3.6) $ & $ 77.2(4.2) $ & $ 71.3(5.0) $ & $ \textbf{89.0}(2.6) $  \\
\hline
\end{tabular}
\end{sc}
\end{scriptsize}
\end{center}
\vskip -0.17in
\end{table}

\cutsubsectionup
\subsection{I-Maze: Description and Results}
\label{sec:exp-imaze}

\paragraph{Task.} 
Our I-Maze task was inspired by T-Mazes which have been used in animal cognition experiments~\cite{olton1979mazes}. Maps for this task (see Figure~\ref{fig:map-i-maze}) have an indicator at the top that has equal chance of being yellow or green. If the indicator is yellow, the red block gives +1 reward and the blue block gives -1 reward; if the indicator is green, the red block gives -1 and the blue block gives +1 reward. Thus, the agent should memorize the color of the indicator at the beginning while it is in view and visit the correct goal depending on the indicator-color. We varied the length of the vertical corridor to $l=\{5,7,9\}$ during training. The last 12 frames were given as input for all architectures, and the size of memory for our architectures was 11.

\cutparagraphup
\paragraph{Performance on the training set.} 
We observed two stages of behavior during learning from all the architectures: 1) early in the training the discount factor and time penalty led to the agent to take a chance by visiting any goal, and 2) later in the training the agent goes to the correct goal by learning the correlation between the indicator and the goal. As seen in the learning curves in Figure~\ref{fig:plot-imaze}, our architectures converge more quickly than DQN and DRQN to the correct behavior. In particular, we observed that DRQN takes many more epochs to reach the second stage after the first stage has been reached. This is possibly due to the long time interval between seeing the indicator and the goals.
Besides, the indicator block is important only when the agent is at the bottom end of the vertical corridor and needs to decide which way to go (see Figure~\ref{fig:map-i-maze}). In other words, the indicator information does not affect the agent's decision making along its way to the end of the corridor. This makes it even more difficult for DRQN to retain the indicator information for a long time. On the other hand, our architectures can handle these problems by storing the history of observations into memory and retrieving such information when it is important, based on the context.

\cutparagraphup
\paragraph{Generalization performance.} 
To investigate generalization performance, we evaluated the architectures on maps that have vertical corridor lengths $\{4,6,8,10,15,20,25,30,35,40\}$ that were not present in the training maps. 
More specifically, testing on $\{6, 8\}$ sizes of maps and the rest of the sizes of maps can evaluate \textit{interpolation} and \textit{extrapolation} performance, respectively~\cite{schaul2015universal}. 
Since some unseen maps are larger than the training maps, we used 50 last frames as input during evaluation on the unseen maps for all architectures except for DQN, which can take only 12 frames as discussed in the experimental setup. The size of memory for our architectures is set to 49. The performance on the unseen set of maps is visualized in Figure~\ref{fig:plot-imaze-test}. Although the generalization performances of all architectures are highly variable even after training performance converges, it can be seen that FRMQN consistently outperforms the other architectures in terms of average reward. To further investigate the performance for different lengths of the vertical corridor, we measured the performance on each size of map in Table~\ref{table:i-maze}. It turns out that all architectures perform well on $\{6,8\}$ sizes of maps, which indicates that they can interpolate within the training set of maps. However, our architectures extrapolate to larger maps significantly better than the two baselines. 

\cutparagraphup
\paragraph{Analysis of memory retrieval.}
Figure~\ref{fig:play-imaze} visualizes FRMQN's memory retrieval on a large I-Maze, where FRMQN sharply retrieves the indicator information only when it reaches the end of the corridor where it then makes a decision of which goal block to visit. 
This is a reasonable strategy because the indicator information is important only when it is at the end of the vertical corridor. This qualitative result implies that FRMQN learned a general strategy that looks for the indicator, goes to the end of the corridor, and retrieves the indicator information when it decides which goal block to visit. We observed similar policies learned by MQN and RMQN, but the memory attention for the indicator was not as sharp as FRMQN's attention and so they visit wrong goals in larger I-Mazes more often.

\begin{table}
\vspace*{-0.08in}
\caption{Performance on pattern matching. The entries represent the probability of visiting the correct goal block for each set of maps with standard error. The performance reported is averages over 10 runs and 10 best-performing parameters for each run. }
\label{table:matching}
\begin{center}
\begin{small}
\begin{sc}
\begin{tabular}{l|c|c}
\hline
 & Train & Unseen \\
\hline 
DQN & $62.9\%$ $(\pm 3.4\%) $ & $60.1\%$ $(\pm 2.8\%)$  \\
DRQN & $49.7\%$ $(\pm 0.2\%)$ & $49.2\%$ $(\pm 0.2\%)$ \\
MQN & $99.0\%$ $(\pm 0.2\%) $ & $69.3\%$ $(\pm 1.5\%)$ \\
RMQN & $82.5\%$ $(\pm 2.5\%)$ & $62.3\%$ $(\pm 1.5\%)$ \\ 
FRMQN & $\textbf{100.0\%}$ $(\pm 0.0\%)$ & $\textbf{91.8\%}$ $(\pm 1.0\%)$ \\
\hline
\end{tabular}
\end{sc}
\end{small}
\end{center}
\vspace*{-0.3in}
\end{table}

The results on I-Maze shown above suggest that solving a task on a set of maps does not guarantee solving the same task on similar but unseen maps, and such generalization performance highly depends on the feature representation learned by deep neural networks. The extrapolation result shows that context-dependent memory retrieval in our architectures is important for learning a general strategy when the importance of an observational-event depends highly on the temporal context.


\cutsubsectionup
\subsection{Pattern Matching: Description and Results}
\label{sec:exp-pattern-matching}
\cutsubsectiondown

\paragraph{Task.} 
As illustrated in Figure~\ref{fig:map-pattern-matching}, this map consists of two $3\times3$ rooms. The visual patterns of the two rooms are either identical or different with equal probability. If the two rooms have the exact same color patterns, the agent should visit the blue block. If the rooms have different color patterns, the agent should visit the red block. The agent receives a +1 reward if it visits the correct block and a -1 reward if it visits the wrong block. This pattern matching task requires more complex reasoning (comparing two visual patterns given at different time steps) than the I-Maze task above.
We generated 500 training and 500 unseen maps in such a way that there is little overlap between the two sets of visual patterns. Details of the map generation process are described in the supplementary material.
The last 10 frames were given as input for all architectures, and the size of memory was set to 9.

\cutparagraphup
\paragraph{Performance on the training set.}
The results plotted in Figure~\ref{fig:plot-match} and Table~\ref{table:matching} show that MQN and FRMQN successfully learned to go to the correct goal block for all runs in the training maps. We observed that DRQN always learned a sub-optimal policy that goes to any goal regardless of the visual patterns of the two rooms. Another observation is the training performances of DQN and RMQN are a bit unstable; they often learned the same sub-optimal policy, whereas MQN and FRMQN consistently learned to go to the correct goal across different runs. We hypothesize that it is not trivial for a neural network to compare two visual patterns observed in different time-steps unless the network can model high-order interactions between two specific observations for visual matching, which might be the reason why DQN and DRQN fail more often. Context-dependent memory retrieval mechanism in our architectures can alleviate this problem by retrieving two visual patterns corresponding to the observations of the two rooms before decision making.

\paragraph{Generalization performance.}
Table~\ref{table:matching} and Figure~\ref{fig:plot-match-test} show that FRMQN achieves the highest success rate on the unseen set of maps. Interestingly, MQN fails to generalize to unseen visual patterns. We observed that MQN pays attention to the two visual patterns before choosing one of the goals through its memory retrieval. However, since the retrieved memory is just a convex combination of two visual patterns, it is hard for MQN to compare the similarity between them. Thus, we believe that MQN simply overfits to the training maps by memorizing the weighted sum of pairs of visual patterns in the training set of maps. On the other hand, FRMQN can utilize retrieved memory as well as its recurrent connections to compare visual patterns over time. 

\begin{figure}
	\centering
    \begin{subfigure}{0.99\linewidth}
	    \includegraphics[width=\linewidth]{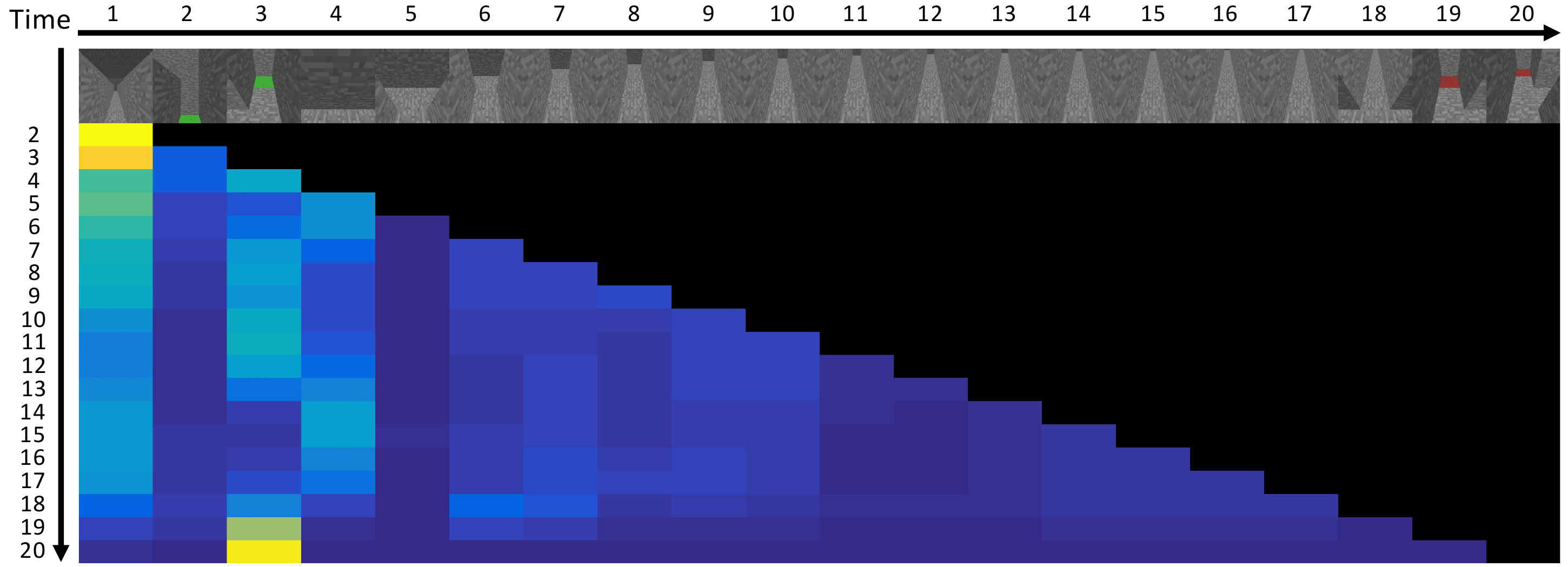} 
	\caption{I-Maze (\S\ref{sec:exp-imaze})} %
	\label{fig:play-imaze}
    \end{subfigure}
    \begin{subfigure}{0.48\linewidth}
	    \centering
	    \includegraphics[width=\linewidth]{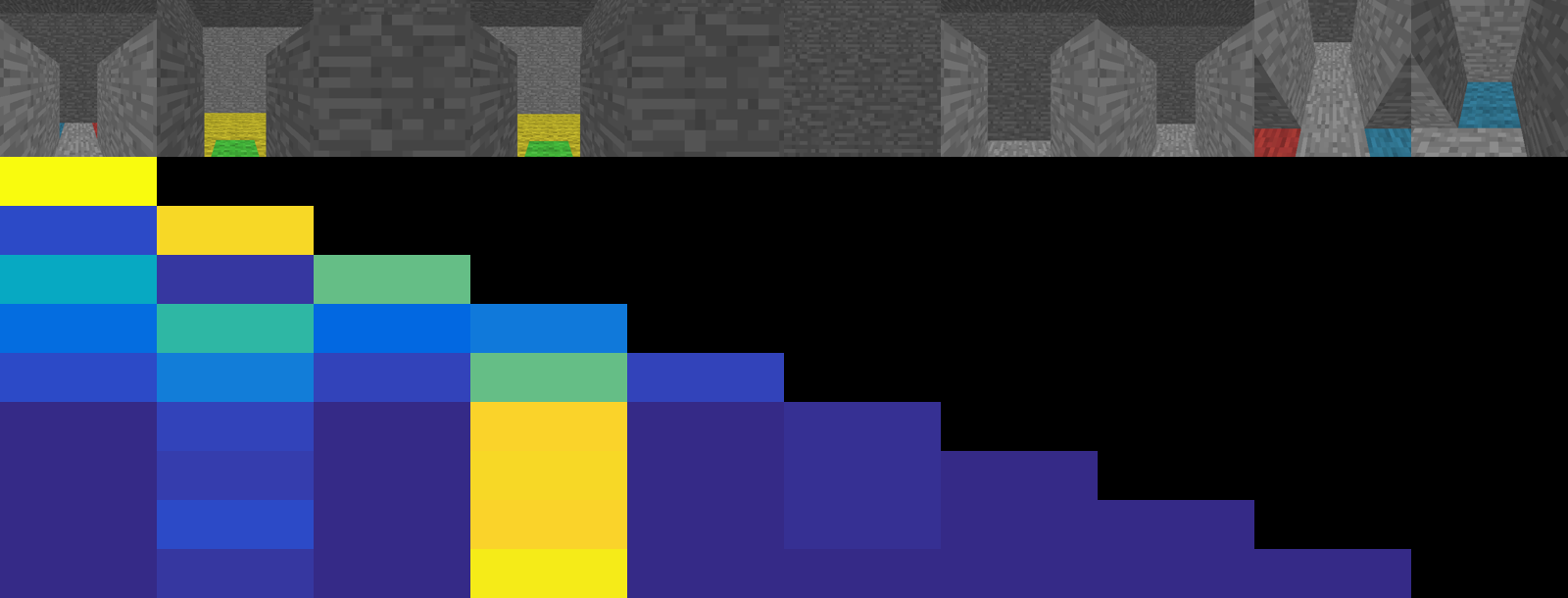} 
	    \caption{Pattern matching (\S\ref{sec:exp-pattern-matching})} %
    	\label{fig:play-matching}
    \end{subfigure}
    \hfill
     \begin{subfigure}{0.48\linewidth}
	    \centering
	    \includegraphics[width=\linewidth]{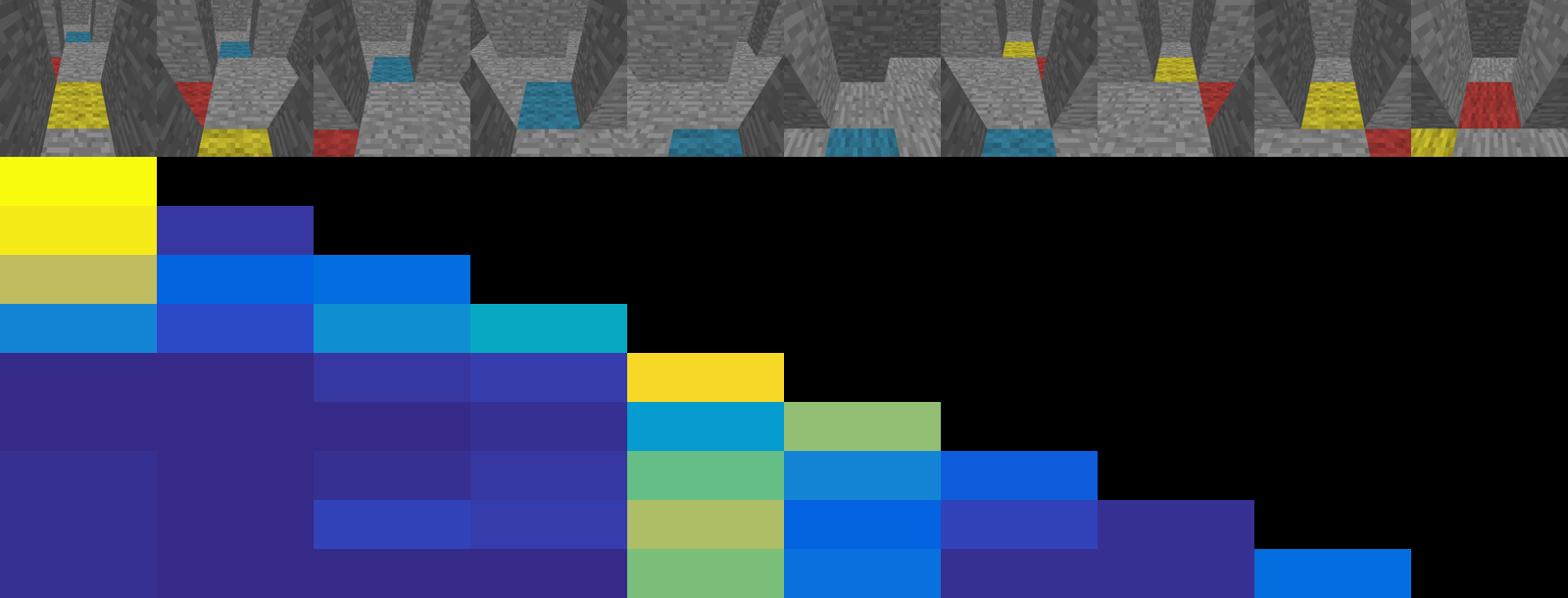} 
	    \caption{Sequential w/ Ind (\S\ref{sec:exp-random-mazes})} %
     	\label{fig:play-seq-i}
     \end{subfigure}
    \vspace{-5pt}
    \caption{Visualization of FRMQN's memory retrieval. Each figure shows a trajectory of FRMQN at the top row, and the following rows visualize attention weights over time. (a) The agent looks at the indicator, goes to the end of the corridor, and retrieves the indicator frame before visiting the goal block. (b) The agent looks at both rooms at the beginning and gradually switches attention weights from one room to another room as it approaches the goal blocks. (c) The agent pays attention to the indicator (yellow) and the first goal block (blue).} 
    \label{fig:game-play}
    \vspace{-15pt}
\end{figure}

\cutparagraphup
\paragraph{Analysis of memory retrieval.}
An example of FRMQN's memory retrieval is visualized in Figure~\ref{fig:play-matching}. FRMQN pays attention to both rooms, gradually moving weight from one to the other as time progresses, which means that the context vector is repeatedly refined based on the encoded features of the room retrieved through its feedback connections. Given this visualization and its good generalization performance, we hypothesize that FRMQN utilizes its feedback connection to compare the two visual features over time rather than comparing them at a single time-step. This result supports our view that feedback connections can play an important role in tasks where more complex reasoning is required with retrieved memories. 

\begin{table*}
\caption{Performance on random maze. The `Size' column lists the size of each set of maps. The entries in the `Reward', `Success', and `Fail' columns are average rewards, success rates, and failure rates measured from 4 runs. We picked the 10 best parameters based on performance on unseen maps for each run and evaluated them on 1000 episodes. `Success' represents the number of correctly completed episodes divided by the total number of episodes, and `Fail' represents the number of incorrectly completed episodes divided by the total number of episodes (e.g., visiting goals in reverse order in sequential goal tasks). The standard errors are lower than $0.03$, $1.5\%$, $1.0\%$ for all average rewards, success rates, and failure rates respectively. }
\vspace{-5pt}
\label{table:rand}
\vskip 0.1in
\begin{center}
\begin{scriptsize}
\begin{sc}
\setlength{\tabcolsep}{1.2pt}
\begin{tabular}{l|l|c|ccc|ccc|ccc|ccc|ccc}
\hline
\multirow{2}{*}{Task} & \multirow{2}{*}{Type} & \multirow{2}{*}{Size} & \multicolumn{3}{c|}{DQN} & \multicolumn{3}{c|}{DRQN} & \multicolumn{3}{c|}{MQN} & \multicolumn{3}{c|}{RMQN} & \multicolumn{3}{c}{FRMQN} \\ \cline{4-18}
 & & & reward & success & fail & reward & success & fail & reward & success & fail & reward & success & fail & reward & success & fail \\ 
\hline
\multirow{3}{*}{Single} & train &4-8 & $0.31$ & $90.4\%$ & $0.6\%$ & $\textbf{0.45}$ & $94.5\%$ & $0.1\%$ & $0.01$ & $78.8\%$ & $0.4\%$ & $\textbf{0.49}$ & $95.7\%$ & $0.1\%$ & $\textbf{0.46}$ & $94.6\%$ & $0.3\%$  \\ 
 & unseen & 4-8 & $0.22$ & $87.3\%$ & $0.7\%$ & $0.23$ & $86.6\%$ & $0.2\%$ & $0.02$ & $79.4\%$ & $0.3\%$ & $\textbf{0.30}$ & $89.4\%$ & $0.3\%$ & $\textbf{0.26}$ & $88.0\%$ & $0.5\%$  \\ 
 & unseen-l & 9-14 & $\textbf{-0.28}$ & $70.0\%$ & $0.3\%$ & $-0.40$ & $63.0\%$ & $0.1\%$ & $-0.63$ & $54.3\%$ & $0.4\%$ & $\textbf{-0.28}$ & $69.3\%$ & $0.1\%$ & $\textbf{-0.28}$ & $69.0\%$ & $0.1\%$  \\  \hline
\multirow{3}{*}{Seq} & train &5-7 & $-0.60$ & $47.6\%$ & $0.8\%$ & $-0.08$ & $66.0\%$ & $0.6\%$ & $-0.48$ & $52.1\%$ & $0.1\%$ & $\textbf{0.21}$ & $77.0\%$ & $0.2\%$ & $\textbf{0.22}$ & $77.6\%$ & $0.2\%$  \\ 
 & unseen & 5-7 & $-0.66$ & $45.0\%$ & $1.0\%$ & $-0.54$ & $48.5\%$ & $0.9\%$ & $-0.59$ & $48.4\%$ & $0.1\%$ & $\textbf{-0.13}$ & $64.3\%$ & $0.1\%$ & $\textbf{-0.18}$ & $63.1\%$ & $0.3\%$  \\ 
 & unseen-l & 8-10 & $-0.82$ & $36.6\%$ & $1.4\%$ & $-0.89$ & $32.6\%$ & $1.0\%$ & $-0.77$ & $38.9\%$ & $0.6\%$ & $\textbf{-0.43}$ & $49.6\%$ & $1.1\%$ & $\textbf{-0.42}$ & $50.8\%$ & $1.0\%$  \\  \hline
\multirow{3}{*}{Single+I} & train &5-7 & $-0.04$ & $79.3\%$ & $6.3\%$ & $0.23$ & $87.9\%$ & $1.2\%$ & $0.11$ & $83.9\%$ & $0.7\%$ & $\textbf{0.34}$ & $91.7\%$ & $0.8\%$ & $0.24$ & $88.0\%$ & $1.4\%$  \\ 
 & unseen & 5-7 & $-0.41$ & $64.8\%$ & $16.1\%$ & $-0.46$ & $61.0\%$ & $13.4\%$ & $-0.46$ & $64.2\%$ & $7.8\%$ & $\textbf{-0.27}$ & $70.0\%$ & $10.2\%$ & $\textbf{-0.23}$ & $71.8\%$ & $8.2\%$  \\ 
 & unseen-l & 8-10 & $-0.74$ & $49.4\%$ & $31.6\%$ & $-0.98$ & $38.5\%$ & $28.3\%$ & $-0.66$ & $55.5\%$ & $17.1\%$ & $\textbf{-0.39}$ & $63.4\%$ & $20.4\%$ & $\textbf{-0.43}$ & $63.4\%$ & $17.2\%$  \\  \hline
\multirow{3}{*}{Seq+I} & train &4-6 & $-0.13$ & $68.0\%$ & $7.0\%$ & $0.25$ & $78.5\%$ & $1.1\%$ & $-0.07$ & $67.7\%$ & $2.3\%$ & $0.37$ & $83.7\%$ & $1.0\%$ & $\textbf{0.48}$ & $87.4\%$ & $0.9\%$  \\ 
 & unseen & 4-6 & $-0.58$ & $54.5\%$ & $14.5\%$ & $-0.65$ & $48.8\%$ & $9.7\%$ & $-0.71$ & $47.3\%$ & $7.2\%$ & $\textbf{-0.32}$ & $62.4\%$ & $7.2\%$ & $\textbf{-0.28}$ & $63.8\%$ & $7.5\%$  \\ 
 & unseen-l & 7-9 & $-0.95$ & $39.1\%$ & $17.8\%$ & $-1.14$ & $30.2\%$ & $13.1\%$ & $-1.04$ & $34.4\%$ & $9.9\%$ & $\textbf{-0.60}$ & $49.5\%$ & $12.5\%$ & $\textbf{-0.54}$ & $51.5\%$ & $12.9\%$  \\  \hline
\end{tabular}
\end{sc}
\end{scriptsize}
\end{center}
\vskip -0.27in
\end{table*}

\cutsubsectionup
\subsection{Random Mazes: Description and Results}
\label{sec:exp-random-mazes}
\cutsubsectiondown

\paragraph{Task.} 
A random maze task consists of randomly generated walls and goal locations as shown in Figure~\ref{fig:map-rand} and~\ref{fig:map-rand-i}. We present 4 classes of tasks using random mazes.
\vspace{-10pt}
\begin{itemize}[leftmargin=*]
  \setlength\itemsep{0em}
  \item \textbf{Single Goal}: The task is to visit the blue block which gives +1 reward while avoiding the red block that gives -1 reward. 
  \item \textbf{Sequential Goals}: The task is to visit the red block first and then the blue block later which gives +0.5 and +1 reward respectively. If an agent visits the colored blocks in the reverse order, it receives -0.5 and -1 reward respectively.
  \item \textbf{Single Goal with Indicator}: If the indicator is yellow, the task is to visit the red block. If the indicator is green, the task is to visit the blue block. Visiting the correct block results in +1 reward and visiting the incorrect block results in -1 reward. 
  \item \textbf{Sequential Goals with Indicator}: If the indicator is yellow, the task is to visit the blue block first and then the red block. If the indicator is green, the task is to visit the red block first and then the blue block. Visiting the blocks in the correct order results in +0.5 for the first block and +1 reward for the second block. Visiting the blocks in the reverse order results in -0.5 and -1 reward respectively.
\end{itemize}
\vspace{-10pt}
We randomly generated 1000 maps used for training and two types of unseen evaluation sets of maps: 1000 maps of the same sizes present in the training maps and 1000 larger maps. The last 10 frames were given as input for all architectures, and the size of memory was set to 9. 
\cutparagraphdown

\cutparagraphup
\paragraph{Performance on the training set.}
\cutparagraphdown
In this task, the agent not only needs to remember important information while traversing the maps (e.g., an indicator) but it also has to search for the goals as different maps have different obstacle and goal locations. Table~\ref{table:rand} shows that RMQN and FRMQN achieve higher asymptotic performances than the other architectures on the training set of maps.

\begin{figure}
    \centering
    \includegraphics[width=0.65\linewidth]{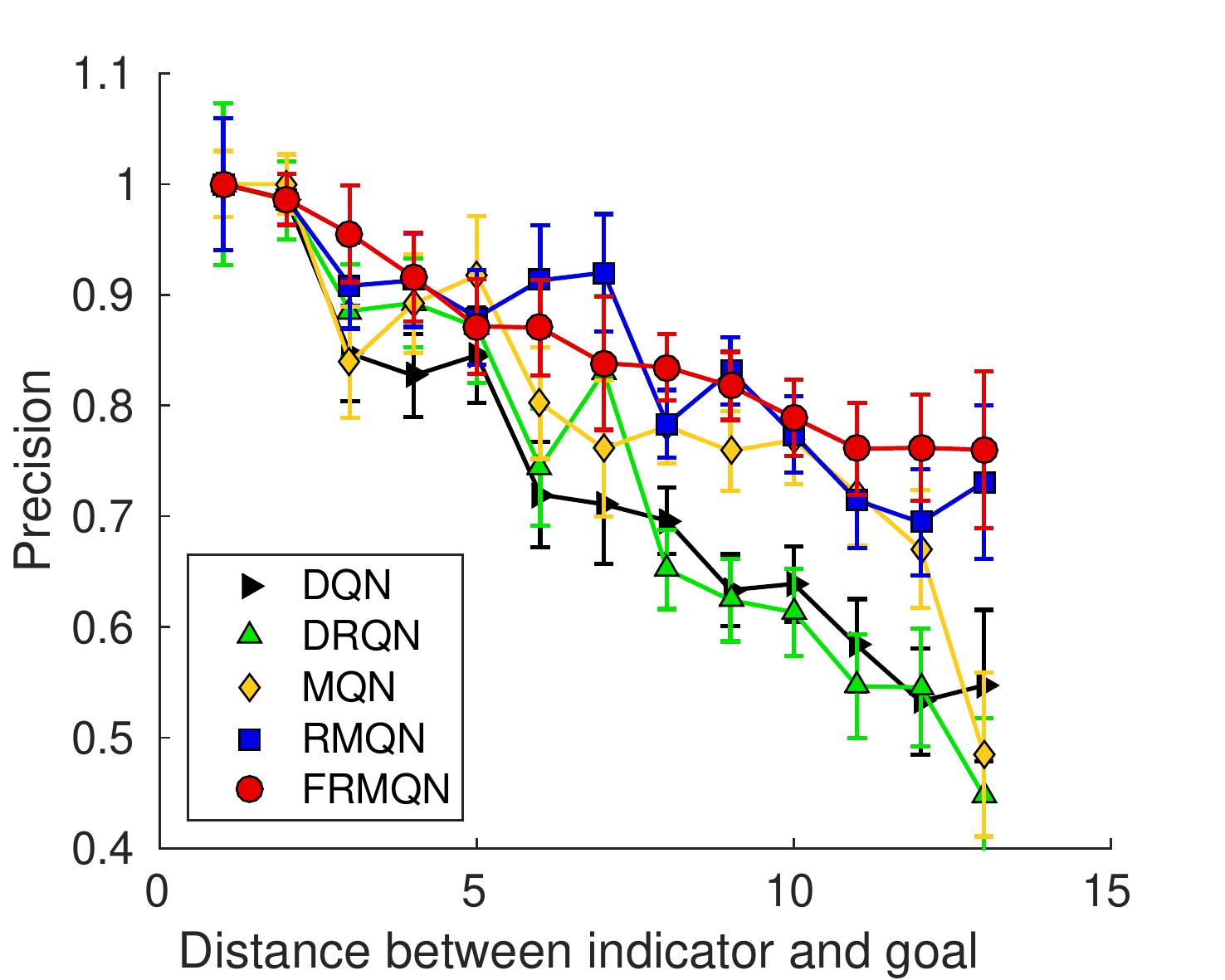}
    \vspace{-10pt}
	\caption{Precision vs. distance. X-axis represents the distance between indicator and goal in Single Goal with Indicator task. Y-axis represents the number of correct goal visits divided by the total number of goal visits. }
	\label{fig:plot-dist}
	 \vspace{-10pt}
\end{figure}

\cutparagraphup
\paragraph{Generalization performance.}
\cutparagraphdown
For the larger-sized unseen maps, we terminated episodes after 100 steps rather than 50 steps and used a time penalty of $-0.02$ considering their size. During evaluation, we used 10 frames as input for DQN and DRQN and 30 frames for MQN, RMQN, and FRMQN; these choices gave the best results for each architecture.

The results in Table~\ref{table:rand} show that, as expected, the performance of all the architectures worsen in unseen maps. From the learning curves (see Figure~\ref{fig:plot}e-g), we observed that generalization performance on unseen maps does not improve after some epochs, even though training performance is improving. This implies that improving policies on a fixed set of maps does not necessarily guarantee better performance on new environments. However, RMQN and FRMQN generalize better than the other architectures in most of the tasks. In particular, compared to the other architectures, DRQN's performance is significantly degraded on unseen maps. In addition, while DQN shows good generalization performance on the Single Goal task which primarily requires search, on the other tasks it tends to go to any goal regardless of important information (e.g., color of indicator). This can be seen through the higher failure rate (the number of incorrectly completed episodes divided by the total number of episodes) of DQN on indicator tasks in Table~\ref{table:rand}.

To investigate how well the architectures handle partial observability, we measured precision (proportion of correct goal visits to all goal visits) versus the distance between goal and indicator in Single Goal with Indicator task, which is visualized in Figure~\ref{fig:plot-dist}. Notably, the gap between our architectures (RMQN and FRMQN) and the other architectures becomes larger as the distance increases. This result implies that our architectures are better at handling partial observability than the other architectures, because large distance between indicator and goal is more likely to introduce deeper partial observability (i.e., long-term dependency). 

Compared to MQN, the RMQN and FRMQN architectures achieve better generalization performance which suggests that the recurrent connections in the latter two architectures are a crucial component for handling random topologies. In addition, FRMQN and RMQN achieve similar performances, which implies that the feedback connection may not be always helpful in these tasks. We note that given a retrieved memory (e.g., indicator), the reasoning required for these tasks is simpler than the reasoning required for Pattern Matching task. 

\cutparagraphup
\paragraph{Analysis of memory retrieval.}
\cutparagraphdown
An example of memory retrieval in FRMQN is visualized in Figure~\ref{fig:play-seq-i}. It retrieves memory that contains important information (e.g., indicator) before it visits a goal block. The memory retrieval strategy is reasonable and is an evidence that the proposed architectures make it easier to generalize to large-scale environments by better handling partial observability.

\cutsectionup
\section{Discussion}
\cutsectiondown
In this paper, we introduced three classes of cognition-inspired tasks in Minecraft and compared the performance of two existing architectures with three architectures that we proposed here. We emphasize that unlike most evaluations of RL algorithms, we trained and evaluated architectures on disjoint sets of maps so as to specifically consider the applicability of learned value functions to unseen (interpolation and extrapolation) maps. 




In summary, our main empirical result is that context-dependent memory retrieval, particularly with a feedback connection from the retrieved memory, can more effectively solve our set of tasks that require control of active perception and external physical movement actions. Our architectures, particularly FRQMN, also show superior ability relative to the baseline architectures when learning value functions whose behavior generalizes better from training to unseen environments. In future work, we intend to take advantage of the flexibility of the Minecraft domain to construct even more challenging cognitive tasks to further evaluate our architectures.



\cutparagraphup
\clearpage
\section*{Acknowledgement}
This work was supported by NSF grant IIS-1526059. Any opinions, findings, conclusions, or recommendations expressed here are those of the authors and do not necessarily reflect the views of the sponsor.
\bibliographystyle{icml2016}
\bibliography{references}
\clearpage
\appendix

\section{Implementation Details}
\subsection{Hyperparameters}
For all architectures, the first convolution layer consists of $32, 4\times 4,$ filters with a stride of 2 and a padding of 1. The second convolution layer consists of $64, 4\times 4,$ filters with a stride of 2 and a padding of 1. In Deep Q-Learning, batch size of 32 and discount factor of $0.99$ are used. We used a replay memory size of $10^6$ for random mazes and $5\times10^4$ for I-Maze and Pattern Matching tasks. We linearly interpolated $\epsilon$ from $1$ to $0.1$ for the initial $10^6$ steps in the $\epsilon$-greedy policy. We chose the best learning rate from $\{0.0001, 0.00025, 0.0005, 0.001\}$ that does not lead to value function explosion depending on the tasks and architectures. The chosen learning rates are shown in Table~\ref{table:lr}. The parameter is updated after every $4$ steps. RMSProp was used with a momentum of 0.95 and a momentum of squared gradients of 0.95. Gradients were clipped at $l2$-norm of 20 to prevent divergence. We used ``soft'' target Q-network updates with a momentum of 0.999 as suggested by~\cite{lillicrap2015continuous}.

\subsection{Map Generation for Pattern Matching}
There are a total of 512 possible visual patterns in a $3\times 3$ room with blocks of two colors. We randomly picked 250 patterns and generated two maps for each pattern: one that contains the same pattern in two rooms and another that has a different randomly generated pattern in one of the rooms that is randomly selected. This produces 500 maps, 250 with identical rooms, and 250 with different rooms, which are used for training. For evaluating generalization, we picked another exclusive set of 250 visual patterns, and generated 500 maps by following the same procedure.

\newpage
\begin{table}[ht!]
\caption{Learning rates. }
\vspace{-5pt}
\label{table:lr}
\vskip 0.1in
\begin{center}
\begin{small}
\begin{sc}
\setlength{\tabcolsep}{2.4pt}
\begin{tabular}{l|ccccc}
\hline
\abovespace\belowspace
Task & DQN & DRQN & MQN & RMQN & FRMQN  \\
\hline \abovespace
I-Maze & 0.00025 & 0.0005 & 0.0005 & 0.0005 & 0.0005 \\
Matching & 0.00025 & 0.001 & 0.0005 & 0.0005 & 0.0005 \\
Single & 0.0001 & 0.00025 & 0.0001 & 0.00025 & 0.00025 \\
Seq & 0.00025 & 0.0005 & 0.00025 & 0.00025 & 0.00025 \\
Single+I & 0.0001 & 0.0005 & 0.00025 & 0.0005 & 0.00025  \\ \belowspace
Seq+I & 0.00025 & 0.001 & 0.00025 & 0.00025 & 0.0005  \\
\hline
\end{tabular}
\end{sc}
\end{small}
\end{center}
\vskip -0.1in
\end{table}

\clearpage
\begin{figure*}
    \centering
    \begin{subfigure}{0.24\linewidth}
	    \centering
	    \includegraphics[width=\linewidth]{figures/plots/error_imaze_train.pdf} 
	    \caption{I-Maze (Train)}
	\end{subfigure}
    \begin{subfigure}{0.24\linewidth}
		\centering
	    \includegraphics[width=\linewidth]{figures/plots/error_imaze_test.pdf} 
	    \caption{I-Maze (Unseen)}
	\end{subfigure}
     \begin{subfigure}{0.24\linewidth}
 		\centering
 	    \includegraphics[width=\linewidth]{figures/plots/error_match_train.pdf} 
 	    \caption{Pattern Matching (Train)}
 	\end{subfigure}
     \begin{subfigure}{0.24\linewidth}
 		\centering
 	    \includegraphics[width=\linewidth]{figures/plots/error_match_test.pdf} 
 	    \caption{Pattern Matching (Unseen)}
	\end{subfigure}
	\begin{subfigure}{0.24\linewidth}
		\centering
	    \includegraphics[width=\linewidth]{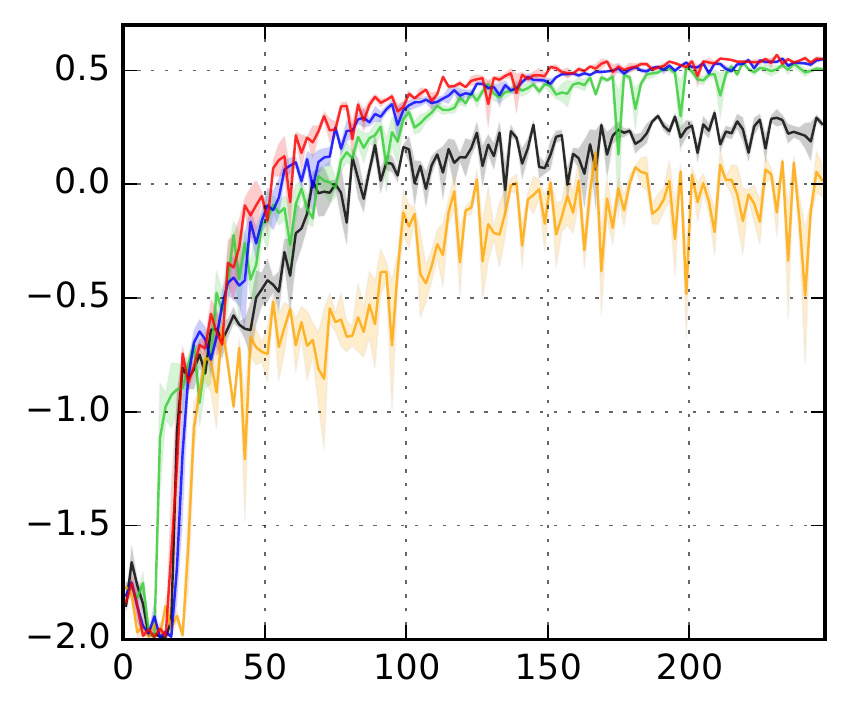} 
	    \caption{Single (Train)}
 	\end{subfigure}
    \begin{subfigure}{0.24\linewidth}
 		\centering
 	     \includegraphics[width=\linewidth]{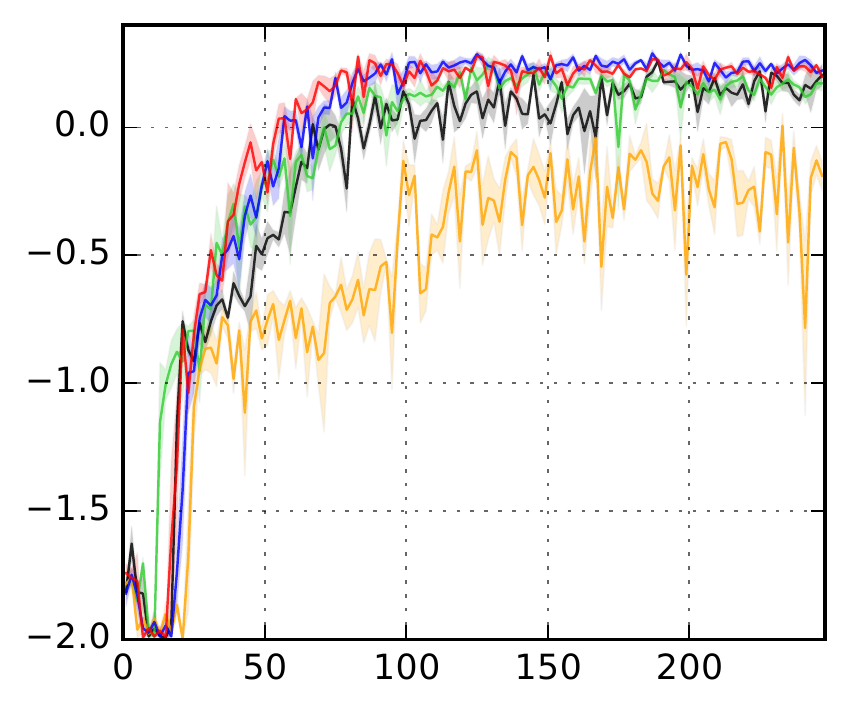} 
	    \caption{Single (Unseen)}
    \end{subfigure}
    \begin{subfigure}{0.24\linewidth}
 		\centering
 	     \includegraphics[width=\linewidth]{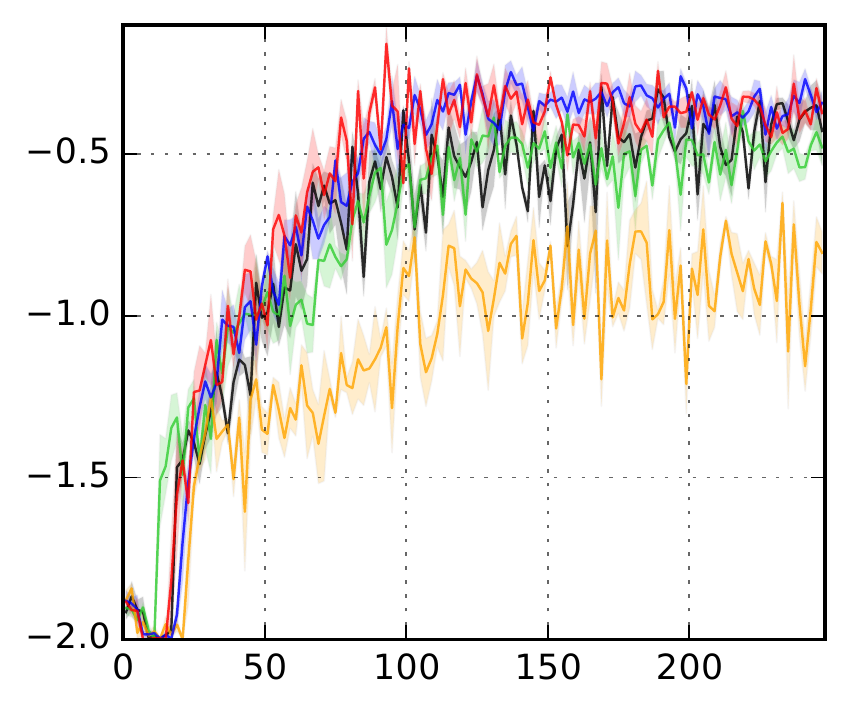} 
	    \caption{Single (Unseen-L)}
    \end{subfigure}
    \begin{subfigure}{0.24\linewidth}
 		\centering
 	    \raisebox{15mm}{\includegraphics[width=0.7\linewidth]{figures/plots/legend.pdf}}
    \end{subfigure}
    	\begin{subfigure}{0.24\linewidth}
		\centering
	    \includegraphics[width=\linewidth]{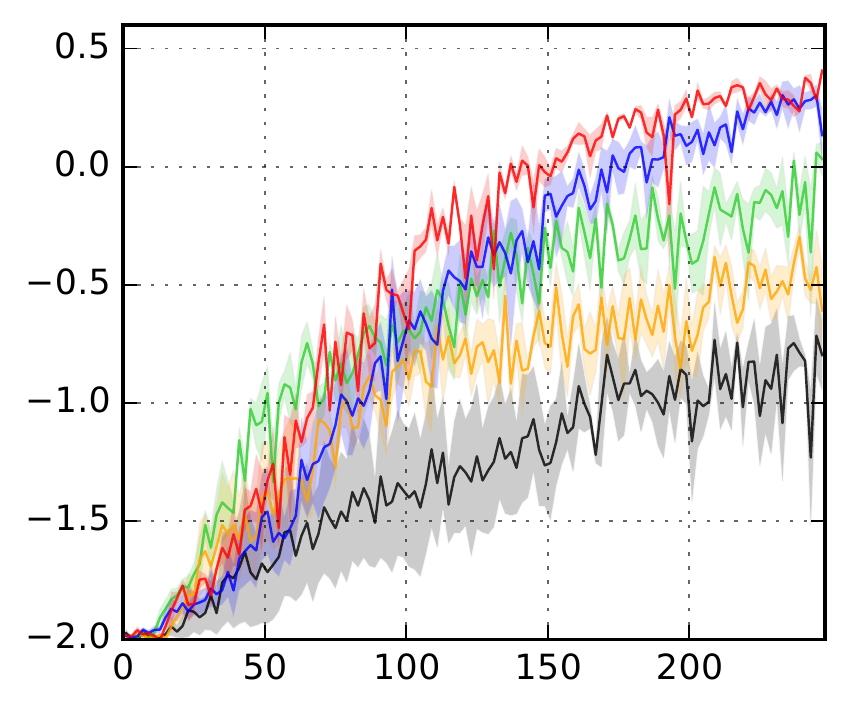} 
	    \caption{Sequential (Train)}
 	\end{subfigure}
    \begin{subfigure}{0.24\linewidth}
 		\centering
 	     \includegraphics[width=\linewidth]{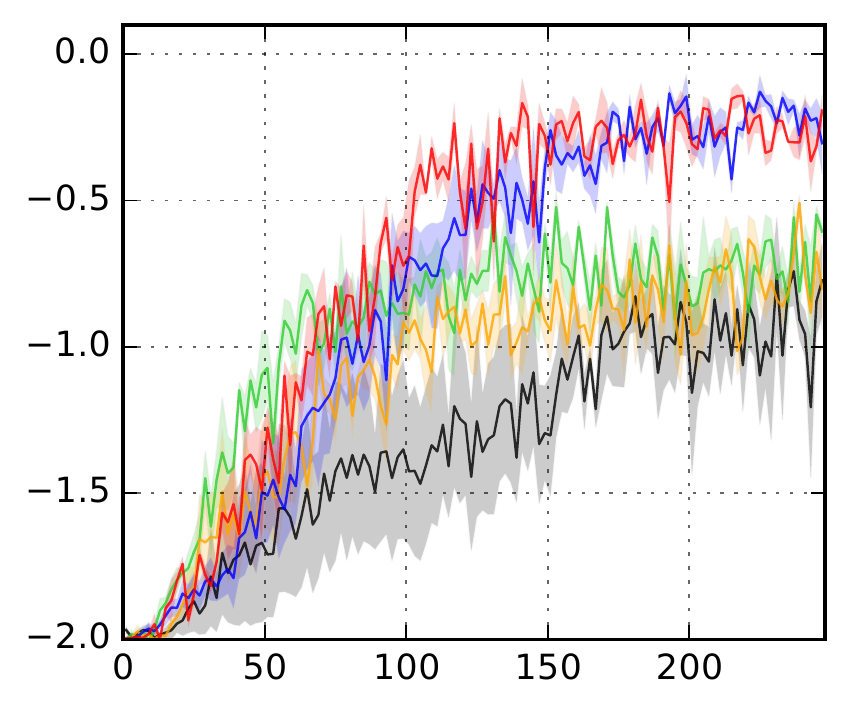} 
	    \caption{Sequential (Unseen)}
    \end{subfigure}
    \begin{subfigure}{0.24\linewidth}
 		\centering
 	     \includegraphics[width=\linewidth]{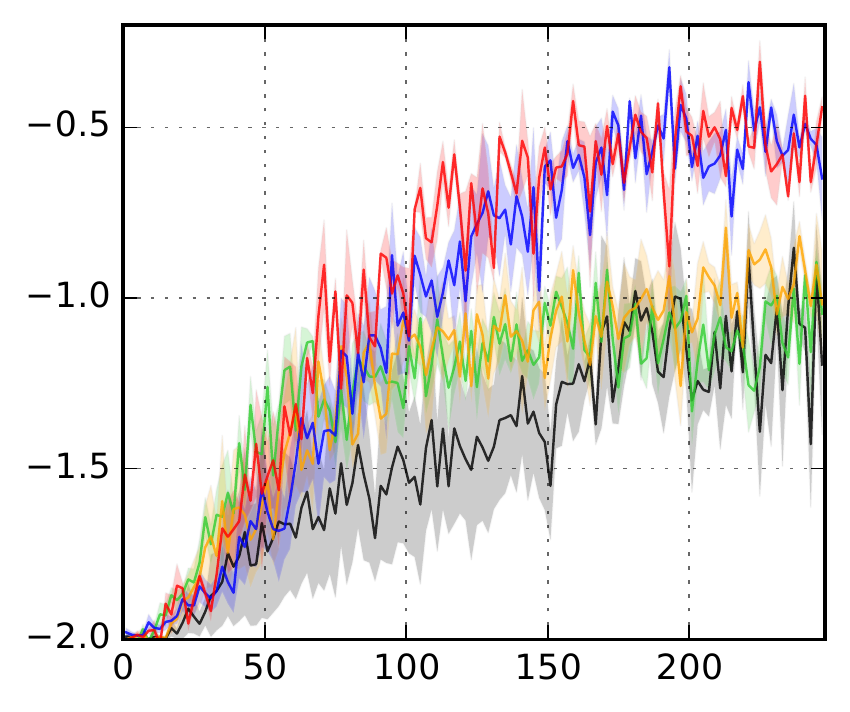} 
	    \caption{Sequential (Unseen-L)}
    \end{subfigure}
    \begin{subfigure}{0.24\linewidth}
 		\hfill
    \end{subfigure}
    	\begin{subfigure}{0.24\linewidth}
		\centering
	    \includegraphics[width=\linewidth]{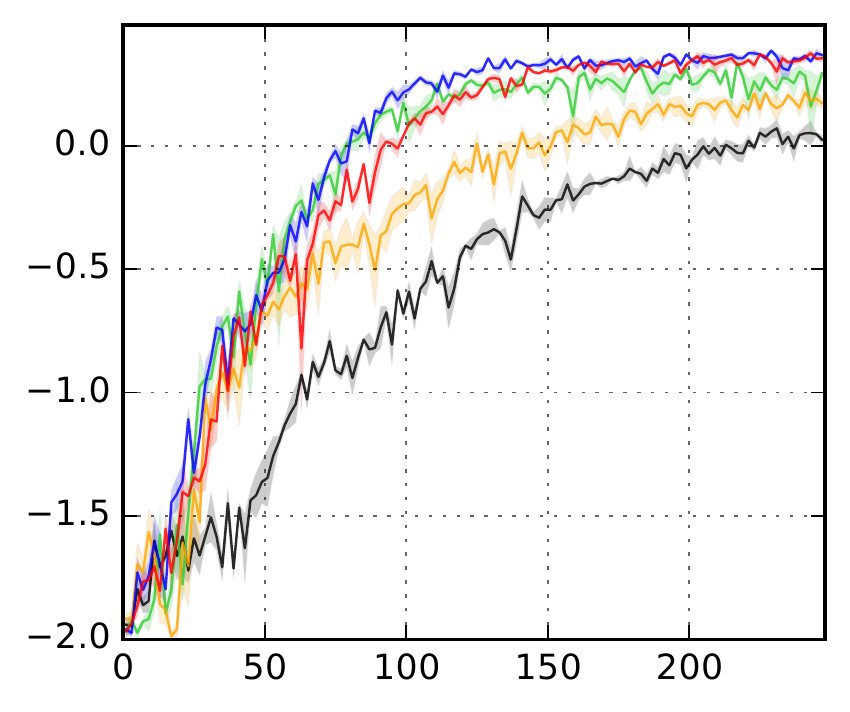} 
	    \caption{Single+I (Train)}
 	\end{subfigure}
    \begin{subfigure}{0.24\linewidth}
 		\centering
 	     \includegraphics[width=\linewidth]{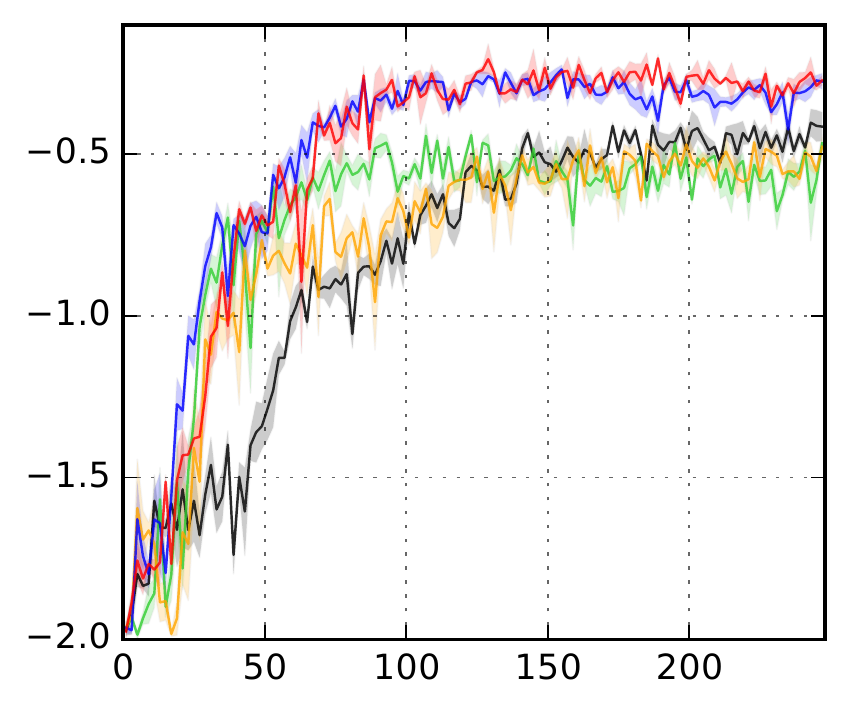} 
	    \caption{Single+I (Unseen)}
    \end{subfigure}
    \begin{subfigure}{0.24\linewidth}
 		\centering
 	     \includegraphics[width=\linewidth]{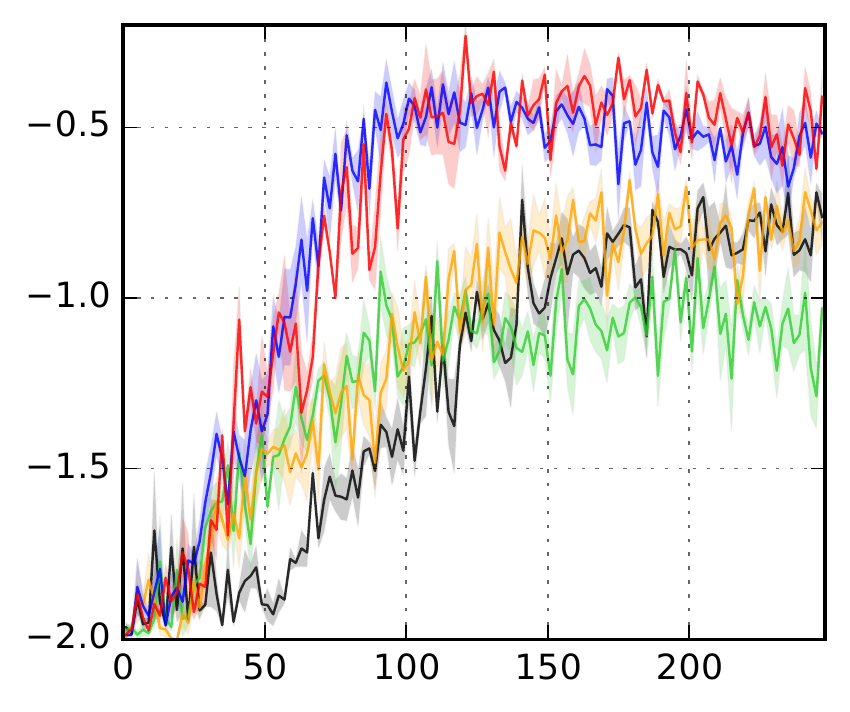} 
	    \caption{Single+I (Unseen-L)}
    \end{subfigure}
    \begin{subfigure}{0.24\linewidth}
 		\hfill
    \end{subfigure}
    \begin{subfigure}{0.24\linewidth}
		\centering
	    \includegraphics[width=\linewidth]{figures/plots/error_seq_indicator_train.pdf} 
	    \caption{Sequential+I (Train)}
 	\end{subfigure}
    \begin{subfigure}{0.24\linewidth}
 		\centering
 	     \includegraphics[width=\linewidth]{figures/plots/error_seq_indicator_test.pdf} 
	    \caption{Sequential+I (Unseen)}
    \end{subfigure}
    \begin{subfigure}{0.24\linewidth}
 		\centering
 	     \includegraphics[width=\linewidth]{figures/plots/error_seq_indicator_large.pdf} 
	    \caption{Sequential+I (Unseen-L)}
    \end{subfigure}
    \begin{subfigure}{0.24\linewidth}
 		\hfill
    \end{subfigure}
    \vspace{-5pt}
    \caption{Learning curves. X-axis and y-axis correspond to the number of training epochs (1 epoch = 10K steps) and the average reward respectively. For I-Maze, `Unseen' represents unseen maps with different sizes. For Pattern Matching, `Unseen' represents maps with different visual patterns. For the rest plots, `Unseen' and `Unseen-L' indicate unseen topologies with the same sizes and larger sizes of maps, respectively. The performance was measured from 4 runs for random mazes and 10 runs for I-Maze and Pattern Matching. } 
\label{fig:full-plot}
\end{figure*}
\clearpage
\begin{figure*}[ht!]
    \small
    \setlength{\tabcolsep}{1pt}
    \def\arraystretch{1}
    \begin{tabular}{llllllllllll}
    \begin{subfigure}{0.08\linewidth}
	    \includegraphics[width=1\linewidth]{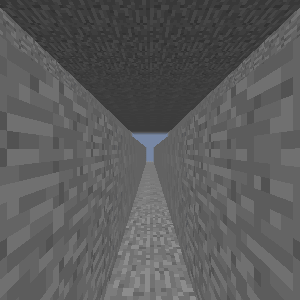} 
   		\includegraphics[width=1\linewidth]{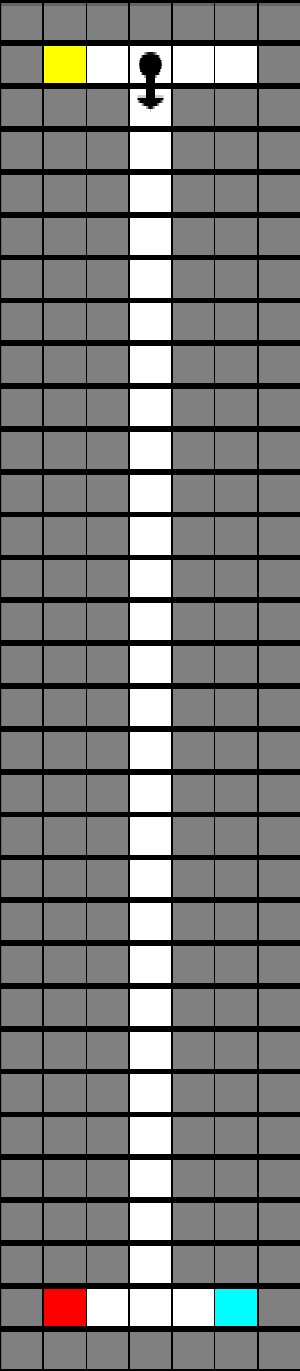} 
   		\caption*{t = 1}
	\end{subfigure} 
	& 
    \begin{subfigure}{0.08\linewidth}
	    \includegraphics[width=1\linewidth]{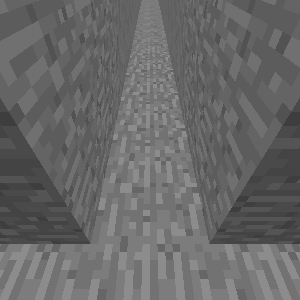} 
   		\includegraphics[width=1\linewidth]{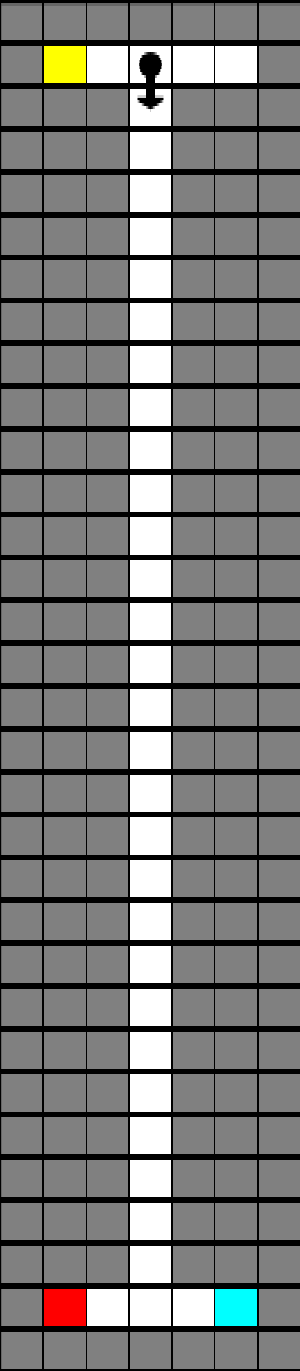} 
   		\caption*{t = 2}
	\end{subfigure} 
	& 
    \begin{subfigure}{0.08\linewidth}
	    \includegraphics[width=1\linewidth]{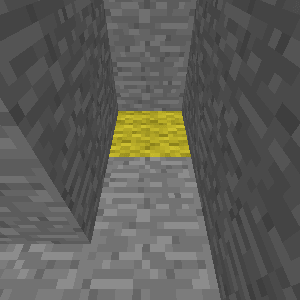} 
   		\includegraphics[width=1\linewidth]{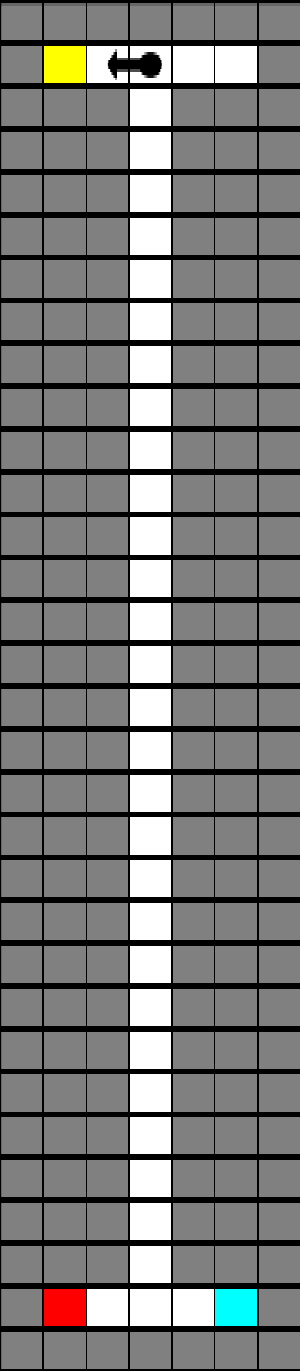} 
   		\caption*{t = 3}
	\end{subfigure} 
	& 
    \begin{subfigure}{0.08\linewidth}
	    \includegraphics[width=1\linewidth]{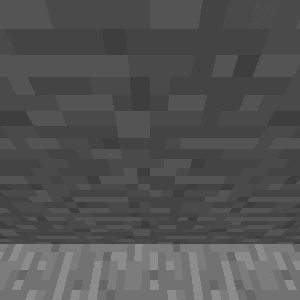} 
   		\includegraphics[width=1\linewidth]{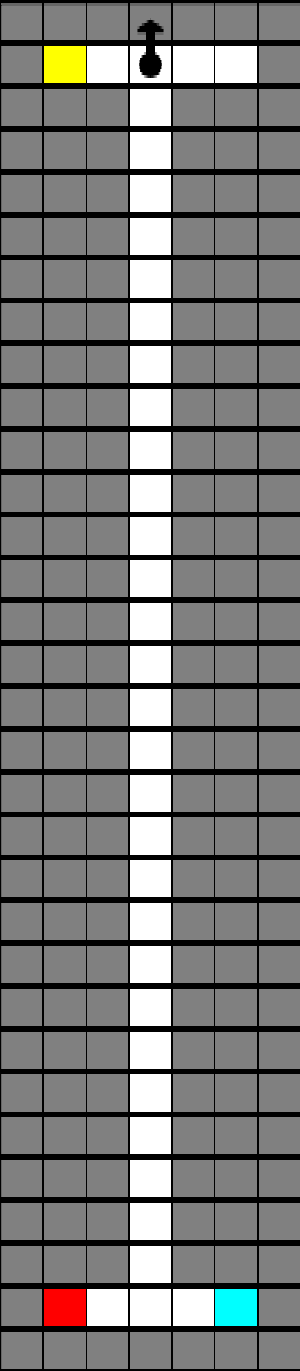} 
   		\caption*{t = 4}
	\end{subfigure}
	& 
    \begin{subfigure}{0.08\linewidth}
	    \includegraphics[width=1\linewidth]{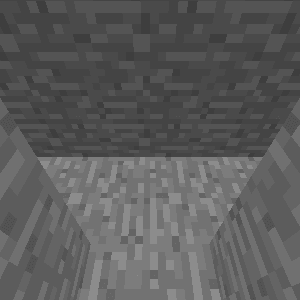} 
   		\includegraphics[width=1\linewidth]{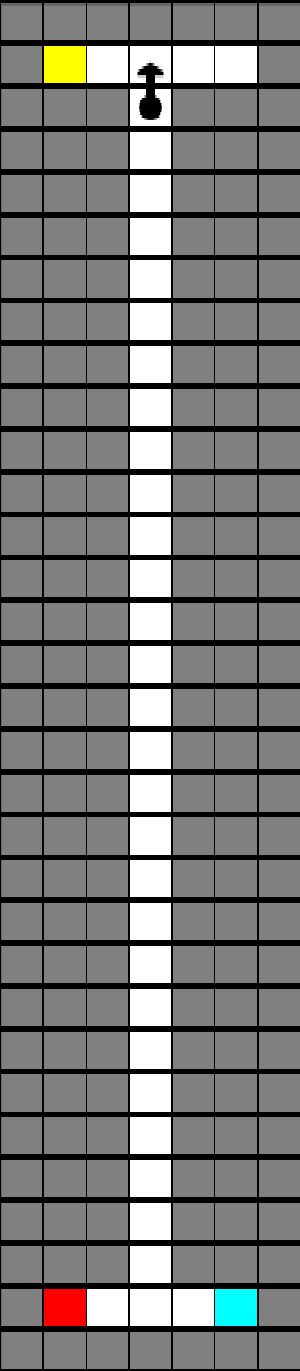} 
   		\caption*{t = 5}
	\end{subfigure}
	& 
    \begin{subfigure}{0.08\linewidth}
	    \includegraphics[width=1\linewidth]{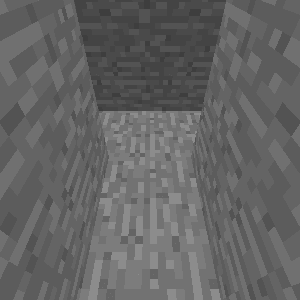} 
   		\includegraphics[width=1\linewidth]{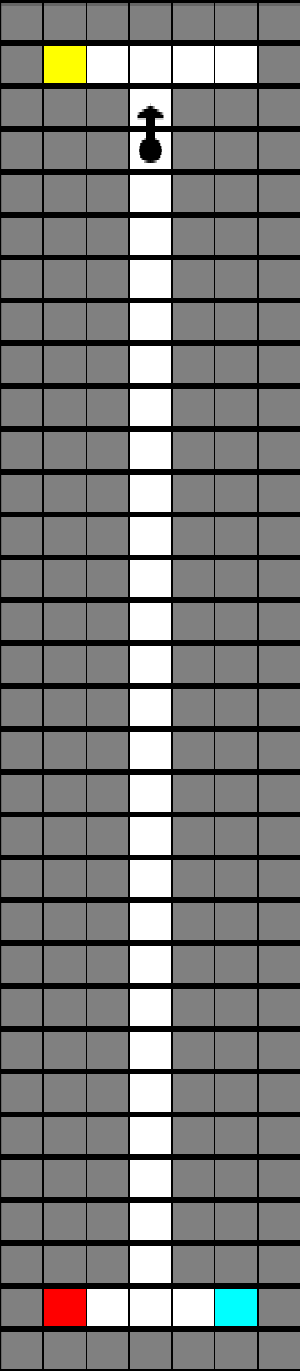} 
   		\caption*{t = 6}
	\end{subfigure} 
	& 
    \begin{subfigure}{0.08\linewidth}
	    \includegraphics[width=1\linewidth]{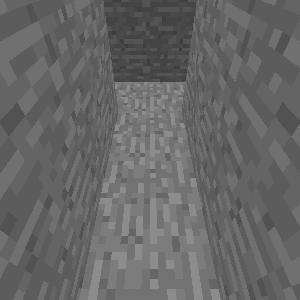} 
   		\includegraphics[width=1\linewidth]{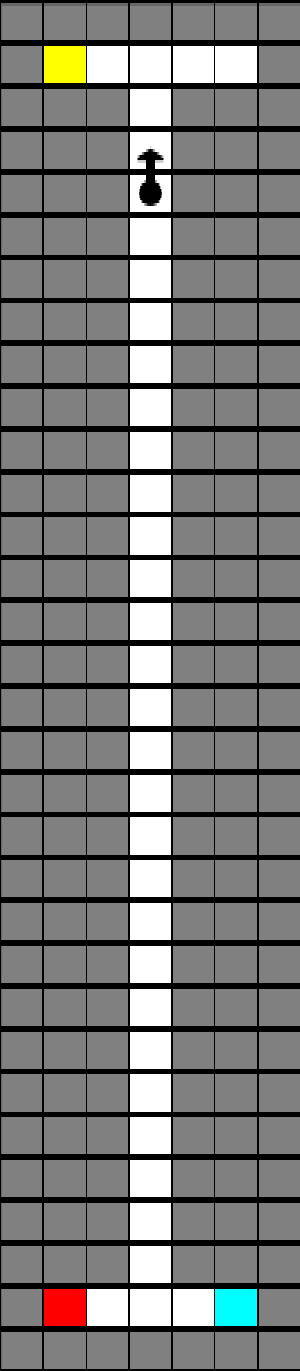} 
   		\caption*{t = 7}
	\end{subfigure} 
	& 
    \begin{subfigure}{0.08\linewidth}
	    \includegraphics[width=1\linewidth]{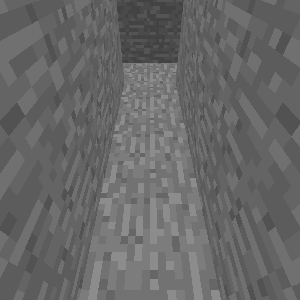} 
   		\includegraphics[width=1\linewidth]{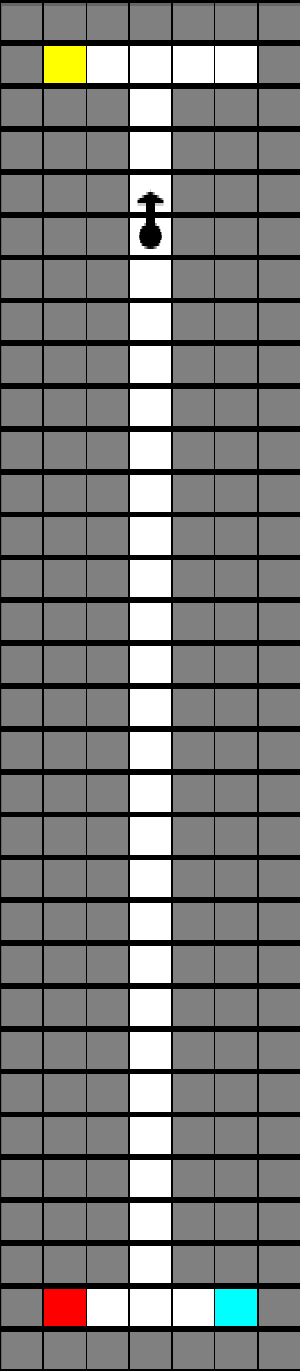} 
   		\caption*{t = 8}
	\end{subfigure} 
	& 
    \begin{subfigure}{0.08\linewidth}
	    \includegraphics[width=1\linewidth]{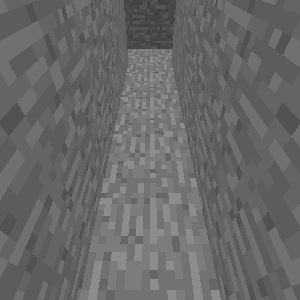} 
   		\includegraphics[width=1\linewidth]{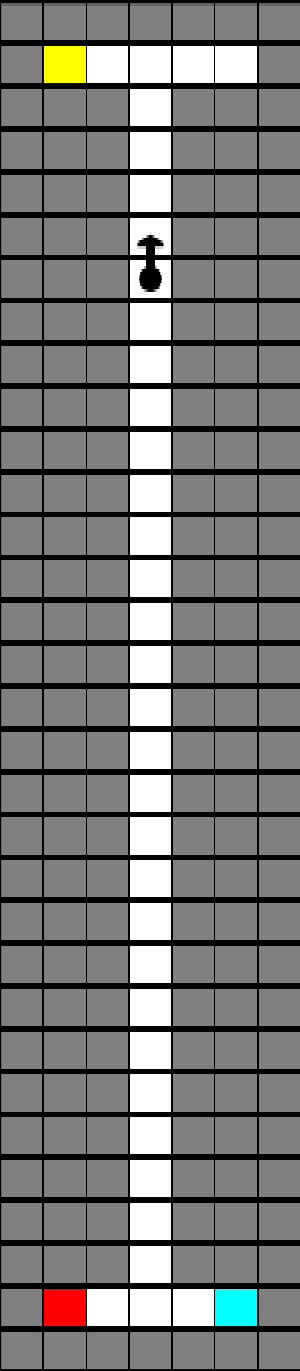} 
   		\caption*{t = 9}
	\end{subfigure}
	& 
    \begin{subfigure}{0.08\linewidth}
	    \includegraphics[width=1\linewidth]{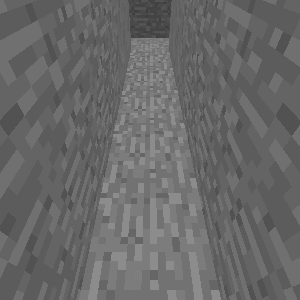} 
   		\includegraphics[width=1\linewidth]{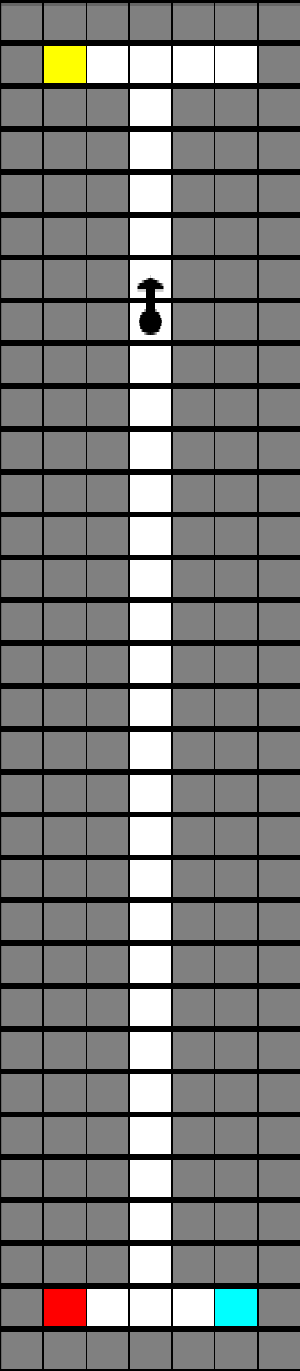} 
   		\caption*{t = 10}
	\end{subfigure}
	& 
    \begin{subfigure}{0.08\linewidth}
	    \includegraphics[width=1\linewidth]{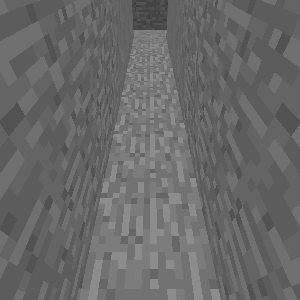} 
   		\includegraphics[width=1\linewidth]{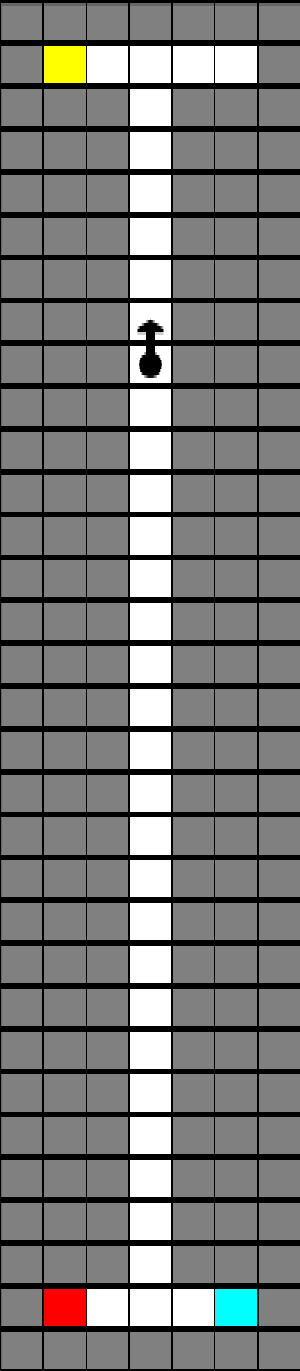} 
   		\caption*{t = 11}
	\end{subfigure} 
	&
    \begin{subfigure}{0.08\linewidth}
	    \includegraphics[width=1\linewidth]{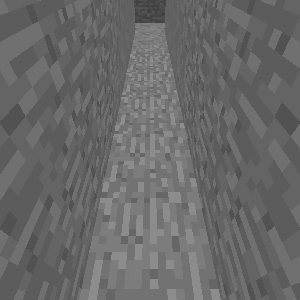} 
   		\includegraphics[width=1\linewidth]{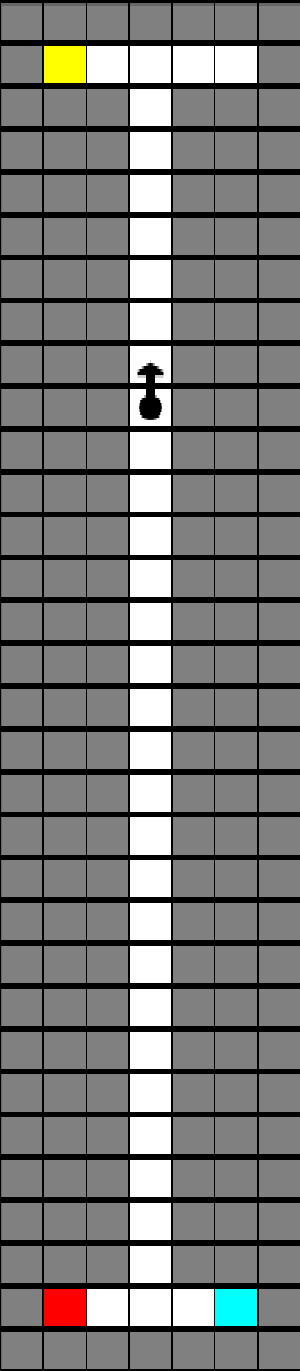} 
   		\caption*{t = 12}
	\end{subfigure} 
	\\
	\\
	\begin{subfigure}{0.08\linewidth}
	    \includegraphics[width=1\linewidth]{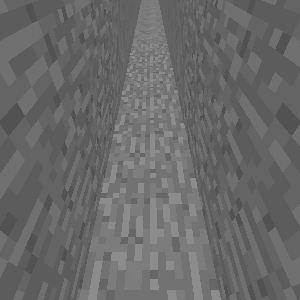} 
   		\includegraphics[width=1\linewidth]{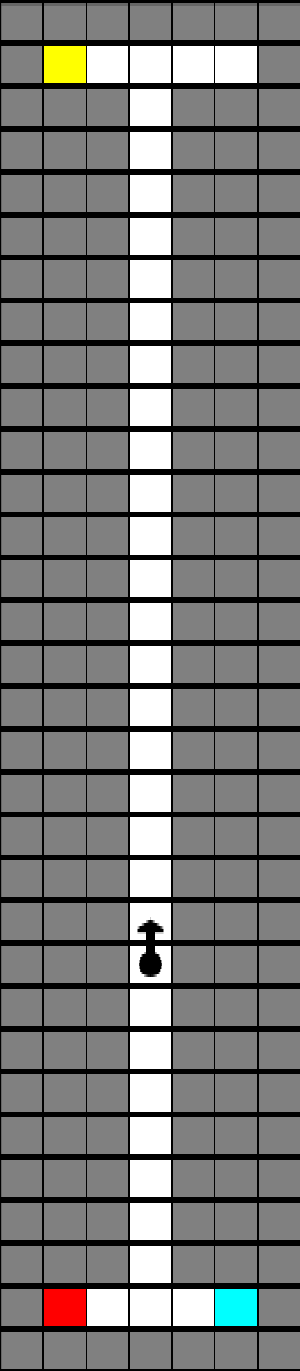} 
   		\caption*{t = 25}
	\end{subfigure}
	&
	\begin{subfigure}{0.08\linewidth}
	    \includegraphics[width=1\linewidth]{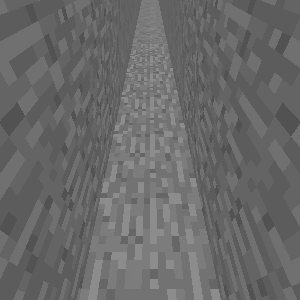} 
   		\includegraphics[width=1\linewidth]{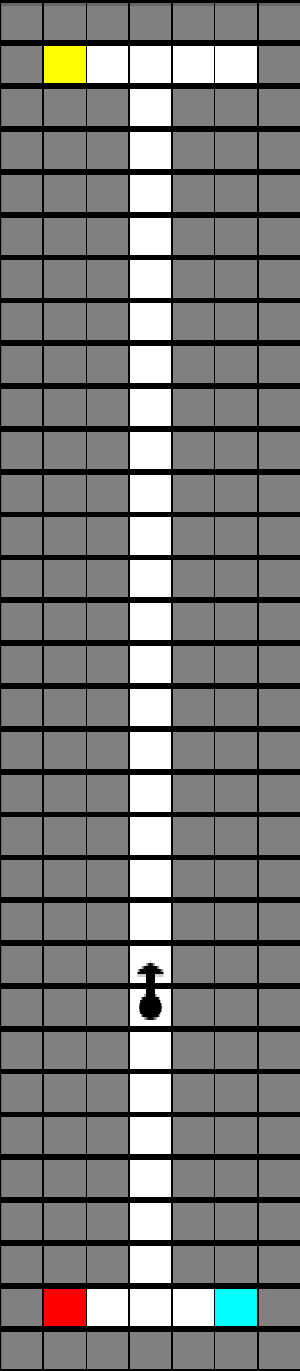} 
   		\caption*{t = 26}
	\end{subfigure}
	&
	\begin{subfigure}{0.08\linewidth}
	    \includegraphics[width=1\linewidth]{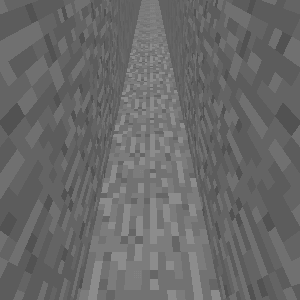} 
   		\includegraphics[width=1\linewidth]{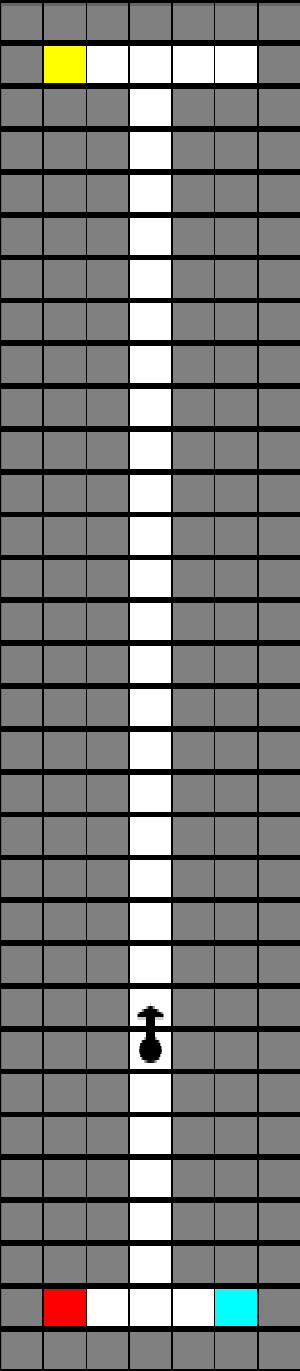} 
   		\caption*{t = 27}
	\end{subfigure}
	&
	\begin{subfigure}{0.08\linewidth}
	    \includegraphics[width=1\linewidth]{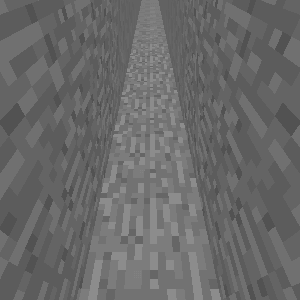} 
   		\includegraphics[width=1\linewidth]{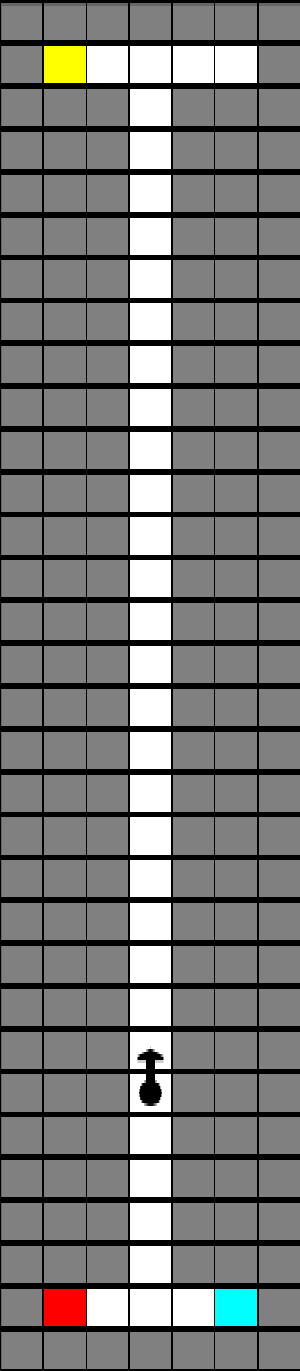} 
   		\caption*{t = 28}
	\end{subfigure}
	&
	\begin{subfigure}{0.08\linewidth}
	    \includegraphics[width=1\linewidth]{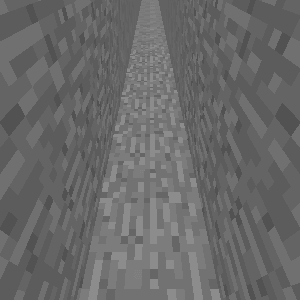} 
   		\includegraphics[width=1\linewidth]{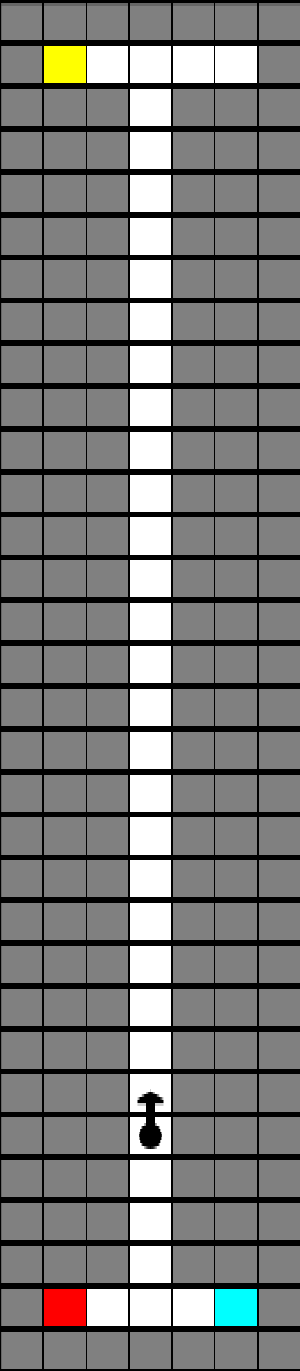} 
   		\caption*{t = 29}
	\end{subfigure}
	&
	\begin{subfigure}{0.08\linewidth}
	    \includegraphics[width=1\linewidth]{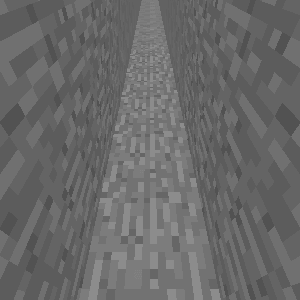} 
   		\includegraphics[width=1\linewidth]{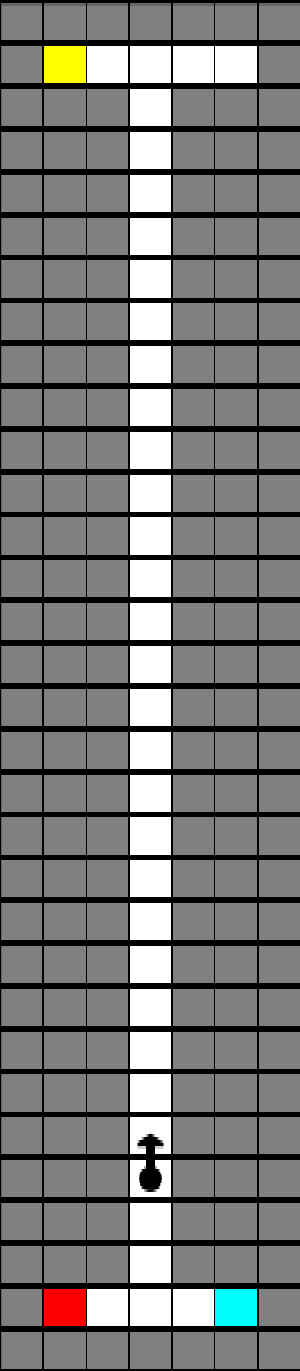} 
   		\caption*{t = 30}
	\end{subfigure}
	&
	\begin{subfigure}{0.08\linewidth}
	    \includegraphics[width=1\linewidth]{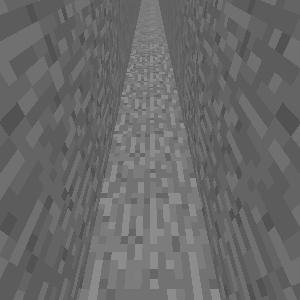} 
   		\includegraphics[width=1\linewidth]{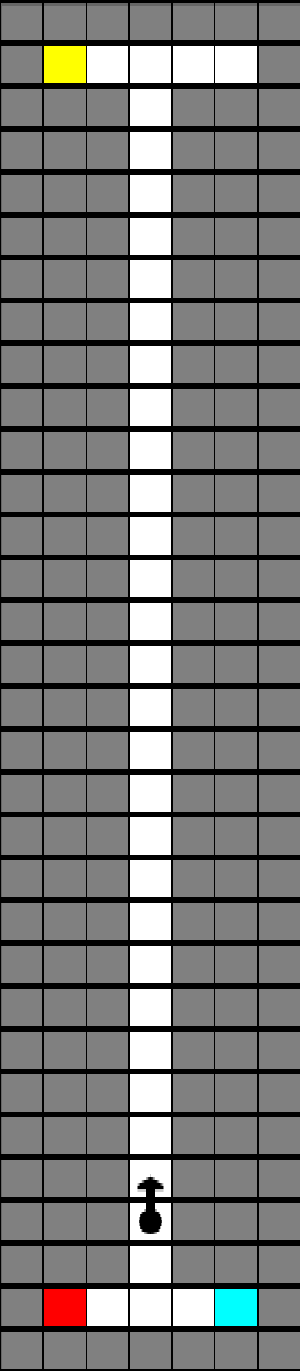} 
   		\caption*{t = 31}
	\end{subfigure}
	&
	\begin{subfigure}{0.08\linewidth}
	    \includegraphics[width=1\linewidth]{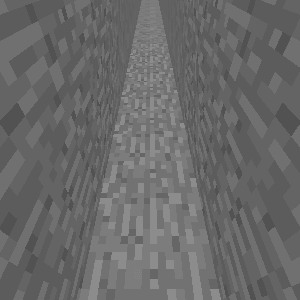} 
   		\includegraphics[width=1\linewidth]{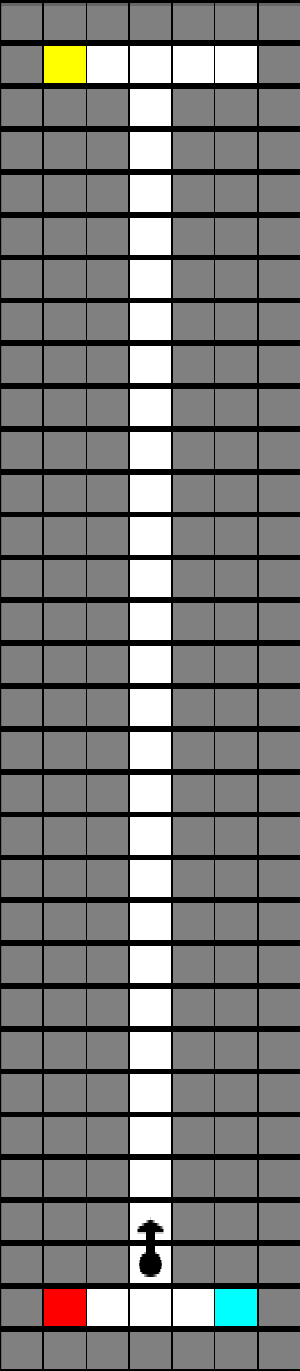} 
   		\caption*{t = 32}
	\end{subfigure}
	&
	\begin{subfigure}{0.08\linewidth}
	    \includegraphics[width=1\linewidth]{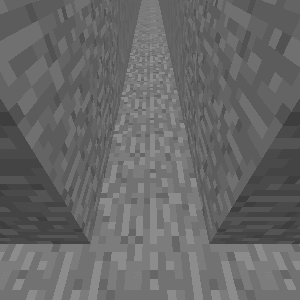} 
   		\includegraphics[width=1\linewidth]{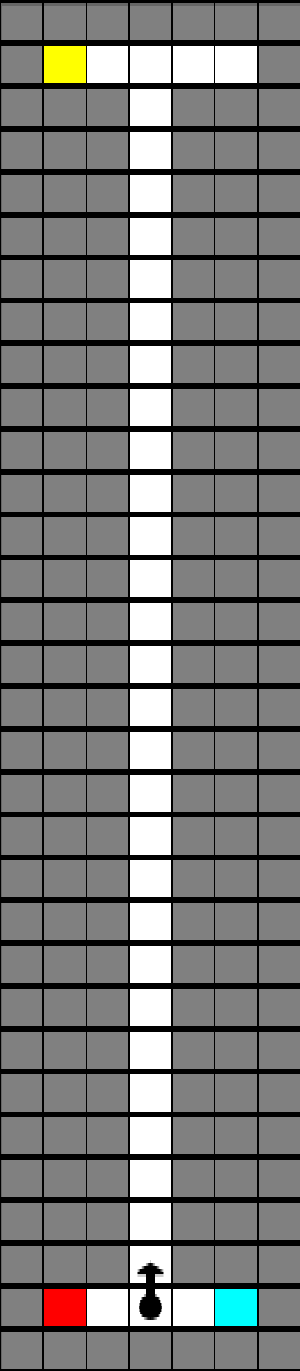} 
   		\caption*{t = 33}
	\end{subfigure}
	&
	\begin{subfigure}{0.08\linewidth}
	    \includegraphics[width=1\linewidth]{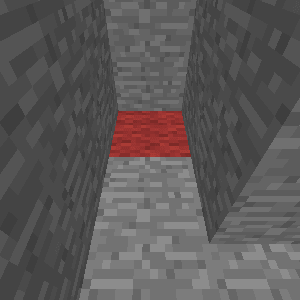} 
   		\includegraphics[width=1\linewidth]{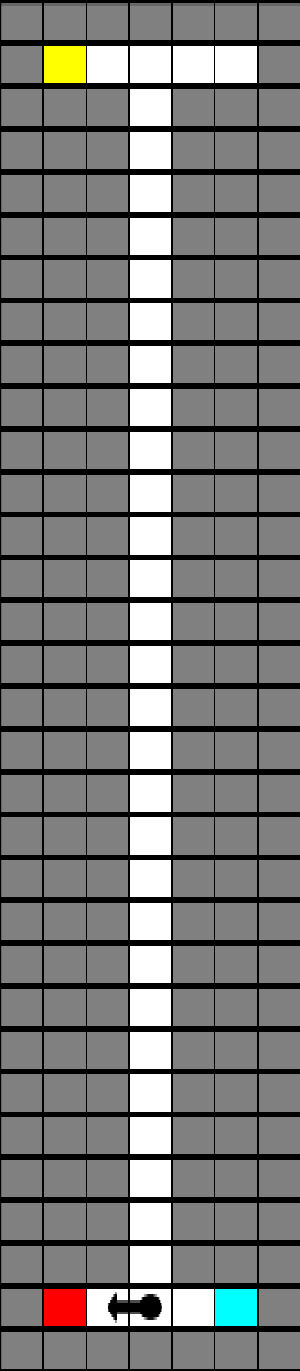} 
   		\caption*{t = 34}
	\end{subfigure}
	&
	\begin{subfigure}{0.08\linewidth}
	    \includegraphics[width=1\linewidth]{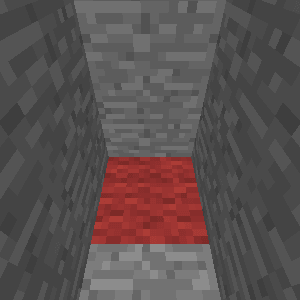} 
   		\includegraphics[width=1\linewidth]{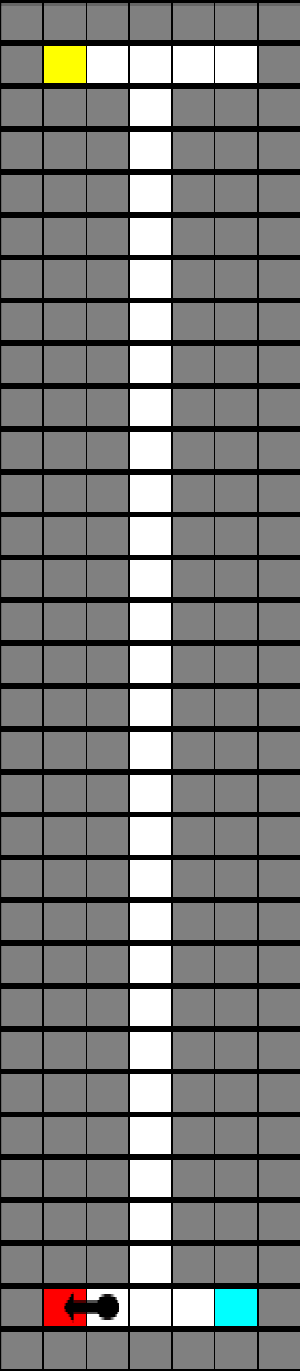} 
   		\caption*{t = 35}
	\end{subfigure}
	&
	\begin{subfigure}{0.08\linewidth}
	    \includegraphics[width=1\linewidth]{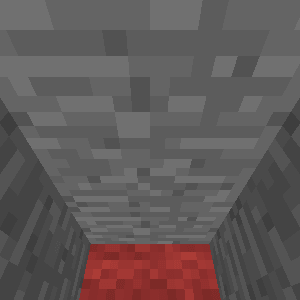} 
   		\includegraphics[width=1\linewidth]{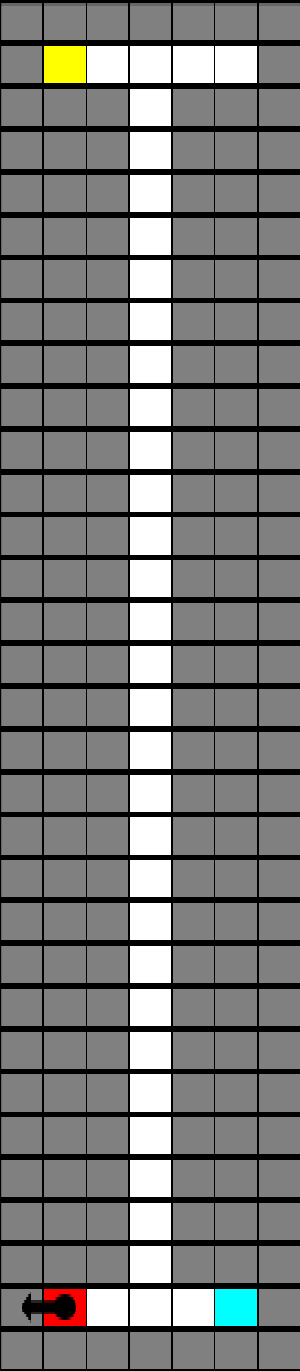} 
   		\caption*{t = 36}
	\end{subfigure}
	\end{tabular}
    \caption{FRMQN's play on an unseen and larger I-maze. The agent successfully completes the task by visiting the red block given the yellow indicator. } 
	\label{play-i-maze-yellow}
\end{figure*}
\begin{figure*}
    \small
    \setlength{\tabcolsep}{1pt}
    \def\arraystretch{1}
    \begin{tabular}{llllllllllll}
    \begin{subfigure}{0.08\linewidth}
	    \includegraphics[width=1\linewidth]{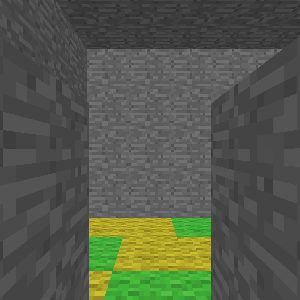} 
   		\includegraphics[width=1\linewidth]{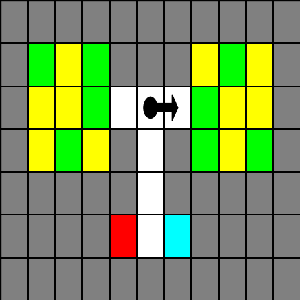} 
   		\caption*{t = 1}
	\end{subfigure} 
	& 
    \begin{subfigure}{0.08\linewidth}
	    \includegraphics[width=1\linewidth]{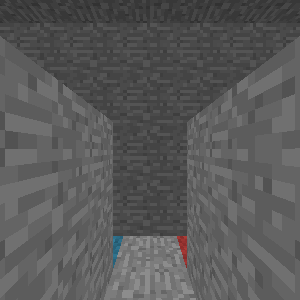} 
   		\includegraphics[width=1\linewidth]{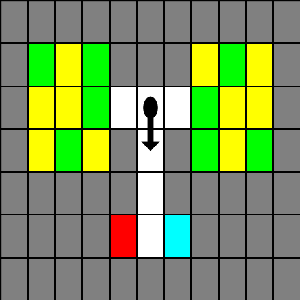} 
   		\caption*{t = 2}
	\end{subfigure} 
	& 
    \begin{subfigure}{0.08\linewidth}
	    \includegraphics[width=1\linewidth]{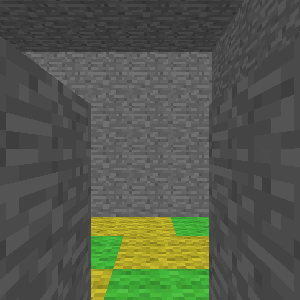} 
   		\includegraphics[width=1\linewidth]{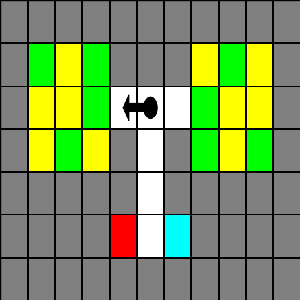} 
   		\caption*{t = 3}
	\end{subfigure} 
	& 
    \begin{subfigure}{0.08\linewidth}
	    \includegraphics[width=1\linewidth]{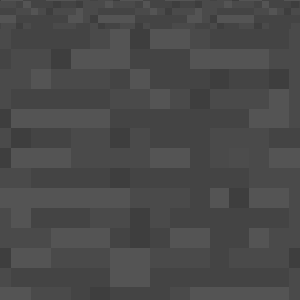} 
   		\includegraphics[width=1\linewidth]{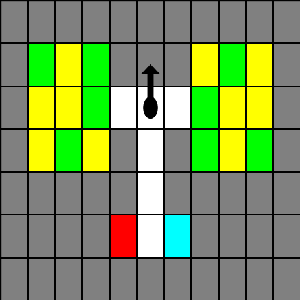} 
   		\caption*{t = 4}
	\end{subfigure}
	& 
    \begin{subfigure}{0.08\linewidth}
	    \includegraphics[width=1\linewidth]{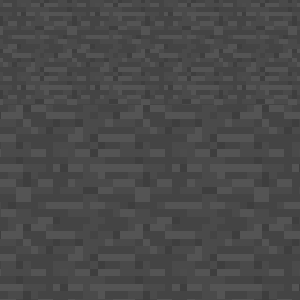} 
   		\includegraphics[width=1\linewidth]{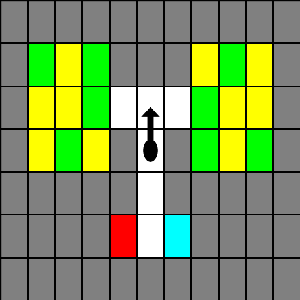} 
   		\caption*{t = 5}
	\end{subfigure}
	& 
    \begin{subfigure}{0.08\linewidth}
	    \includegraphics[width=1\linewidth]{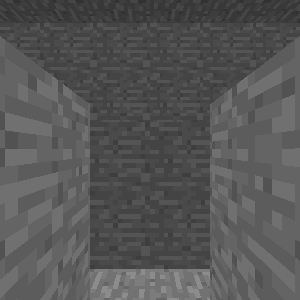} 
   		\includegraphics[width=1\linewidth]{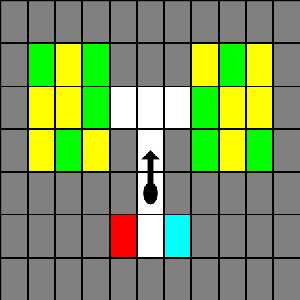} 
   		\caption*{t = 6}
	\end{subfigure} 
	& 
    \begin{subfigure}{0.08\linewidth}
	    \includegraphics[width=1\linewidth]{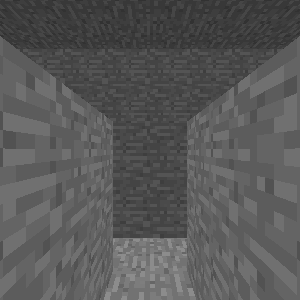} 
   		\includegraphics[width=1\linewidth]{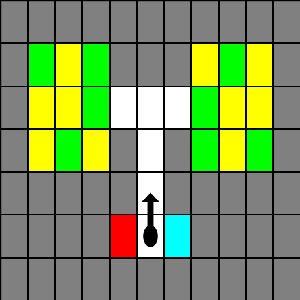} 
   		\caption*{t = 7}
	\end{subfigure} 
	& 
    \begin{subfigure}{0.08\linewidth}
	    \includegraphics[width=1\linewidth]{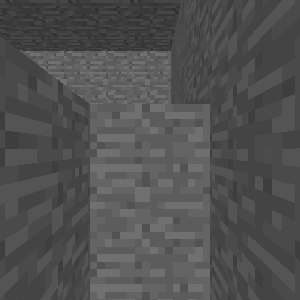} 
   		\includegraphics[width=1\linewidth]{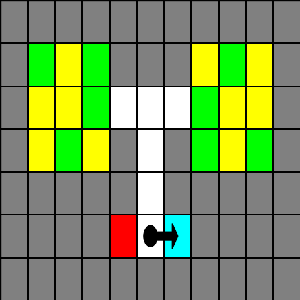} 
   		\caption*{t = 8}
	\end{subfigure} 
	& 
    \begin{subfigure}{0.08\linewidth}
	    \includegraphics[width=1\linewidth]{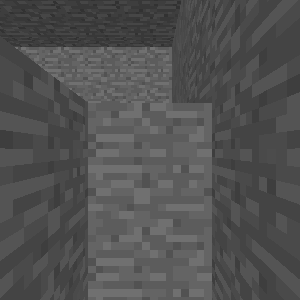} 
   		\includegraphics[width=1\linewidth]{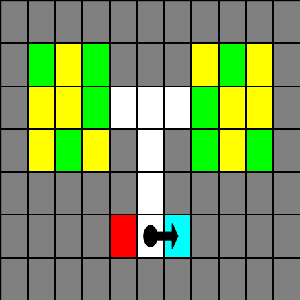} 
   		\caption*{t = 9}
	\end{subfigure}
	& 
    \begin{subfigure}{0.08\linewidth}
	    \includegraphics[width=1\linewidth]{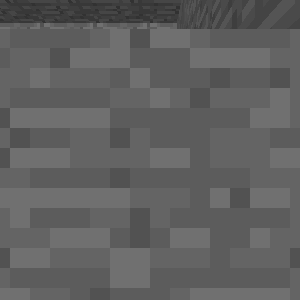} 
   		\includegraphics[width=1\linewidth]{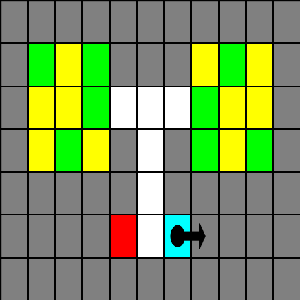} 
   		\caption*{t = 10}
	\end{subfigure}
	\end{tabular}
    \caption{FRMQN's play on a Pattern Matching task. The agent starts by looking at one room and then turns twice to look at the other room. Upon observing the two rooms, the agent uses backward actions repeatedly to move along the vertical corridor. Finally, once it is at the end of the corridor, it decides to turn and move forward to the blue block as the visual patterns of the rooms were identical. Note that the agent's performance is near optimal.} 
	\label{play-match-true}
\end{figure*}

\begin{figure*}
    \small
    \setlength{\tabcolsep}{1pt}
    \def\arraystretch{1}
    \begin{tabular}{llllllllllll}
    \begin{subfigure}{0.08\linewidth}
	    \includegraphics[width=1\linewidth]{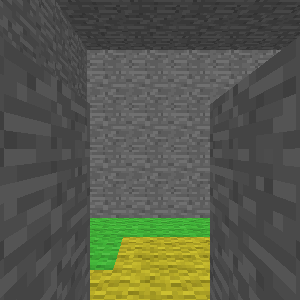} 
   		\includegraphics[width=1\linewidth]{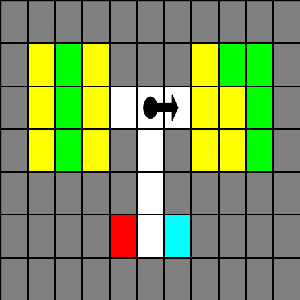} 
   		\caption*{t = 1}
	\end{subfigure} 
	& 
    \begin{subfigure}{0.08\linewidth}
	    \includegraphics[width=1\linewidth]{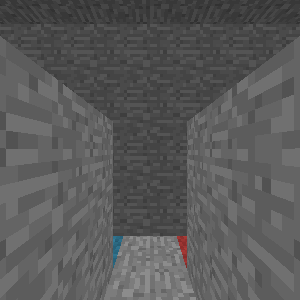} 
   		\includegraphics[width=1\linewidth]{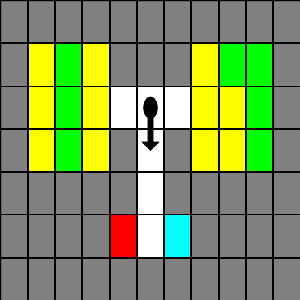}
   		\caption*{t = 2}
	\end{subfigure} 
	& 
    \begin{subfigure}{0.08\linewidth}
	    \includegraphics[width=1\linewidth]{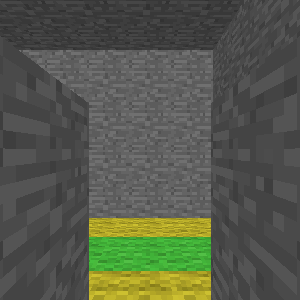} 
   		\includegraphics[width=1\linewidth]{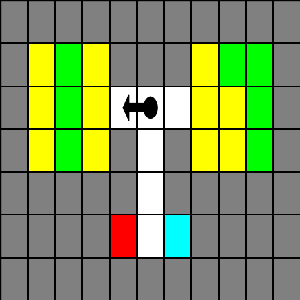} 
   		\caption*{t = 3}
	\end{subfigure} 
	& 
    \begin{subfigure}{0.08\linewidth}
	    \includegraphics[width=1\linewidth]{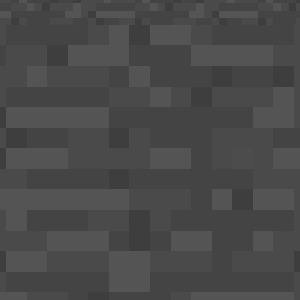} 
   		\includegraphics[width=1\linewidth]{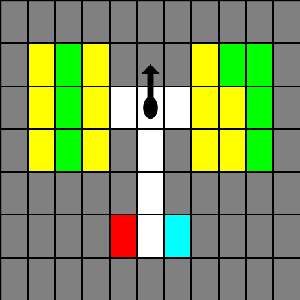} 
   		\caption*{t = 4}
	\end{subfigure}
	& 
    \begin{subfigure}{0.08\linewidth}
	    \includegraphics[width=1\linewidth]{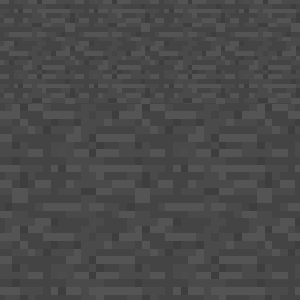} 
   		\includegraphics[width=1\linewidth]{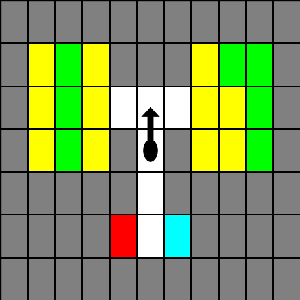} 
   		\caption*{t = 5}
	\end{subfigure}
	& 
    \begin{subfigure}{0.08\linewidth}
	    \includegraphics[width=1\linewidth]{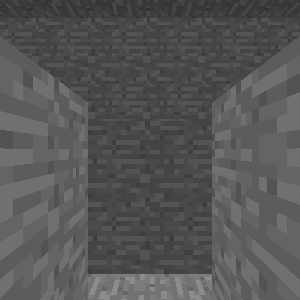} 
   		\includegraphics[width=1\linewidth]{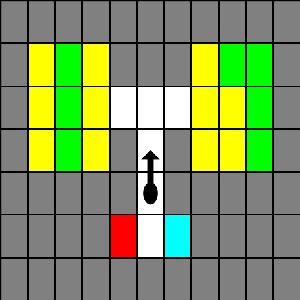} 
   		\caption*{t = 6}
	\end{subfigure} 
	& 
    \begin{subfigure}{0.08\linewidth}
	    \includegraphics[width=1\linewidth]{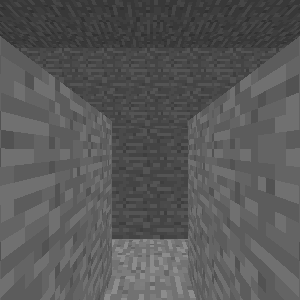} 
   		\includegraphics[width=1\linewidth]{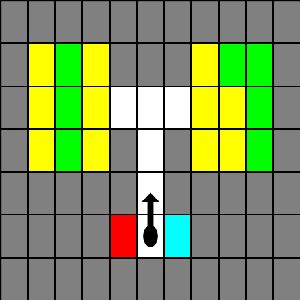}
   		\caption*{t = 7}
	\end{subfigure} 
	& 
    \begin{subfigure}{0.08\linewidth}
	    \includegraphics[width=1\linewidth]{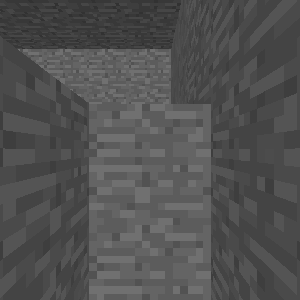} 
   		\includegraphics[width=1\linewidth]{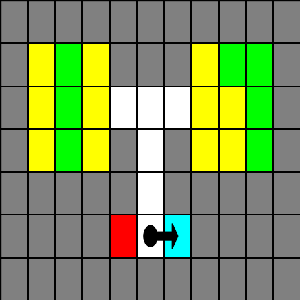} 
   		\caption*{t = 8}
	\end{subfigure} 
	& 
    \begin{subfigure}{0.08\linewidth}
	    \includegraphics[width=1\linewidth]{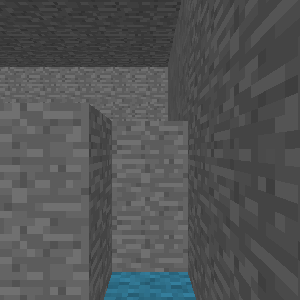} 
   		\includegraphics[width=1\linewidth]{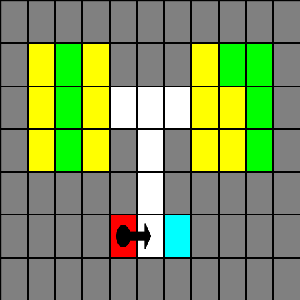} 
   		\caption*{t = 9}
	\end{subfigure}
	\end{tabular}
    \caption{FRMQN's play on a Pattern Matching task. The agent successfully goes to the red goal, given that the visual patterns of the two rooms were different.} 
	\label{play-match-false}
\end{figure*}
\begin{figure*}
    \small
    \setlength{\tabcolsep}{1pt}
    \def\arraystretch{1}
    \begin{tabular}{llllllllllll}
    \begin{subfigure}{0.08\linewidth}
	    \includegraphics[width=1\linewidth]{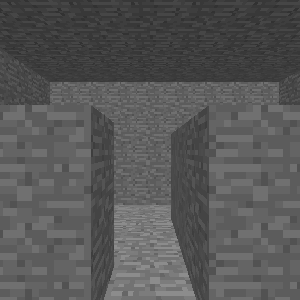} 
   		\includegraphics[width=1\linewidth]{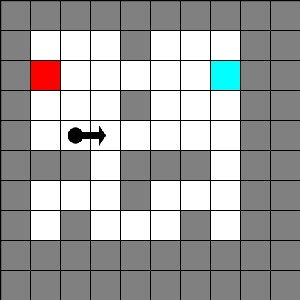} 
		\caption*{t = 1}
	\end{subfigure} 
	& 
    \begin{subfigure}{0.08\linewidth}
	    \includegraphics[width=1\linewidth]{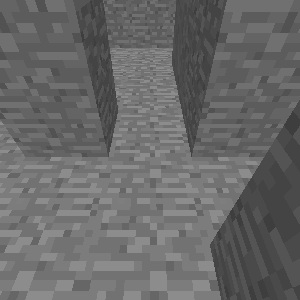} 
   		\includegraphics[width=1\linewidth]{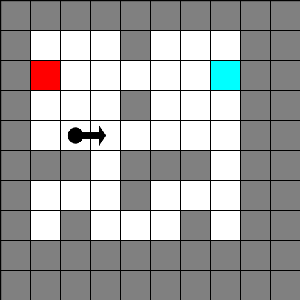} 
   		\caption*{t = 2}
	\end{subfigure} 
	& 
    \begin{subfigure}{0.08\linewidth}
	    \includegraphics[width=1\linewidth]{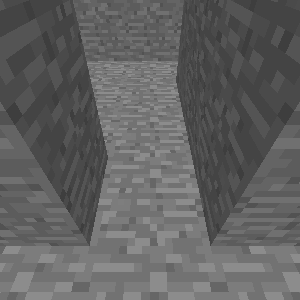} 
   		\includegraphics[width=1\linewidth]{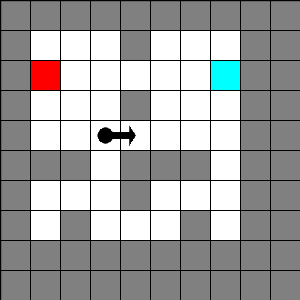} 
   		\caption*{t = 3}
	\end{subfigure} 
	& 
    \begin{subfigure}{0.08\linewidth}
	    \includegraphics[width=1\linewidth]{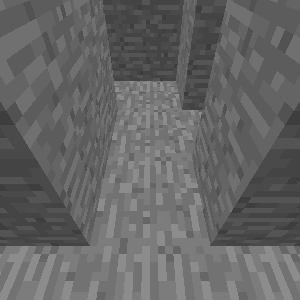} 
   		\includegraphics[width=1\linewidth]{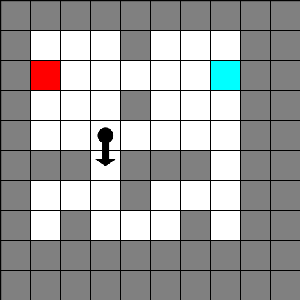} 
   		\caption*{t = 4}
	\end{subfigure}
	& 
    \begin{subfigure}{0.08\linewidth}
	    \includegraphics[width=1\linewidth]{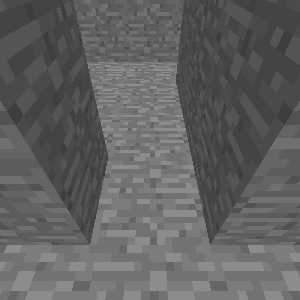} 
   		\includegraphics[width=1\linewidth]{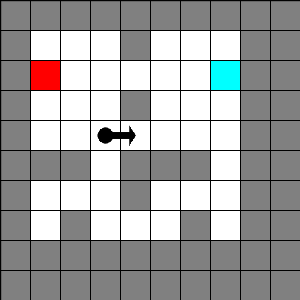} 
   		\caption*{t = 5}
	\end{subfigure}
	& 
    \begin{subfigure}{0.08\linewidth}
	    \includegraphics[width=1\linewidth]{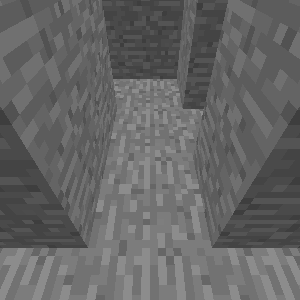} 
   		\includegraphics[width=1\linewidth]{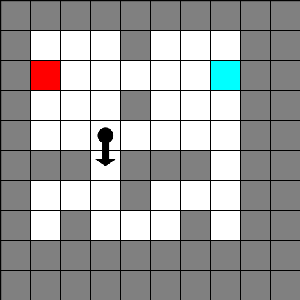} 
   		\caption*{t = 6}
	\end{subfigure} 
	& 
    \begin{subfigure}{0.08\linewidth}
	    \includegraphics[width=1\linewidth]{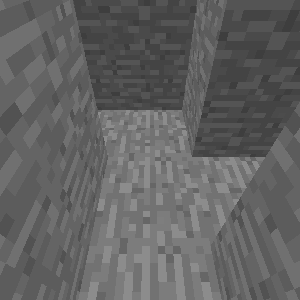} 
   		\includegraphics[width=1\linewidth]{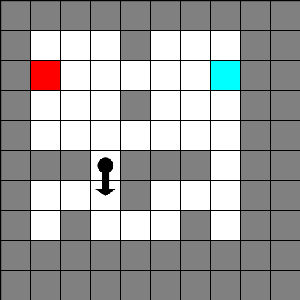} 
   		\caption*{t = 7}
	\end{subfigure} 
	& 
    \begin{subfigure}{0.08\linewidth}
	    \includegraphics[width=1\linewidth]{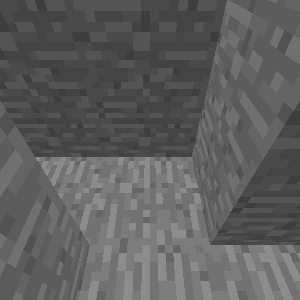} 
   		\includegraphics[width=1\linewidth]{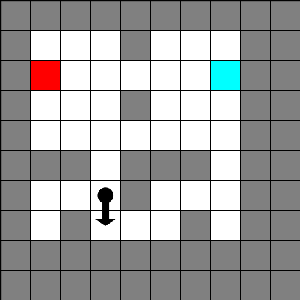} 
   		\caption*{t = 8}
	\end{subfigure} 
	& 
    \begin{subfigure}{0.08\linewidth}
	    \includegraphics[width=1\linewidth]{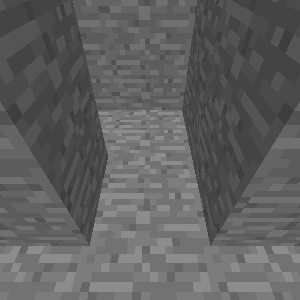} 
   		\includegraphics[width=1\linewidth]{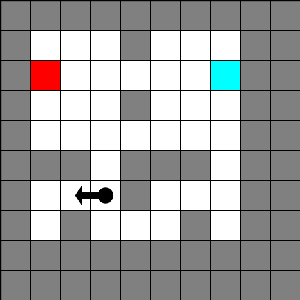} 
   		\caption*{t = 9}
	\end{subfigure}
	& 
    \begin{subfigure}{0.08\linewidth}
	    \includegraphics[width=1\linewidth]{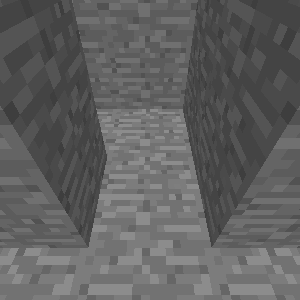} 
   		\includegraphics[width=1\linewidth]{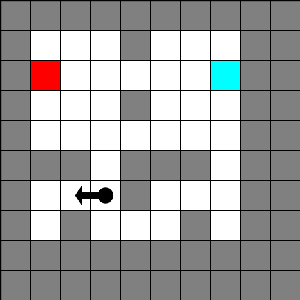} 
   		\caption*{t = 10}
	\end{subfigure}
	& 
    \begin{subfigure}{0.08\linewidth}
	    \includegraphics[width=1\linewidth]{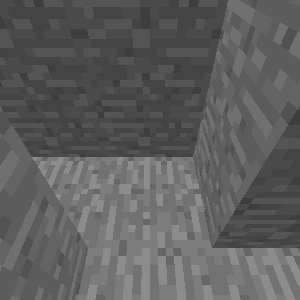} 
   		\includegraphics[width=1\linewidth]{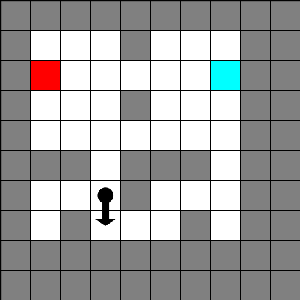} 
   		\caption*{t = 11}
	\end{subfigure} 
	&
    \begin{subfigure}{0.08\linewidth}
	    \includegraphics[width=1\linewidth]{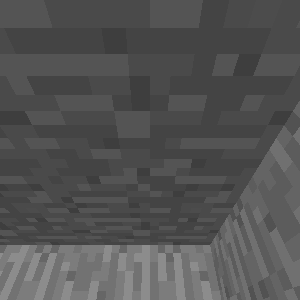} 
   		\includegraphics[width=1\linewidth]{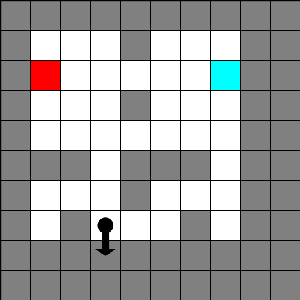} 
   		\caption*{t = 12}
	\end{subfigure} 
	\\
	\\
    \begin{subfigure}{0.08\linewidth}
	    \includegraphics[width=1\linewidth]{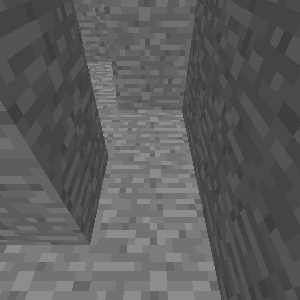} 
   		\includegraphics[width=1\linewidth]{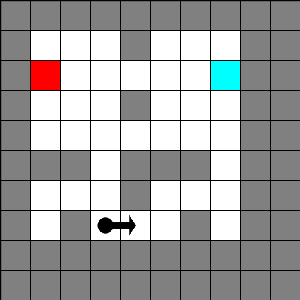} 
   		\caption*{t = 13}
	\end{subfigure} 
	& 
    \begin{subfigure}{0.08\linewidth}
	    \includegraphics[width=1\linewidth]{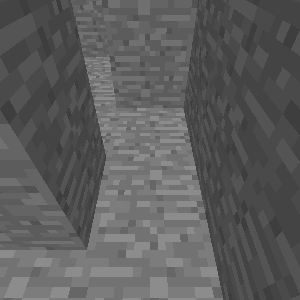} 
   		\includegraphics[width=1\linewidth]{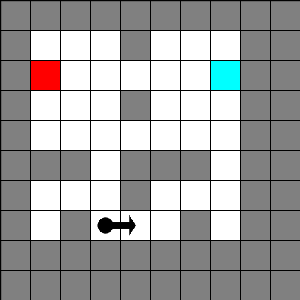} 
   		\caption*{t = 14}
	\end{subfigure} 
	& 
    \begin{subfigure}{0.08\linewidth}
	    \includegraphics[width=1\linewidth]{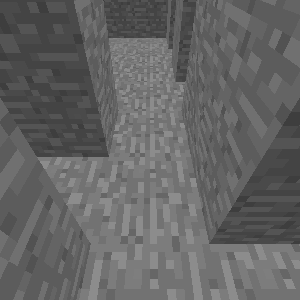} 
   		\includegraphics[width=1\linewidth]{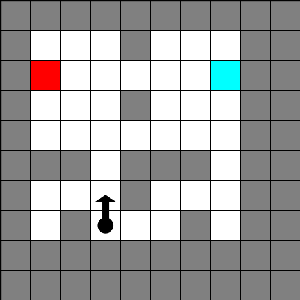} 
   		\caption*{t = 15}
	\end{subfigure} 
	& 
    \begin{subfigure}{0.08\linewidth}
	    \includegraphics[width=1\linewidth]{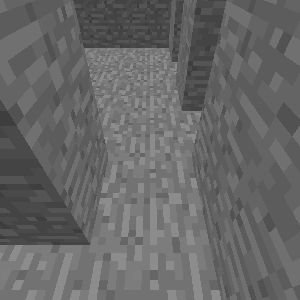} 
   		\includegraphics[width=1\linewidth]{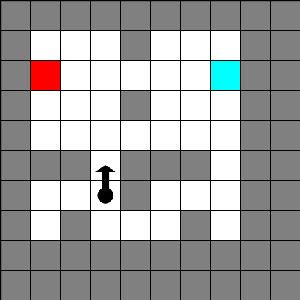} 
   		\caption*{t = 16}
	\end{subfigure}
	& 
    \begin{subfigure}{0.08\linewidth}
	    \includegraphics[width=1\linewidth]{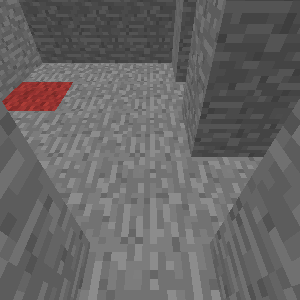} 
   		\includegraphics[width=1\linewidth]{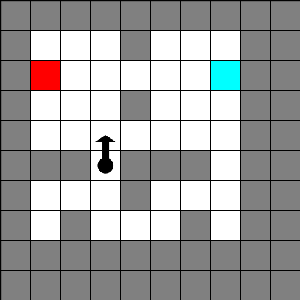}
   		\caption*{t = 17} 
	\end{subfigure}
	& 
    \begin{subfigure}{0.08\linewidth}
	    \includegraphics[width=1\linewidth]{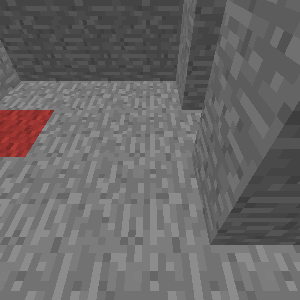} 
   		\includegraphics[width=1\linewidth]{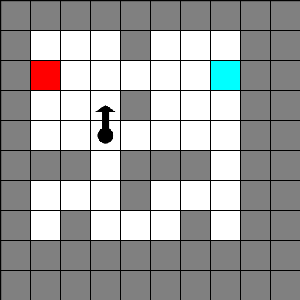} 
   		\caption*{t = 18}
	\end{subfigure} 
	& 
    \begin{subfigure}{0.08\linewidth}
	    \includegraphics[width=1\linewidth]{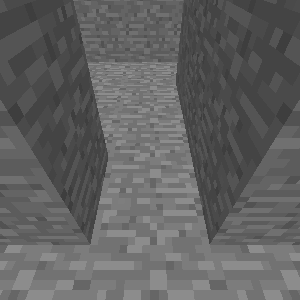} 
   		\includegraphics[width=1\linewidth]{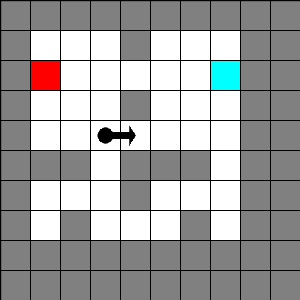} 
   		\caption*{t = 19}
	\end{subfigure} 
	& 
    \begin{subfigure}{0.08\linewidth}
	    \includegraphics[width=1\linewidth]{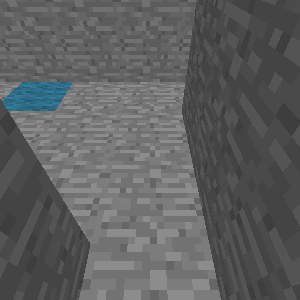} 
   		\includegraphics[width=1\linewidth]{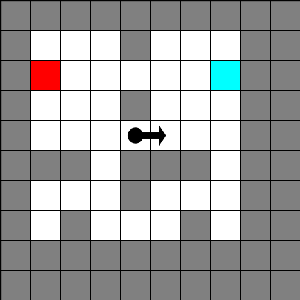} 
   		\caption*{t = 20}
	\end{subfigure} 
	& 
    \begin{subfigure}{0.08\linewidth}
	    \includegraphics[width=1\linewidth]{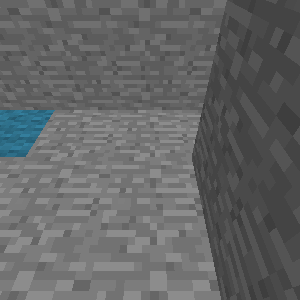} 
   		\includegraphics[width=1\linewidth]{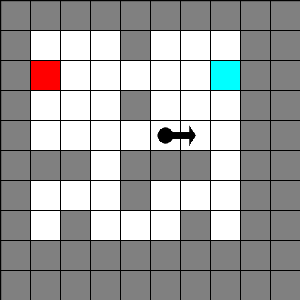} 
   		\caption*{t = 21}
	\end{subfigure}
	& 
    \begin{subfigure}{0.08\linewidth}
	    \includegraphics[width=1\linewidth]{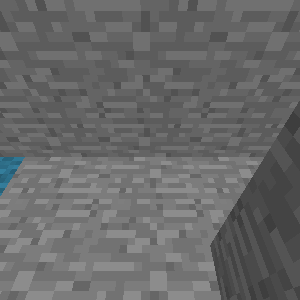} 
   		\includegraphics[width=1\linewidth]{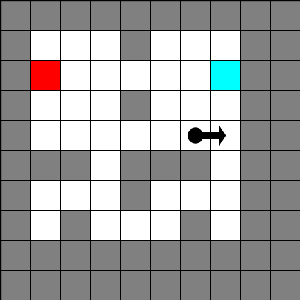} 
   		\caption*{t = 22}
	\end{subfigure}
	& 
    \begin{subfigure}{0.08\linewidth}
	    \includegraphics[width=1\linewidth]{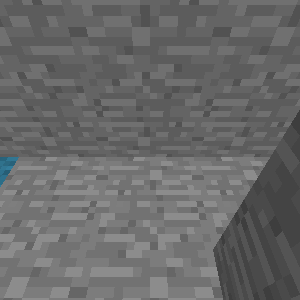} 
   		\includegraphics[width=1\linewidth]{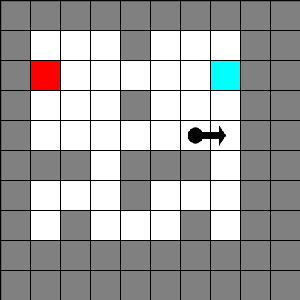} 
   		\caption*{t = 23}
	\end{subfigure} 
	&
    \begin{subfigure}{0.08\linewidth}
	    \includegraphics[width=1\linewidth]{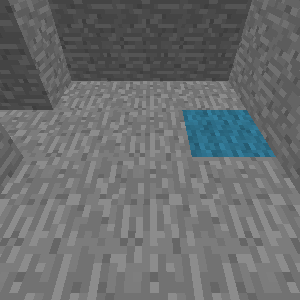} 
   		\includegraphics[width=1\linewidth]{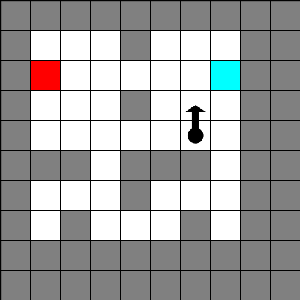} 
   		\caption*{t = 24}
	\end{subfigure} 
	\\
	\\
	\begin{subfigure}{0.08\linewidth}
	    \includegraphics[width=1\linewidth]{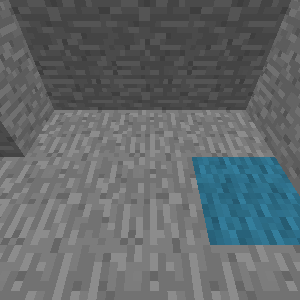} 
   		\includegraphics[width=1\linewidth]{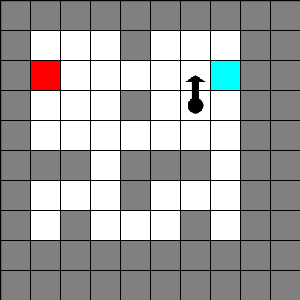} 
   		\caption*{t = 25}
	\end{subfigure} 
	& 
    \begin{subfigure}{0.08\linewidth}
	    \includegraphics[width=1\linewidth]{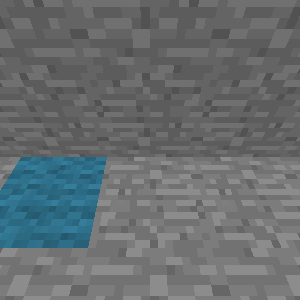} 
   		\includegraphics[width=1\linewidth]{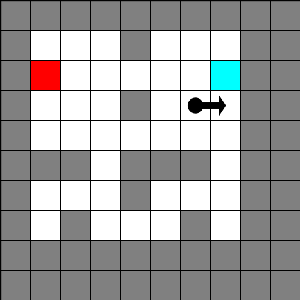} 
   		\caption*{t = 26}
	\end{subfigure} 
	& 
    \begin{subfigure}{0.08\linewidth}
	    \includegraphics[width=1\linewidth]{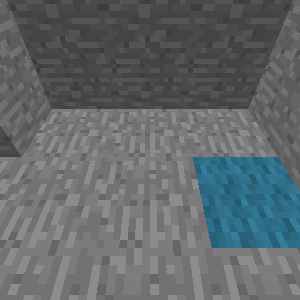} 
   		\includegraphics[width=1\linewidth]{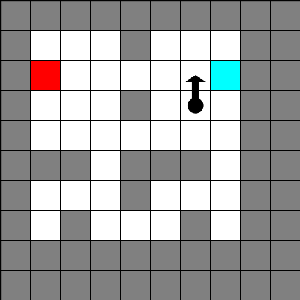} 
   		\caption*{t = 27}
	\end{subfigure} 
	& 
    \begin{subfigure}{0.08\linewidth}
	    \includegraphics[width=1\linewidth]{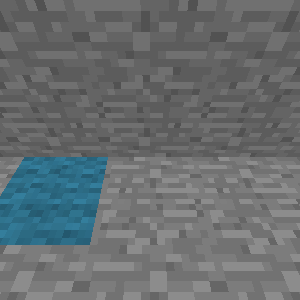} 
   		\includegraphics[width=1\linewidth]{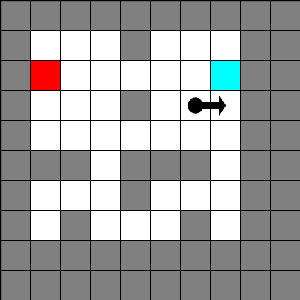} 
   		\caption*{t = 28}
	\end{subfigure}
	& 
    \begin{subfigure}{0.08\linewidth}
	    \includegraphics[width=1\linewidth]{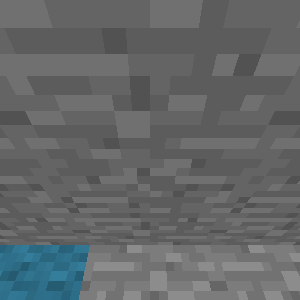} 
   		\includegraphics[width=1\linewidth]{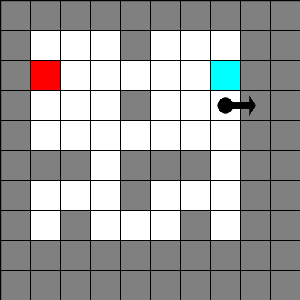} 
   		\caption*{t = 29}
	\end{subfigure}
	& 
    \begin{subfigure}{0.08\linewidth}
	    \includegraphics[width=1\linewidth]{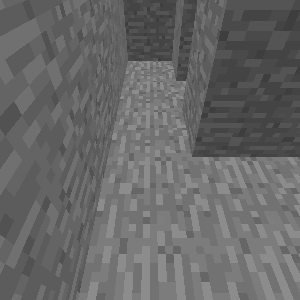} 
   		\includegraphics[width=1\linewidth]{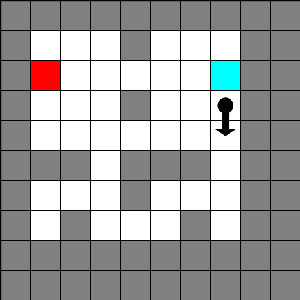} 
   		\caption*{t = 30}
	\end{subfigure} 
	& 
    \begin{subfigure}{0.08\linewidth}
	    \includegraphics[width=1\linewidth]{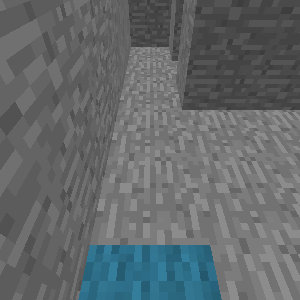} 
   		\includegraphics[width=1\linewidth]{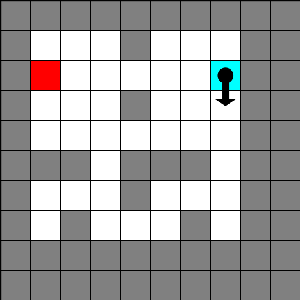} 
   		\caption*{t = 31}
	\end{subfigure} 
	\end{tabular}
    \caption{FRMQN's play on an unseen random maze with Single Goal task. As the case with most of the tasks, the agent starts by looking down quickly to see the important stimuli (e.g., goal blocks and indicators) more clearly. The agent then looks around it's vicinity and explores corridors. As soon as the agent sees the blue block, it goes to the block and successfully complete the task.} 
	\label{play-single}
\end{figure*}

\begin{figure*}
\small
\setlength{\tabcolsep}{1pt}
\def\arraystretch{1}
\begin{tabular}{llllllllllll}
\begin{subfigure}{0.08\linewidth}
\includegraphics[width=1\linewidth]{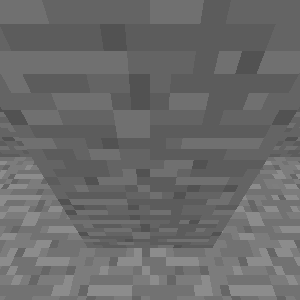}
\includegraphics[width=1\linewidth]{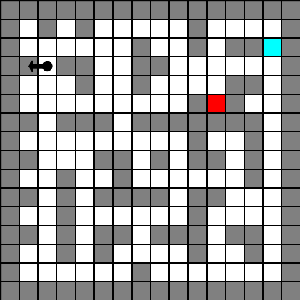}
\caption*{t = 1}
\end{subfigure}
&
\begin{subfigure}{0.08\linewidth}
\includegraphics[width=1\linewidth]{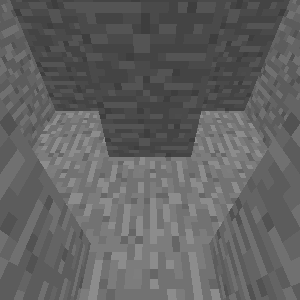}
\includegraphics[width=1\linewidth]{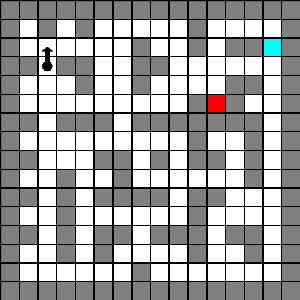}
\caption*{t = 2}
\end{subfigure}
&
\begin{subfigure}{0.08\linewidth}
\includegraphics[width=1\linewidth]{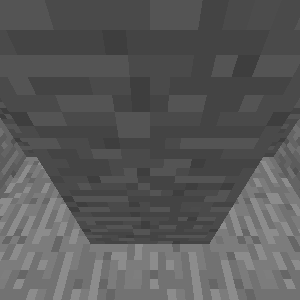}
\includegraphics[width=1\linewidth]{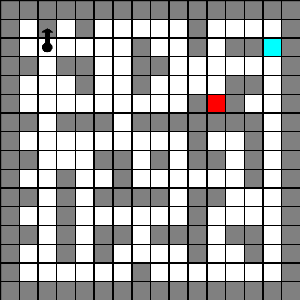}
\caption*{t = 3}
\end{subfigure}
&
\begin{subfigure}{0.08\linewidth}
\includegraphics[width=1\linewidth]{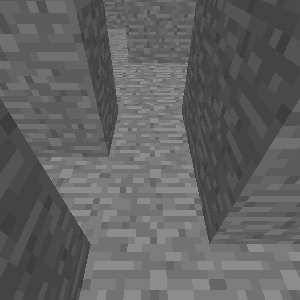}
\includegraphics[width=1\linewidth]{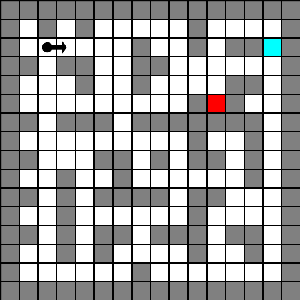}
\caption*{t = 4}
\end{subfigure}
&
\begin{subfigure}{0.08\linewidth}
\includegraphics[width=1\linewidth]{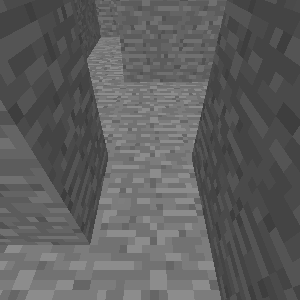}
\includegraphics[width=1\linewidth]{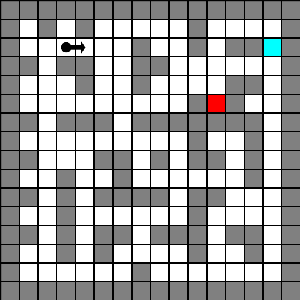}
\caption*{t = 5}
\end{subfigure}
&
\begin{subfigure}{0.08\linewidth}
\includegraphics[width=1\linewidth]{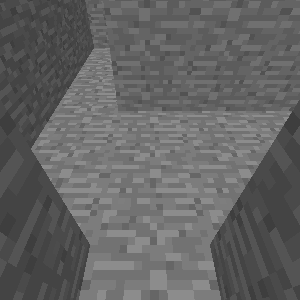}
\includegraphics[width=1\linewidth]{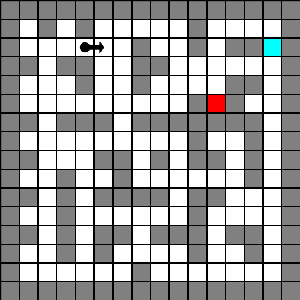}
\caption*{t = 6}
\end{subfigure}
&
\begin{subfigure}{0.08\linewidth}
\includegraphics[width=1\linewidth]{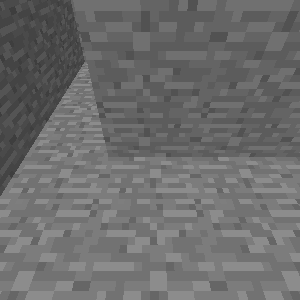}
\includegraphics[width=1\linewidth]{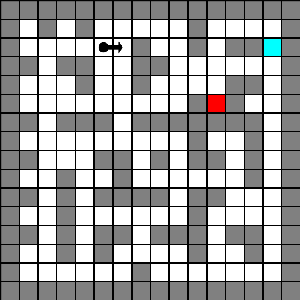}
\caption*{t = 7}
\end{subfigure}
&
\begin{subfigure}{0.08\linewidth}
\includegraphics[width=1\linewidth]{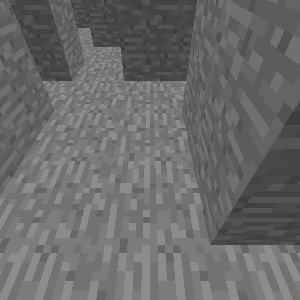}
\includegraphics[width=1\linewidth]{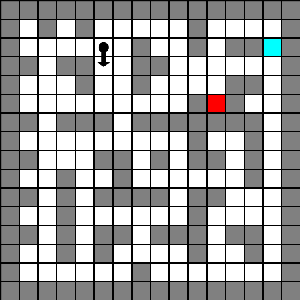}
\caption*{t = 8}
\end{subfigure}
&
\begin{subfigure}{0.08\linewidth}
\includegraphics[width=1\linewidth]{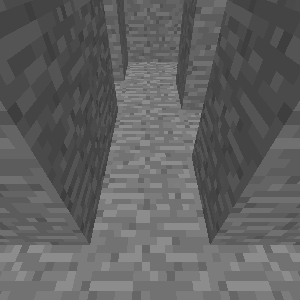}
\includegraphics[width=1\linewidth]{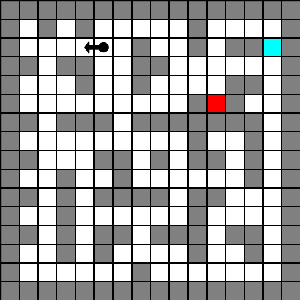}
\caption*{t = 9}
\end{subfigure}
&
\begin{subfigure}{0.08\linewidth}
\includegraphics[width=1\linewidth]{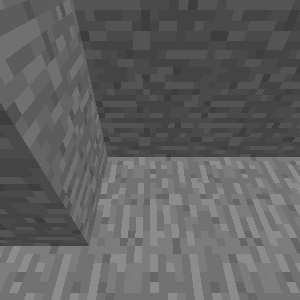}
\includegraphics[width=1\linewidth]{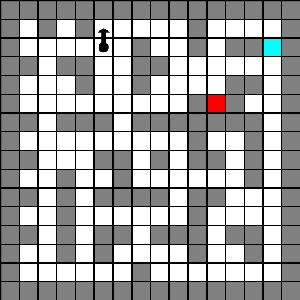}
\caption*{t = 10}
\end{subfigure}
&
\begin{subfigure}{0.08\linewidth}
\includegraphics[width=1\linewidth]{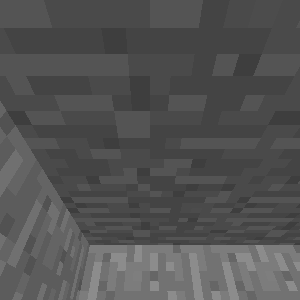}
\includegraphics[width=1\linewidth]{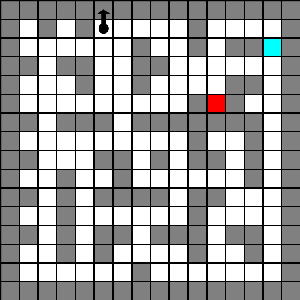}
\caption*{t = 11}
\end{subfigure}
&
\begin{subfigure}{0.08\linewidth}
\includegraphics[width=1\linewidth]{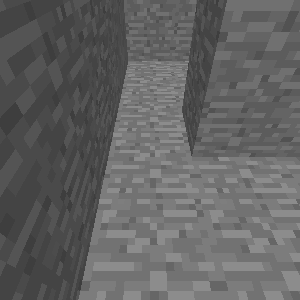}
\includegraphics[width=1\linewidth]{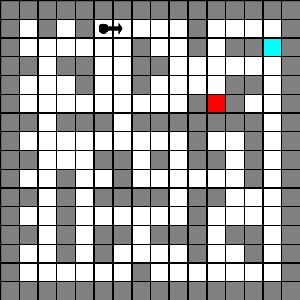}
\caption*{t = 12}
\end{subfigure}
\\
\\
\begin{subfigure}{0.08\linewidth}
\includegraphics[width=1\linewidth]{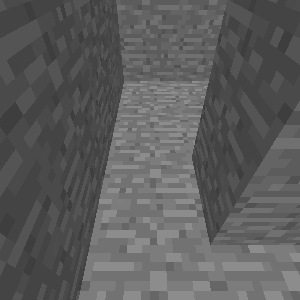}
\includegraphics[width=1\linewidth]{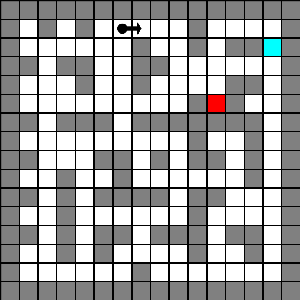}
\caption*{t = 13}
\end{subfigure}
&
\begin{subfigure}{0.08\linewidth}
\includegraphics[width=1\linewidth]{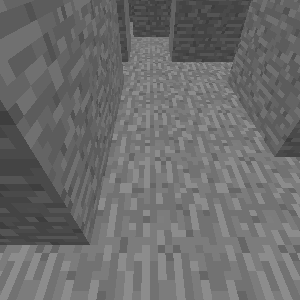}
\includegraphics[width=1\linewidth]{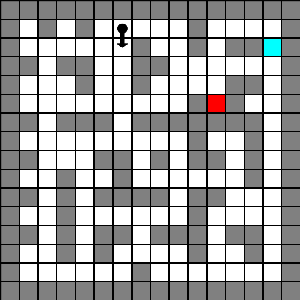}
\caption*{t = 14}
\end{subfigure}
&
\begin{subfigure}{0.08\linewidth}
\includegraphics[width=1\linewidth]{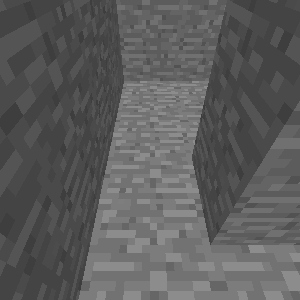}
\includegraphics[width=1\linewidth]{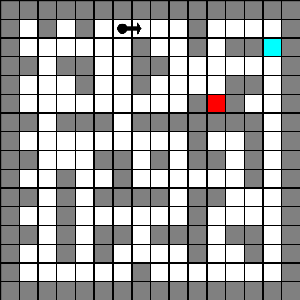}
\caption*{t = 15}
\end{subfigure}
&
\begin{subfigure}{0.08\linewidth}
\includegraphics[width=1\linewidth]{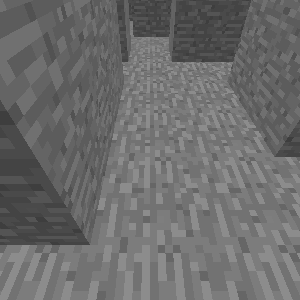}
\includegraphics[width=1\linewidth]{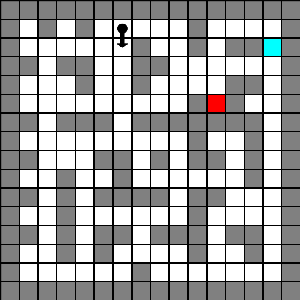}
\caption*{t = 16}
\end{subfigure}
&
\begin{subfigure}{0.08\linewidth}
\includegraphics[width=1\linewidth]{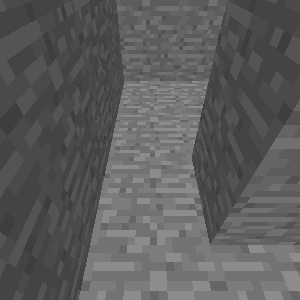}
\includegraphics[width=1\linewidth]{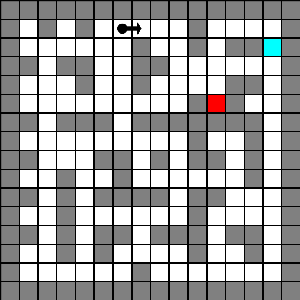}
\caption*{t = 17}
\end{subfigure}
&
\begin{subfigure}{0.08\linewidth}
\includegraphics[width=1\linewidth]{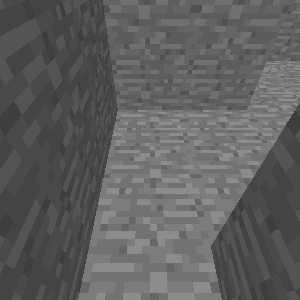}
\includegraphics[width=1\linewidth]{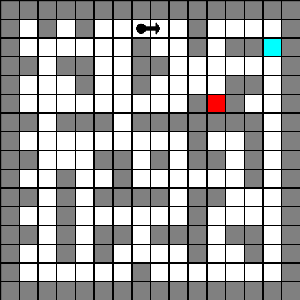}
\caption*{t = 18}
\end{subfigure}
&
\begin{subfigure}{0.08\linewidth}
\includegraphics[width=1\linewidth]{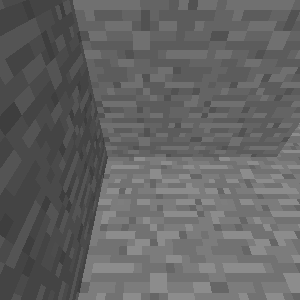}
\includegraphics[width=1\linewidth]{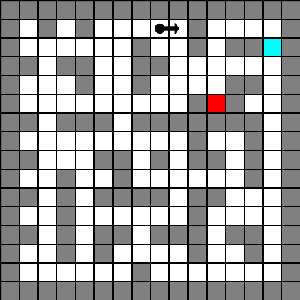}
\caption*{t = 19}
\end{subfigure}
&
\begin{subfigure}{0.08\linewidth}
\includegraphics[width=1\linewidth]{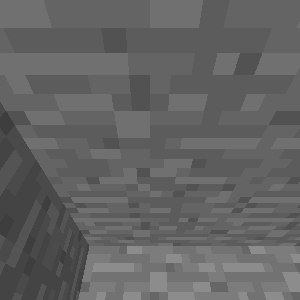}
\includegraphics[width=1\linewidth]{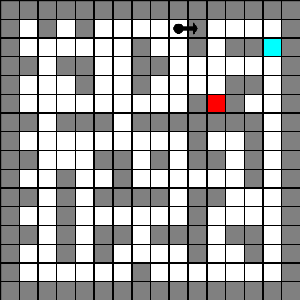}
\caption*{t = 20}
\end{subfigure}
&
\begin{subfigure}{0.08\linewidth}
\includegraphics[width=1\linewidth]{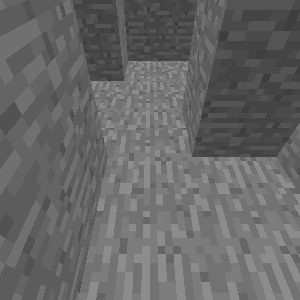}
\includegraphics[width=1\linewidth]{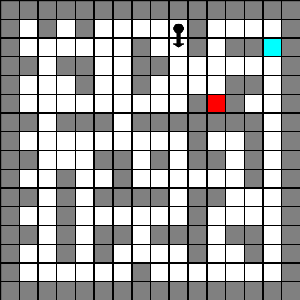}
\caption*{t = 21}
\end{subfigure}
&
\begin{subfigure}{0.08\linewidth}
\includegraphics[width=1\linewidth]{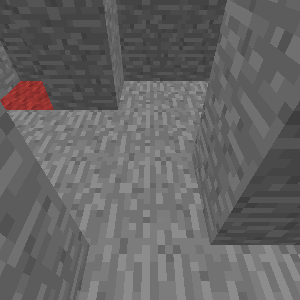}
\includegraphics[width=1\linewidth]{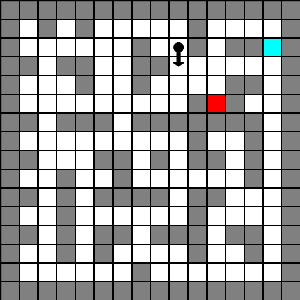}
\caption*{t = 22}
\end{subfigure}
&
\begin{subfigure}{0.08\linewidth}
\includegraphics[width=1\linewidth]{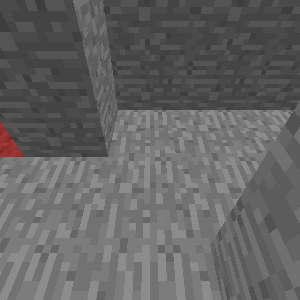}
\includegraphics[width=1\linewidth]{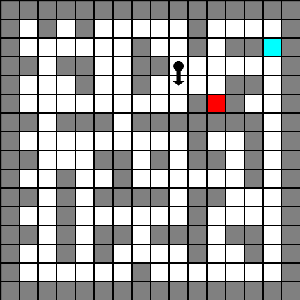}
\caption*{t = 23}
\end{subfigure}
&
\begin{subfigure}{0.08\linewidth}
\includegraphics[width=1\linewidth]{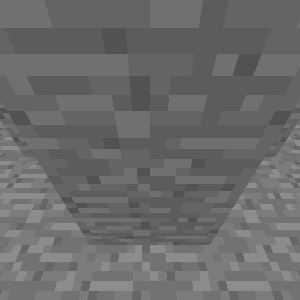}
\includegraphics[width=1\linewidth]{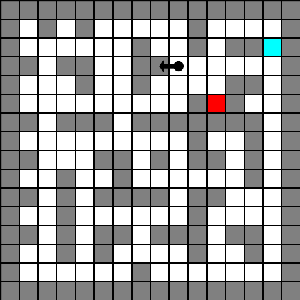}
\caption*{t = 24}
\end{subfigure}
\\
\\
\begin{subfigure}{0.08\linewidth}
\includegraphics[width=1\linewidth]{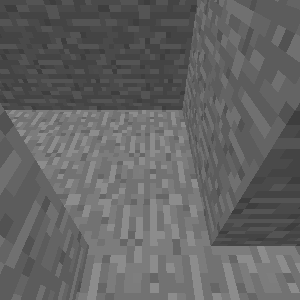}
\includegraphics[width=1\linewidth]{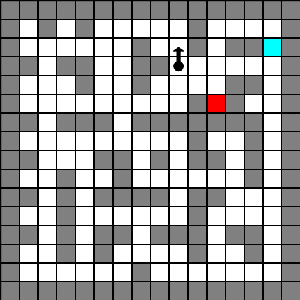}
\caption*{t = 25}
\end{subfigure}
&
\begin{subfigure}{0.08\linewidth}
\includegraphics[width=1\linewidth]{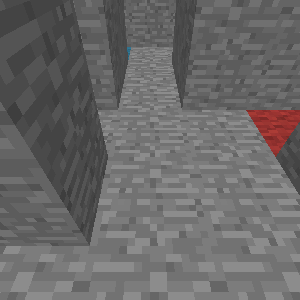}
\includegraphics[width=1\linewidth]{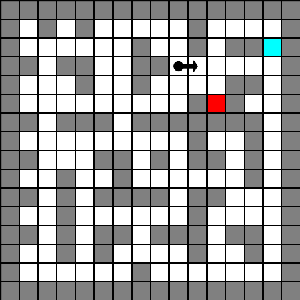}
\caption*{t = 26}
\end{subfigure}
&
\begin{subfigure}{0.08\linewidth}
\includegraphics[width=1\linewidth]{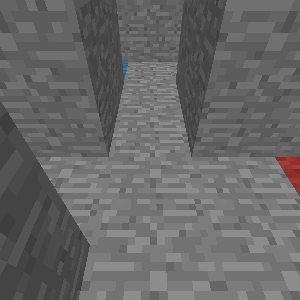}
\includegraphics[width=1\linewidth]{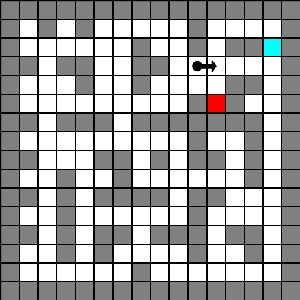}
\caption*{t = 27}
\end{subfigure}
&
\begin{subfigure}{0.08\linewidth}
\includegraphics[width=1\linewidth]{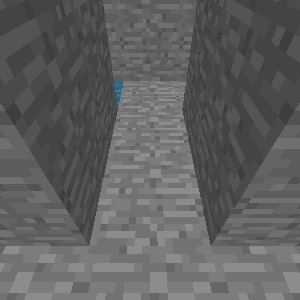}
\includegraphics[width=1\linewidth]{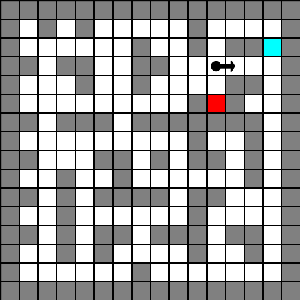}
\caption*{t = 28}
\end{subfigure}
&
\begin{subfigure}{0.08\linewidth}
\includegraphics[width=1\linewidth]{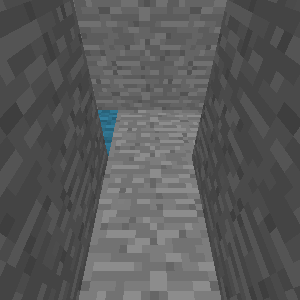}
\includegraphics[width=1\linewidth]{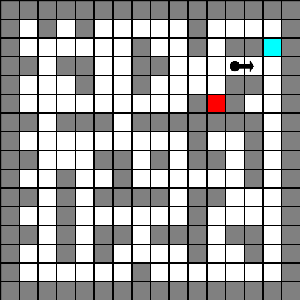}
\caption*{t = 29}
\end{subfigure}
&
\begin{subfigure}{0.08\linewidth}
\includegraphics[width=1\linewidth]{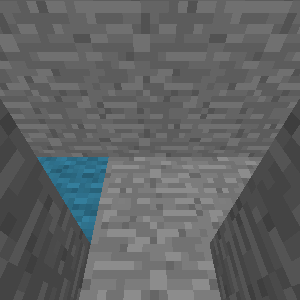}
\includegraphics[width=1\linewidth]{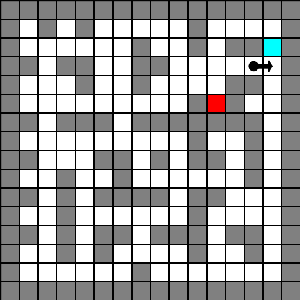}
\caption*{t = 30}
\end{subfigure}
&
\begin{subfigure}{0.08\linewidth}
\includegraphics[width=1\linewidth]{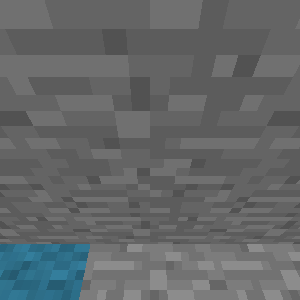}
\includegraphics[width=1\linewidth]{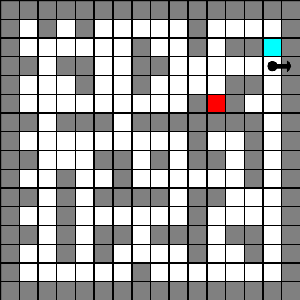}
\caption*{t = 31}
\end{subfigure}
&
\begin{subfigure}{0.08\linewidth}
\includegraphics[width=1\linewidth]{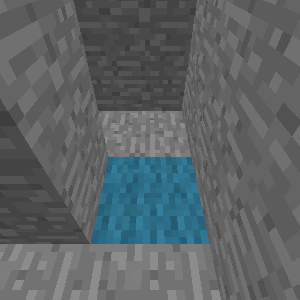}
\includegraphics[width=1\linewidth]{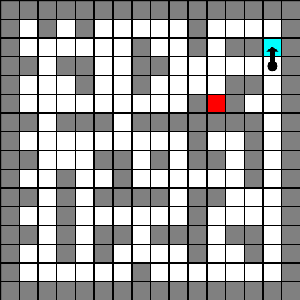}
\caption*{t = 32}
\end{subfigure}
&
\begin{subfigure}{0.08\linewidth}
\includegraphics[width=1\linewidth]{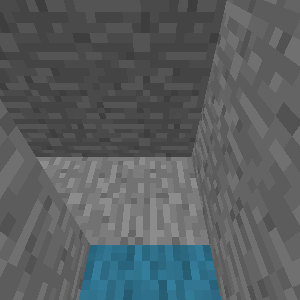}
\includegraphics[width=1\linewidth]{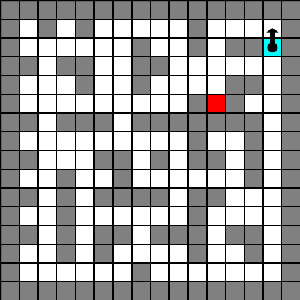}
\caption*{t = 33}
\end{subfigure}
\end{tabular}
\caption{FRMQN's play in an unseen and larger random maze with Single Goal task. The agent explores the map from the top-left side to top-right side, and successfully finds and visits the blue block. }
\label{single_l_success}
\end{figure*}

\begin{figure*}
\small
\setlength{\tabcolsep}{1pt}
\def\arraystretch{1}
\begin{tabular}{llllllllllll}
\begin{subfigure}{0.08\linewidth}
\includegraphics[width=1\linewidth]{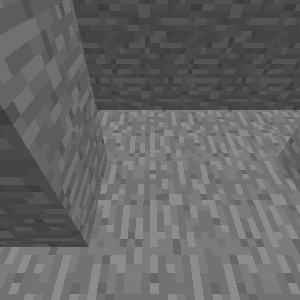}
\includegraphics[width=1\linewidth]{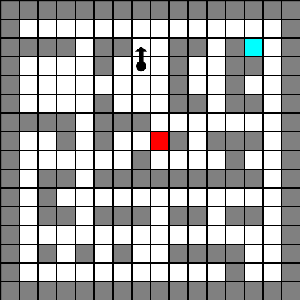}
\caption*{t = 1}
\end{subfigure}
&
\begin{subfigure}{0.08\linewidth}
\includegraphics[width=1\linewidth]{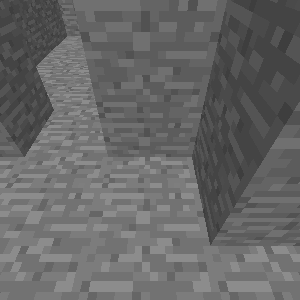}
\includegraphics[width=1\linewidth]{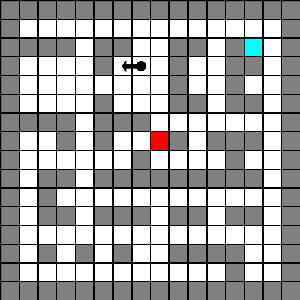}
\caption*{t = 2}
\end{subfigure}
&
\begin{subfigure}{0.08\linewidth}
\includegraphics[width=1\linewidth]{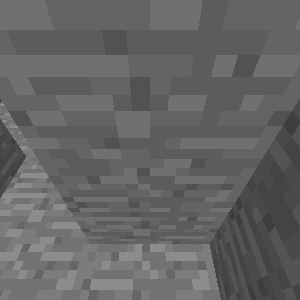}
\includegraphics[width=1\linewidth]{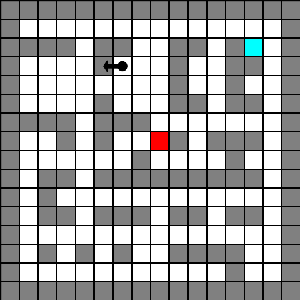}
\caption*{t = 3}
\end{subfigure}
&
\begin{subfigure}{0.08\linewidth}
\includegraphics[width=1\linewidth]{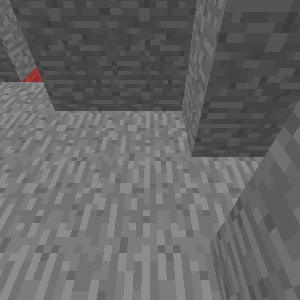}
\includegraphics[width=1\linewidth]{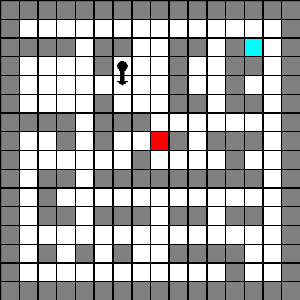}
\caption*{t = 4}
\end{subfigure}
&
\begin{subfigure}{0.08\linewidth}
\includegraphics[width=1\linewidth]{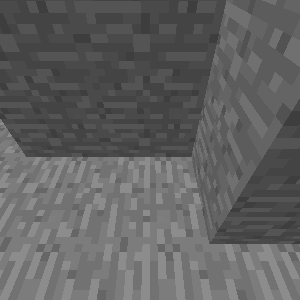}
\includegraphics[width=1\linewidth]{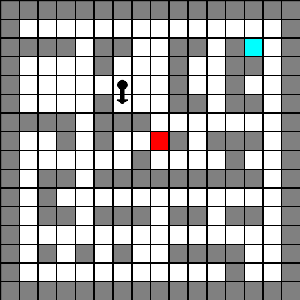}
\caption*{t = 5}
\end{subfigure}
&
\begin{subfigure}{0.08\linewidth}
\includegraphics[width=1\linewidth]{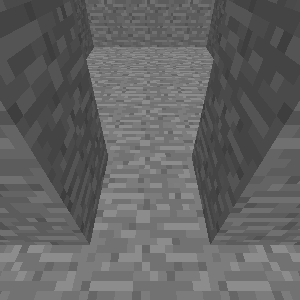}
\includegraphics[width=1\linewidth]{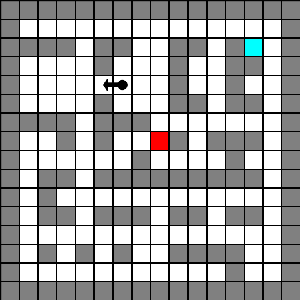}
\caption*{t = 6}
\end{subfigure}
&
\begin{subfigure}{0.08\linewidth}
\includegraphics[width=1\linewidth]{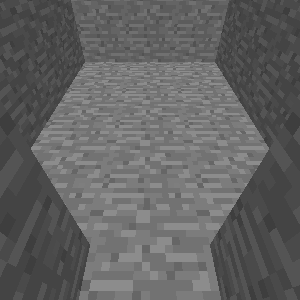}
\includegraphics[width=1\linewidth]{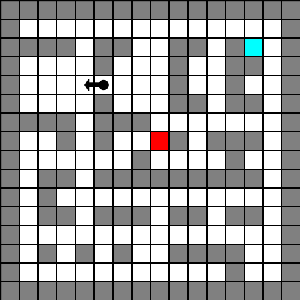}
\caption*{t = 7}
\end{subfigure}
&
\begin{subfigure}{0.08\linewidth}
\includegraphics[width=1\linewidth]{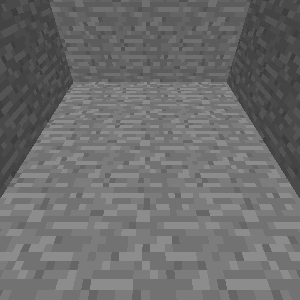}
\includegraphics[width=1\linewidth]{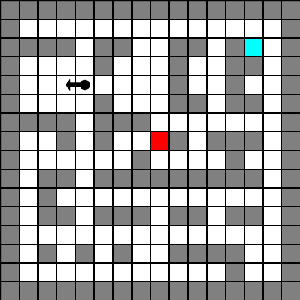}
\caption*{t = 8}
\end{subfigure}
&
\begin{subfigure}{0.08\linewidth}
\includegraphics[width=1\linewidth]{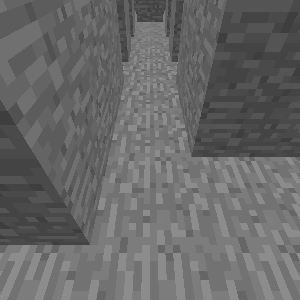}
\includegraphics[width=1\linewidth]{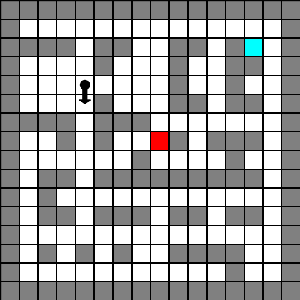}
\caption*{t = 9}
\end{subfigure}
&
\begin{subfigure}{0.08\linewidth}
\includegraphics[width=1\linewidth]{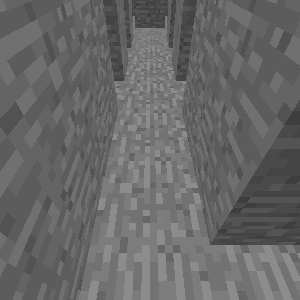}
\includegraphics[width=1\linewidth]{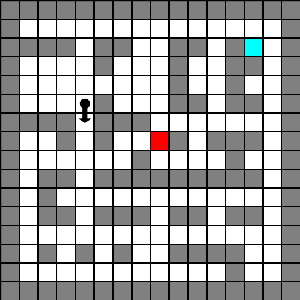}
\caption*{t = 10}
\end{subfigure}
&
\begin{subfigure}{0.08\linewidth}
\includegraphics[width=1\linewidth]{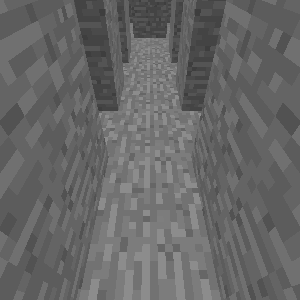}
\includegraphics[width=1\linewidth]{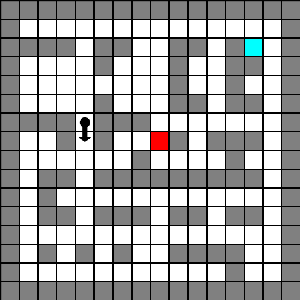}
\caption*{t = 11}
\end{subfigure}
&
\begin{subfigure}{0.08\linewidth}
\includegraphics[width=1\linewidth]{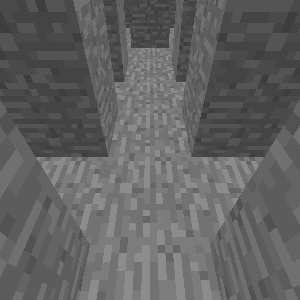}
\includegraphics[width=1\linewidth]{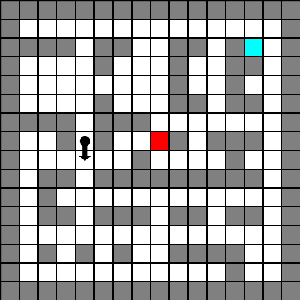}
\caption*{t = 12}
\end{subfigure}
\\
\\
\begin{subfigure}{0.08\linewidth}
\includegraphics[width=1\linewidth]{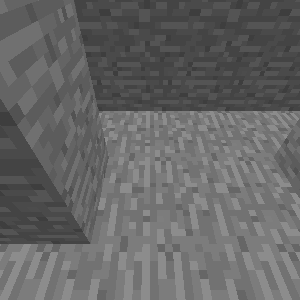}
\includegraphics[width=1\linewidth]{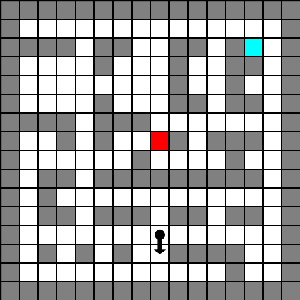}
\caption*{t = 25}
\end{subfigure}
&
\begin{subfigure}{0.08\linewidth}
\includegraphics[width=1\linewidth]{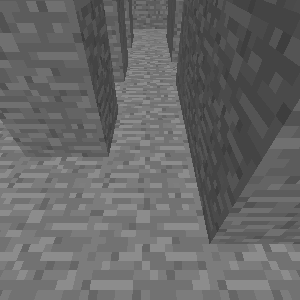}
\includegraphics[width=1\linewidth]{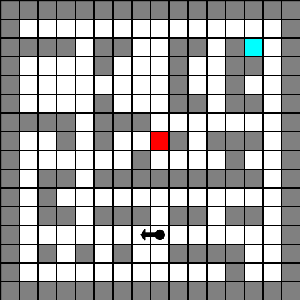}
\caption*{t = 26}
\end{subfigure}
&
\begin{subfigure}{0.08\linewidth}
\includegraphics[width=1\linewidth]{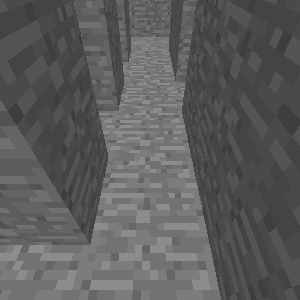}
\includegraphics[width=1\linewidth]{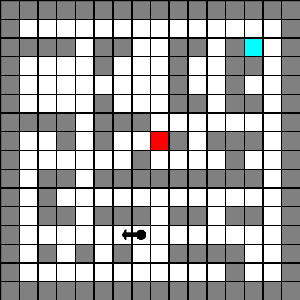}
\caption*{t = 27}
\end{subfigure}
&
\begin{subfigure}{0.08\linewidth}
\includegraphics[width=1\linewidth]{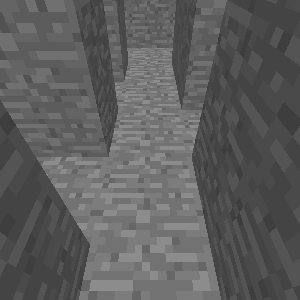}
\includegraphics[width=1\linewidth]{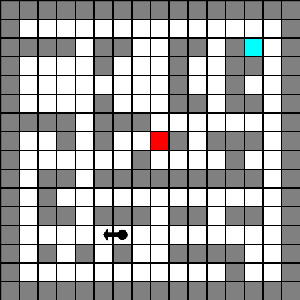}
\caption*{t = 28}
\end{subfigure}
&
\begin{subfigure}{0.08\linewidth}
\includegraphics[width=1\linewidth]{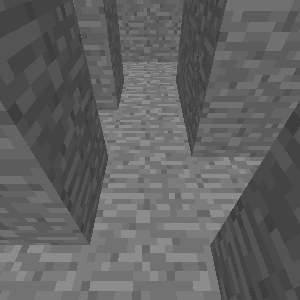}
\includegraphics[width=1\linewidth]{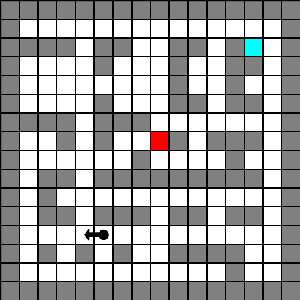}
\caption*{t = 29}
\end{subfigure}
&
\begin{subfigure}{0.08\linewidth}
\includegraphics[width=1\linewidth]{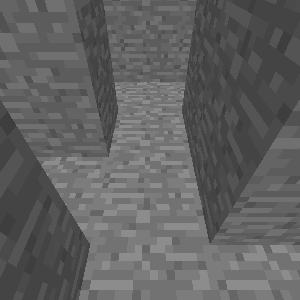}
\includegraphics[width=1\linewidth]{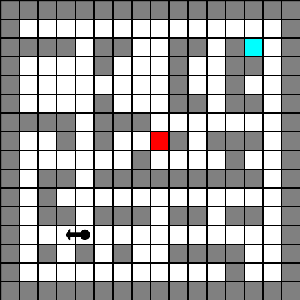}
\caption*{t = 30}
\end{subfigure}
&
\begin{subfigure}{0.08\linewidth}
\includegraphics[width=1\linewidth]{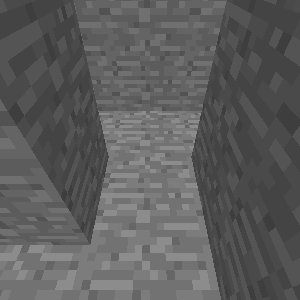}
\includegraphics[width=1\linewidth]{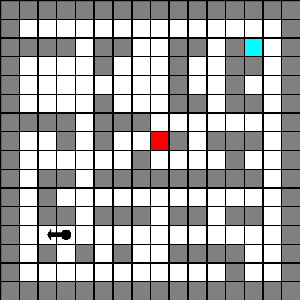}
\caption*{t = 31}
\end{subfigure}
&
\begin{subfigure}{0.08\linewidth}
\includegraphics[width=1\linewidth]{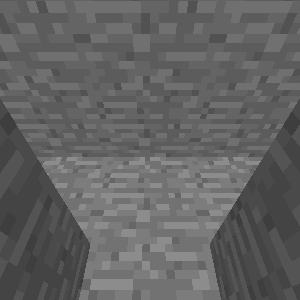}
\includegraphics[width=1\linewidth]{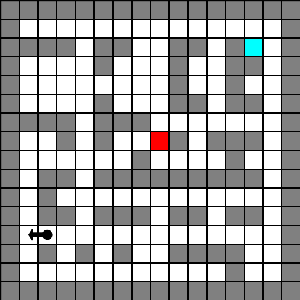}
\caption*{t = 32}
\end{subfigure}
&
\begin{subfigure}{0.08\linewidth}
\includegraphics[width=1\linewidth]{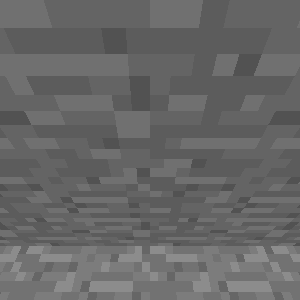}
\includegraphics[width=1\linewidth]{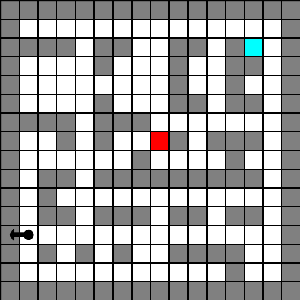}
\caption*{t = 33}
\end{subfigure}
&
\begin{subfigure}{0.08\linewidth}
\includegraphics[width=1\linewidth]{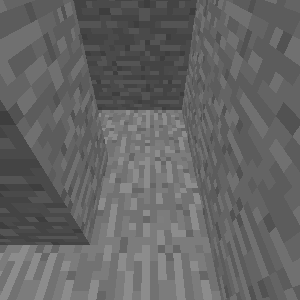}
\includegraphics[width=1\linewidth]{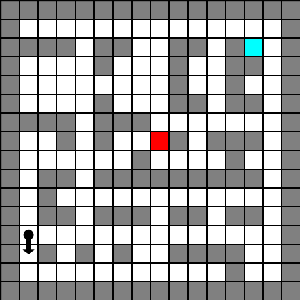}
\caption*{t = 34}
\end{subfigure}
&
\begin{subfigure}{0.08\linewidth}
\includegraphics[width=1\linewidth]{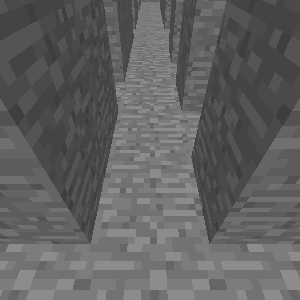}
\includegraphics[width=1\linewidth]{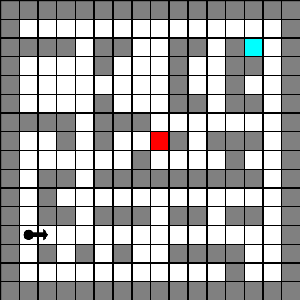}
\caption*{t = 35}
\end{subfigure}
&
\begin{subfigure}{0.08\linewidth}
\includegraphics[width=1\linewidth]{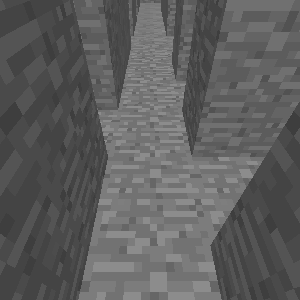}
\includegraphics[width=1\linewidth]{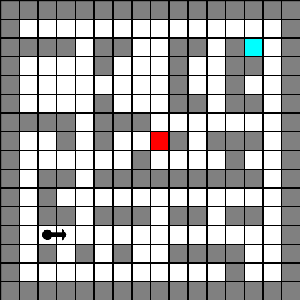}
\caption*{t = 36}
\end{subfigure}
\\
\\
\begin{subfigure}{0.08\linewidth}
\includegraphics[width=1\linewidth]{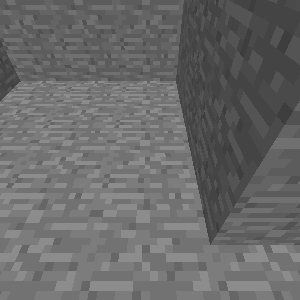}
\includegraphics[width=1\linewidth]{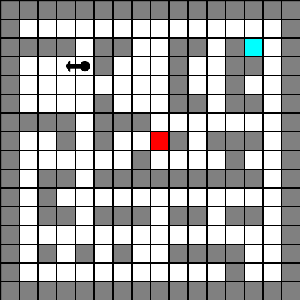}
\caption*{t = 49}
\end{subfigure}
&
\begin{subfigure}{0.08\linewidth}
\includegraphics[width=1\linewidth]{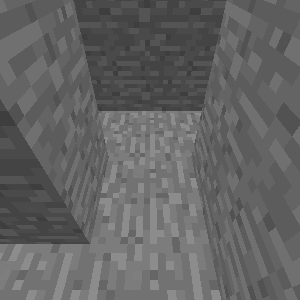}
\includegraphics[width=1\linewidth]{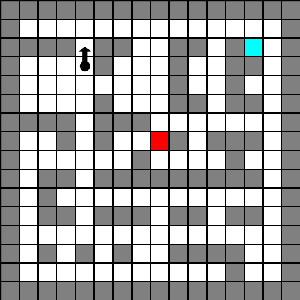}
\caption*{t = 50}
\end{subfigure}
&
\begin{subfigure}{0.08\linewidth}
\includegraphics[width=1\linewidth]{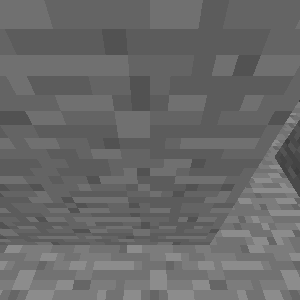}
\includegraphics[width=1\linewidth]{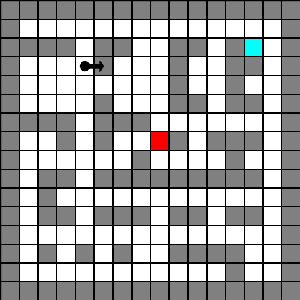}
\caption*{t = 51}
\end{subfigure}
&
\begin{subfigure}{0.08\linewidth}
\includegraphics[width=1\linewidth]{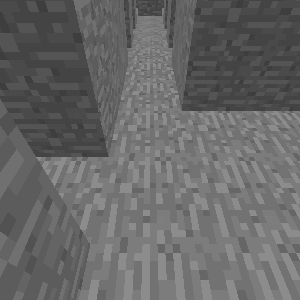}
\includegraphics[width=1\linewidth]{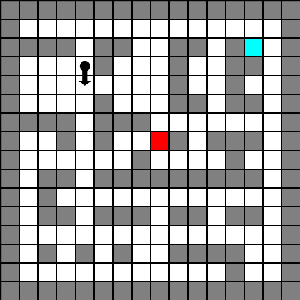}
\caption*{t = 52}
\end{subfigure}
&
\begin{subfigure}{0.08\linewidth}
\includegraphics[width=1\linewidth]{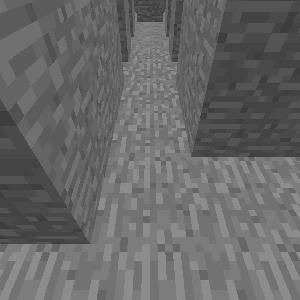}
\includegraphics[width=1\linewidth]{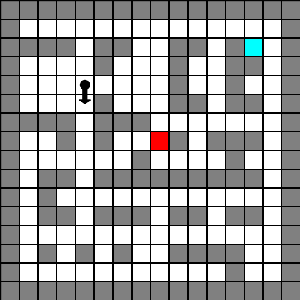}
\caption*{t = 53}
\end{subfigure}
&
\begin{subfigure}{0.08\linewidth}
\includegraphics[width=1\linewidth]{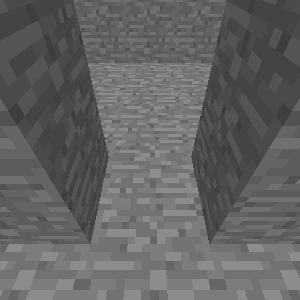}
\includegraphics[width=1\linewidth]{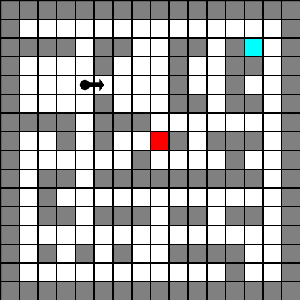}
\caption*{t = 54}
\end{subfigure}
&
\begin{subfigure}{0.08\linewidth}
\includegraphics[width=1\linewidth]{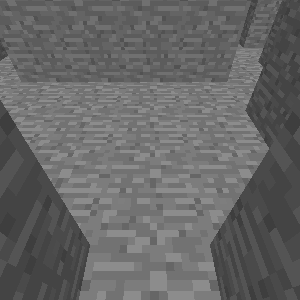}
\includegraphics[width=1\linewidth]{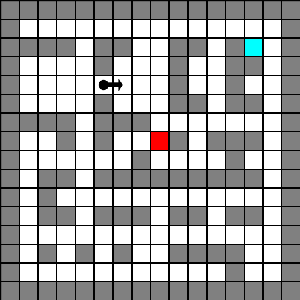}
\caption*{t = 55}
\end{subfigure}
&
\begin{subfigure}{0.08\linewidth}
\includegraphics[width=1\linewidth]{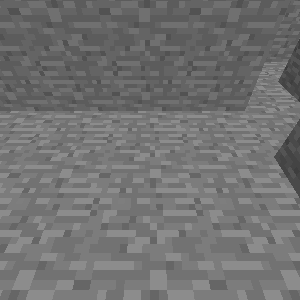}
\includegraphics[width=1\linewidth]{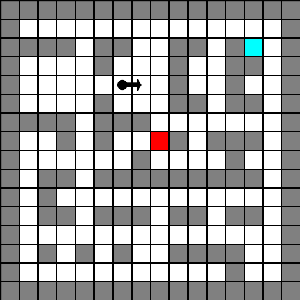}
\caption*{t = 56}
\end{subfigure}
&
\begin{subfigure}{0.08\linewidth}
\includegraphics[width=1\linewidth]{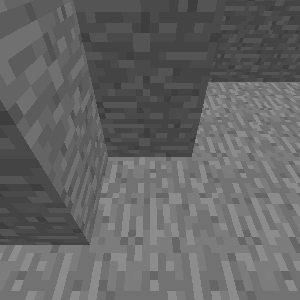}
\includegraphics[width=1\linewidth]{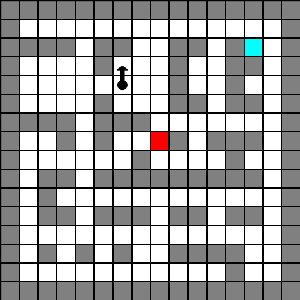}
\caption*{t = 57}
\end{subfigure}
&
\begin{subfigure}{0.08\linewidth}
\includegraphics[width=1\linewidth]{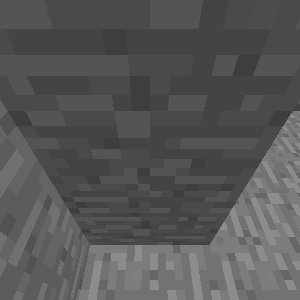}
\includegraphics[width=1\linewidth]{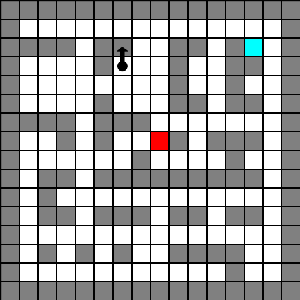}
\caption*{t = 58}
\end{subfigure}
&
\begin{subfigure}{0.08\linewidth}
\includegraphics[width=1\linewidth]{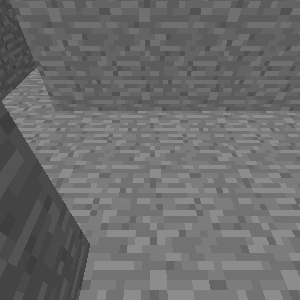}
\includegraphics[width=1\linewidth]{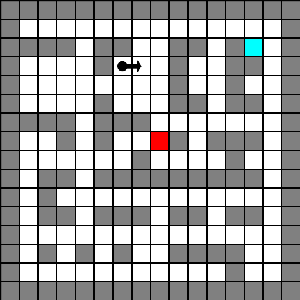}
\caption*{t = 59}
\end{subfigure}
&
\begin{subfigure}{0.08\linewidth}
\includegraphics[width=1\linewidth]{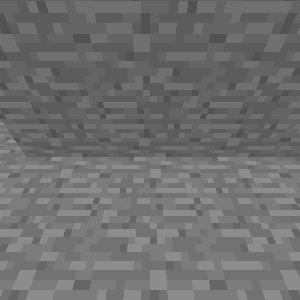}
\includegraphics[width=1\linewidth]{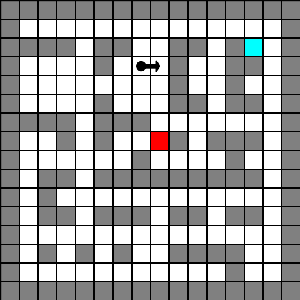}
\caption*{t = 60}
\end{subfigure}
\\
\\
\begin{subfigure}{0.08\linewidth}
\includegraphics[width=1\linewidth]{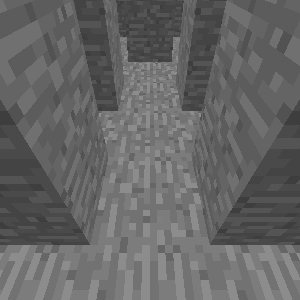}
\includegraphics[width=1\linewidth]{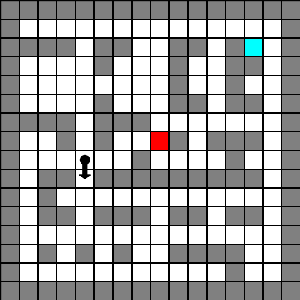}
\caption*{t = 73}
\end{subfigure}
&
\begin{subfigure}{0.08\linewidth}
\includegraphics[width=1\linewidth]{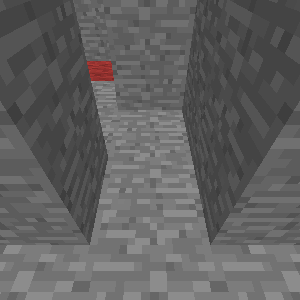}
\includegraphics[width=1\linewidth]{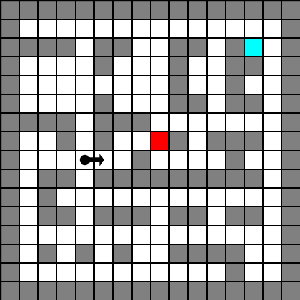}
\caption*{t = 74}
\end{subfigure}
&
\begin{subfigure}{0.08\linewidth}
\includegraphics[width=1\linewidth]{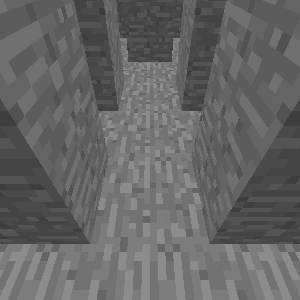}
\includegraphics[width=1\linewidth]{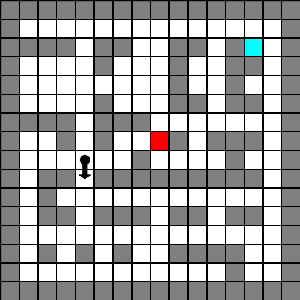}
\caption*{t = 75}
\end{subfigure}
&
\begin{subfigure}{0.08\linewidth}
\includegraphics[width=1\linewidth]{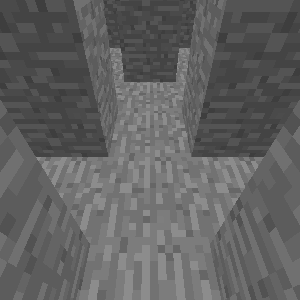}
\includegraphics[width=1\linewidth]{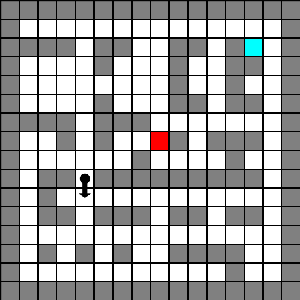}
\caption*{t = 76}
\end{subfigure}
&
\begin{subfigure}{0.08\linewidth}
\includegraphics[width=1\linewidth]{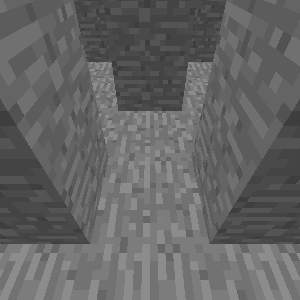}
\includegraphics[width=1\linewidth]{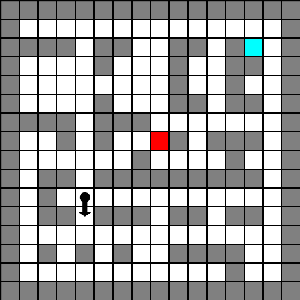}
\caption*{t = 77}
\end{subfigure}
&
\begin{subfigure}{0.08\linewidth}
\includegraphics[width=1\linewidth]{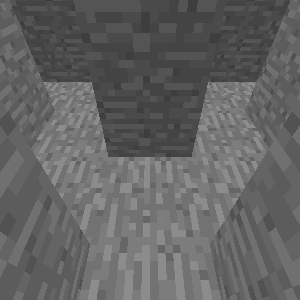}
\includegraphics[width=1\linewidth]{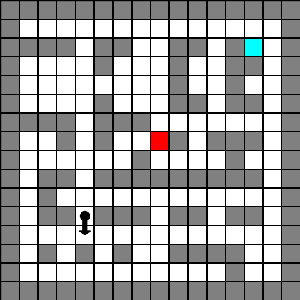}
\caption*{t = 78}
\end{subfigure}
&
\begin{subfigure}{0.08\linewidth}
\includegraphics[width=1\linewidth]{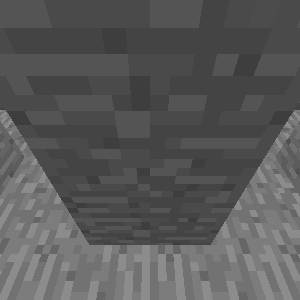}
\includegraphics[width=1\linewidth]{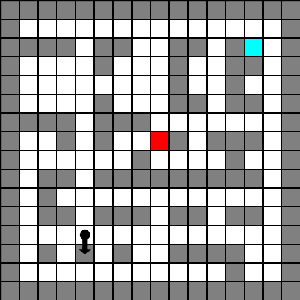}
\caption*{t = 79}
\end{subfigure}
&
\begin{subfigure}{0.08\linewidth}
\includegraphics[width=1\linewidth]{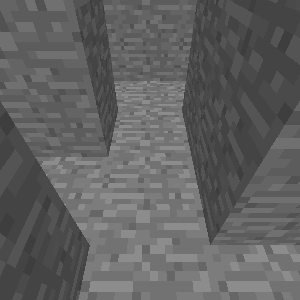}
\includegraphics[width=1\linewidth]{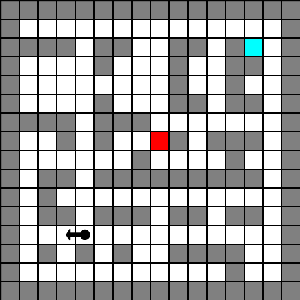}
\caption*{t = 80}
\end{subfigure}
&
\begin{subfigure}{0.08\linewidth}
\includegraphics[width=1\linewidth]{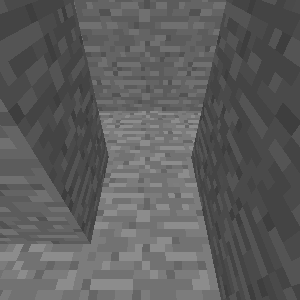}
\includegraphics[width=1\linewidth]{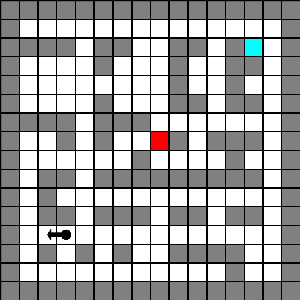}
\caption*{t = 81}
\end{subfigure}
&
\begin{subfigure}{0.08\linewidth}
\includegraphics[width=1\linewidth]{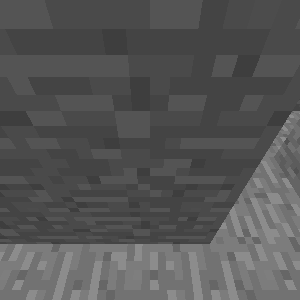}
\includegraphics[width=1\linewidth]{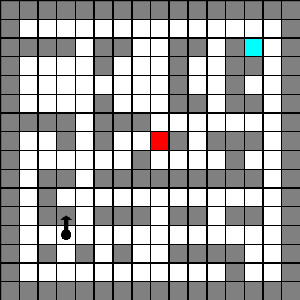}
\caption*{t = 82}
\end{subfigure}
&
\begin{subfigure}{0.08\linewidth}
\includegraphics[width=1\linewidth]{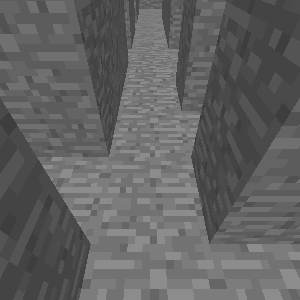}
\includegraphics[width=1\linewidth]{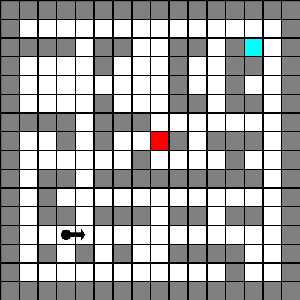}
\caption*{t = 83}
\end{subfigure}
&
\begin{subfigure}{0.08\linewidth}
\includegraphics[width=1\linewidth]{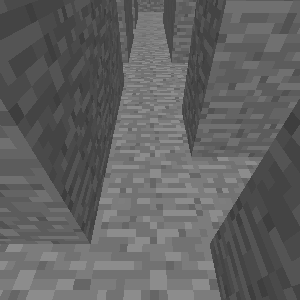}
\includegraphics[width=1\linewidth]{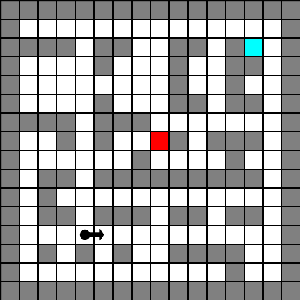}
\caption*{t = 84}
\end{subfigure}
\\
\\
\begin{subfigure}{0.08\linewidth}
\includegraphics[width=1\linewidth]{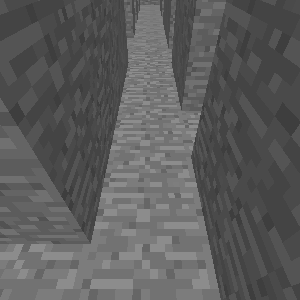}
\includegraphics[width=1\linewidth]{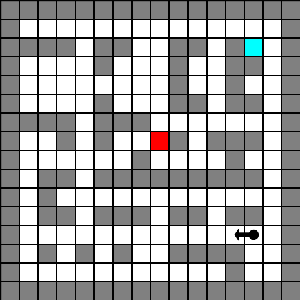}
\caption*{t = 97}
\end{subfigure}
&
\begin{subfigure}{0.08\linewidth}
\includegraphics[width=1\linewidth]{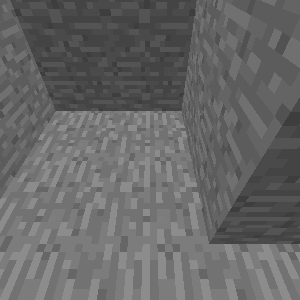}
\includegraphics[width=1\linewidth]{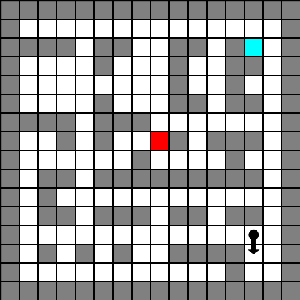}
\caption*{t = 98}
\end{subfigure}
&
\begin{subfigure}{0.08\linewidth}
\includegraphics[width=1\linewidth]{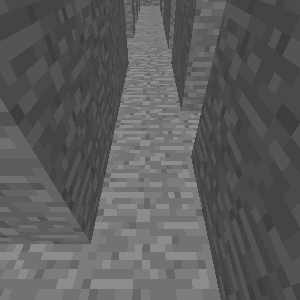}
\includegraphics[width=1\linewidth]{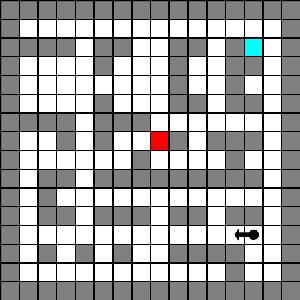}
\caption*{t = 99}
\end{subfigure}
&
\begin{subfigure}{0.08\linewidth}
\includegraphics[width=1\linewidth]{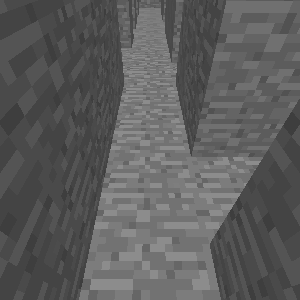}
\includegraphics[width=1\linewidth]{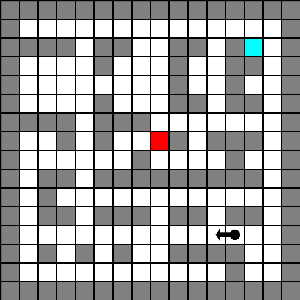}
\caption*{t = 100}
\end{subfigure}
\end{tabular}
\caption{FRMQN's play in an unseen and larger random maze with Single Goal task. Even though the agent explores the entire map in a reasonable way, it fails to find the blue block within 100 steps. This can occur occasionally, especially when the map is quite large and intrinsically complex (e.g., contains many walls giving rise to deep partial observability).}
\label{single_l_fail}
\end{figure*}
\begin{figure*}
    \small
    \setlength{\tabcolsep}{1pt}
    \def\arraystretch{1}
    \begin{tabular}{llllllllllll}
    \begin{subfigure}{0.08\linewidth}
	    \includegraphics[width=1\linewidth]{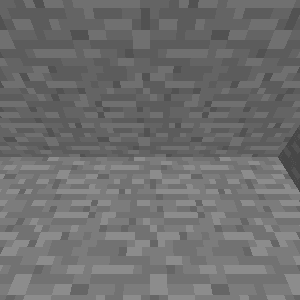} 
   		\includegraphics[width=1\linewidth]{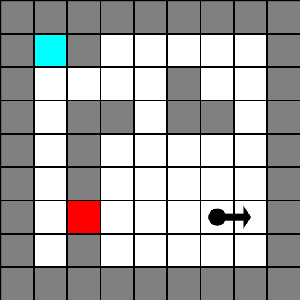} 
   		\caption*{t = 1}
	\end{subfigure} 
	& 
    \begin{subfigure}{0.08\linewidth}
	    \includegraphics[width=1\linewidth]{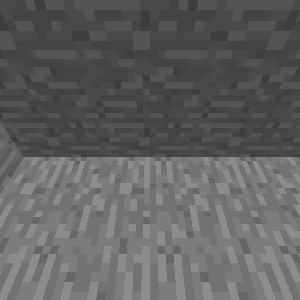} 
   		\includegraphics[width=1\linewidth]{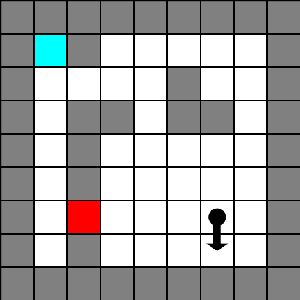} 
   		\caption*{t = 2}
	\end{subfigure} 
	& 
    \begin{subfigure}{0.08\linewidth}
	    \includegraphics[width=1\linewidth]{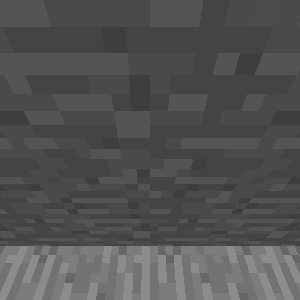} 
   		\includegraphics[width=1\linewidth]{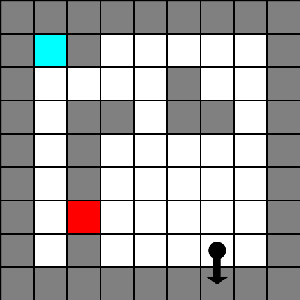} 
   		\caption*{t = 3}
	\end{subfigure} 
	& 
    \begin{subfigure}{0.08\linewidth}
	    \includegraphics[width=1\linewidth]{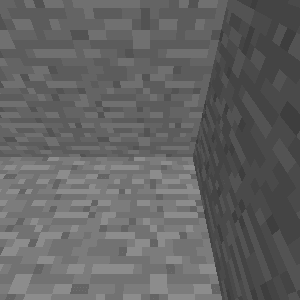} 
   		\includegraphics[width=1\linewidth]{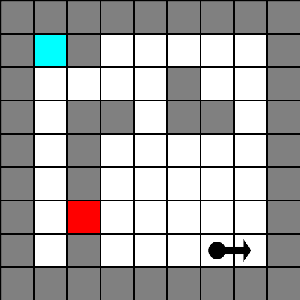} 
   		\caption*{t = 4}
	\end{subfigure}
	& 
    \begin{subfigure}{0.08\linewidth}
	    \includegraphics[width=1\linewidth]{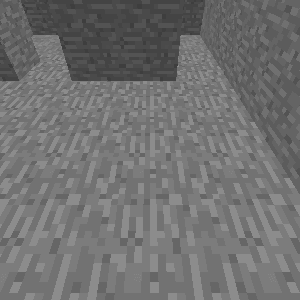} 
   		\includegraphics[width=1\linewidth]{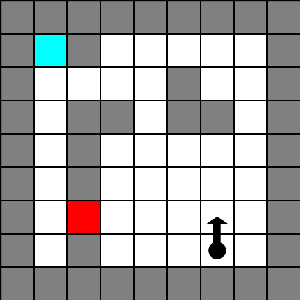}
   		\caption*{t = 5}
	\end{subfigure}
	& 
    \begin{subfigure}{0.08\linewidth}
	    \includegraphics[width=1\linewidth]{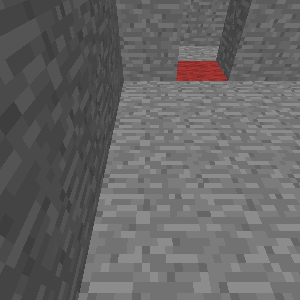} 
   		\includegraphics[width=1\linewidth]{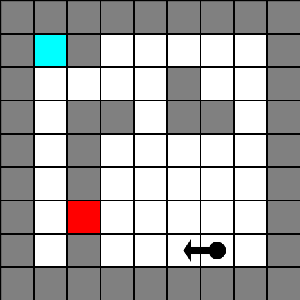} 
   		\caption*{t = 6}
	\end{subfigure} 
	& 
    \begin{subfigure}{0.08\linewidth}
	    \includegraphics[width=1\linewidth]{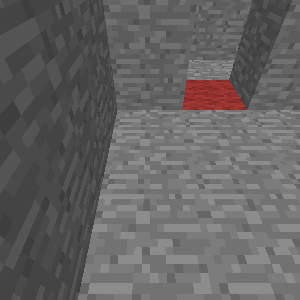} 
   		\includegraphics[width=1\linewidth]{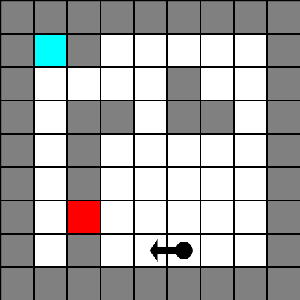} 
   		\caption*{t = 7}
	\end{subfigure} 
	& 
    \begin{subfigure}{0.08\linewidth}
	    \includegraphics[width=1\linewidth]{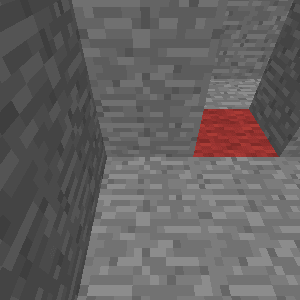} 
   		\includegraphics[width=1\linewidth]{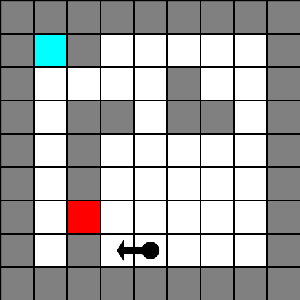} 
   		\caption*{t = 8}
	\end{subfigure} 
	& 
    \begin{subfigure}{0.08\linewidth}
	    \includegraphics[width=1\linewidth]{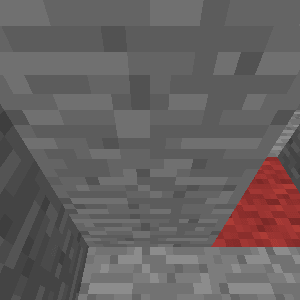} 
   		\includegraphics[width=1\linewidth]{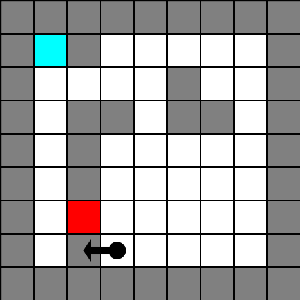} 
   		\caption*{t = 9}
	\end{subfigure}
	& 
    \begin{subfigure}{0.08\linewidth}
	    \includegraphics[width=1\linewidth]{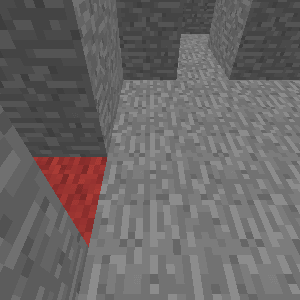} 
   		\includegraphics[width=1\linewidth]{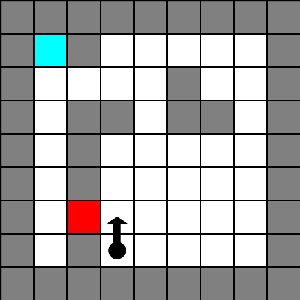} 
   		\caption*{t = 10}
	\end{subfigure}
	& 
    \begin{subfigure}{0.08\linewidth}
	    \includegraphics[width=1\linewidth]{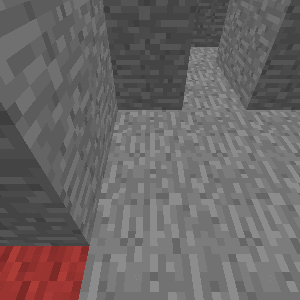} 
   		\includegraphics[width=1\linewidth]{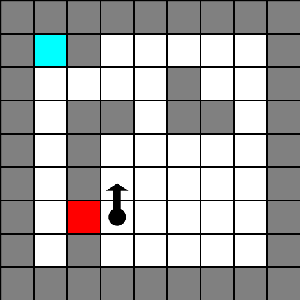} 
   		\caption*{t = 11}
	\end{subfigure} 
	&
    \begin{subfigure}{0.08\linewidth}
	    \includegraphics[width=1\linewidth]{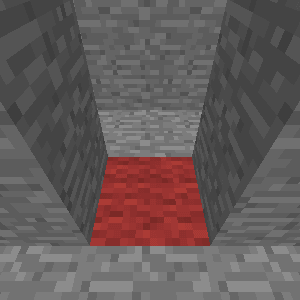} 
   		\includegraphics[width=1\linewidth]{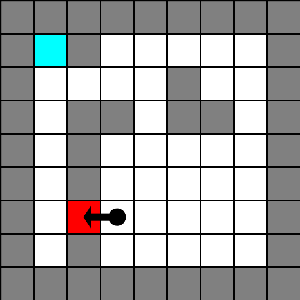} 
   		\caption*{t = 12}
	\end{subfigure} 
	\\
	\\
    \begin{subfigure}{0.08\linewidth}
	    \includegraphics[width=1\linewidth]{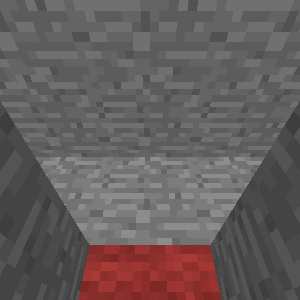} 
   		\includegraphics[width=1\linewidth]{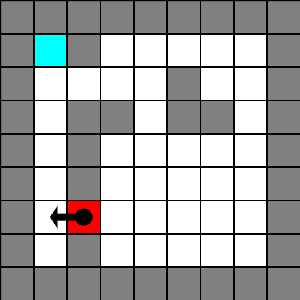} 
   		\caption*{t = 13}
	\end{subfigure} 
	& 
    \begin{subfigure}{0.08\linewidth}
	    \includegraphics[width=1\linewidth]{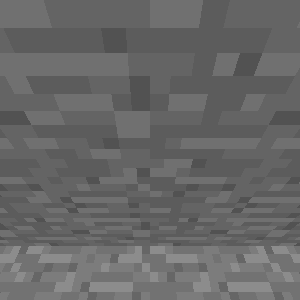} 
   		\includegraphics[width=1\linewidth]{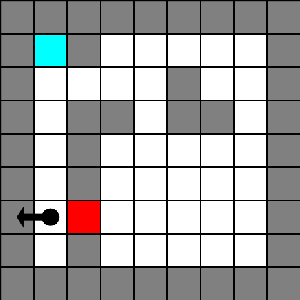} 
   		\caption*{t = 14}
	\end{subfigure} 
	& 
    \begin{subfigure}{0.08\linewidth}
	    \includegraphics[width=1\linewidth]{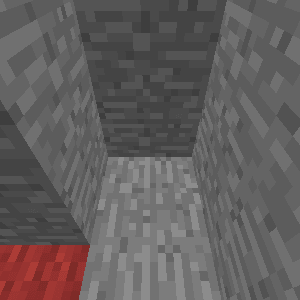} 
   		\includegraphics[width=1\linewidth]{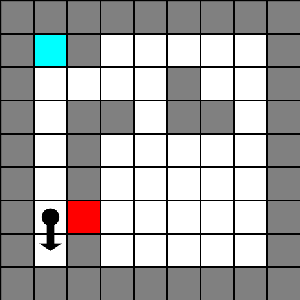} 
   		\caption*{t = 15}
	\end{subfigure} 
	& 
    \begin{subfigure}{0.08\linewidth}
	    \includegraphics[width=1\linewidth]{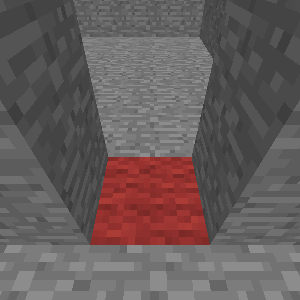} 
   		\includegraphics[width=1\linewidth]{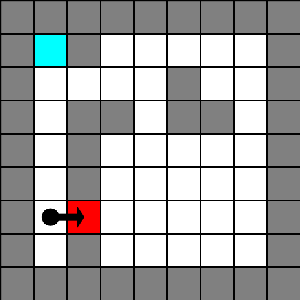} 
   		\caption*{t = 16}
	\end{subfigure}
	& 
    \begin{subfigure}{0.08\linewidth}
	    \includegraphics[width=1\linewidth]{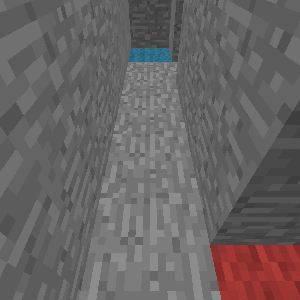} 
   		\includegraphics[width=1\linewidth]{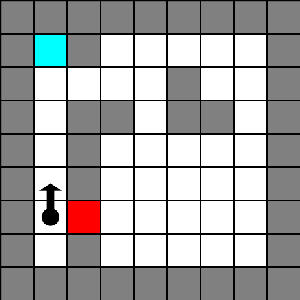} 
   		\caption*{t = 17}
	\end{subfigure}
	& 
    \begin{subfigure}{0.08\linewidth}
	    \includegraphics[width=1\linewidth]{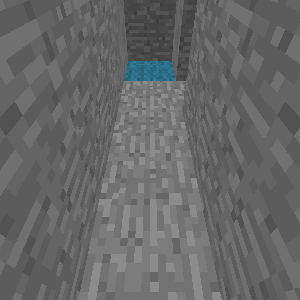} 
   		\includegraphics[width=1\linewidth]{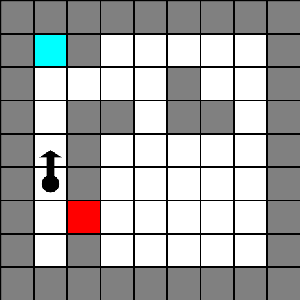} 
   		\caption*{t = 18}
	\end{subfigure} 
	& 
    \begin{subfigure}{0.08\linewidth}
	    \includegraphics[width=1\linewidth]{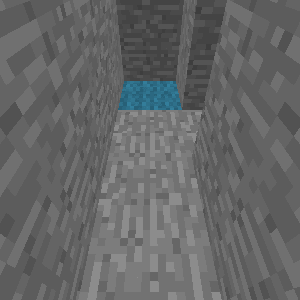} 
   		\includegraphics[width=1\linewidth]{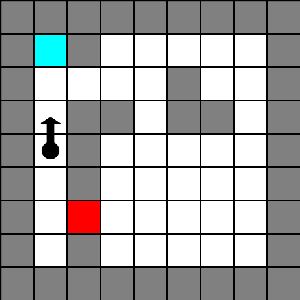} 
   		\caption*{t = 19}
	\end{subfigure} 
	& 
    \begin{subfigure}{0.08\linewidth}
	    \includegraphics[width=1\linewidth]{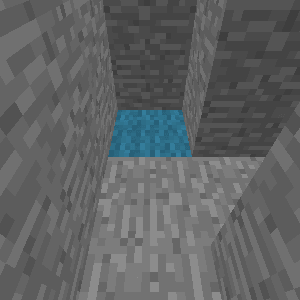} 
   		\includegraphics[width=1\linewidth]{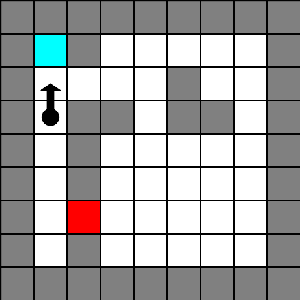} 
   		\caption*{t = 20}
	\end{subfigure} 
	& 
    \begin{subfigure}{0.08\linewidth}
	    \includegraphics[width=1\linewidth]{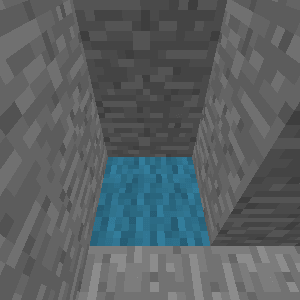} 
   		\includegraphics[width=1\linewidth]{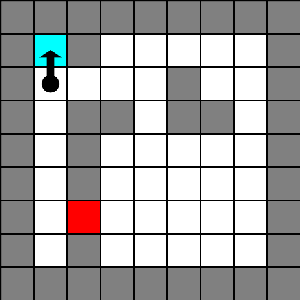} 
   		\caption*{t = 21}
	\end{subfigure}
	& 
    \begin{subfigure}{0.08\linewidth}
	    \includegraphics[width=1\linewidth]{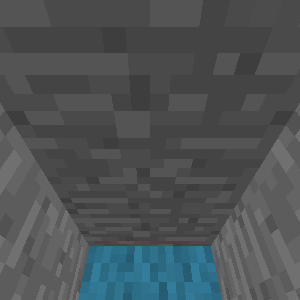} 
   		\includegraphics[width=1\linewidth]{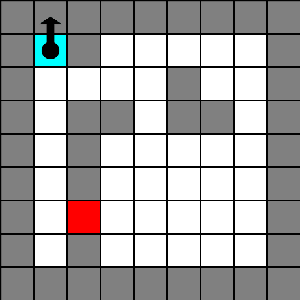} 
   		\caption*{t = 22}
	\end{subfigure}
	\end{tabular}
    \caption{FRMQN's play on a random maze with Sequential Goals task. As the case with most of the other tasks, the agent begins by looking down. It then looks around for the red block (t=1-6). Upon finding the red block, it visits it (t=13). The agents then looks for the blue block (t=14-18), successfully finds it, and hence complete the sequence and task (t=22). Notably the agent does not keep searching for the red block after visiting the red block. This is where memory can be crucial.} 
	\label{play-seq}
\end{figure*}

\begin{figure*}
\small
\setlength{\tabcolsep}{1pt}
\def\arraystretch{1}
\begin{tabular}{llllllllllll}
\begin{subfigure}{0.08\linewidth}
\includegraphics[width=1\linewidth]{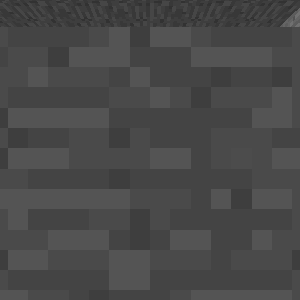}
\includegraphics[width=1\linewidth]{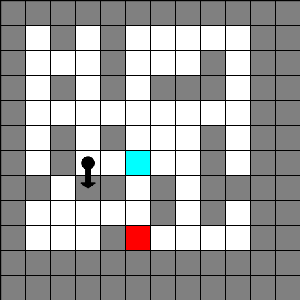}
\caption*{t = 1}
\end{subfigure}
&
\begin{subfigure}{0.08\linewidth}
\includegraphics[width=1\linewidth]{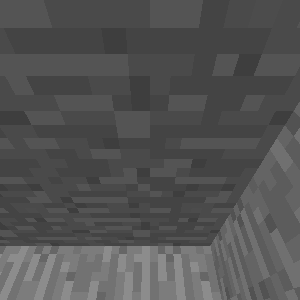}
\includegraphics[width=1\linewidth]{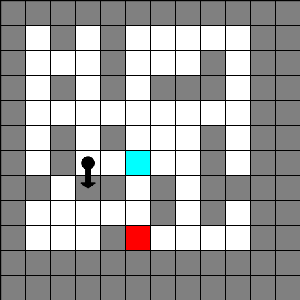}
\caption*{t = 2}
\end{subfigure}
&
\begin{subfigure}{0.08\linewidth}
\includegraphics[width=1\linewidth]{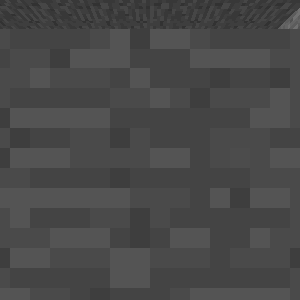}
\includegraphics[width=1\linewidth]{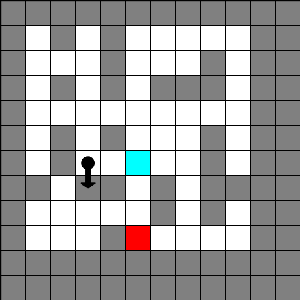}
\caption*{t = 3}
\end{subfigure}
&
\begin{subfigure}{0.08\linewidth}
\includegraphics[width=1\linewidth]{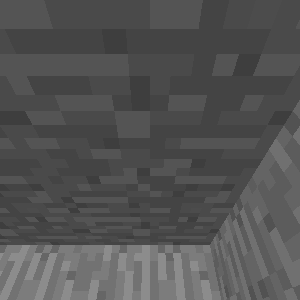}
\includegraphics[width=1\linewidth]{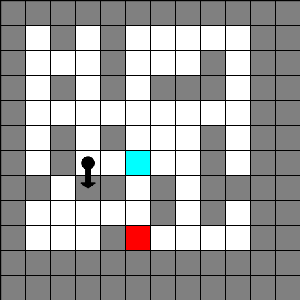}
\caption*{t = 4}
\end{subfigure}
&
\begin{subfigure}{0.08\linewidth}
\includegraphics[width=1\linewidth]{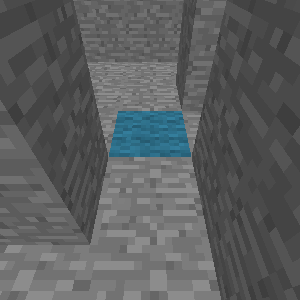}
\includegraphics[width=1\linewidth]{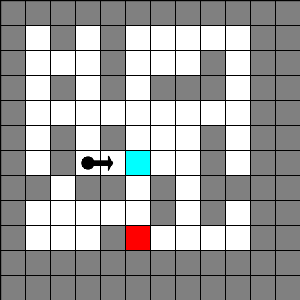}
\caption*{t = 5}
\end{subfigure}
&
\begin{subfigure}{0.08\linewidth}
\includegraphics[width=1\linewidth]{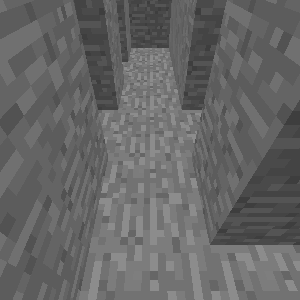}
\includegraphics[width=1\linewidth]{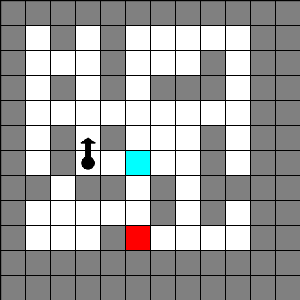}
\caption*{t = 6}
\end{subfigure}
&
\begin{subfigure}{0.08\linewidth}
\includegraphics[width=1\linewidth]{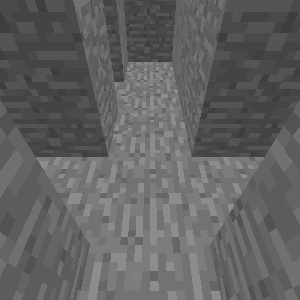}
\includegraphics[width=1\linewidth]{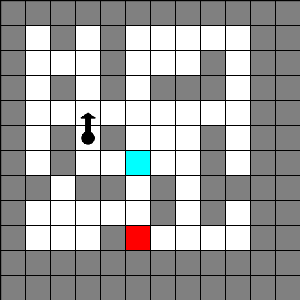}
\caption*{t = 7}
\end{subfigure}
&
\begin{subfigure}{0.08\linewidth}
\includegraphics[width=1\linewidth]{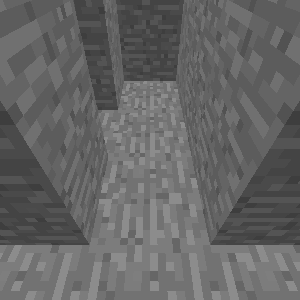}
\includegraphics[width=1\linewidth]{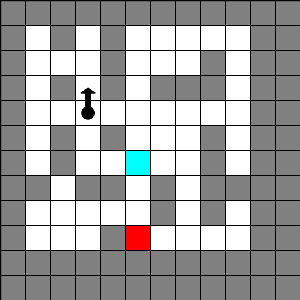}
\caption*{t = 8}
\end{subfigure}
&
\begin{subfigure}{0.08\linewidth}
\includegraphics[width=1\linewidth]{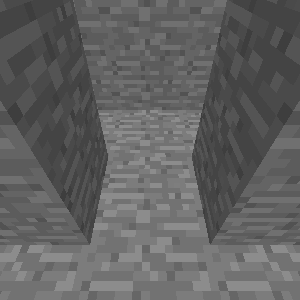}
\includegraphics[width=1\linewidth]{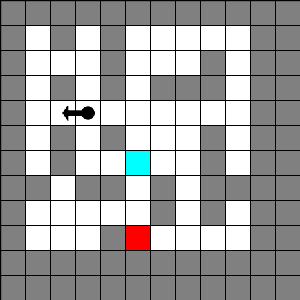}
\caption*{t = 9}
\end{subfigure}
&
\begin{subfigure}{0.08\linewidth}
\includegraphics[width=1\linewidth]{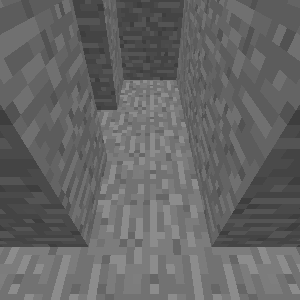}
\includegraphics[width=1\linewidth]{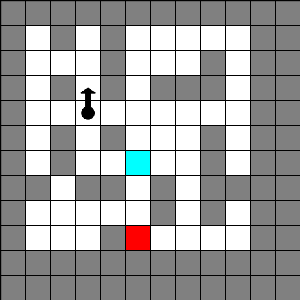}
\caption*{t = 10}
\end{subfigure}
&
\begin{subfigure}{0.08\linewidth}
\includegraphics[width=1\linewidth]{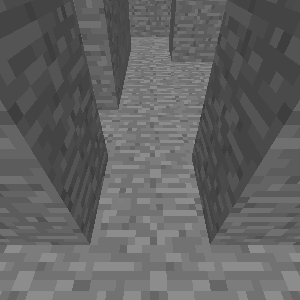}
\includegraphics[width=1\linewidth]{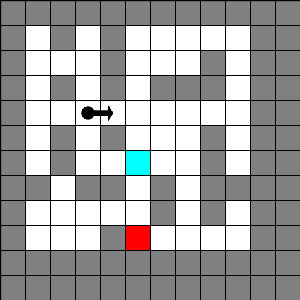}
\caption*{t = 11}
\end{subfigure}
&
\begin{subfigure}{0.08\linewidth}
\includegraphics[width=1\linewidth]{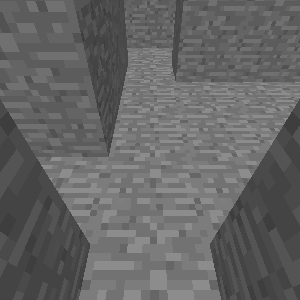}
\includegraphics[width=1\linewidth]{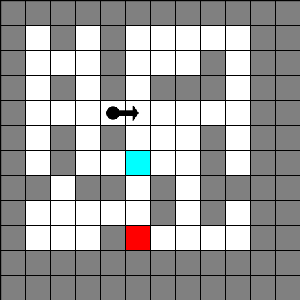}
\caption*{t = 12}
\end{subfigure}
\\
\\
\begin{subfigure}{0.08\linewidth}
\includegraphics[width=1\linewidth]{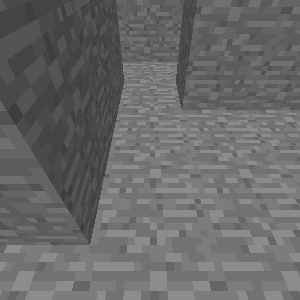}
\includegraphics[width=1\linewidth]{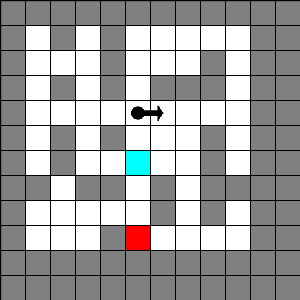}
\caption*{t = 13}
\end{subfigure}
&
\begin{subfigure}{0.08\linewidth}
\includegraphics[width=1\linewidth]{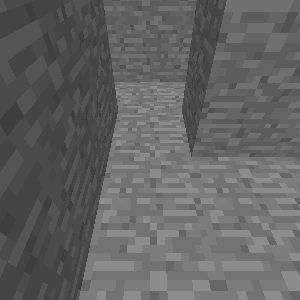}
\includegraphics[width=1\linewidth]{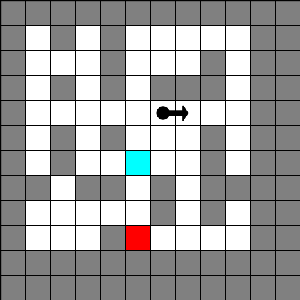}
\caption*{t = 14}
\end{subfigure}
&
\begin{subfigure}{0.08\linewidth}
\includegraphics[width=1\linewidth]{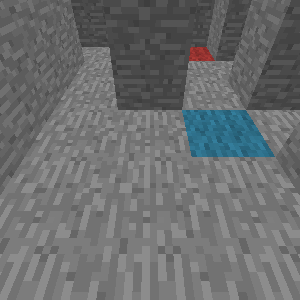}
\includegraphics[width=1\linewidth]{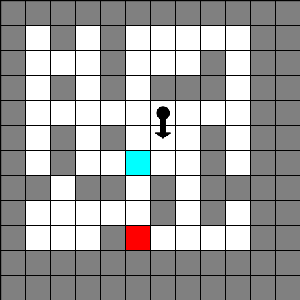}
\caption*{t = 15}
\end{subfigure}
&
\begin{subfigure}{0.08\linewidth}
\includegraphics[width=1\linewidth]{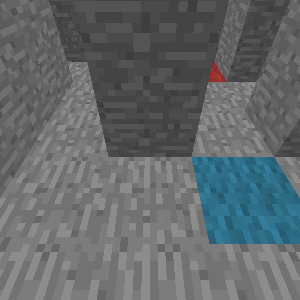}
\includegraphics[width=1\linewidth]{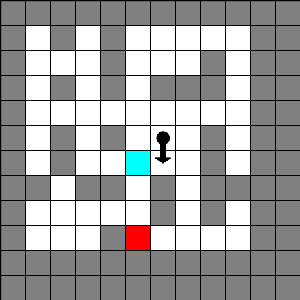}
\caption*{t = 16}
\end{subfigure}
&
\begin{subfigure}{0.08\linewidth}
\includegraphics[width=1\linewidth]{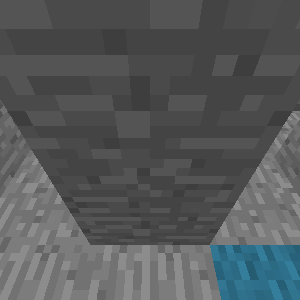}
\includegraphics[width=1\linewidth]{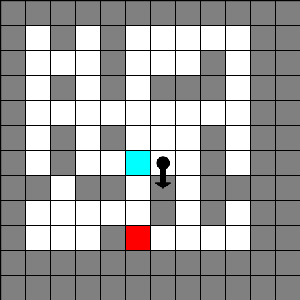}
\caption*{t = 17}
\end{subfigure}
&
\begin{subfigure}{0.08\linewidth}
\includegraphics[width=1\linewidth]{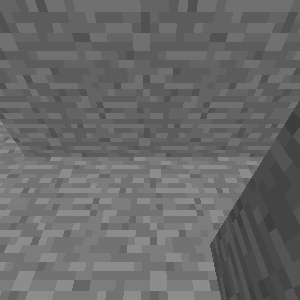}
\includegraphics[width=1\linewidth]{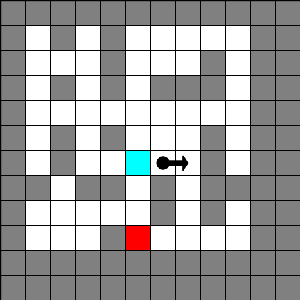}
\caption*{t = 18}
\end{subfigure}
&
\begin{subfigure}{0.08\linewidth}
\includegraphics[width=1\linewidth]{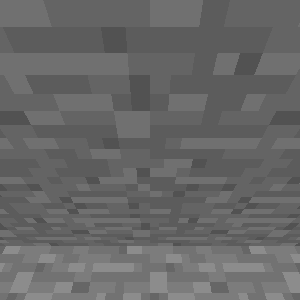}
\includegraphics[width=1\linewidth]{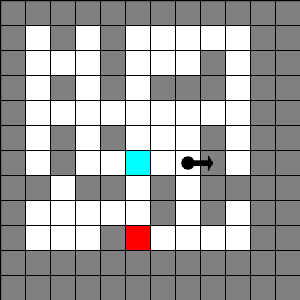}
\caption*{t = 19}
\end{subfigure}
&
\begin{subfigure}{0.08\linewidth}
\includegraphics[width=1\linewidth]{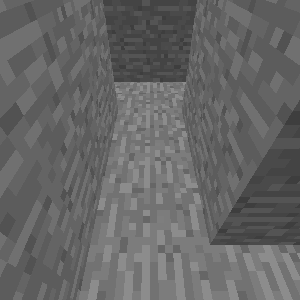}
\includegraphics[width=1\linewidth]{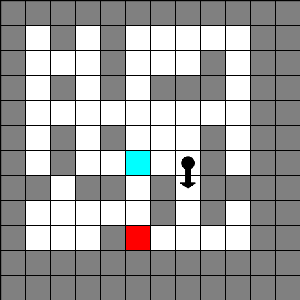}
\caption*{t = 20}
\end{subfigure}
&
\begin{subfigure}{0.08\linewidth}
\includegraphics[width=1\linewidth]{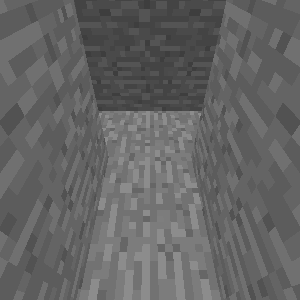}
\includegraphics[width=1\linewidth]{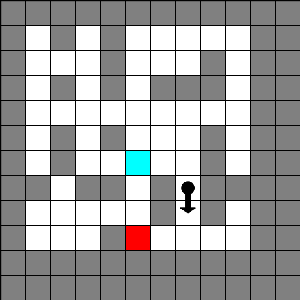}
\caption*{t = 21}
\end{subfigure}
&
\begin{subfigure}{0.08\linewidth}
\includegraphics[width=1\linewidth]{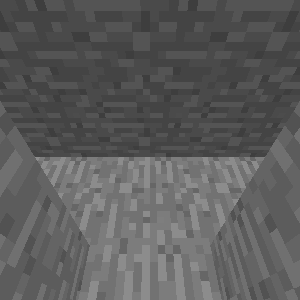}
\includegraphics[width=1\linewidth]{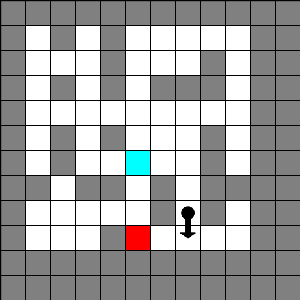}
\caption*{t = 22}
\end{subfigure}
&
\begin{subfigure}{0.08\linewidth}
\includegraphics[width=1\linewidth]{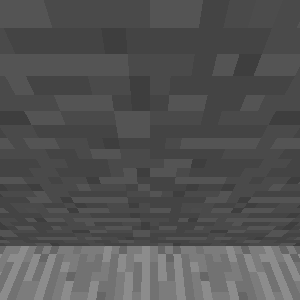}
\includegraphics[width=1\linewidth]{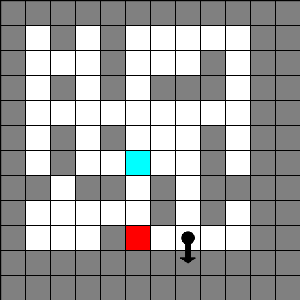}
\caption*{t = 23}
\end{subfigure}
&
\begin{subfigure}{0.08\linewidth}
\includegraphics[width=1\linewidth]{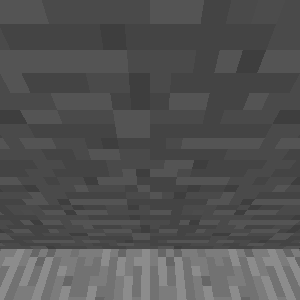}
\includegraphics[width=1\linewidth]{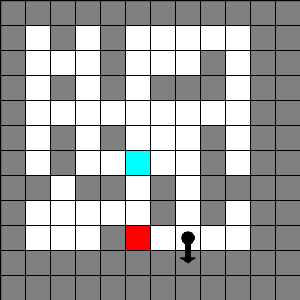}
\caption*{t = 24}
\end{subfigure}
\\
\\
\begin{subfigure}{0.08\linewidth}
\includegraphics[width=1\linewidth]{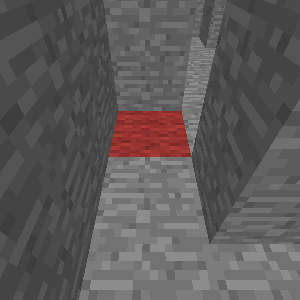}
\includegraphics[width=1\linewidth]{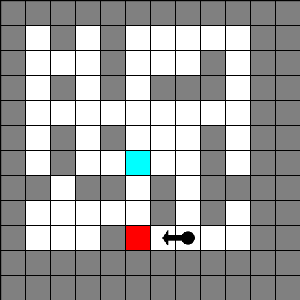}
\caption*{t = 25}
\end{subfigure}
&
\begin{subfigure}{0.08\linewidth}
\includegraphics[width=1\linewidth]{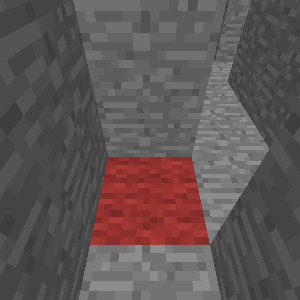}
\includegraphics[width=1\linewidth]{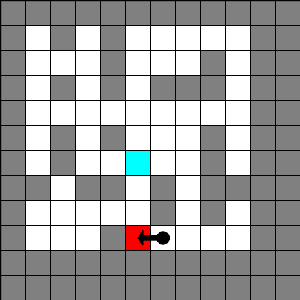}
\caption*{t = 26}
\end{subfigure}
&
\begin{subfigure}{0.08\linewidth}
\includegraphics[width=1\linewidth]{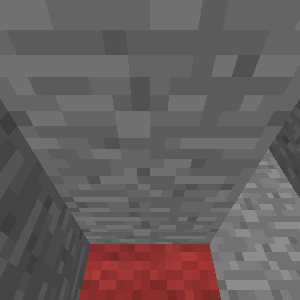}
\includegraphics[width=1\linewidth]{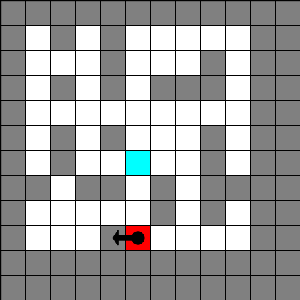}
\caption*{t = 27}
\end{subfigure}
&
\begin{subfigure}{0.08\linewidth}
\includegraphics[width=1\linewidth]{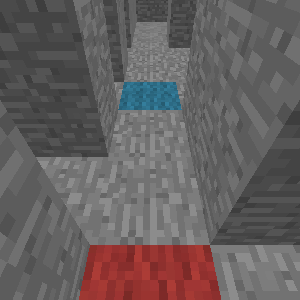}
\includegraphics[width=1\linewidth]{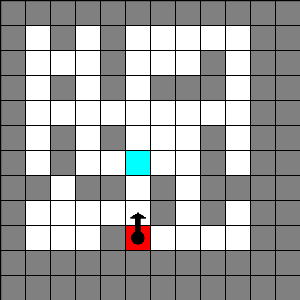}
\caption*{t = 28}
\end{subfigure}
&
\begin{subfigure}{0.08\linewidth}
\includegraphics[width=1\linewidth]{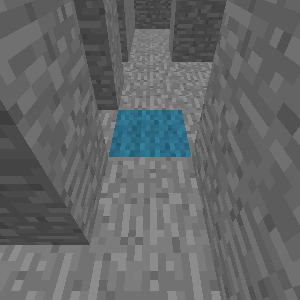}
\includegraphics[width=1\linewidth]{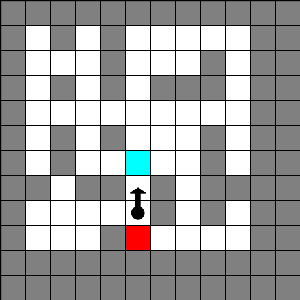}
\caption*{t = 29}
\end{subfigure}
&
\begin{subfigure}{0.08\linewidth}
\includegraphics[width=1\linewidth]{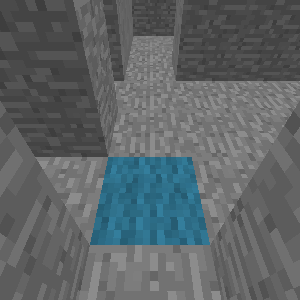}
\includegraphics[width=1\linewidth]{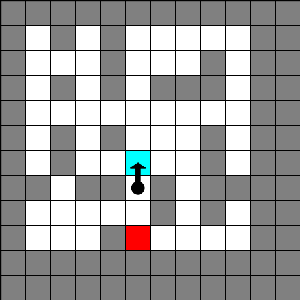}
\caption*{t = 30}
\end{subfigure}
&
\begin{subfigure}{0.08\linewidth}
\includegraphics[width=1\linewidth]{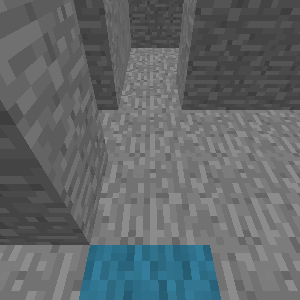}
\includegraphics[width=1\linewidth]{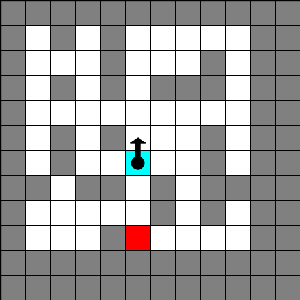}
\caption*{t = 31}
\end{subfigure}
\end{tabular}
\caption{FRMQN's play on an unseen and larger random maze with Sequential Goals task. With the context of the visual observations containing the blue block (t=5,15,16,17), the agent avoids the blue block and keeps searching for the red block based on its memory because it has not visited the red block. After finding and visiting the red block (t=28), it directly goes to the blue block (t=31), completing the sequence in the correct order, and hence the task.}
\label{seq_l_success}
\end{figure*}

\begin{figure*}
\small
\setlength{\tabcolsep}{1pt}
\def\arraystretch{1}
\begin{tabular}{llllllllllll}
\begin{subfigure}{0.08\linewidth}
\includegraphics[width=1\linewidth]{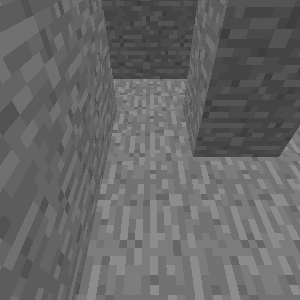}
\includegraphics[width=1\linewidth]{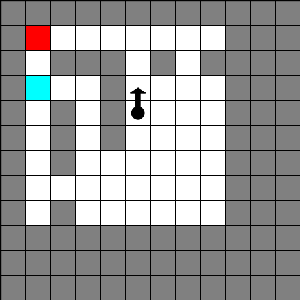}
\caption*{t = 1}
\end{subfigure}
&
\begin{subfigure}{0.08\linewidth}
\includegraphics[width=1\linewidth]{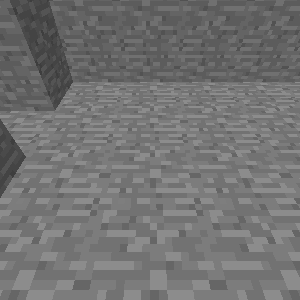}
\includegraphics[width=1\linewidth]{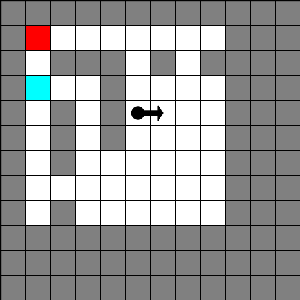}
\caption*{t = 2}
\end{subfigure}
&
\begin{subfigure}{0.08\linewidth}
\includegraphics[width=1\linewidth]{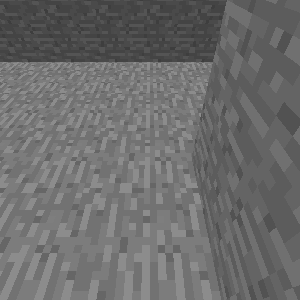}
\includegraphics[width=1\linewidth]{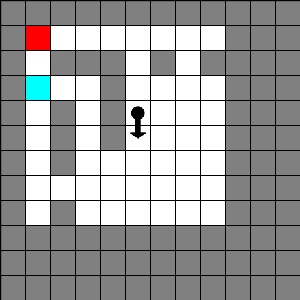}
\caption*{t = 3}
\end{subfigure}
&
\begin{subfigure}{0.08\linewidth}
\includegraphics[width=1\linewidth]{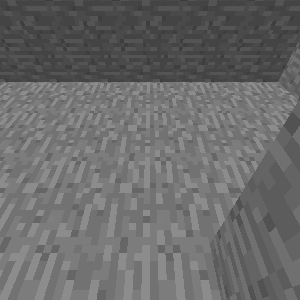}
\includegraphics[width=1\linewidth]{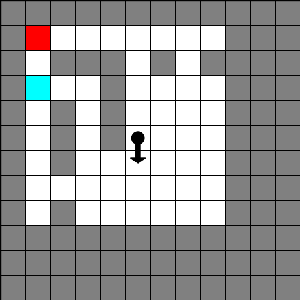}
\caption*{t = 4}
\end{subfigure}
&
\begin{subfigure}{0.08\linewidth}
\includegraphics[width=1\linewidth]{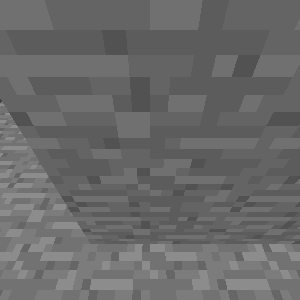}
\includegraphics[width=1\linewidth]{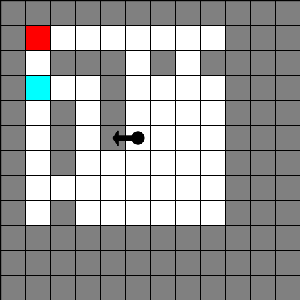}
\caption*{t = 5}
\end{subfigure}
&
\begin{subfigure}{0.08\linewidth}
\includegraphics[width=1\linewidth]{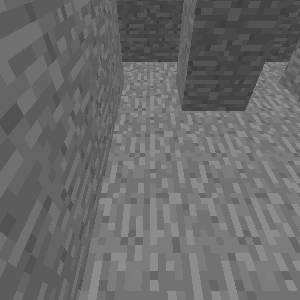}
\includegraphics[width=1\linewidth]{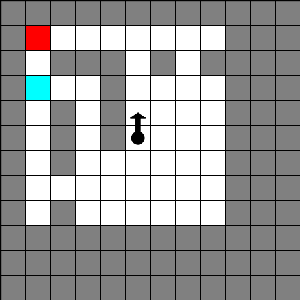}
\caption*{t = 6}
\end{subfigure}
&
\begin{subfigure}{0.08\linewidth}
\includegraphics[width=1\linewidth]{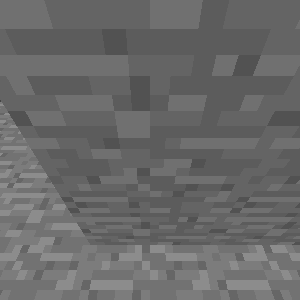}
\includegraphics[width=1\linewidth]{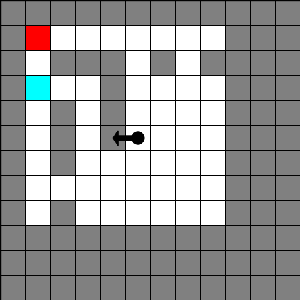}
\caption*{t = 7}
\end{subfigure}
&
\begin{subfigure}{0.08\linewidth}
\includegraphics[width=1\linewidth]{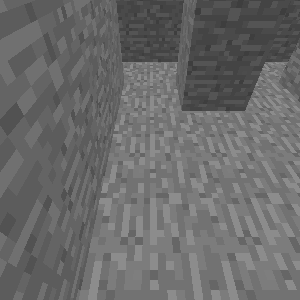}
\includegraphics[width=1\linewidth]{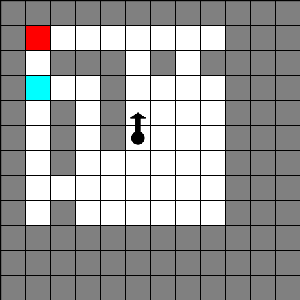}
\caption*{t = 8}
\end{subfigure}
&
\begin{subfigure}{0.08\linewidth}
\includegraphics[width=1\linewidth]{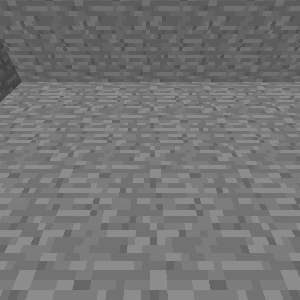}
\includegraphics[width=1\linewidth]{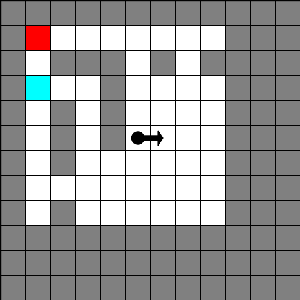}
\caption*{t = 9}
\end{subfigure}
&
\begin{subfigure}{0.08\linewidth}
\includegraphics[width=1\linewidth]{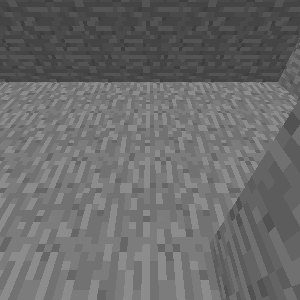}
\includegraphics[width=1\linewidth]{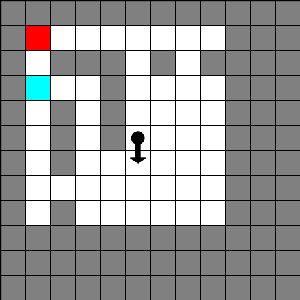}
\caption*{t = 10}
\end{subfigure}
&
\begin{subfigure}{0.08\linewidth}
\includegraphics[width=1\linewidth]{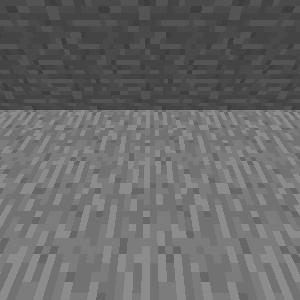}
\includegraphics[width=1\linewidth]{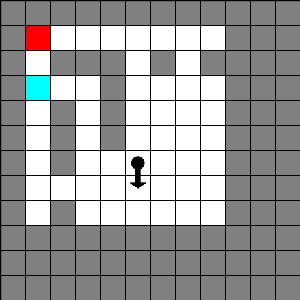}
\caption*{t = 11}
\end{subfigure}
&
\begin{subfigure}{0.08\linewidth}
\includegraphics[width=1\linewidth]{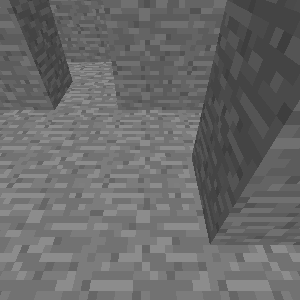}
\includegraphics[width=1\linewidth]{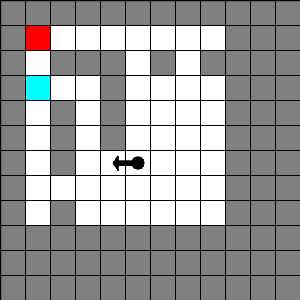}
\caption*{t = 12}
\end{subfigure}
\\
\\
\begin{subfigure}{0.08\linewidth}
\includegraphics[width=1\linewidth]{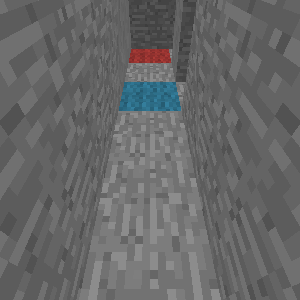}
\includegraphics[width=1\linewidth]{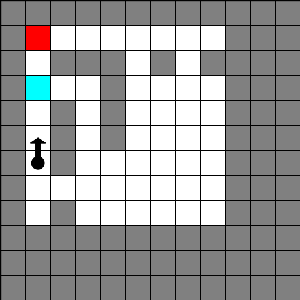}
\caption*{t = 25}
\end{subfigure}
&
\begin{subfigure}{0.08\linewidth}
\includegraphics[width=1\linewidth]{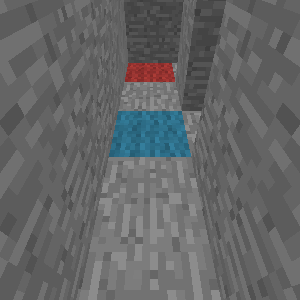}
\includegraphics[width=1\linewidth]{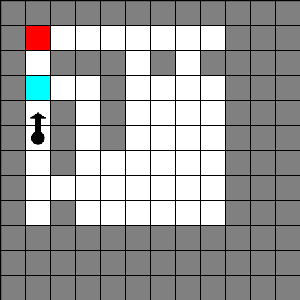}
\caption*{t = 26}
\end{subfigure}
&
\begin{subfigure}{0.08\linewidth}
\includegraphics[width=1\linewidth]{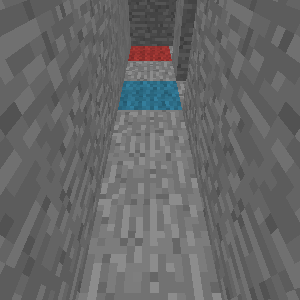}
\includegraphics[width=1\linewidth]{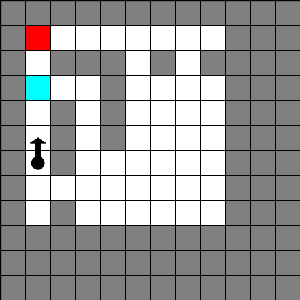}
\caption*{t = 27}
\end{subfigure}
&
\begin{subfigure}{0.08\linewidth}
\includegraphics[width=1\linewidth]{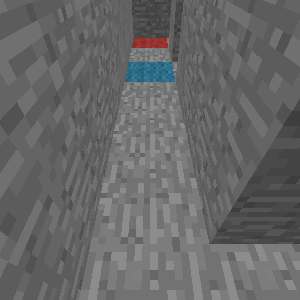}
\includegraphics[width=1\linewidth]{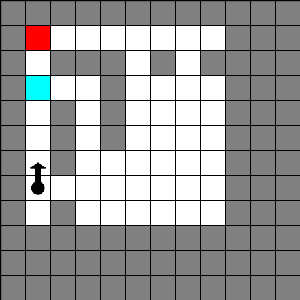}
\caption*{t = 28}
\end{subfigure}
&
\begin{subfigure}{0.08\linewidth}
\includegraphics[width=1\linewidth]{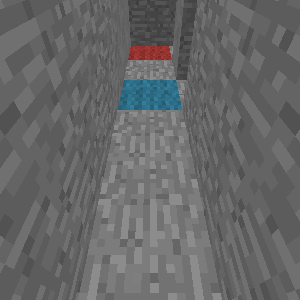}
\includegraphics[width=1\linewidth]{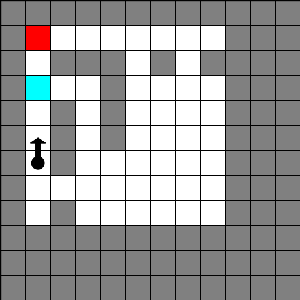}
\caption*{t = 29}
\end{subfigure}
&
\begin{subfigure}{0.08\linewidth}
\includegraphics[width=1\linewidth]{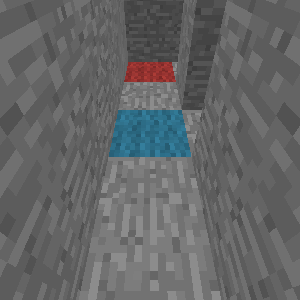}
\includegraphics[width=1\linewidth]{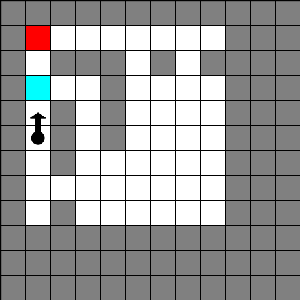}
\caption*{t = 30}
\end{subfigure}
&
\begin{subfigure}{0.08\linewidth}
\includegraphics[width=1\linewidth]{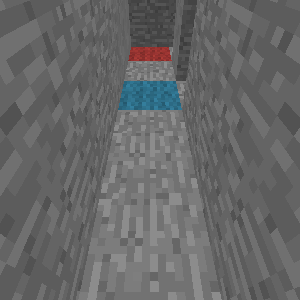}
\includegraphics[width=1\linewidth]{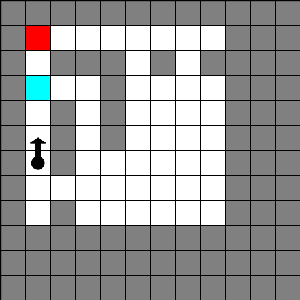}
\caption*{t = 31}
\end{subfigure}
&
\begin{subfigure}{0.08\linewidth}
\includegraphics[width=1\linewidth]{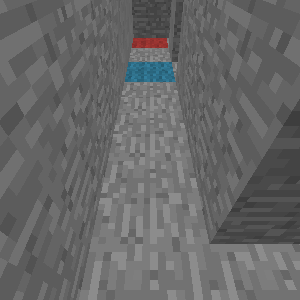}
\includegraphics[width=1\linewidth]{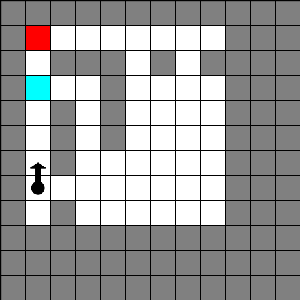}
\caption*{t = 32}
\end{subfigure}
&
\begin{subfigure}{0.08\linewidth}
\includegraphics[width=1\linewidth]{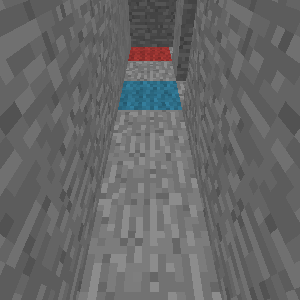}
\includegraphics[width=1\linewidth]{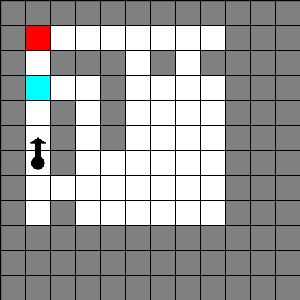}
\caption*{t = 33}
\end{subfigure}
&
\begin{subfigure}{0.08\linewidth}
\includegraphics[width=1\linewidth]{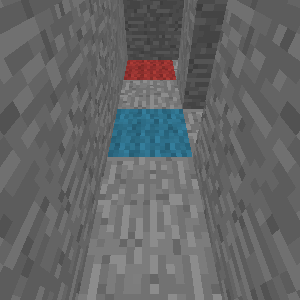}
\includegraphics[width=1\linewidth]{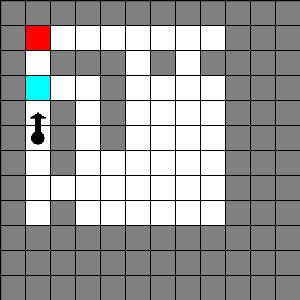}
\caption*{t = 34}
\end{subfigure}
&
\begin{subfigure}{0.08\linewidth}
\includegraphics[width=1\linewidth]{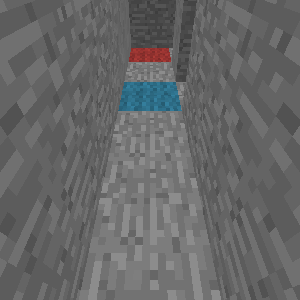}
\includegraphics[width=1\linewidth]{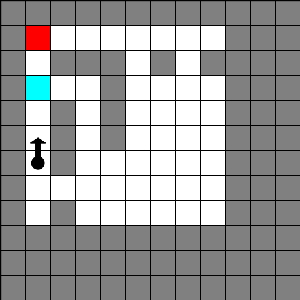}
\caption*{t = 35}
\end{subfigure}
&
\begin{subfigure}{0.08\linewidth}
\includegraphics[width=1\linewidth]{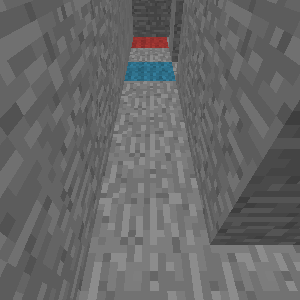}
\includegraphics[width=1\linewidth]{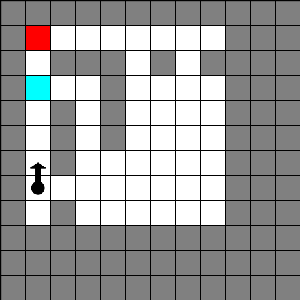}
\caption*{t = 36}
\end{subfigure}
\\
\\
\begin{subfigure}{0.08\linewidth}
\includegraphics[width=1\linewidth]{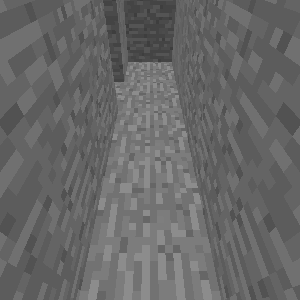}
\includegraphics[width=1\linewidth]{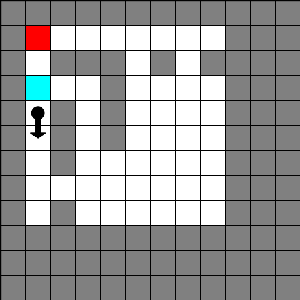}
\caption*{t = 49}
\end{subfigure}
&
\begin{subfigure}{0.08\linewidth}
\includegraphics[width=1\linewidth]{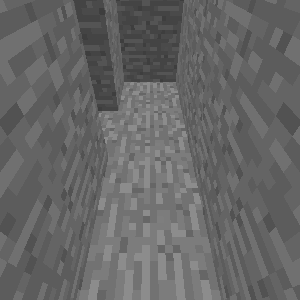}
\includegraphics[width=1\linewidth]{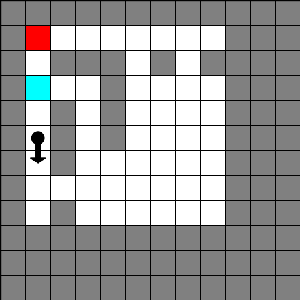}
\caption*{t = 50}
\end{subfigure}
&
\begin{subfigure}{0.08\linewidth}
\includegraphics[width=1\linewidth]{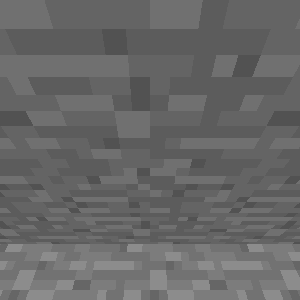}
\includegraphics[width=1\linewidth]{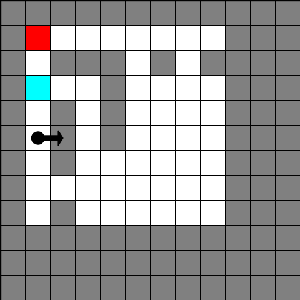}
\caption*{t = 51}
\end{subfigure}
&
\begin{subfigure}{0.08\linewidth}
\includegraphics[width=1\linewidth]{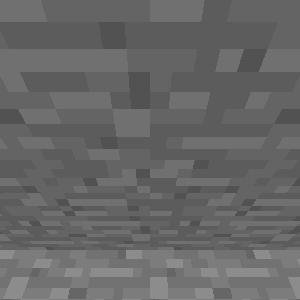}
\includegraphics[width=1\linewidth]{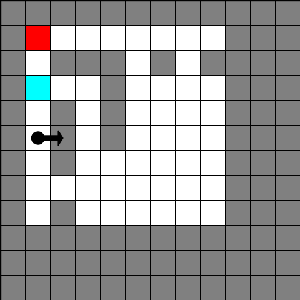}
\caption*{t = 52}
\end{subfigure}
&
\begin{subfigure}{0.08\linewidth}
\includegraphics[width=1\linewidth]{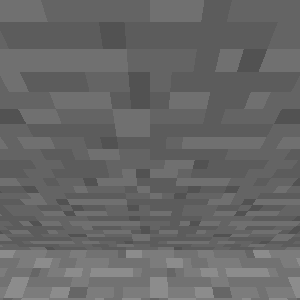}
\includegraphics[width=1\linewidth]{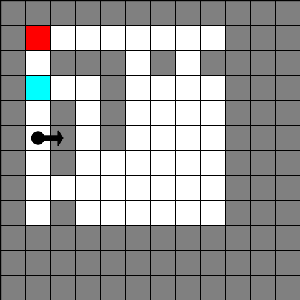}
\caption*{t = 53}
\end{subfigure}
&
\begin{subfigure}{0.08\linewidth}
\includegraphics[width=1\linewidth]{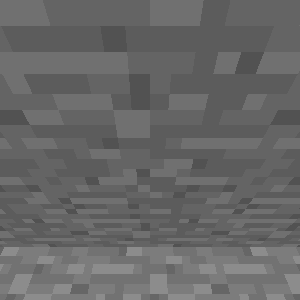}
\includegraphics[width=1\linewidth]{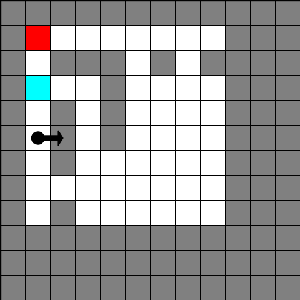}
\caption*{t = 54}
\end{subfigure}
&
\begin{subfigure}{0.08\linewidth}
\includegraphics[width=1\linewidth]{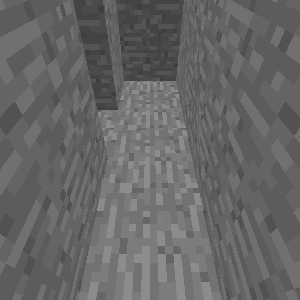}
\includegraphics[width=1\linewidth]{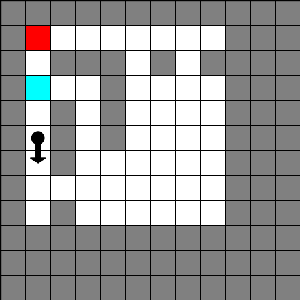}
\caption*{t = 55}
\end{subfigure}
&
\begin{subfigure}{0.08\linewidth}
\includegraphics[width=1\linewidth]{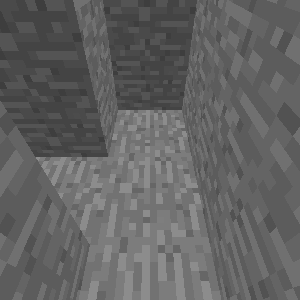}
\includegraphics[width=1\linewidth]{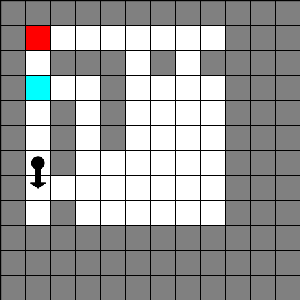}
\caption*{t = 56}
\end{subfigure}
&
\begin{subfigure}{0.08\linewidth}
\includegraphics[width=1\linewidth]{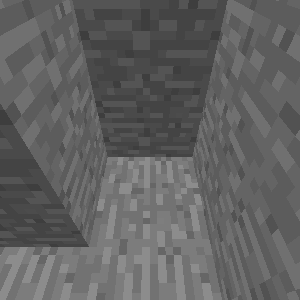}
\includegraphics[width=1\linewidth]{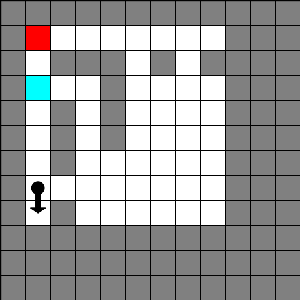}
\caption*{t = 57}
\end{subfigure}
&
\begin{subfigure}{0.08\linewidth}
\includegraphics[width=1\linewidth]{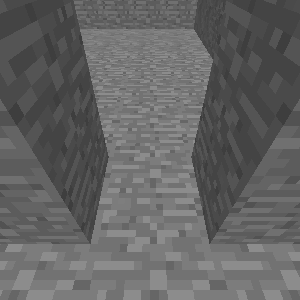}
\includegraphics[width=1\linewidth]{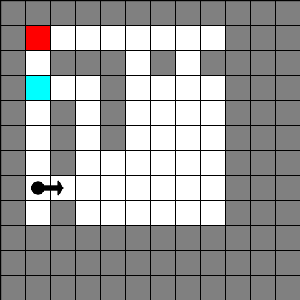}
\caption*{t = 58}
\end{subfigure}
&
\begin{subfigure}{0.08\linewidth}
\includegraphics[width=1\linewidth]{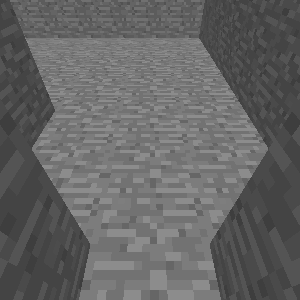}
\includegraphics[width=1\linewidth]{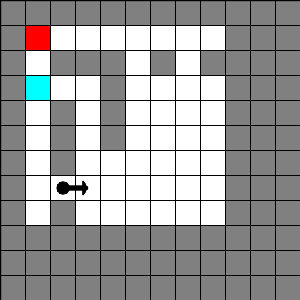}
\caption*{t = 59}
\end{subfigure}
&
\begin{subfigure}{0.08\linewidth}
\includegraphics[width=1\linewidth]{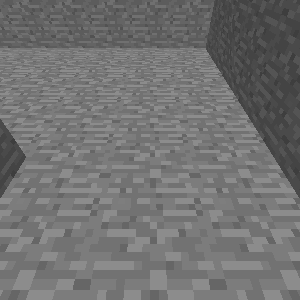}
\includegraphics[width=1\linewidth]{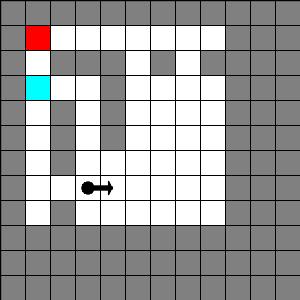}
\caption*{t = 60}
\end{subfigure}
\\
\\
\begin{subfigure}{0.08\linewidth}
\includegraphics[width=1\linewidth]{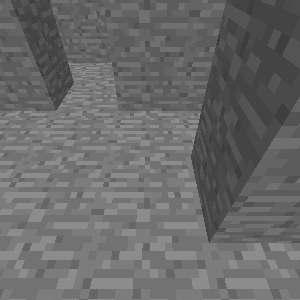}
\includegraphics[width=1\linewidth]{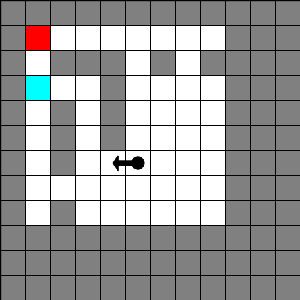}
\caption*{t = 73}
\end{subfigure}
&
\begin{subfigure}{0.08\linewidth}
\includegraphics[width=1\linewidth]{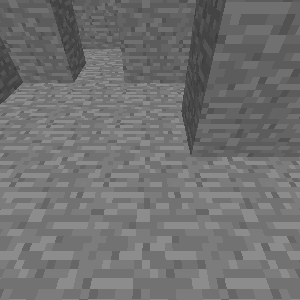}
\includegraphics[width=1\linewidth]{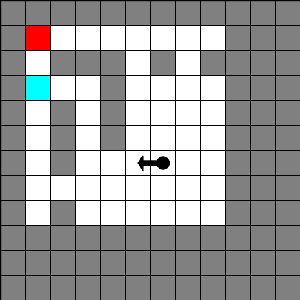}
\caption*{t = 74}
\end{subfigure}
&
\begin{subfigure}{0.08\linewidth}
\includegraphics[width=1\linewidth]{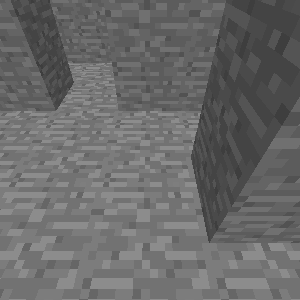}
\includegraphics[width=1\linewidth]{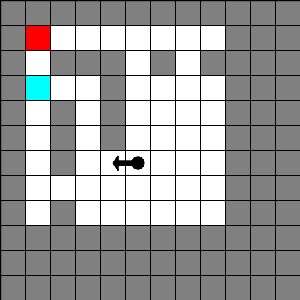}
\caption*{t = 75}
\end{subfigure}
&
\begin{subfigure}{0.08\linewidth}
\includegraphics[width=1\linewidth]{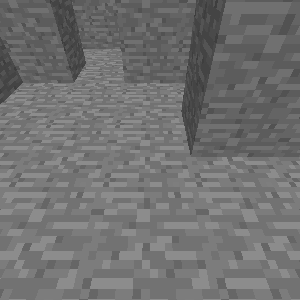}
\includegraphics[width=1\linewidth]{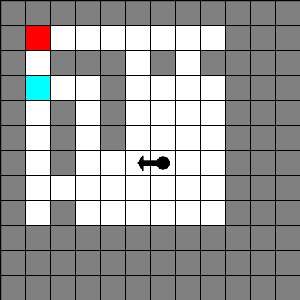}
\caption*{t = 76}
\end{subfigure}
&
\begin{subfigure}{0.08\linewidth}
\includegraphics[width=1\linewidth]{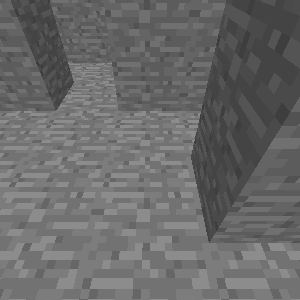}
\includegraphics[width=1\linewidth]{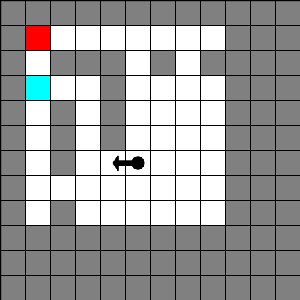}
\caption*{t = 77}
\end{subfigure}
&
\begin{subfigure}{0.08\linewidth}
\includegraphics[width=1\linewidth]{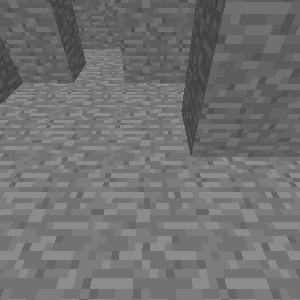}
\includegraphics[width=1\linewidth]{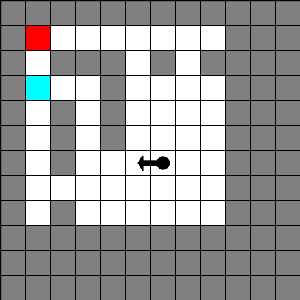}
\caption*{t = 78}
\end{subfigure}
&
\begin{subfigure}{0.08\linewidth}
\includegraphics[width=1\linewidth]{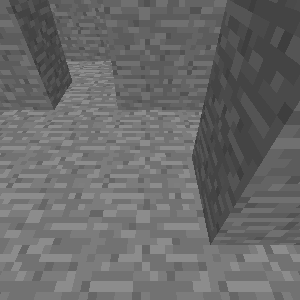}
\includegraphics[width=1\linewidth]{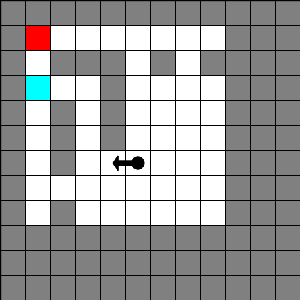}
\caption*{t = 79}
\end{subfigure}
&
\begin{subfigure}{0.08\linewidth}
\includegraphics[width=1\linewidth]{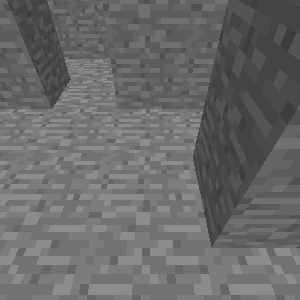}
\includegraphics[width=1\linewidth]{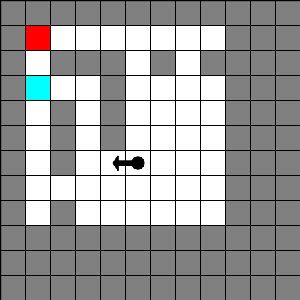}
\caption*{t = 80}
\end{subfigure}
&
\begin{subfigure}{0.08\linewidth}
\includegraphics[width=1\linewidth]{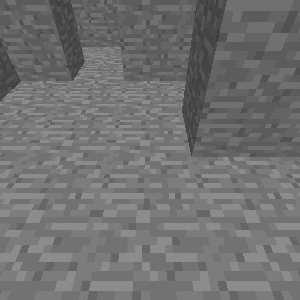}
\includegraphics[width=1\linewidth]{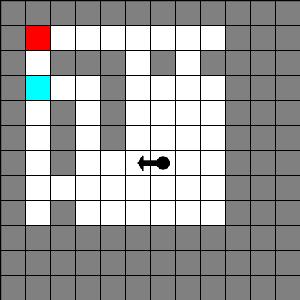}
\caption*{t = 81}
\end{subfigure}
&
\begin{subfigure}{0.08\linewidth}
\includegraphics[width=1\linewidth]{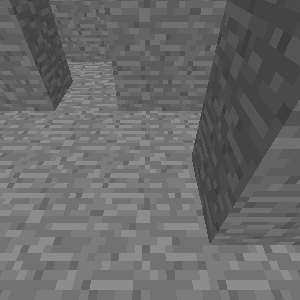}
\includegraphics[width=1\linewidth]{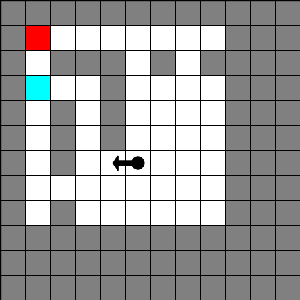}
\caption*{t = 82}
\end{subfigure}
&
\begin{subfigure}{0.08\linewidth}
\includegraphics[width=1\linewidth]{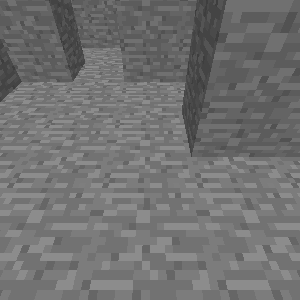}
\includegraphics[width=1\linewidth]{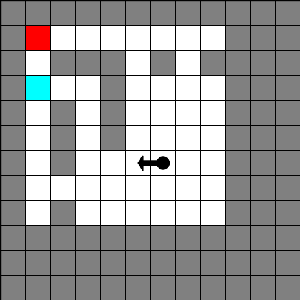}
\caption*{t = 83}
\end{subfigure}
&
\begin{subfigure}{0.08\linewidth}
\includegraphics[width=1\linewidth]{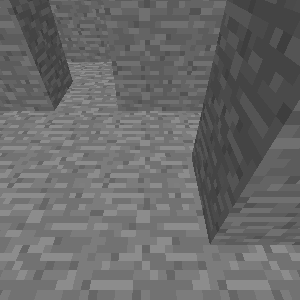}
\includegraphics[width=1\linewidth]{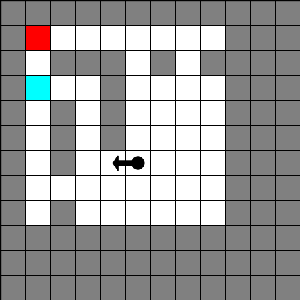}
\caption*{t = 84}
\end{subfigure}
\\
\\
\begin{subfigure}{0.08\linewidth}
\includegraphics[width=1\linewidth]{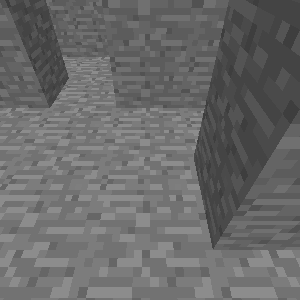}
\includegraphics[width=1\linewidth]{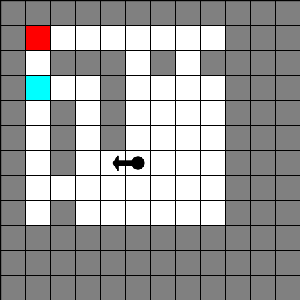}
\caption*{t = 85}
\end{subfigure}
&
\begin{subfigure}{0.08\linewidth}
\includegraphics[width=1\linewidth]{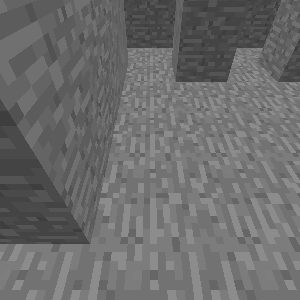}
\includegraphics[width=1\linewidth]{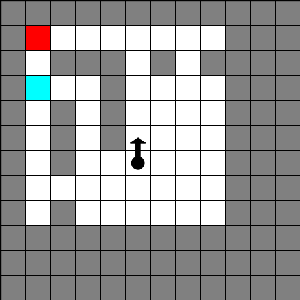}
\caption*{t = 86}
\end{subfigure}
&
\begin{subfigure}{0.08\linewidth}
\includegraphics[width=1\linewidth]{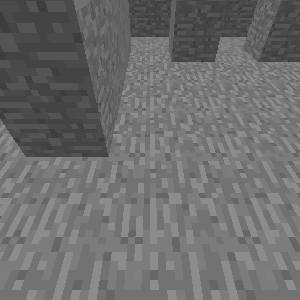}
\includegraphics[width=1\linewidth]{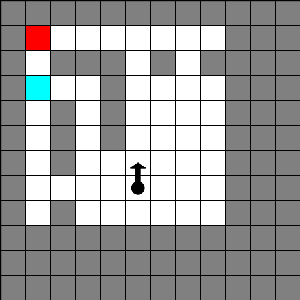}
\caption*{t = 87}
\end{subfigure}
&
\begin{subfigure}{0.08\linewidth}
\includegraphics[width=1\linewidth]{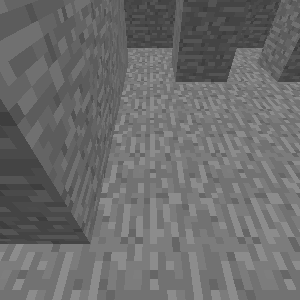}
\includegraphics[width=1\linewidth]{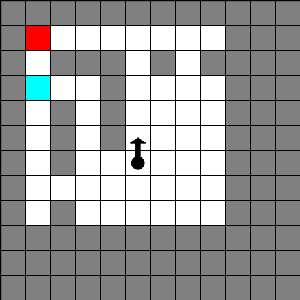}
\caption*{t = 88}
\end{subfigure}
&
\begin{subfigure}{0.08\linewidth}
\includegraphics[width=1\linewidth]{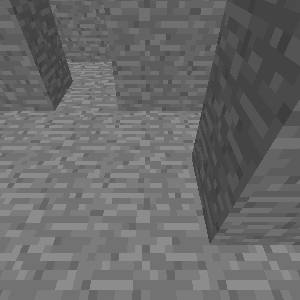}
\includegraphics[width=1\linewidth]{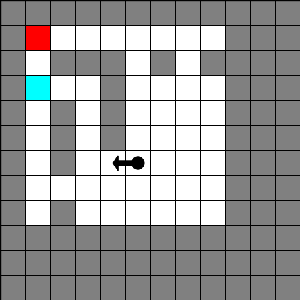}
\caption*{t = 89}
\end{subfigure}
&
\begin{subfigure}{0.08\linewidth}
\includegraphics[width=1\linewidth]{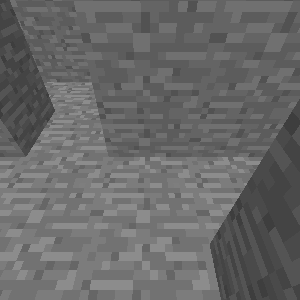}
\includegraphics[width=1\linewidth]{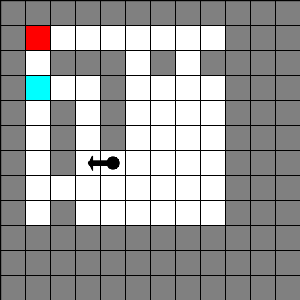}
\caption*{t = 90}
\end{subfigure}
&
\begin{subfigure}{0.08\linewidth}
\includegraphics[width=1\linewidth]{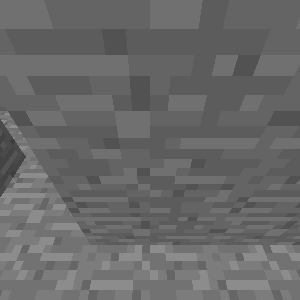}
\includegraphics[width=1\linewidth]{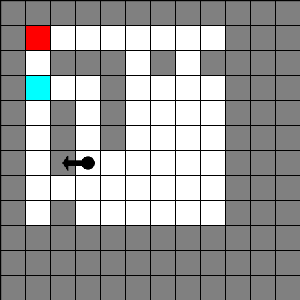}
\caption*{t = 91}
\end{subfigure}
&
\begin{subfigure}{0.08\linewidth}
\includegraphics[width=1\linewidth]{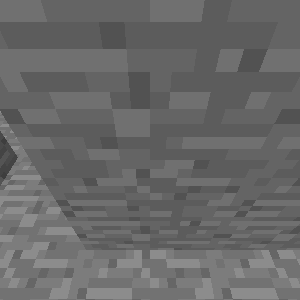}
\includegraphics[width=1\linewidth]{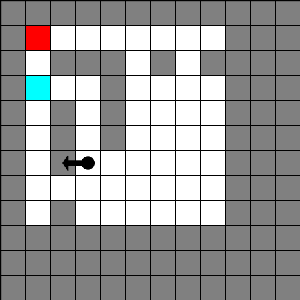}
\caption*{t = 92}
\end{subfigure}
&
\begin{subfigure}{0.08\linewidth}
\includegraphics[width=1\linewidth]{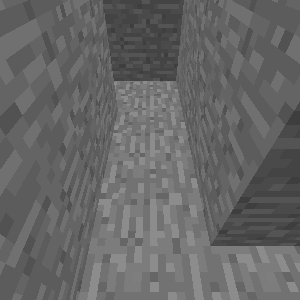}
\includegraphics[width=1\linewidth]{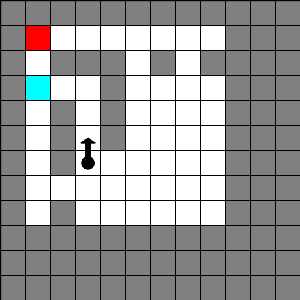}
\caption*{t = 93}
\end{subfigure}
&
\begin{subfigure}{0.08\linewidth}
\includegraphics[width=1\linewidth]{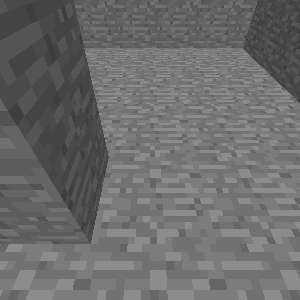}
\includegraphics[width=1\linewidth]{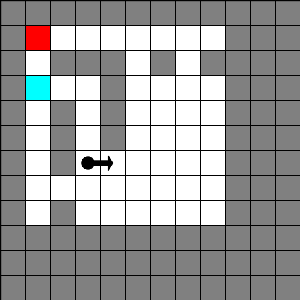}
\caption*{t = 94}
\end{subfigure}
&
\begin{subfigure}{0.08\linewidth}
\includegraphics[width=1\linewidth]{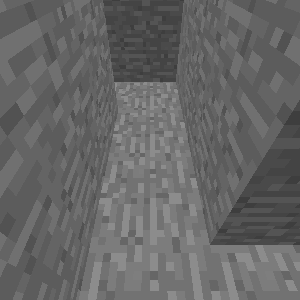}
\includegraphics[width=1\linewidth]{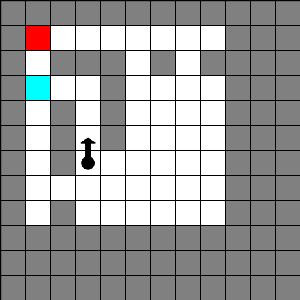}
\caption*{t = 95}
\end{subfigure}
&
\begin{subfigure}{0.08\linewidth}
\includegraphics[width=1\linewidth]{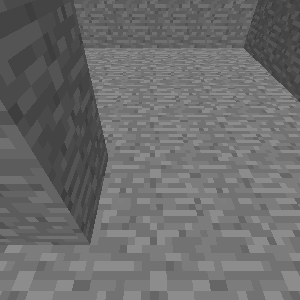}
\includegraphics[width=1\linewidth]{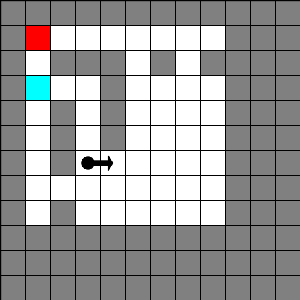}
\caption*{t = 96}
\end{subfigure}
\\
\\
\begin{subfigure}{0.08\linewidth}
\includegraphics[width=1\linewidth]{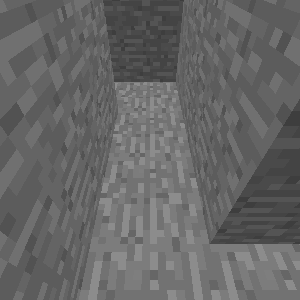}
\includegraphics[width=1\linewidth]{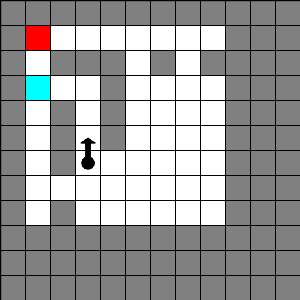}
\caption*{t = 97}
\end{subfigure}
&
\begin{subfigure}{0.08\linewidth}
\includegraphics[width=1\linewidth]{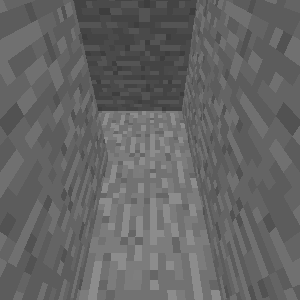}
\includegraphics[width=1\linewidth]{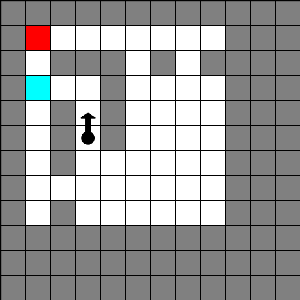}
\caption*{t = 98}
\end{subfigure}
&
\begin{subfigure}{0.08\linewidth}
\includegraphics[width=1\linewidth]{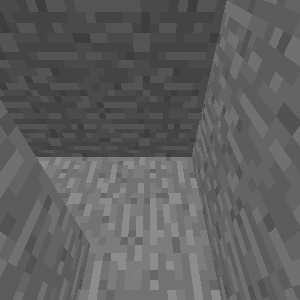}
\includegraphics[width=1\linewidth]{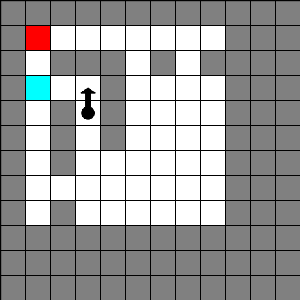}
\caption*{t = 99}
\end{subfigure}
&
\begin{subfigure}{0.08\linewidth}
\includegraphics[width=1\linewidth]{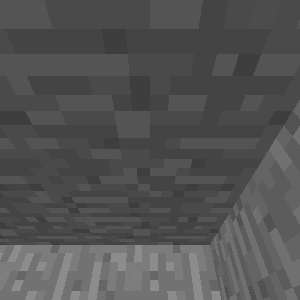}
\includegraphics[width=1\linewidth]{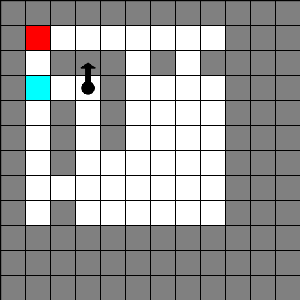}
\caption*{t = 100}
\end{subfigure}
\end{tabular}
\caption{FRMQN's play on an unseen and larger random maze with Sequential Goals task. At t=25, the agent finds both the red and blue blocks. However, visiting the red block (the first goal in the task) by following the corridor found would lead to visiting the blue block, and result in a negative reward. Thus, the agent attempts to search for another route to reach the red goal (t=49-100). But, the agent fails to find another route within the time limit (t=100). }
\label{seq_l_fail}
\end{figure*}
\begin{figure*}
    \small
    \setlength{\tabcolsep}{1pt}
    \def\arraystretch{1}
    \begin{tabular}{llllllllllll}
    \begin{subfigure}{0.08\linewidth}
	    \includegraphics[width=1\linewidth]{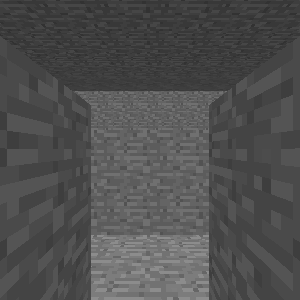} 
   		\includegraphics[width=1\linewidth]{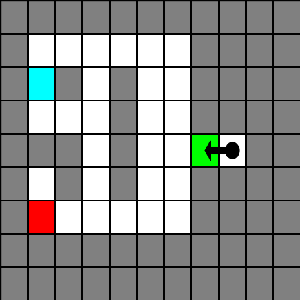} 
   		\caption*{t = 1}
	\end{subfigure} 
	& 
    \begin{subfigure}{0.08\linewidth}
	    \includegraphics[width=1\linewidth]{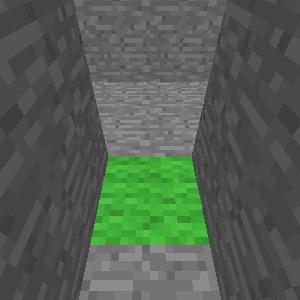} 
   		\includegraphics[width=1\linewidth]{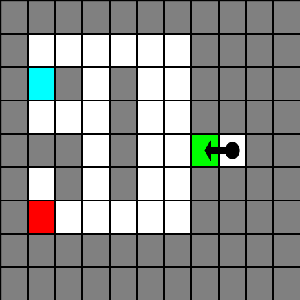} 
   		\caption*{t = 2}
	\end{subfigure} 
	& 
    \begin{subfigure}{0.08\linewidth}
	    \includegraphics[width=1\linewidth]{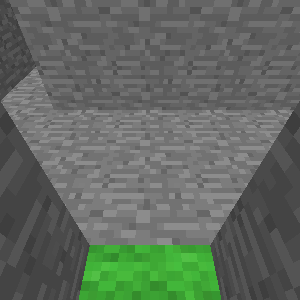} 
   		\includegraphics[width=1\linewidth]{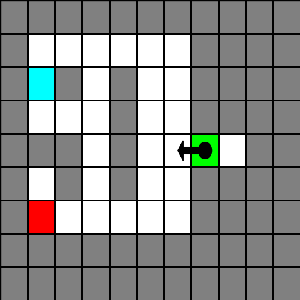} 
   		\caption*{t = 3}
	\end{subfigure} 
	& 
    \begin{subfigure}{0.08\linewidth}
	    \includegraphics[width=1\linewidth]{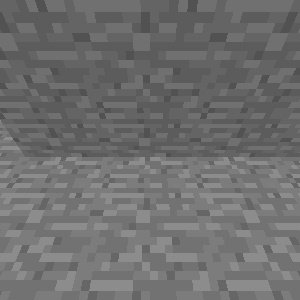} 
   		\includegraphics[width=1\linewidth]{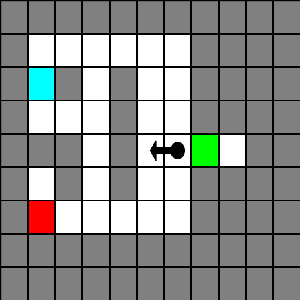} 
   		\caption*{t = 4}
	\end{subfigure}
	& 
    \begin{subfigure}{0.08\linewidth}
	    \includegraphics[width=1\linewidth]{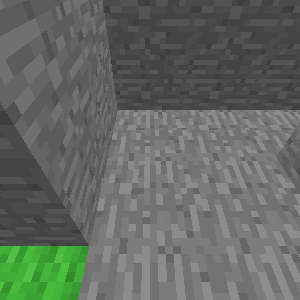} 
   		\includegraphics[width=1\linewidth]{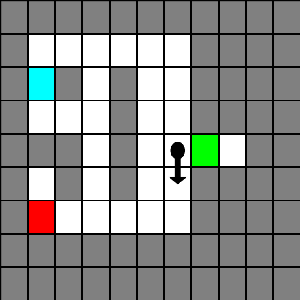} 
   		\caption*{t = 5}
	\end{subfigure}
	& 
    \begin{subfigure}{0.08\linewidth}
	    \includegraphics[width=1\linewidth]{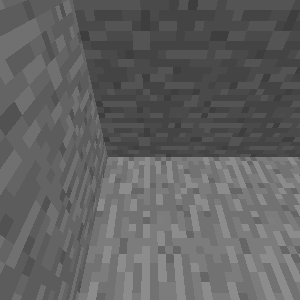} 
   		\includegraphics[width=1\linewidth]{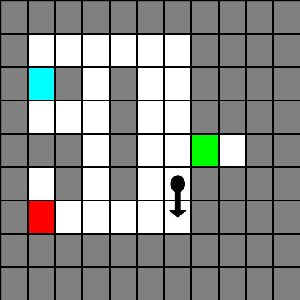} 
   		\caption*{t = 6}
	\end{subfigure} 
	& 
    \begin{subfigure}{0.08\linewidth}
	    \includegraphics[width=1\linewidth]{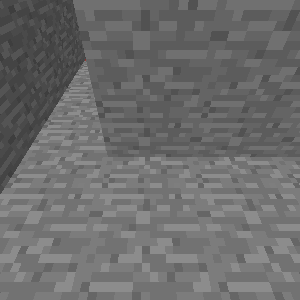} 
   		\includegraphics[width=1\linewidth]{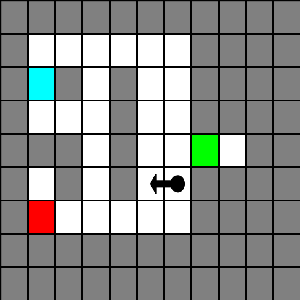} 
   		\caption*{t = 7}
	\end{subfigure} 
	& 
    \begin{subfigure}{0.08\linewidth}
	    \includegraphics[width=1\linewidth]{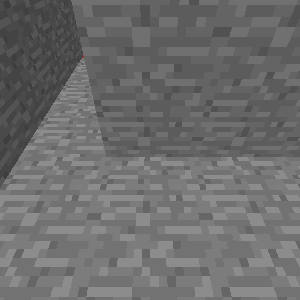} 
   		\includegraphics[width=1\linewidth]{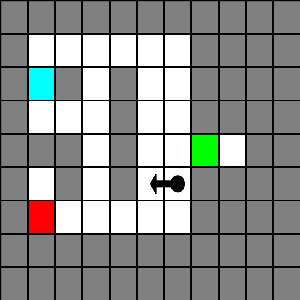} 
   		\caption*{t = 8}
	\end{subfigure} 
	& 
    \begin{subfigure}{0.08\linewidth}
	    \includegraphics[width=1\linewidth]{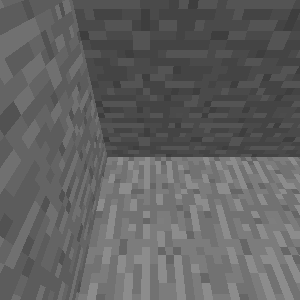} 
   		\includegraphics[width=1\linewidth]{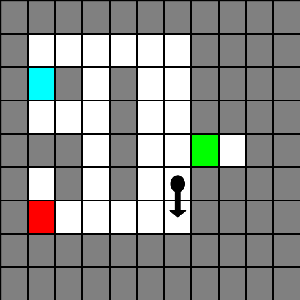} 
   		\caption*{t = 9}
	\end{subfigure}
	& 
    \begin{subfigure}{0.08\linewidth}
	    \includegraphics[width=1\linewidth]{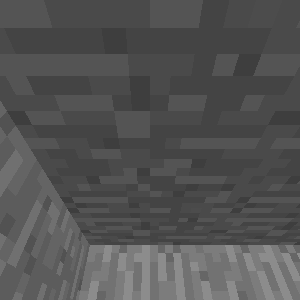} 
   		\includegraphics[width=1\linewidth]{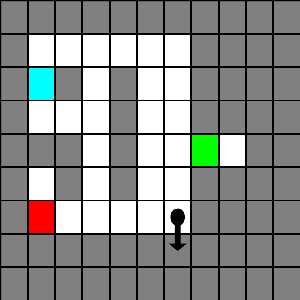} 
   		\caption*{t = 10}
	\end{subfigure}
	& 
    \begin{subfigure}{0.08\linewidth}
	    \includegraphics[width=1\linewidth]{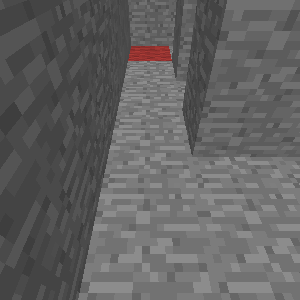} 
   		\includegraphics[width=1\linewidth]{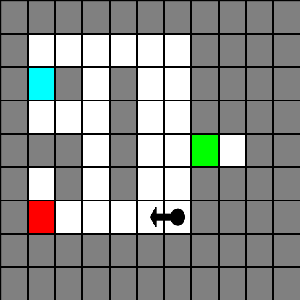} 
   		\caption*{t = 11}
	\end{subfigure} 
	&
    \begin{subfigure}{0.08\linewidth}
	    \includegraphics[width=1\linewidth]{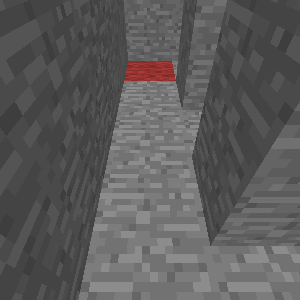} 
   		\includegraphics[width=1\linewidth]{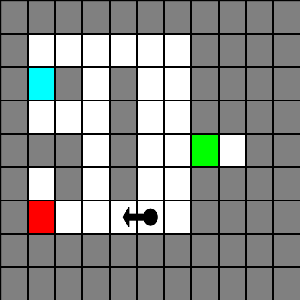} 
   		\caption*{t = 12}
	\end{subfigure} 
	\\
	\\
    \begin{subfigure}{0.08\linewidth}
	    \includegraphics[width=1\linewidth]{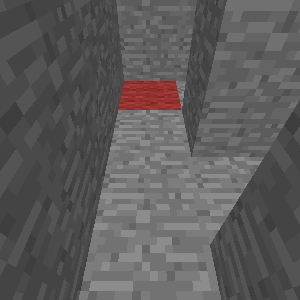} 
   		\includegraphics[width=1\linewidth]{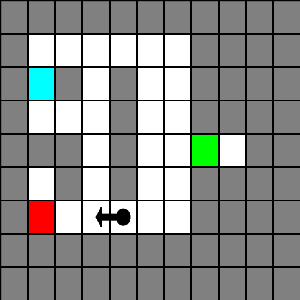} 
   		\caption*{t = 13}
	\end{subfigure} 
	& 
    \begin{subfigure}{0.08\linewidth}
	    \includegraphics[width=1\linewidth]{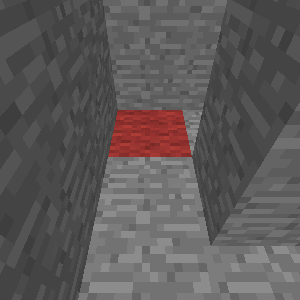} 
   		\includegraphics[width=1\linewidth]{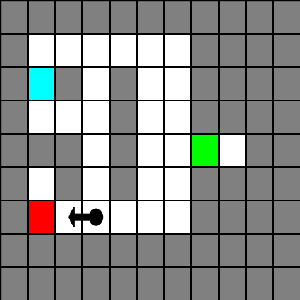} 
   		\caption*{t = 14}
	\end{subfigure} 
	& 
    \begin{subfigure}{0.08\linewidth}
	    \includegraphics[width=1\linewidth]{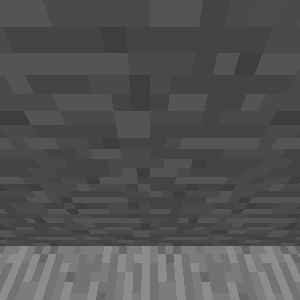} 
   		\includegraphics[width=1\linewidth]{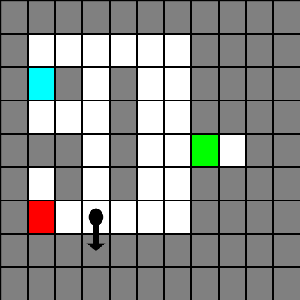} 
   		\caption*{t = 15}
	\end{subfigure} 
	& 
    \begin{subfigure}{0.08\linewidth}
	    \includegraphics[width=1\linewidth]{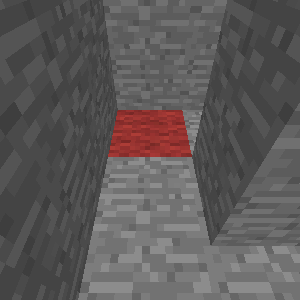} 
   		\includegraphics[width=1\linewidth]{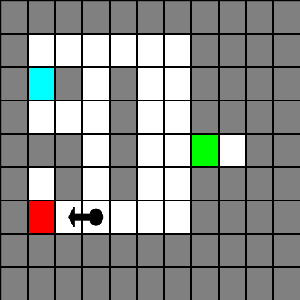} 
   		\caption*{t = 16}
	\end{subfigure}
	& 
    \begin{subfigure}{0.08\linewidth}
	    \includegraphics[width=1\linewidth]{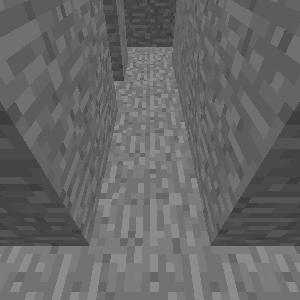} 
   		\includegraphics[width=1\linewidth]{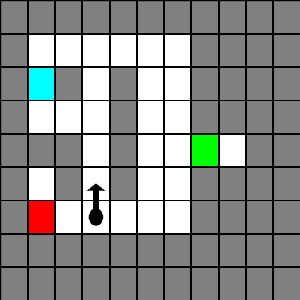} 
   		\caption*{t = 17}
	\end{subfigure}
	& 
    \begin{subfigure}{0.08\linewidth}
	    \includegraphics[width=1\linewidth]{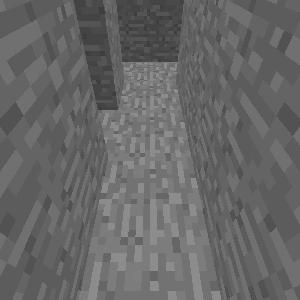} 
   		\includegraphics[width=1\linewidth]{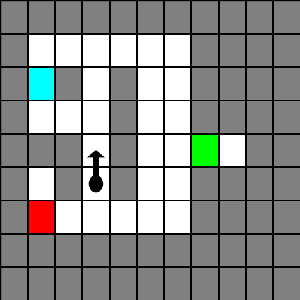} 
   		\caption*{t = 18}
	\end{subfigure} 
	& 
    \begin{subfigure}{0.08\linewidth}
	    \includegraphics[width=1\linewidth]{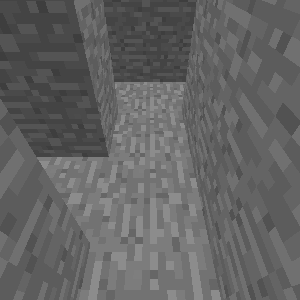} 
   		\includegraphics[width=1\linewidth]{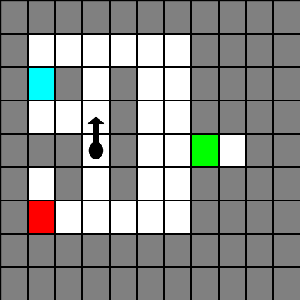} 
   		\caption*{t = 19}
	\end{subfigure} 
	& 
    \begin{subfigure}{0.08\linewidth}
	    \includegraphics[width=1\linewidth]{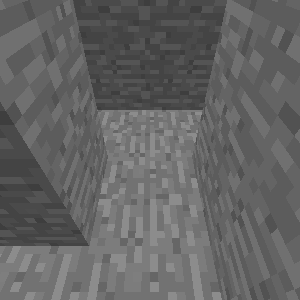} 
   		\includegraphics[width=1\linewidth]{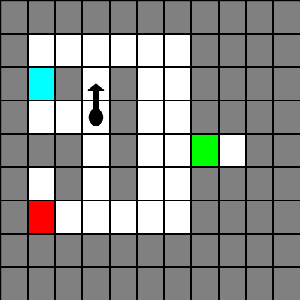} 
   		\caption*{t = 20}
	\end{subfigure} 
	& 
    \begin{subfigure}{0.08\linewidth}
	    \includegraphics[width=1\linewidth]{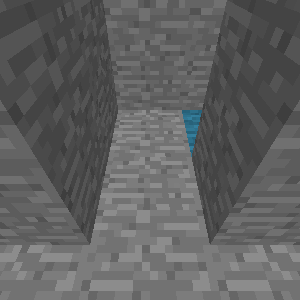} 
   		\includegraphics[width=1\linewidth]{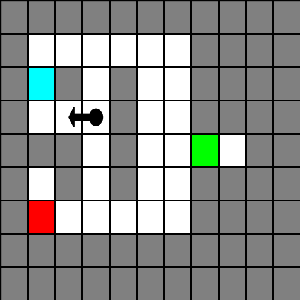} 
   		\caption*{t = 21}
	\end{subfigure}
	& 
    \begin{subfigure}{0.08\linewidth}
	    \includegraphics[width=1\linewidth]{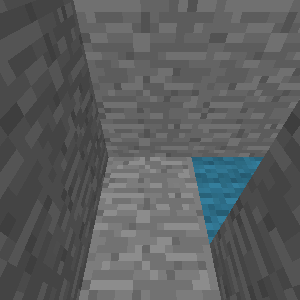} 
   		\includegraphics[width=1\linewidth]{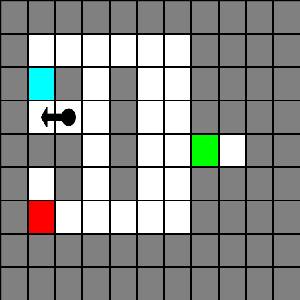} 
   		\caption*{t = 22}
	\end{subfigure}
	& 
    \begin{subfigure}{0.08\linewidth}
	    \includegraphics[width=1\linewidth]{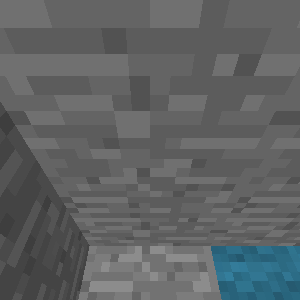} 
   		\includegraphics[width=1\linewidth]{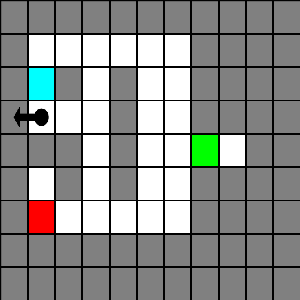} 
   		\caption*{t = 23}
	\end{subfigure} 
	&
    \begin{subfigure}{0.08\linewidth}
	    \includegraphics[width=1\linewidth]{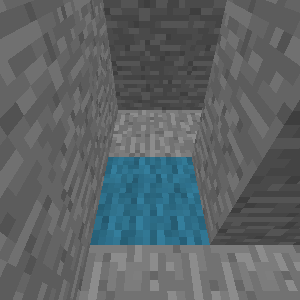} 
   		\includegraphics[width=1\linewidth]{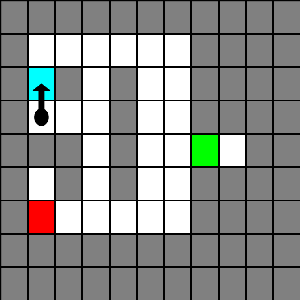}
   		\caption*{t = 24}
	\end{subfigure} 
	\\
	\\
	\begin{subfigure}{0.08\linewidth}
	    \includegraphics[width=1\linewidth]{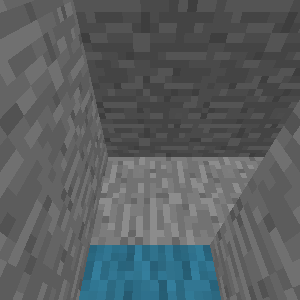} 
   		\includegraphics[width=1\linewidth]{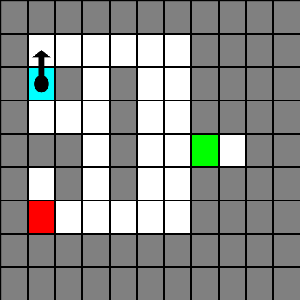} 
   		\caption*{t = 25}
	\end{subfigure} 
	\end{tabular}
    \caption{FRMQN's play in a random maze with Single Goal with Indicator task. Upon seeing that the indicator is green in color (t=2), the agent proceeds to explore the map. During its search, it comes across a corridor with a red block (t=11). This is where memory helps. The agent avoids the red block (having observed that the indicator is green). From this we can infer that the agent utilizes its memory appropriately. Later, it successfully completes the task by finding and visiting the blue block (t=22-25). } 
	\label{play-single-i}
\end{figure*}

\begin{figure*}
    \small
    \setlength{\tabcolsep}{1pt}
    \def\arraystretch{1}
    \begin{tabular}{llllllllllll}
    \begin{subfigure}{0.08\linewidth}
	    \includegraphics[width=1\linewidth]{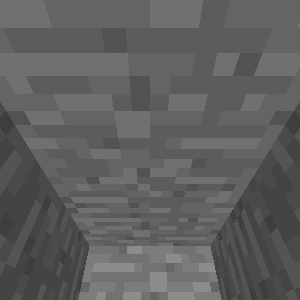} 
   		\includegraphics[width=1\linewidth]{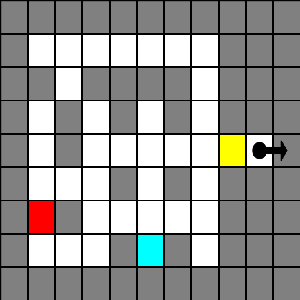} 
   		\caption*{t = 1}
	\end{subfigure} 
	& 
    \begin{subfigure}{0.08\linewidth}
	    \includegraphics[width=1\linewidth]{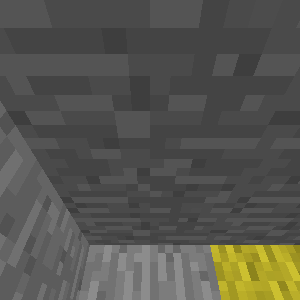} 
   		\includegraphics[width=1\linewidth]{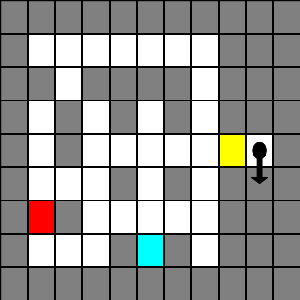} 
   		\caption*{t = 2}
	\end{subfigure} 
	& 
    \begin{subfigure}{0.08\linewidth}
	    \includegraphics[width=1\linewidth]{figures/video_single_i_2/0002.png} 
   		\includegraphics[width=1\linewidth]{figures/video_single_i_2/0002_t.png}
   		\caption*{t = 3}
	\end{subfigure} 
	& 
    \begin{subfigure}{0.08\linewidth}
	    \includegraphics[width=1\linewidth]{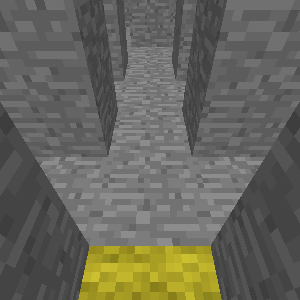} 
   		\includegraphics[width=1\linewidth]{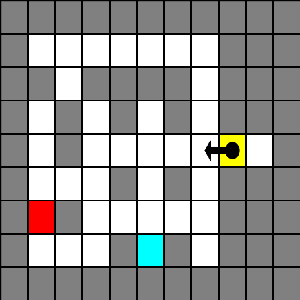} 
   		\caption*{t = 4}
	\end{subfigure}
	& 
    \begin{subfigure}{0.08\linewidth}
	    \includegraphics[width=1\linewidth]{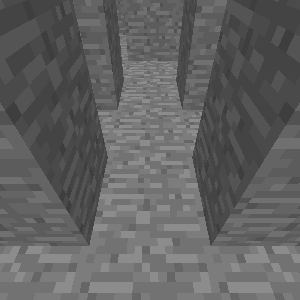} 
   		\includegraphics[width=1\linewidth]{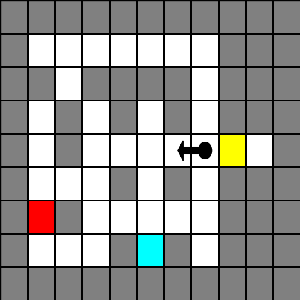} 
   		\caption*{t = 5}
	\end{subfigure}
	& 
    \begin{subfigure}{0.08\linewidth}
	    \includegraphics[width=1\linewidth]{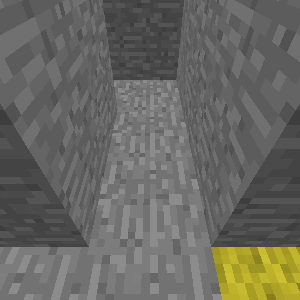} 
   		\includegraphics[width=1\linewidth]{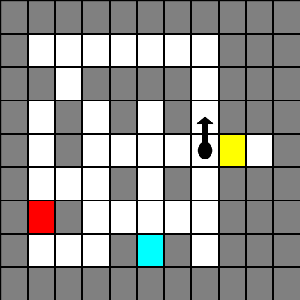} 
   		\caption*{t = 6}
	\end{subfigure} 
	& 
    \begin{subfigure}{0.08\linewidth}
	    \includegraphics[width=1\linewidth]{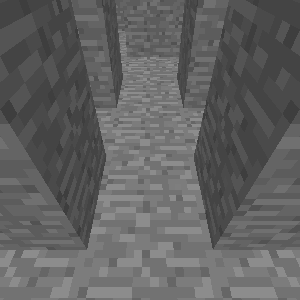} 
   		\includegraphics[width=1\linewidth]{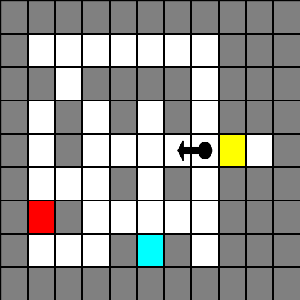} 
   		\caption*{t = 7}
	\end{subfigure} 
	& 
    \begin{subfigure}{0.08\linewidth}
	    \includegraphics[width=1\linewidth]{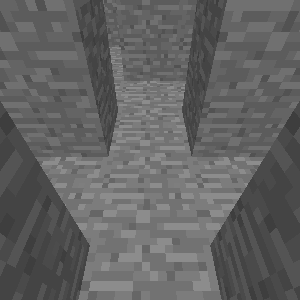} 
   		\includegraphics[width=1\linewidth]{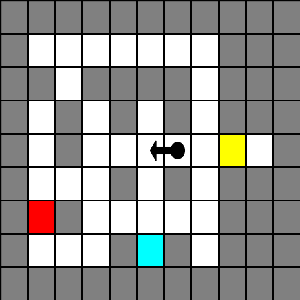} 
   		\caption*{t = 8}
	\end{subfigure} 
	& 
    \begin{subfigure}{0.08\linewidth}
	    \includegraphics[width=1\linewidth]{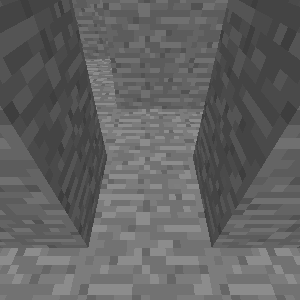} 
   		\includegraphics[width=1\linewidth]{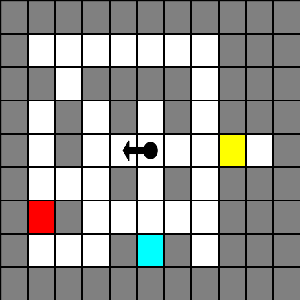} 
   		\caption*{t = 9}
	\end{subfigure}
	& 
    \begin{subfigure}{0.08\linewidth}
	    \includegraphics[width=1\linewidth]{figures/video_single_i_2/0009.png} 
   		\includegraphics[width=1\linewidth]{figures/video_single_i_2/0009_t.png}
   		\caption*{t = 10}
	\end{subfigure}
	& 
    \begin{subfigure}{0.08\linewidth}
	    \includegraphics[width=1\linewidth]{figures/video_single_i_2/0010.png} 
   		\includegraphics[width=1\linewidth]{figures/video_single_i_2/0010_t.png} 
   		\caption*{t = 11}
	\end{subfigure} 
	&
    \begin{subfigure}{0.08\linewidth}
	    \includegraphics[width=1\linewidth]{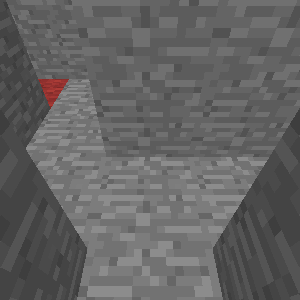} 
   		\includegraphics[width=1\linewidth]{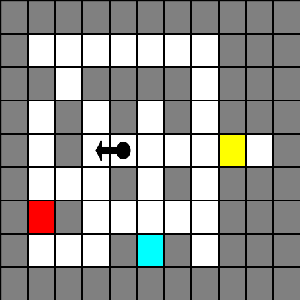} 
   		\caption*{t = 12}
	\end{subfigure} 
	\\
	\\
    \begin{subfigure}{0.08\linewidth}
	    \includegraphics[width=1\linewidth]{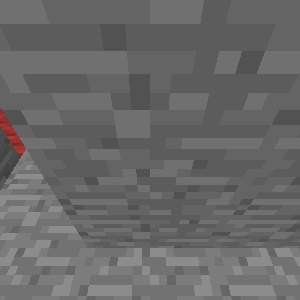} 
   		\includegraphics[width=1\linewidth]{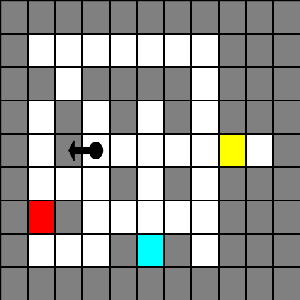} 
   		\caption*{t = 13}
	\end{subfigure} 
	& 
    \begin{subfigure}{0.08\linewidth}
	    \includegraphics[width=1\linewidth]{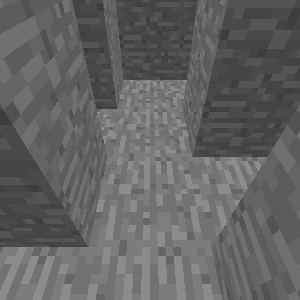} 
   		\includegraphics[width=1\linewidth]{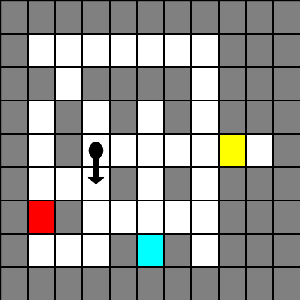} 
   		\caption*{t = 14}
	\end{subfigure} 
	& 
    \begin{subfigure}{0.08\linewidth}
	    \includegraphics[width=1\linewidth]{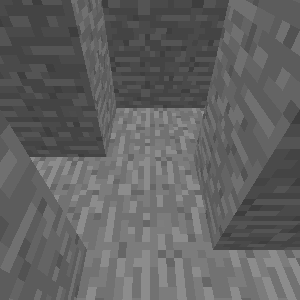} 
   		\includegraphics[width=1\linewidth]{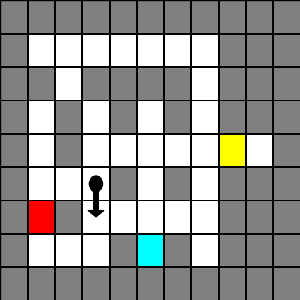} 
   		\caption*{t = 15}
	\end{subfigure} 
	& 
    \begin{subfigure}{0.08\linewidth}
	    \includegraphics[width=1\linewidth]{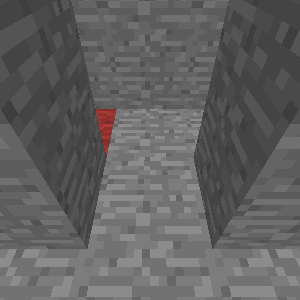} 
   		\includegraphics[width=1\linewidth]{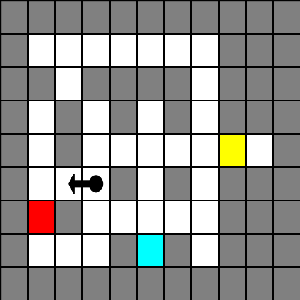} 
   		\caption*{t = 16}
	\end{subfigure}
	& 
    \begin{subfigure}{0.08\linewidth}
	    \includegraphics[width=1\linewidth]{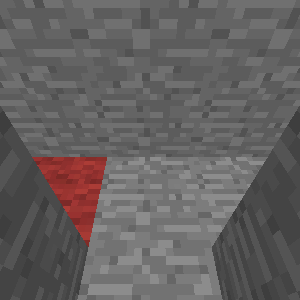} 
   		\includegraphics[width=1\linewidth]{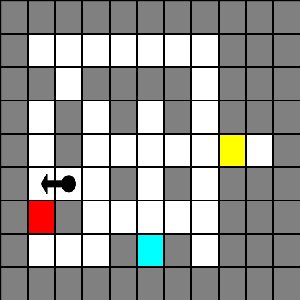} 
   		\caption*{t = 17}
	\end{subfigure}
	& 
    \begin{subfigure}{0.08\linewidth}
	    \includegraphics[width=1\linewidth]{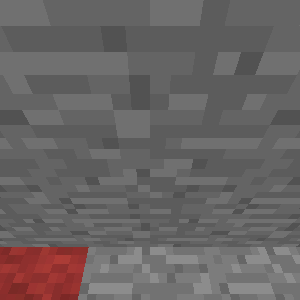} 
   		\includegraphics[width=1\linewidth]{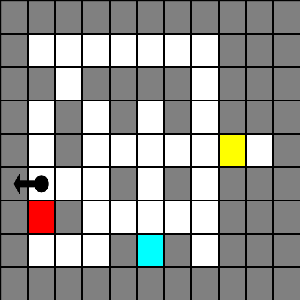} 
   		\caption*{t = 18}
	\end{subfigure} 
	& 
    \begin{subfigure}{0.08\linewidth}
	    \includegraphics[width=1\linewidth]{figures/video_single_i_2/0018.png} 
   		\includegraphics[width=1\linewidth]{figures/video_single_i_2/0018_t.png} 
   		\caption*{t = 20}
	\end{subfigure} 
	& 
    \begin{subfigure}{0.08\linewidth}
	    \includegraphics[width=1\linewidth]{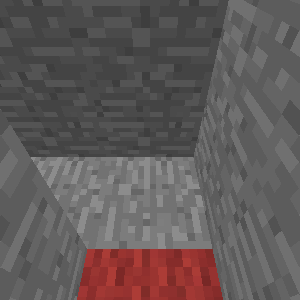} 
   		\includegraphics[width=1\linewidth]{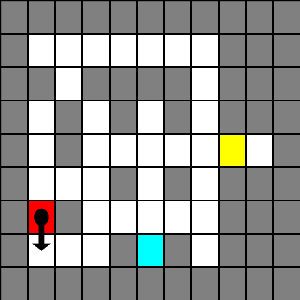} 
   		\caption*{t = 21}
	\end{subfigure} 
	\end{tabular}
    \caption{FRMQN's play in a random maze with Single Goal with Indicator task. The agent visits the red block correctly, given the yellow indicator. } 
	\label{play-single-i-2}
\end{figure*}

\begin{figure*}
\small
\setlength{\tabcolsep}{1pt}
\def\arraystretch{1}
\begin{tabular}{llllllllllll}
\begin{subfigure}{0.08\linewidth}
\includegraphics[width=1\linewidth]{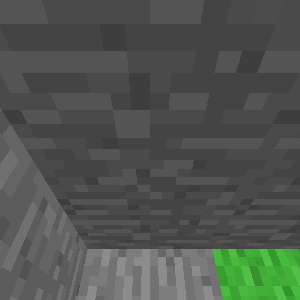}
\includegraphics[width=1\linewidth]{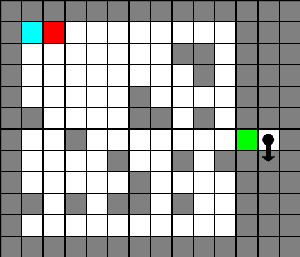}
\caption*{t = 1}
\end{subfigure}
&
\begin{subfigure}{0.08\linewidth}
\includegraphics[width=1\linewidth]{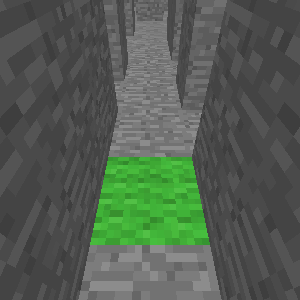}
\includegraphics[width=1\linewidth]{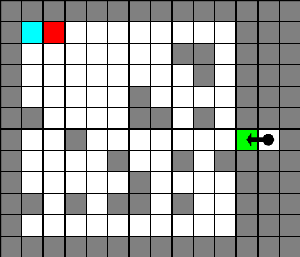}
\caption*{t = 2}
\end{subfigure}
&
\begin{subfigure}{0.08\linewidth}
\includegraphics[width=1\linewidth]{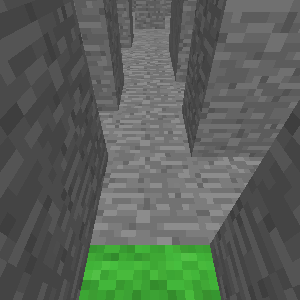}
\includegraphics[width=1\linewidth]{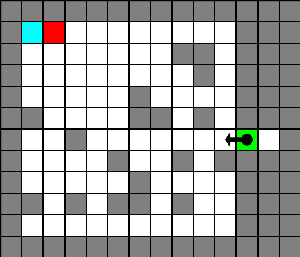}
\caption*{t = 3}
\end{subfigure}
&
\begin{subfigure}{0.08\linewidth}
\includegraphics[width=1\linewidth]{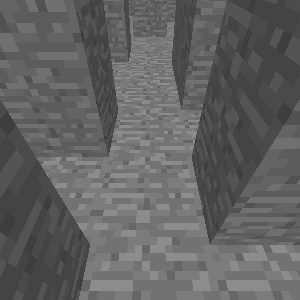}
\includegraphics[width=1\linewidth]{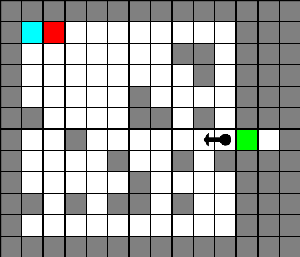}
\caption*{t = 4}
\end{subfigure}
&
\begin{subfigure}{0.08\linewidth}
\includegraphics[width=1\linewidth]{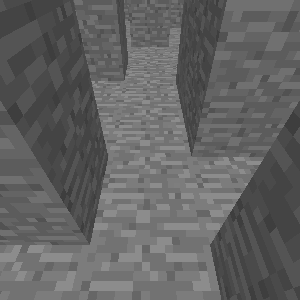}
\includegraphics[width=1\linewidth]{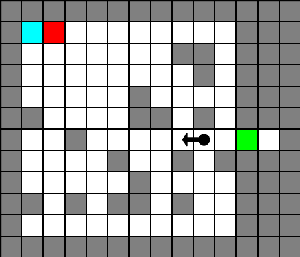}
\caption*{t = 5}
\end{subfigure}
&
\begin{subfigure}{0.08\linewidth}
\includegraphics[width=1\linewidth]{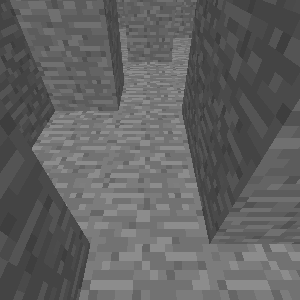}
\includegraphics[width=1\linewidth]{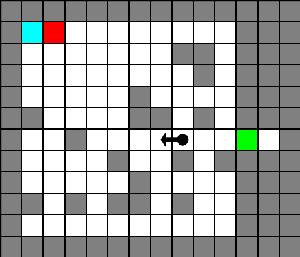}
\caption*{t = 6}
\end{subfigure}
&
\begin{subfigure}{0.08\linewidth}
\includegraphics[width=1\linewidth]{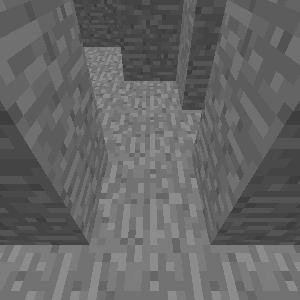}
\includegraphics[width=1\linewidth]{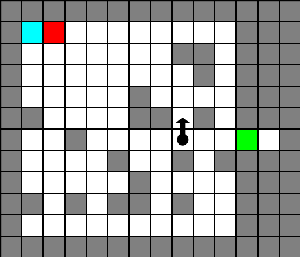}
\caption*{t = 7}
\end{subfigure}
&
\begin{subfigure}{0.08\linewidth}
\includegraphics[width=1\linewidth]{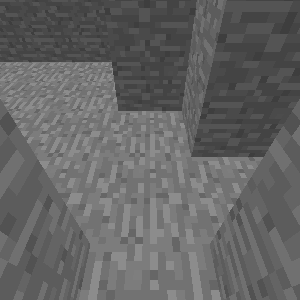}
\includegraphics[width=1\linewidth]{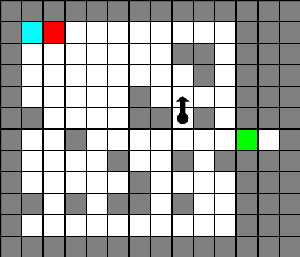}
\caption*{t = 8}
\end{subfigure}
&
\begin{subfigure}{0.08\linewidth}
\includegraphics[width=1\linewidth]{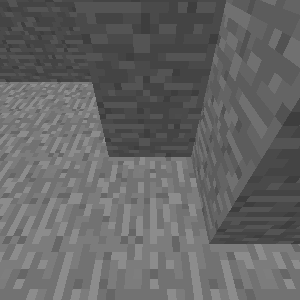}
\includegraphics[width=1\linewidth]{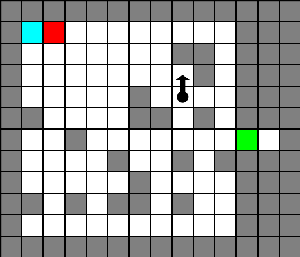}
\caption*{t = 9}
\end{subfigure}
&
\begin{subfigure}{0.08\linewidth}
\includegraphics[width=1\linewidth]{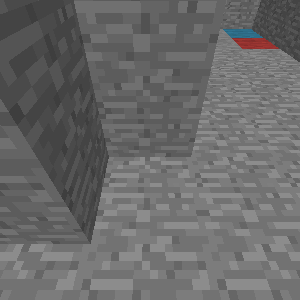}
\includegraphics[width=1\linewidth]{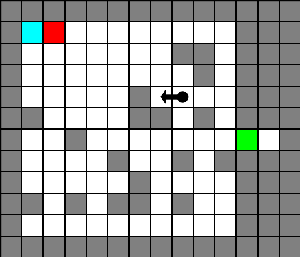}
\caption*{t = 10}
\end{subfigure}
&
\begin{subfigure}{0.08\linewidth}
\includegraphics[width=1\linewidth]{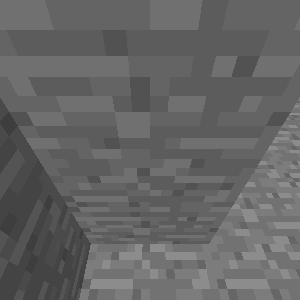}
\includegraphics[width=1\linewidth]{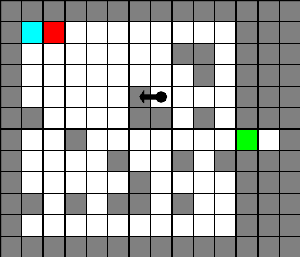}
\caption*{t = 11}
\end{subfigure}
&
\begin{subfigure}{0.08\linewidth}
\includegraphics[width=1\linewidth]{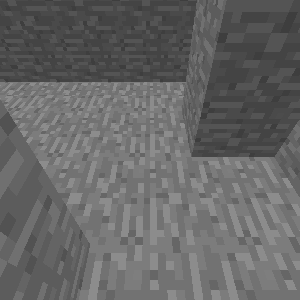}
\includegraphics[width=1\linewidth]{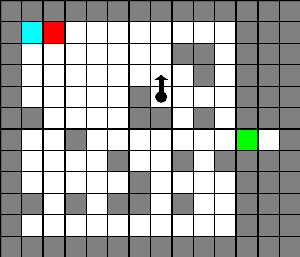}
\caption*{t = 12}
\end{subfigure}
\\
\\
\begin{subfigure}{0.08\linewidth}
\includegraphics[width=1\linewidth]{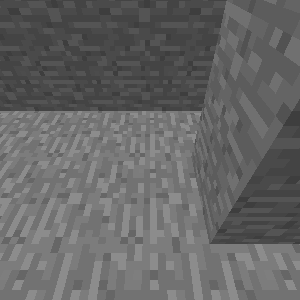}
\includegraphics[width=1\linewidth]{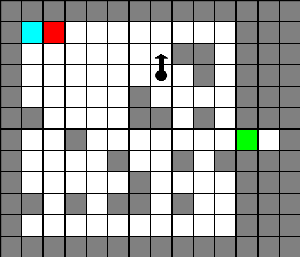}
\caption*{t = 13}
\end{subfigure}
&
\begin{subfigure}{0.08\linewidth}
\includegraphics[width=1\linewidth]{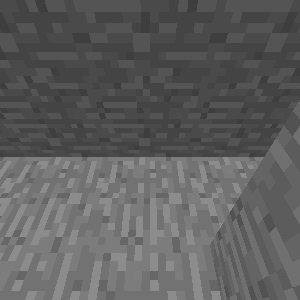}
\includegraphics[width=1\linewidth]{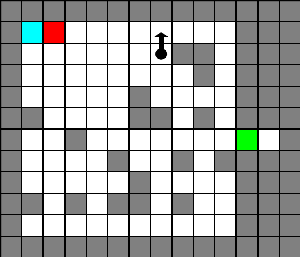}
\caption*{t = 14}
\end{subfigure}
&
\begin{subfigure}{0.08\linewidth}
\includegraphics[width=1\linewidth]{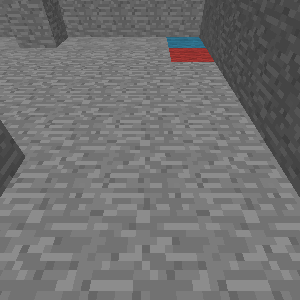}
\includegraphics[width=1\linewidth]{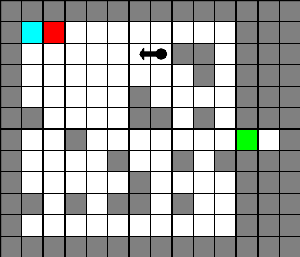}
\caption*{t = 15}
\end{subfigure}
&
\begin{subfigure}{0.08\linewidth}
\includegraphics[width=1\linewidth]{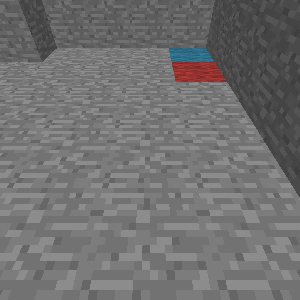}
\includegraphics[width=1\linewidth]{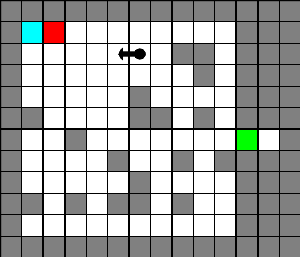}
\caption*{t = 16}
\end{subfigure}
&
\begin{subfigure}{0.08\linewidth}
\includegraphics[width=1\linewidth]{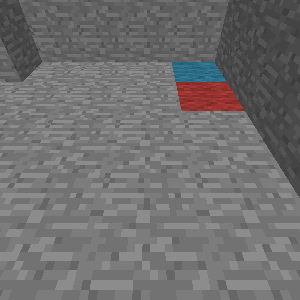}
\includegraphics[width=1\linewidth]{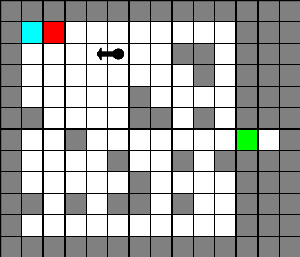}
\caption*{t = 17}
\end{subfigure}
&
\begin{subfigure}{0.08\linewidth}
\includegraphics[width=1\linewidth]{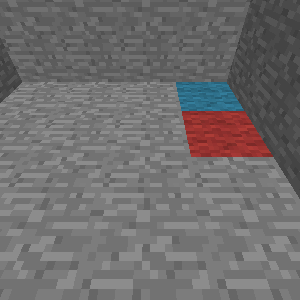}
\includegraphics[width=1\linewidth]{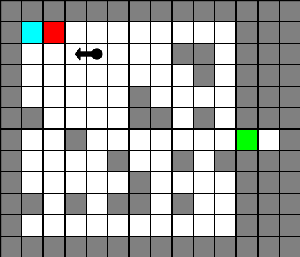}
\caption*{t = 18}
\end{subfigure}
&
\begin{subfigure}{0.08\linewidth}
\includegraphics[width=1\linewidth]{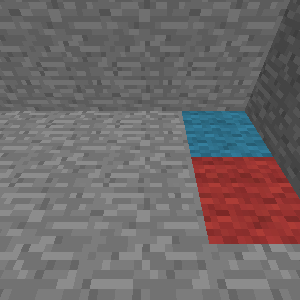}
\includegraphics[width=1\linewidth]{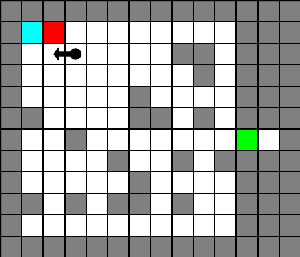}
\caption*{t = 19}
\end{subfigure}
&
\begin{subfigure}{0.08\linewidth}
\includegraphics[width=1\linewidth]{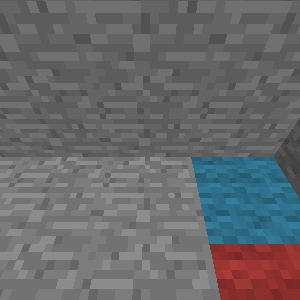}
\includegraphics[width=1\linewidth]{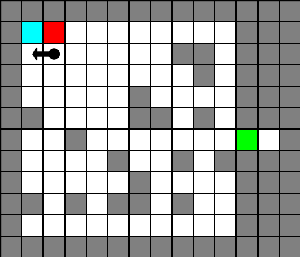}
\caption*{t = 20}
\end{subfigure}
&
\begin{subfigure}{0.08\linewidth}
\includegraphics[width=1\linewidth]{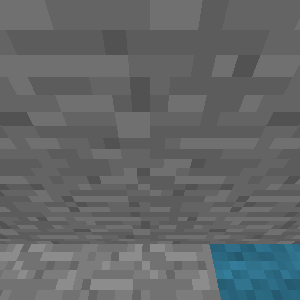}
\includegraphics[width=1\linewidth]{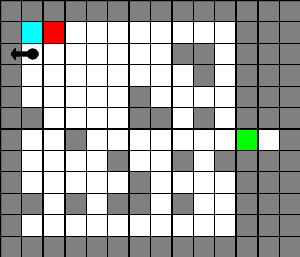}
\caption*{t = 21}
\end{subfigure}
&
\begin{subfigure}{0.08\linewidth}
\includegraphics[width=1\linewidth]{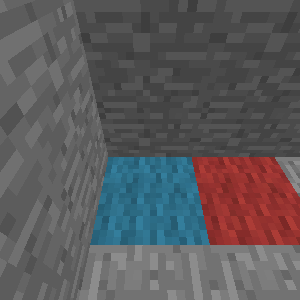}
\includegraphics[width=1\linewidth]{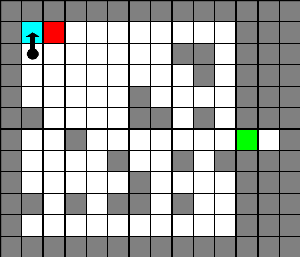}
\caption*{t = 22}
\end{subfigure}
&
\begin{subfigure}{0.08\linewidth}
\includegraphics[width=1\linewidth]{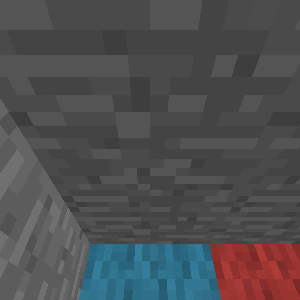}
\includegraphics[width=1\linewidth]{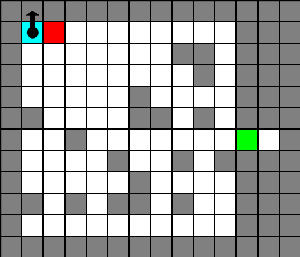}
\caption*{t = 23}
\end{subfigure}
\end{tabular}
\caption{FRMQN's play in an unseen and larger random maze with Single Goal with Indicator task. While the correct goal is far from the indicator, the agent is able to memorize the color of the indicator observed at the beginning (t=3) and visits the correct goal (t=23). }
\label{single_i_l_success}
\end{figure*}

\begin{figure*}
    \small
    \setlength{\tabcolsep}{1pt}
    \def\arraystretch{1}
    \begin{tabular}{llllllllllll}
    \begin{subfigure}{0.08\linewidth}
	    \includegraphics[width=1\linewidth]{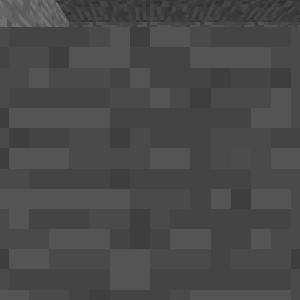} 
   		\includegraphics[width=1\linewidth]{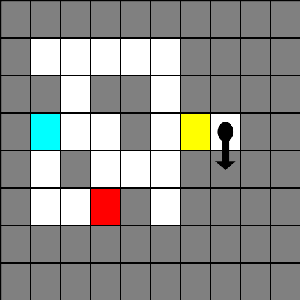} 
   		\caption*{t = 1}
	\end{subfigure} 
	& 
    \begin{subfigure}{0.08\linewidth}
	    \includegraphics[width=1\linewidth]{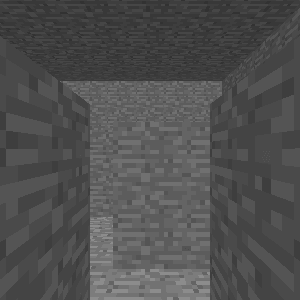} 
   		\includegraphics[width=1\linewidth]{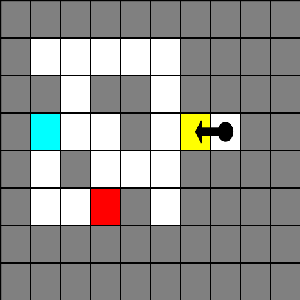} 
   		\caption*{t = 2}
	\end{subfigure} 
	& 
    \begin{subfigure}{0.08\linewidth}
	    \includegraphics[width=1\linewidth]{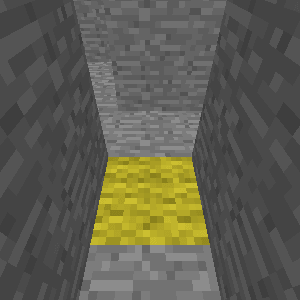} 
   		\includegraphics[width=1\linewidth]{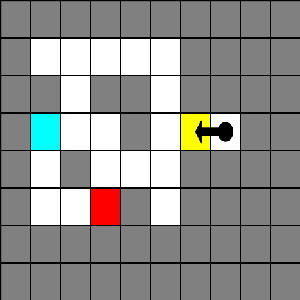} 
   		\caption*{t = 3}
	\end{subfigure} 
	& 
    \begin{subfigure}{0.08\linewidth}
	    \includegraphics[width=1\linewidth]{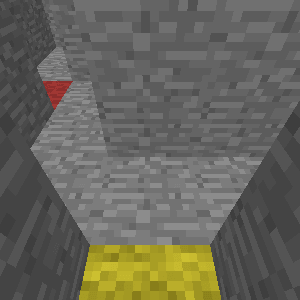} 
   		\includegraphics[width=1\linewidth]{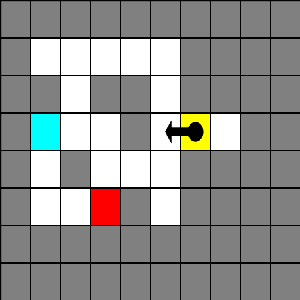} 
   		\caption*{t = 4}
	\end{subfigure}
	& 
    \begin{subfigure}{0.08\linewidth}
	    \includegraphics[width=1\linewidth]{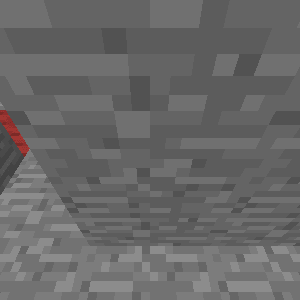} 
   		\includegraphics[width=1\linewidth]{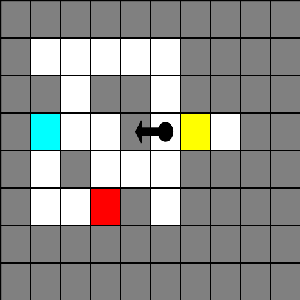} 
   		\caption*{t = 5}
	\end{subfigure}
	& 
    \begin{subfigure}{0.08\linewidth}
	    \includegraphics[width=1\linewidth]{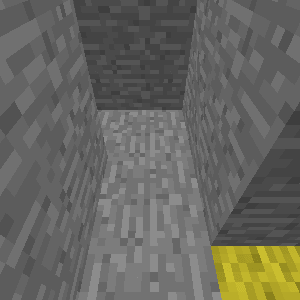} 
   		\includegraphics[width=1\linewidth]{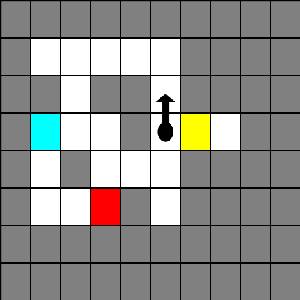} 
   		\caption*{t = 6}
	\end{subfigure} 
	& 
    \begin{subfigure}{0.08\linewidth}
	    \includegraphics[width=1\linewidth]{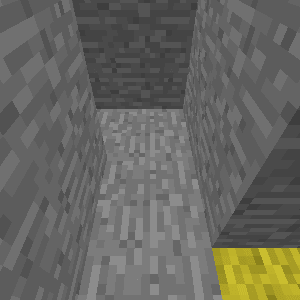} 
   		\includegraphics[width=1\linewidth]{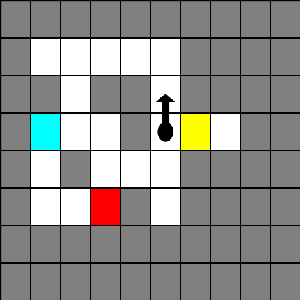} 
   		\caption*{t = 7}
	\end{subfigure} 
	& 
    \begin{subfigure}{0.08\linewidth}
	    \includegraphics[width=1\linewidth]{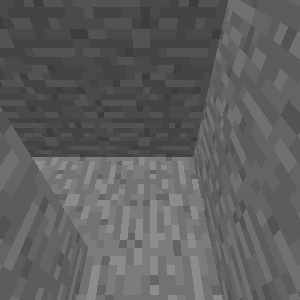} 
   		\includegraphics[width=1\linewidth]{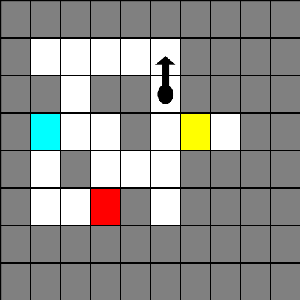} 
   		\caption*{t = 8}
	\end{subfigure} 
	& 
    \begin{subfigure}{0.08\linewidth}
	    \includegraphics[width=1\linewidth]{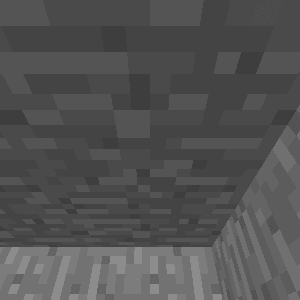} 
   		\includegraphics[width=1\linewidth]{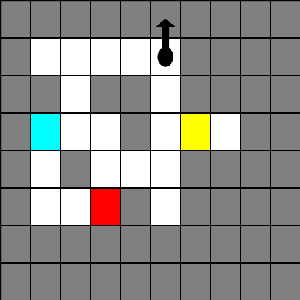} 
   		\caption*{t = 9}
	\end{subfigure}
	& 
    \begin{subfigure}{0.08\linewidth}
	    \includegraphics[width=1\linewidth]{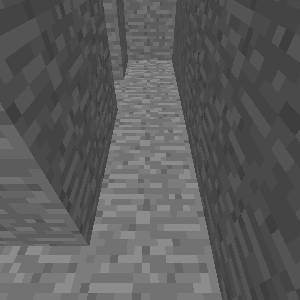} 
   		\includegraphics[width=1\linewidth]{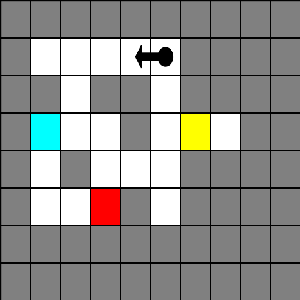} 
   		\caption*{t = 10}
	\end{subfigure}
	& 
    \begin{subfigure}{0.08\linewidth}
	    \includegraphics[width=1\linewidth]{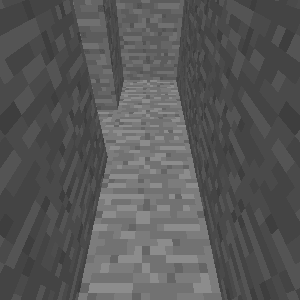} 
   		\includegraphics[width=1\linewidth]{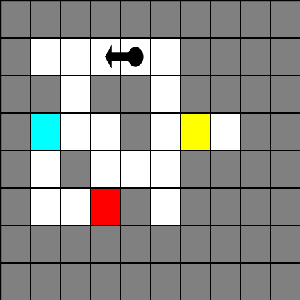} 
   		\caption*{t = 11}
	\end{subfigure} 
	&
    \begin{subfigure}{0.08\linewidth}
	    \includegraphics[width=1\linewidth]{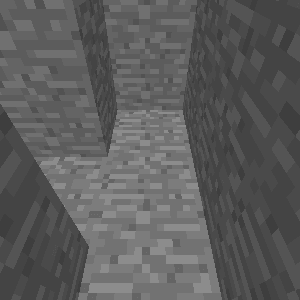} 
   		\includegraphics[width=1\linewidth]{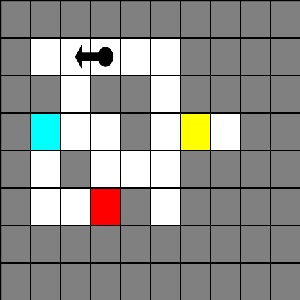} 
   		\caption*{t = 12}
	\end{subfigure} 
	\\
	\\
    \begin{subfigure}{0.08\linewidth}
	    \includegraphics[width=1\linewidth]{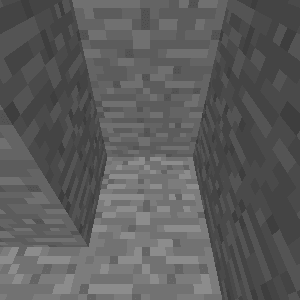} 
   		\includegraphics[width=1\linewidth]{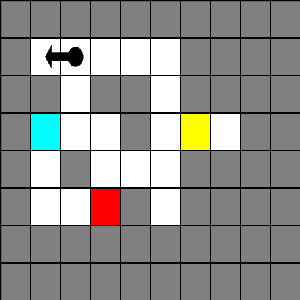} 
   		\caption*{t = 13}
	\end{subfigure} 
	& 
    \begin{subfigure}{0.08\linewidth}
	    \includegraphics[width=1\linewidth]{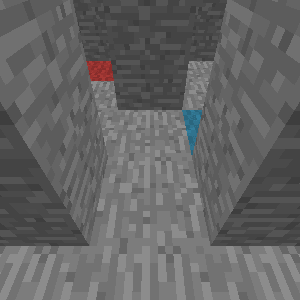} 
   		\includegraphics[width=1\linewidth]{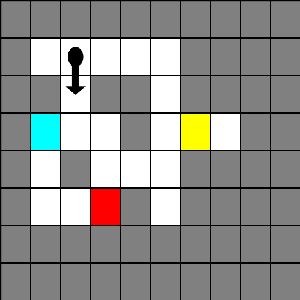} 
   		\caption*{t = 14}
	\end{subfigure} 
	& 
    \begin{subfigure}{0.08\linewidth}
	    \includegraphics[width=1\linewidth]{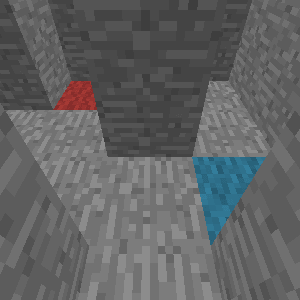} 
   		\includegraphics[width=1\linewidth]{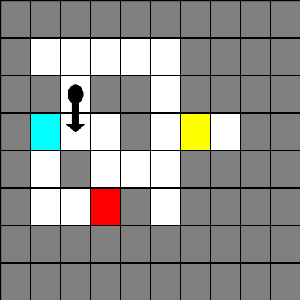} 
   		\caption*{t = 15}
	\end{subfigure} 
	& 
    \begin{subfigure}{0.08\linewidth}
	    \includegraphics[width=1\linewidth]{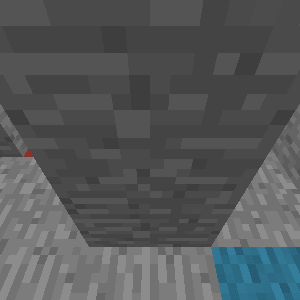} 
   		\includegraphics[width=1\linewidth]{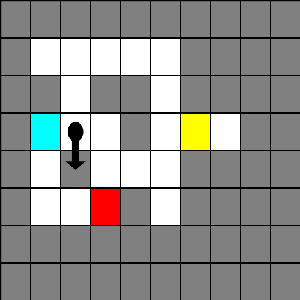} 
   		\caption*{t = 16}
	\end{subfigure}
	& 
    \begin{subfigure}{0.08\linewidth}
	    \includegraphics[width=1\linewidth]{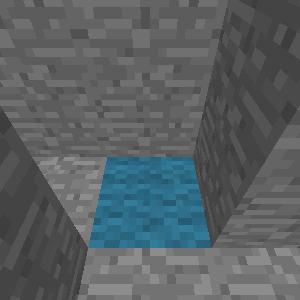} 
   		\includegraphics[width=1\linewidth]{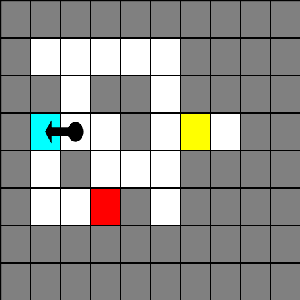} 
   		\caption*{t = 17}
	\end{subfigure}
	& 
    \begin{subfigure}{0.08\linewidth}
	    \includegraphics[width=1\linewidth]{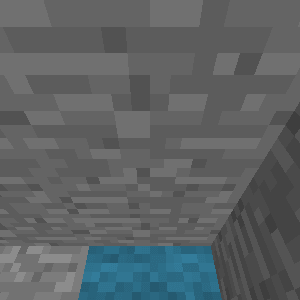} 
   		\includegraphics[width=1\linewidth]{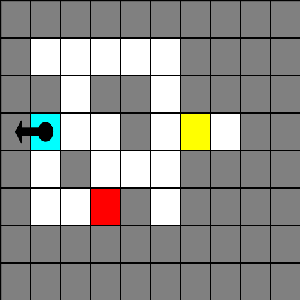} 
   		\caption*{t = 18}
	\end{subfigure} 
	& 
    \begin{subfigure}{0.08\linewidth}
	    \includegraphics[width=1\linewidth]{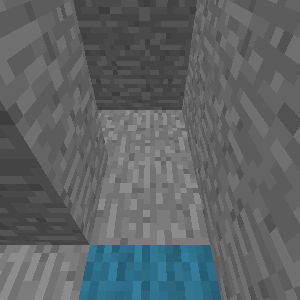} 
   		\includegraphics[width=1\linewidth]{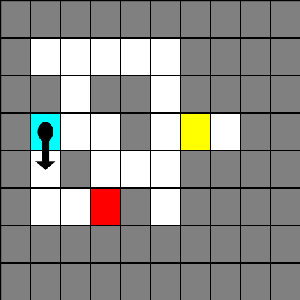} 
   		\caption*{t = 19}
	\end{subfigure} 
	& 
    \begin{subfigure}{0.08\linewidth}
	    \includegraphics[width=1\linewidth]{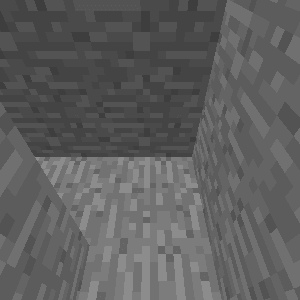} 
   		\includegraphics[width=1\linewidth]{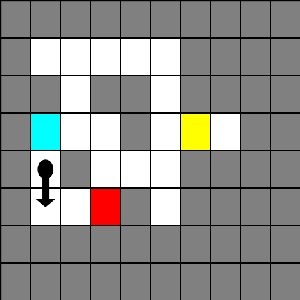} 
   		\caption*{t = 20}
	\end{subfigure} 
	& 
    \begin{subfigure}{0.08\linewidth}
	    \includegraphics[width=1\linewidth]{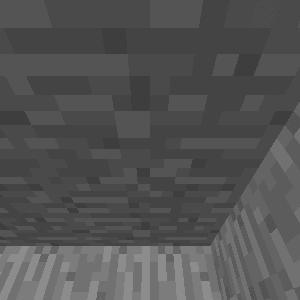} 
   		\includegraphics[width=1\linewidth]{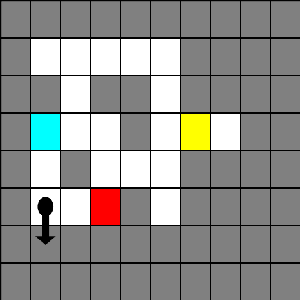} 
   		\caption*{t = 21}
	\end{subfigure}
	& 
    \begin{subfigure}{0.08\linewidth}
	    \includegraphics[width=1\linewidth]{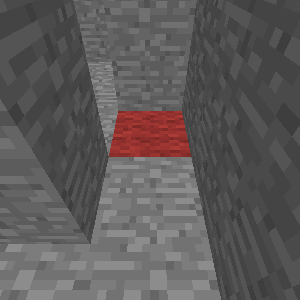} 
   		\includegraphics[width=1\linewidth]{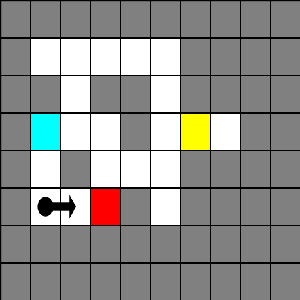} 
   		\caption*{t = 22}
	\end{subfigure}
	& 
    \begin{subfigure}{0.08\linewidth}
	    \includegraphics[width=1\linewidth]{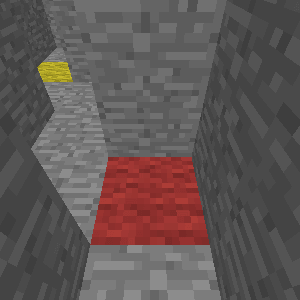} 
   		\includegraphics[width=1\linewidth]{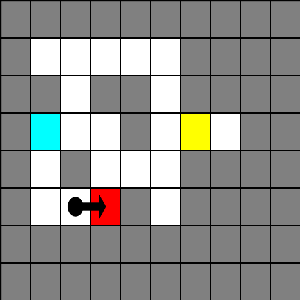} 
   		\caption*{t = 23}
	\end{subfigure} 
	&
    \begin{subfigure}{0.08\linewidth}
	    \includegraphics[width=1\linewidth]{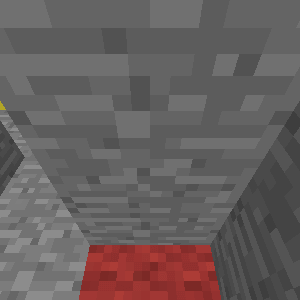} 
   		\includegraphics[width=1\linewidth]{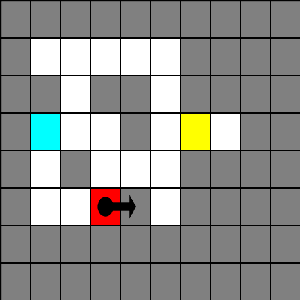} 
   		\caption*{t = 24}
	\end{subfigure} 
	\\
	\\
	\end{tabular}
    \caption{FRMQN's play in a random maze with Sequential Goals with Indicator task. The agent observes that the indicator is yellow at t=3. The agent can see that the red block is near at t=4. Since the task is to first visit the blue block and then the red block if the indicator is yellow, it avoids moving towards the red block (t=6). The agent proceeds by turning and exploring some other corridors. Finally, it gets a glimpse of both the blue and red block (t=14). Using this visual observation along with retrieving from memory visual observations with the indicator present, the agent correctly goes to the blue block and then proceeds to the red block, hence completing the task (t=15-24).} 
	\label{play-seq-i}
\end{figure*}

\begin{figure*}
\small
\setlength{\tabcolsep}{1pt}
\def\arraystretch{1}
\begin{tabular}{llllllllllll}
\begin{subfigure}{0.08\linewidth}
\includegraphics[width=1\linewidth]{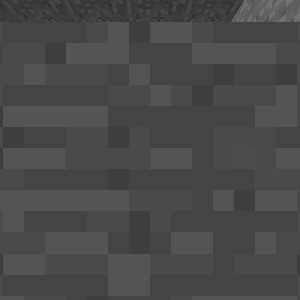}
\includegraphics[width=1\linewidth]{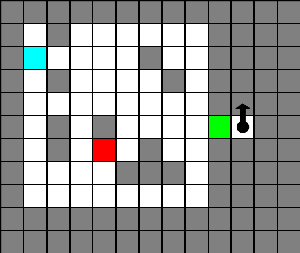}
\caption*{t = 1}
\end{subfigure}
&
\begin{subfigure}{0.08\linewidth}
\includegraphics[width=1\linewidth]{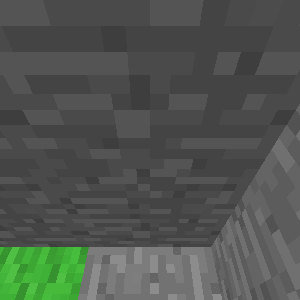}
\includegraphics[width=1\linewidth]{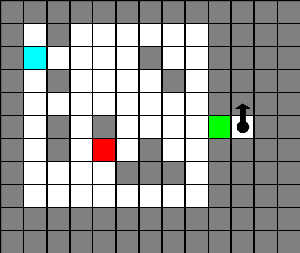}
\caption*{t = 2}
\end{subfigure}
&
\begin{subfigure}{0.08\linewidth}
\includegraphics[width=1\linewidth]{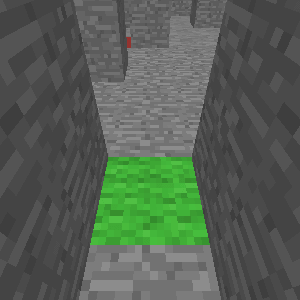}
\includegraphics[width=1\linewidth]{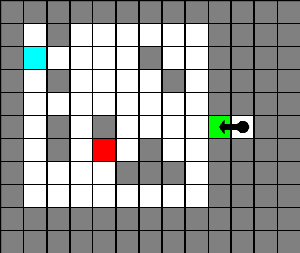}
\caption*{t = 3}
\end{subfigure}
&
\begin{subfigure}{0.08\linewidth}
\includegraphics[width=1\linewidth]{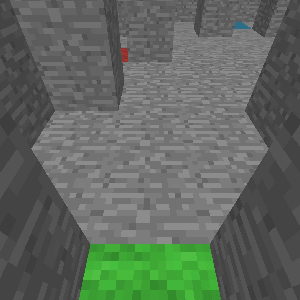}
\includegraphics[width=1\linewidth]{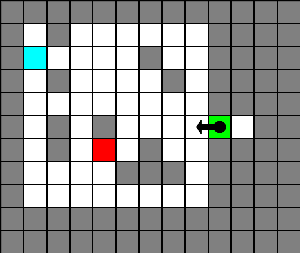}
\caption*{t = 4}
\end{subfigure}
&
\begin{subfigure}{0.08\linewidth}
\includegraphics[width=1\linewidth]{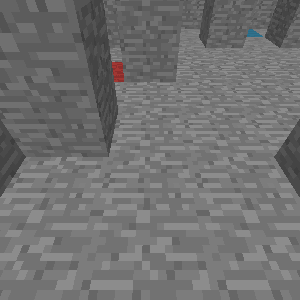}
\includegraphics[width=1\linewidth]{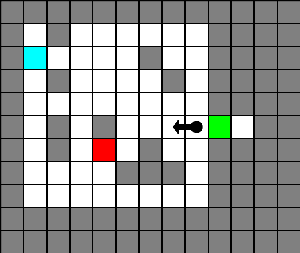}
\caption*{t = 5}
\end{subfigure}
&
\begin{subfigure}{0.08\linewidth}
\includegraphics[width=1\linewidth]{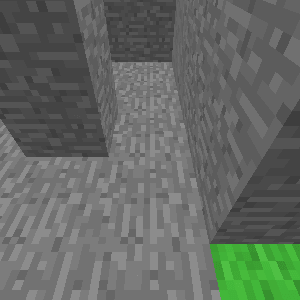}
\includegraphics[width=1\linewidth]{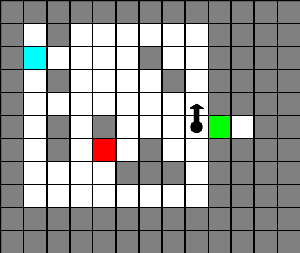}
\caption*{t = 6}
\end{subfigure}
&
\begin{subfigure}{0.08\linewidth}
\includegraphics[width=1\linewidth]{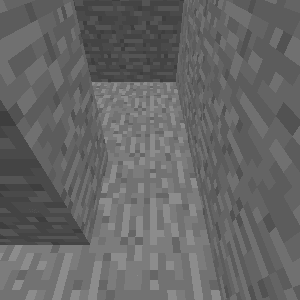}
\includegraphics[width=1\linewidth]{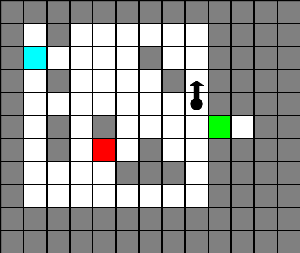}
\caption*{t = 7}
\end{subfigure}
&
\begin{subfigure}{0.08\linewidth}
\includegraphics[width=1\linewidth]{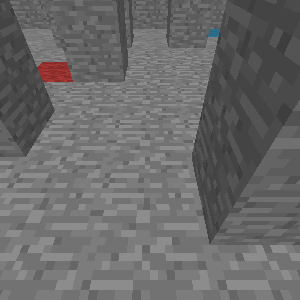}
\includegraphics[width=1\linewidth]{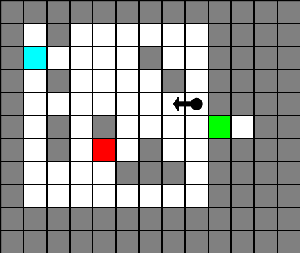}
\caption*{t = 8}
\end{subfigure}
&
\begin{subfigure}{0.08\linewidth}
\includegraphics[width=1\linewidth]{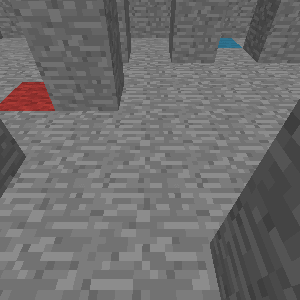}
\includegraphics[width=1\linewidth]{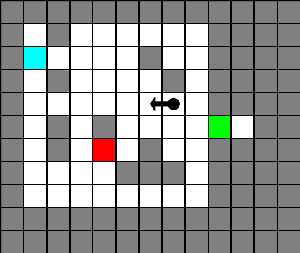}
\caption*{t = 9}
\end{subfigure}
&
\begin{subfigure}{0.08\linewidth}
\includegraphics[width=1\linewidth]{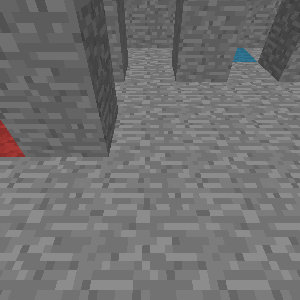}
\includegraphics[width=1\linewidth]{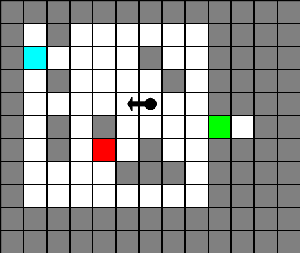}
\caption*{t = 10}
\end{subfigure}
&
\begin{subfigure}{0.08\linewidth}
\includegraphics[width=1\linewidth]{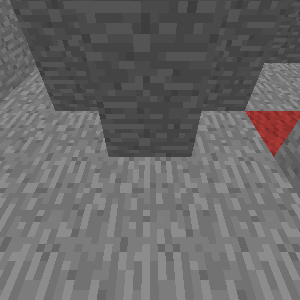}
\includegraphics[width=1\linewidth]{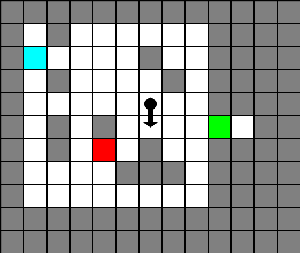}
\caption*{t = 11}
\end{subfigure}
&
\begin{subfigure}{0.08\linewidth}
\includegraphics[width=1\linewidth]{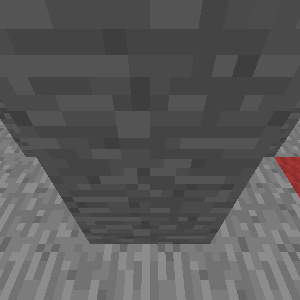}
\includegraphics[width=1\linewidth]{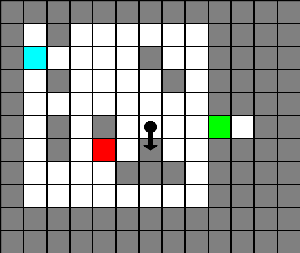}
\caption*{t = 12}
\end{subfigure}
\\
\\
\begin{subfigure}{0.08\linewidth}
\includegraphics[width=1\linewidth]{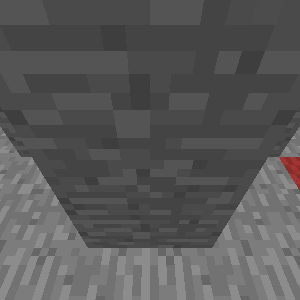}
\includegraphics[width=1\linewidth]{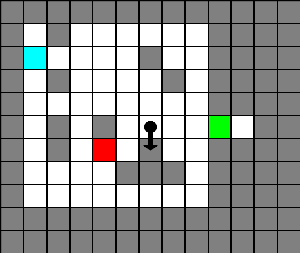}
\caption*{t = 13}
\end{subfigure}
&
\begin{subfigure}{0.08\linewidth}
\includegraphics[width=1\linewidth]{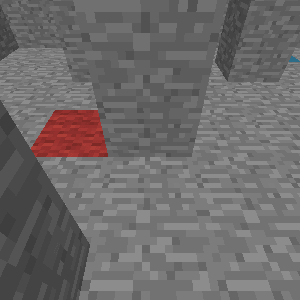}
\includegraphics[width=1\linewidth]{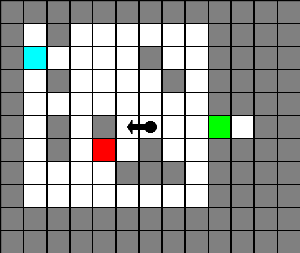}
\caption*{t = 14}
\end{subfigure}
&
\begin{subfigure}{0.08\linewidth}
\includegraphics[width=1\linewidth]{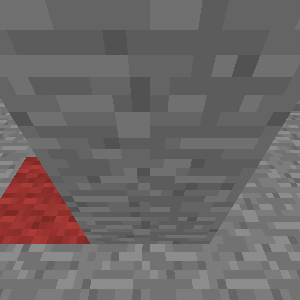}
\includegraphics[width=1\linewidth]{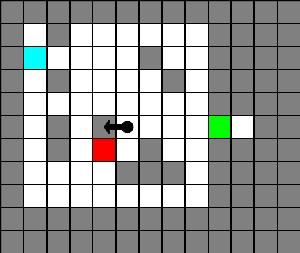}
\caption*{t = 15}
\end{subfigure}
&
\begin{subfigure}{0.08\linewidth}
\includegraphics[width=1\linewidth]{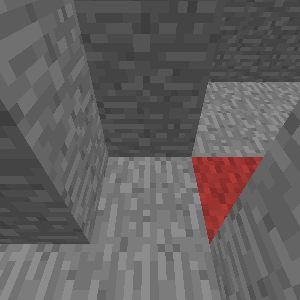}
\includegraphics[width=1\linewidth]{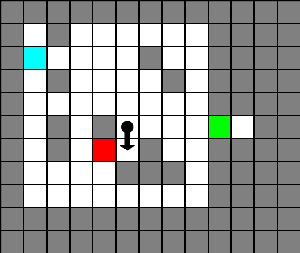}
\caption*{t = 16}
\end{subfigure}
&
\begin{subfigure}{0.08\linewidth}
\includegraphics[width=1\linewidth]{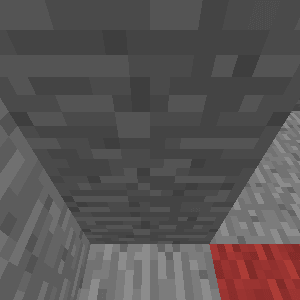}
\includegraphics[width=1\linewidth]{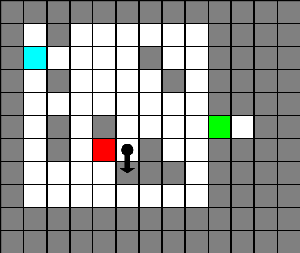}
\caption*{t = 17}
\end{subfigure}
&
\begin{subfigure}{0.08\linewidth}
\includegraphics[width=1\linewidth]{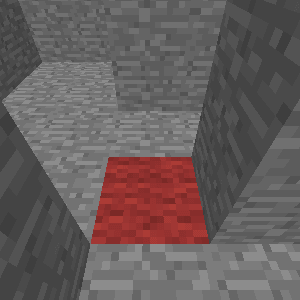}
\includegraphics[width=1\linewidth]{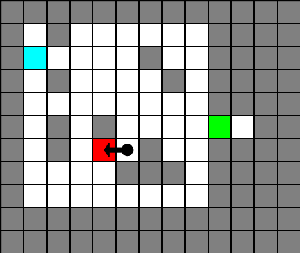}
\caption*{t = 18}
\end{subfigure}
&
\begin{subfigure}{0.08\linewidth}
\includegraphics[width=1\linewidth]{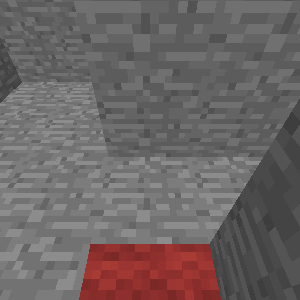}
\includegraphics[width=1\linewidth]{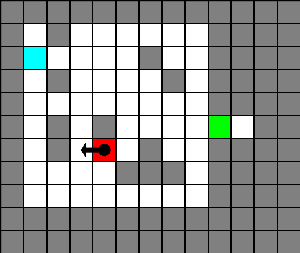}
\caption*{t = 19}
\end{subfigure}
&
\begin{subfigure}{0.08\linewidth}
\includegraphics[width=1\linewidth]{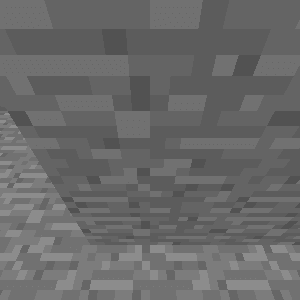}
\includegraphics[width=1\linewidth]{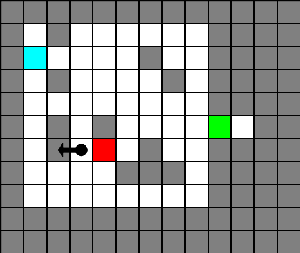}
\caption*{t = 20}
\end{subfigure}
&
\begin{subfigure}{0.08\linewidth}
\includegraphics[width=1\linewidth]{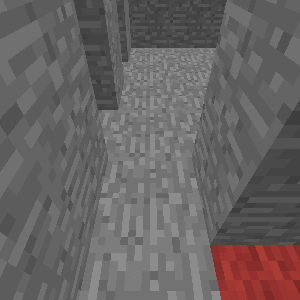}
\includegraphics[width=1\linewidth]{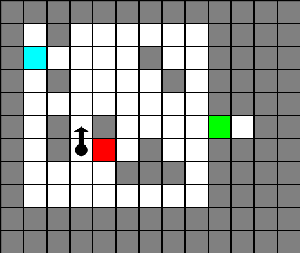}
\caption*{t = 21}
\end{subfigure}
&
\begin{subfigure}{0.08\linewidth}
\includegraphics[width=1\linewidth]{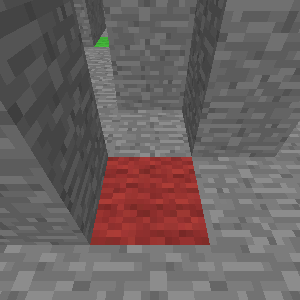}
\includegraphics[width=1\linewidth]{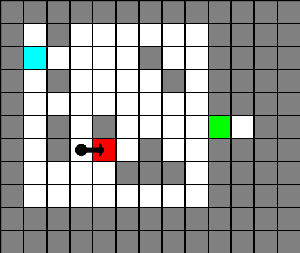}
\caption*{t = 22}
\end{subfigure}
&
\begin{subfigure}{0.08\linewidth}
\includegraphics[width=1\linewidth]{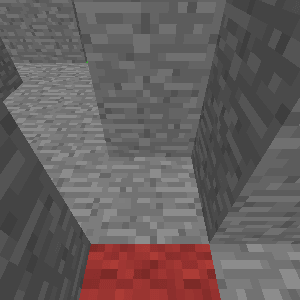}
\includegraphics[width=1\linewidth]{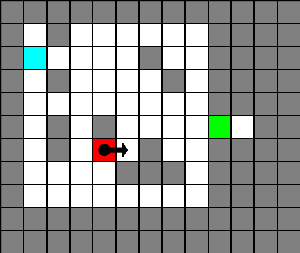}
\caption*{t = 23}
\end{subfigure}
&
\begin{subfigure}{0.08\linewidth}
\includegraphics[width=1\linewidth]{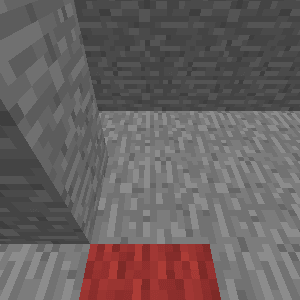}
\includegraphics[width=1\linewidth]{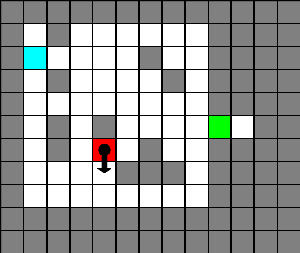}
\caption*{t = 24}
\end{subfigure}
\\
\\
\begin{subfigure}{0.08\linewidth}
\includegraphics[width=1\linewidth]{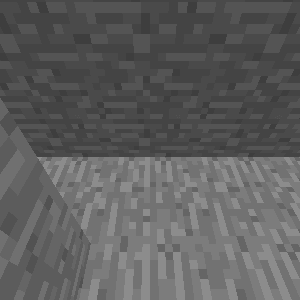}
\includegraphics[width=1\linewidth]{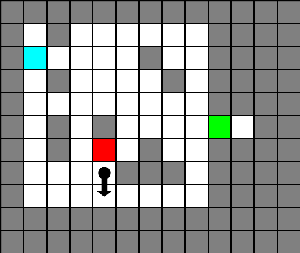}
\caption*{t = 25}
\end{subfigure}
&
\begin{subfigure}{0.08\linewidth}
\includegraphics[width=1\linewidth]{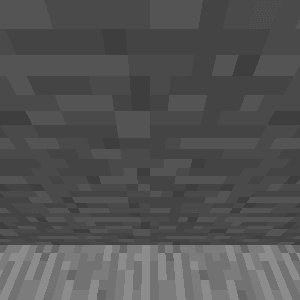}
\includegraphics[width=1\linewidth]{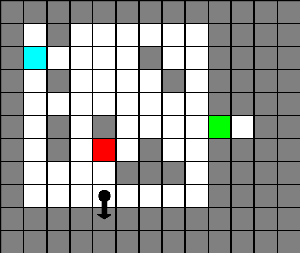}
\caption*{t = 26}
\end{subfigure}
&
\begin{subfigure}{0.08\linewidth}
\includegraphics[width=1\linewidth]{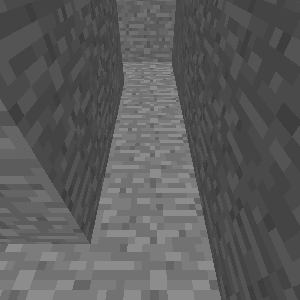}
\includegraphics[width=1\linewidth]{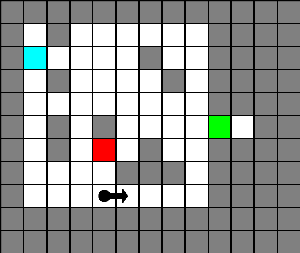}
\caption*{t = 27}
\end{subfigure}
&
\begin{subfigure}{0.08\linewidth}
\includegraphics[width=1\linewidth]{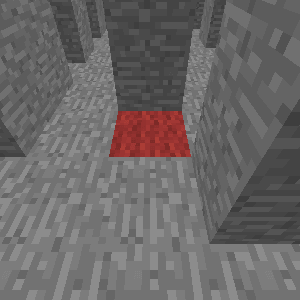}
\includegraphics[width=1\linewidth]{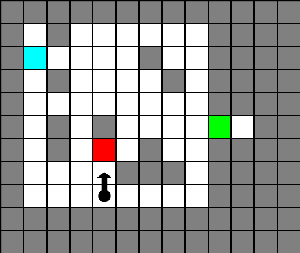}
\caption*{t = 28}
\end{subfigure}
&
\begin{subfigure}{0.08\linewidth}
\includegraphics[width=1\linewidth]{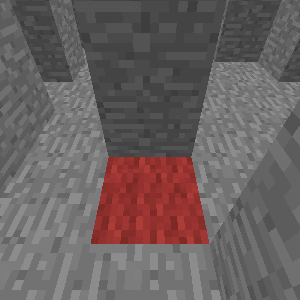}
\includegraphics[width=1\linewidth]{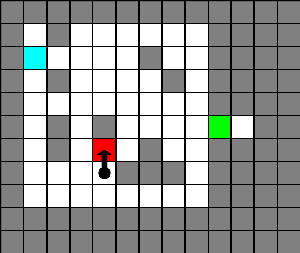}
\caption*{t = 29}
\end{subfigure}
&
\begin{subfigure}{0.08\linewidth}
\includegraphics[width=1\linewidth]{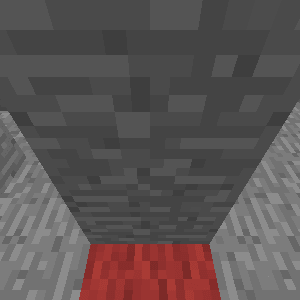}
\includegraphics[width=1\linewidth]{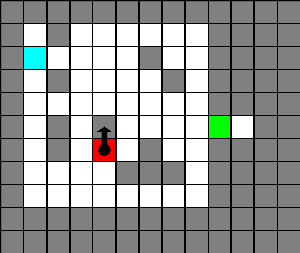}
\caption*{t = 30}
\end{subfigure}
&
\begin{subfigure}{0.08\linewidth}
\includegraphics[width=1\linewidth]{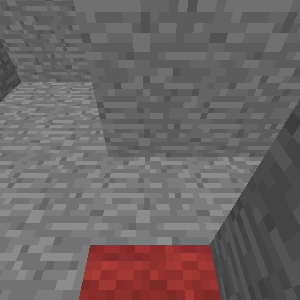}
\includegraphics[width=1\linewidth]{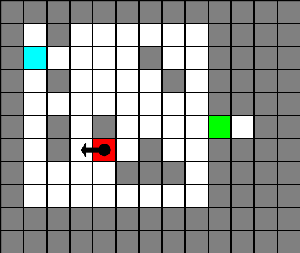}
\caption*{t = 31}
\end{subfigure}
&
\begin{subfigure}{0.08\linewidth}
\includegraphics[width=1\linewidth]{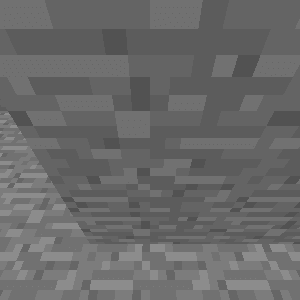}
\includegraphics[width=1\linewidth]{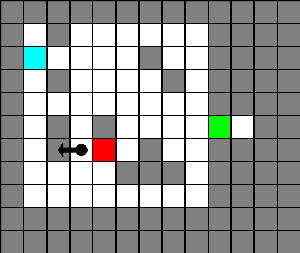}
\caption*{t = 32}
\end{subfigure}
&
\begin{subfigure}{0.08\linewidth}
\includegraphics[width=1\linewidth]{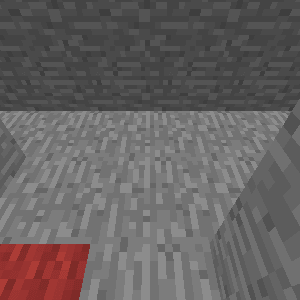}
\includegraphics[width=1\linewidth]{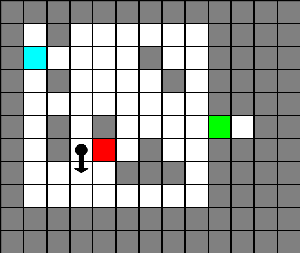}
\caption*{t = 33}
\end{subfigure}
&
\begin{subfigure}{0.08\linewidth}
\includegraphics[width=1\linewidth]{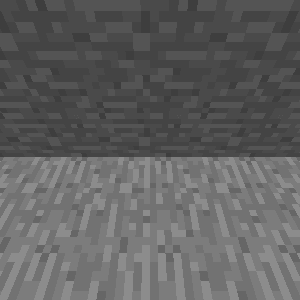}
\includegraphics[width=1\linewidth]{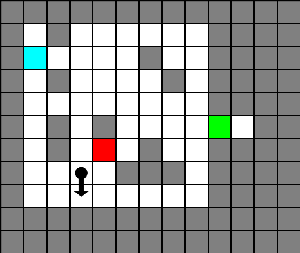}
\caption*{t = 34}
\end{subfigure}
&
\begin{subfigure}{0.08\linewidth}
\includegraphics[width=1\linewidth]{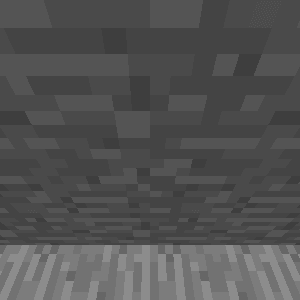}
\includegraphics[width=1\linewidth]{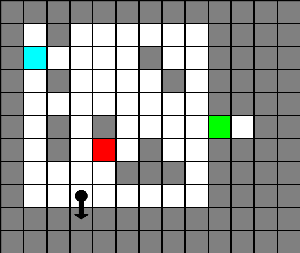}
\caption*{t = 35}
\end{subfigure}
&
\begin{subfigure}{0.08\linewidth}
\includegraphics[width=1\linewidth]{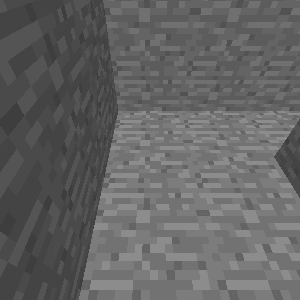}
\includegraphics[width=1\linewidth]{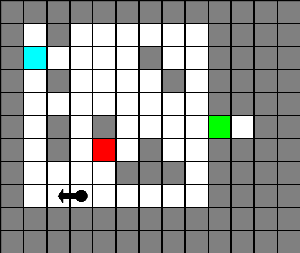}
\caption*{t = 36}
\end{subfigure}
\\
\\
\begin{subfigure}{0.08\linewidth}
\includegraphics[width=1\linewidth]{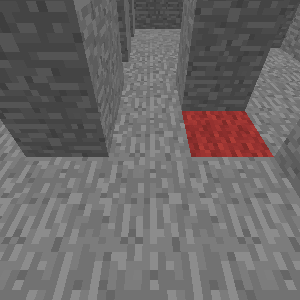}
\includegraphics[width=1\linewidth]{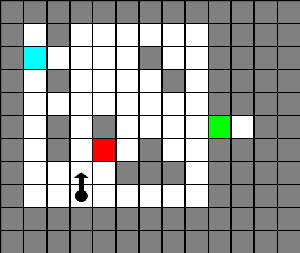}
\caption*{t = 37}
\end{subfigure}
&
\begin{subfigure}{0.08\linewidth}
\includegraphics[width=1\linewidth]{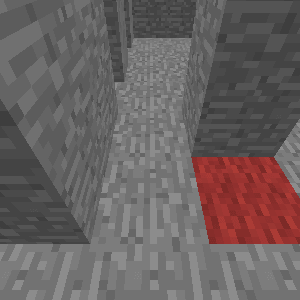}
\includegraphics[width=1\linewidth]{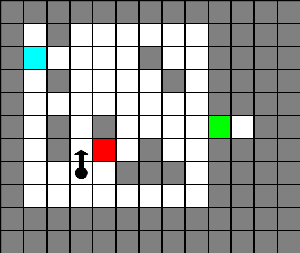}
\caption*{t = 38}
\end{subfigure}
&
\begin{subfigure}{0.08\linewidth}
\includegraphics[width=1\linewidth]{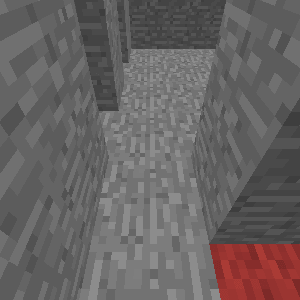}
\includegraphics[width=1\linewidth]{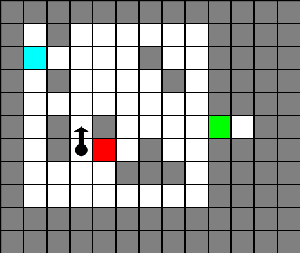}
\caption*{t = 39}
\end{subfigure}
&
\begin{subfigure}{0.08\linewidth}
\includegraphics[width=1\linewidth]{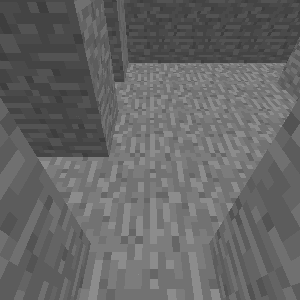}
\includegraphics[width=1\linewidth]{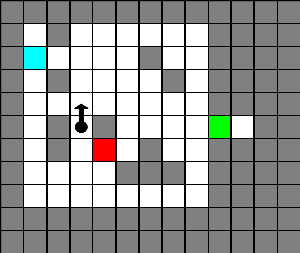}
\caption*{t = 40}
\end{subfigure}
&
\begin{subfigure}{0.08\linewidth}
\includegraphics[width=1\linewidth]{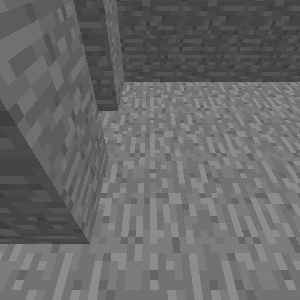}
\includegraphics[width=1\linewidth]{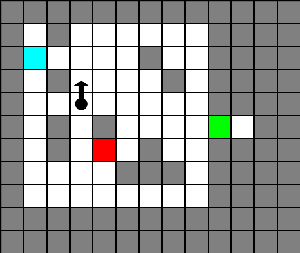}
\caption*{t = 41}
\end{subfigure}
&
\begin{subfigure}{0.08\linewidth}
\includegraphics[width=1\linewidth]{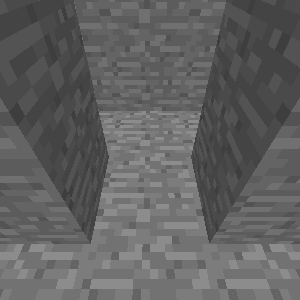}
\includegraphics[width=1\linewidth]{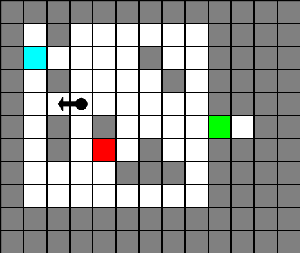}
\caption*{t = 42}
\end{subfigure}
&
\begin{subfigure}{0.08\linewidth}
\includegraphics[width=1\linewidth]{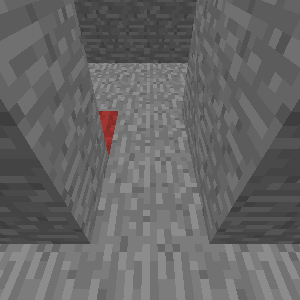}
\includegraphics[width=1\linewidth]{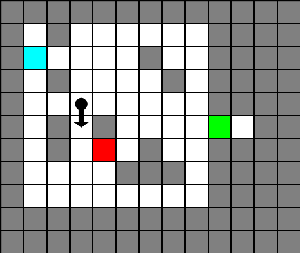}
\caption*{t = 43}
\end{subfigure}
&
\begin{subfigure}{0.08\linewidth}
\includegraphics[width=1\linewidth]{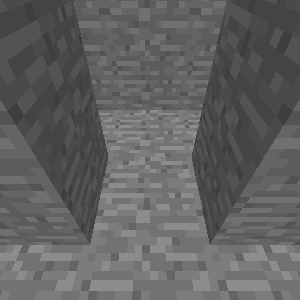}
\includegraphics[width=1\linewidth]{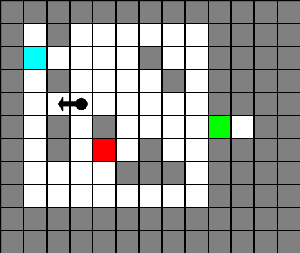}
\caption*{t = 44}
\end{subfigure}
&
\begin{subfigure}{0.08\linewidth}
\includegraphics[width=1\linewidth]{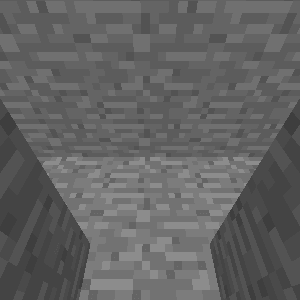}
\includegraphics[width=1\linewidth]{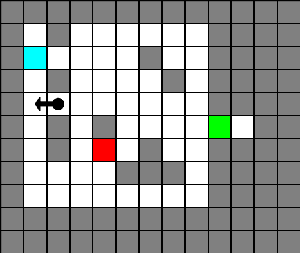}
\caption*{t = 45}
\end{subfigure}
&
\begin{subfigure}{0.08\linewidth}
\includegraphics[width=1\linewidth]{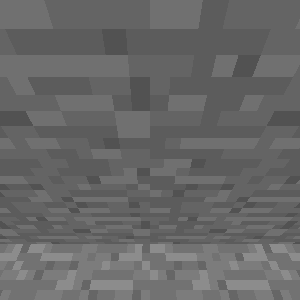}
\includegraphics[width=1\linewidth]{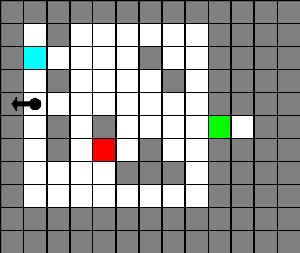}
\caption*{t = 46}
\end{subfigure}
&
\begin{subfigure}{0.08\linewidth}
\includegraphics[width=1\linewidth]{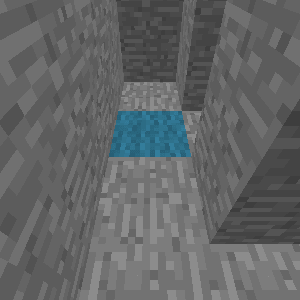}
\includegraphics[width=1\linewidth]{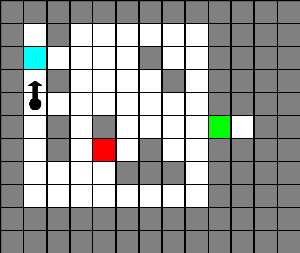}
\caption*{t = 47}
\end{subfigure}
&
\begin{subfigure}{0.08\linewidth}
\includegraphics[width=1\linewidth]{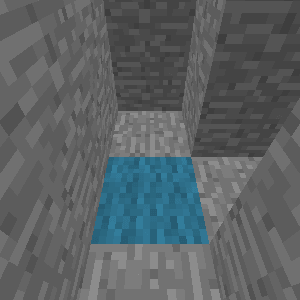}
\includegraphics[width=1\linewidth]{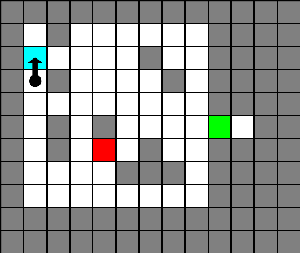}
\caption*{t = 48}
\end{subfigure}
\\
\\
\begin{subfigure}{0.08\linewidth}
\includegraphics[width=1\linewidth]{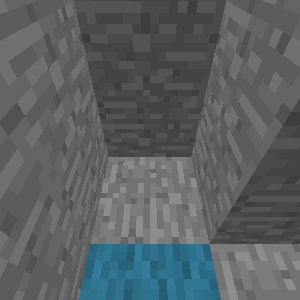}
\includegraphics[width=1\linewidth]{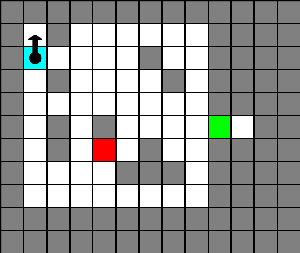}
\caption*{t = 49}
\end{subfigure}
\end{tabular}
\caption{FRMQN's play in an unseen and larger random maze with Sequential Goals with Indicator task. The agent visits the red block first (t=19), given the green indicator (t=3). The agent looks for the blue block which is far apart from the red block (t=19-47). Once the agent finds the blue block, it finishes the task by completing the sequence (t=49). }
\label{seq_i_l_success}
\end{figure*}

\begin{figure*}
\small
\setlength{\tabcolsep}{1pt}
\def\arraystretch{1}
\begin{tabular}{llllllllllll}
\begin{subfigure}{0.08\linewidth}
\includegraphics[width=1\linewidth]{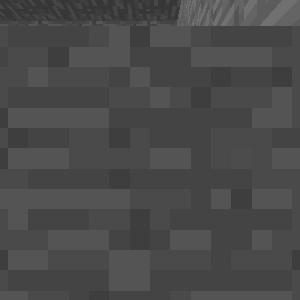}
\includegraphics[width=1\linewidth]{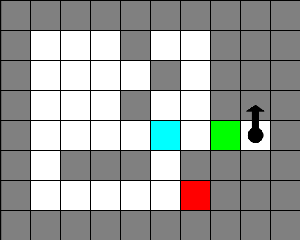}
\caption*{t = 1}
\end{subfigure}
&
\begin{subfigure}{0.08\linewidth}
\includegraphics[width=1\linewidth]{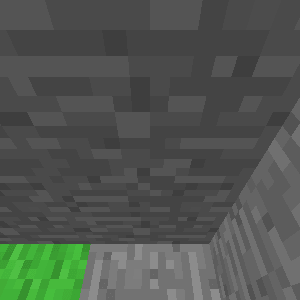}
\includegraphics[width=1\linewidth]{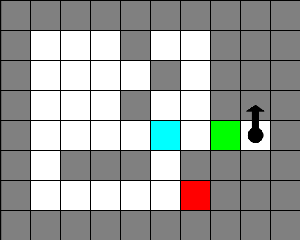}
\caption*{t = 2}
\end{subfigure}
&
\begin{subfigure}{0.08\linewidth}
\includegraphics[width=1\linewidth]{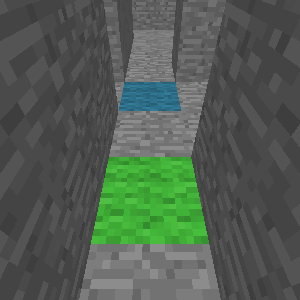}
\includegraphics[width=1\linewidth]{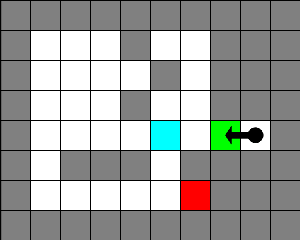}
\caption*{t = 3}
\end{subfigure}
&
\begin{subfigure}{0.08\linewidth}
\includegraphics[width=1\linewidth]{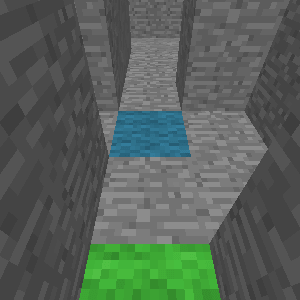}
\includegraphics[width=1\linewidth]{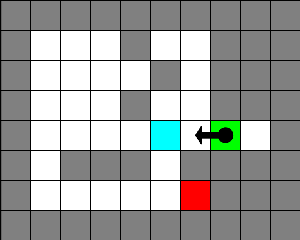}
\caption*{t = 4}
\end{subfigure}
&
\begin{subfigure}{0.08\linewidth}
\includegraphics[width=1\linewidth]{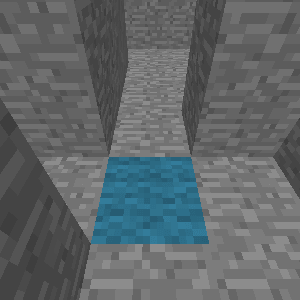}
\includegraphics[width=1\linewidth]{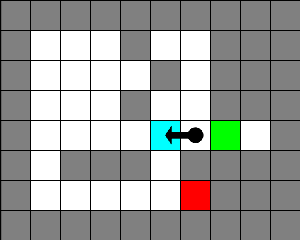}
\caption*{t = 5}
\end{subfigure}
&
\begin{subfigure}{0.08\linewidth}
\includegraphics[width=1\linewidth]{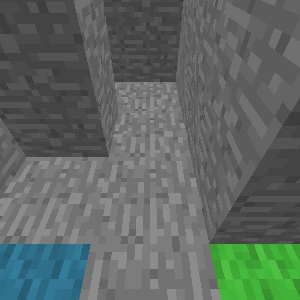}
\includegraphics[width=1\linewidth]{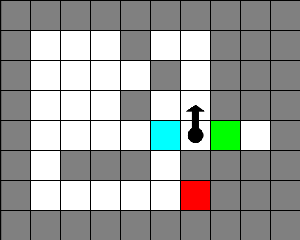}
\caption*{t = 6}
\end{subfigure}
&
\begin{subfigure}{0.08\linewidth}
\includegraphics[width=1\linewidth]{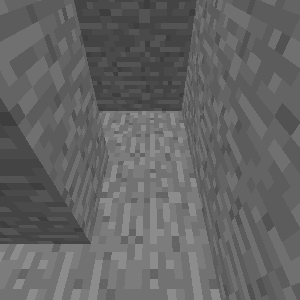}
\includegraphics[width=1\linewidth]{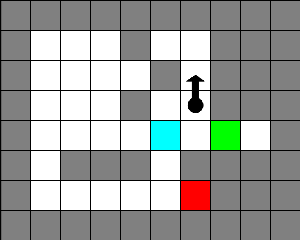}
\caption*{t = 7}
\end{subfigure}
&
\begin{subfigure}{0.08\linewidth}
\includegraphics[width=1\linewidth]{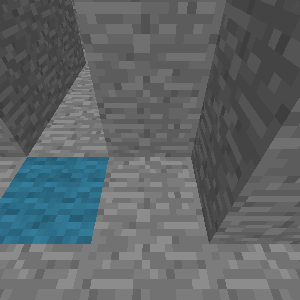}
\includegraphics[width=1\linewidth]{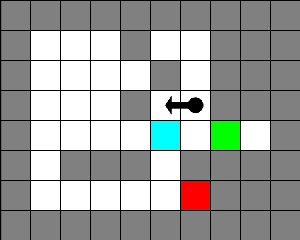}
\caption*{t = 8}
\end{subfigure}
&
\begin{subfigure}{0.08\linewidth}
\includegraphics[width=1\linewidth]{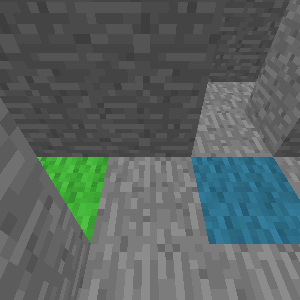}
\includegraphics[width=1\linewidth]{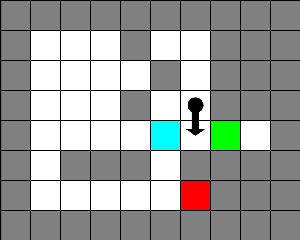}
\caption*{t = 9}
\end{subfigure}
&
\begin{subfigure}{0.08\linewidth}
\includegraphics[width=1\linewidth]{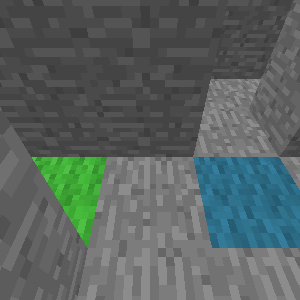}
\includegraphics[width=1\linewidth]{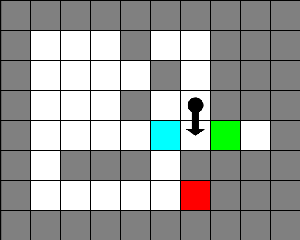}
\caption*{t = 10}
\end{subfigure}
&
\begin{subfigure}{0.08\linewidth}
\includegraphics[width=1\linewidth]{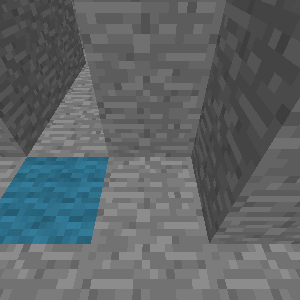}
\includegraphics[width=1\linewidth]{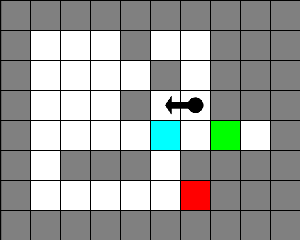}
\caption*{t = 11}
\end{subfigure}
&
\begin{subfigure}{0.08\linewidth}
\includegraphics[width=1\linewidth]{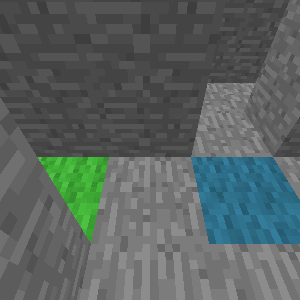}
\includegraphics[width=1\linewidth]{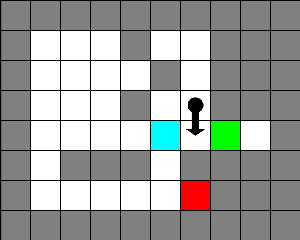}
\caption*{t = 12}
\end{subfigure}
\\
\\
\begin{subfigure}{0.08\linewidth}
\includegraphics[width=1\linewidth]{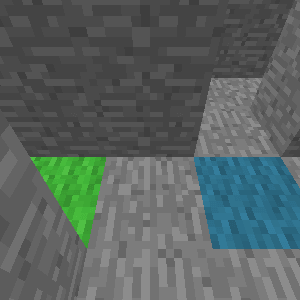}
\includegraphics[width=1\linewidth]{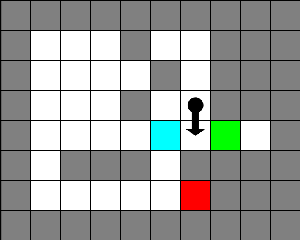}
\caption*{t = 13}
\end{subfigure}
&
\begin{subfigure}{0.08\linewidth}
\includegraphics[width=1\linewidth]{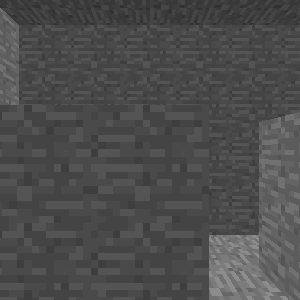}
\includegraphics[width=1\linewidth]{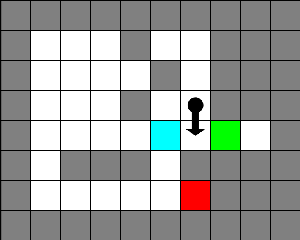}
\caption*{t = 14}
\end{subfigure}
&
\begin{subfigure}{0.08\linewidth}
\includegraphics[width=1\linewidth]{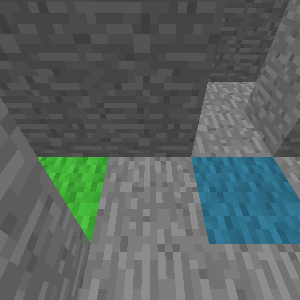}
\includegraphics[width=1\linewidth]{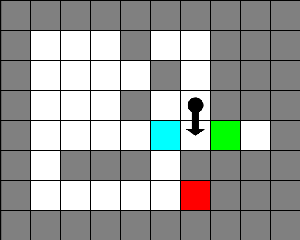}
\caption*{t = 15}
\end{subfigure}
&
\begin{subfigure}{0.08\linewidth}
\includegraphics[width=1\linewidth]{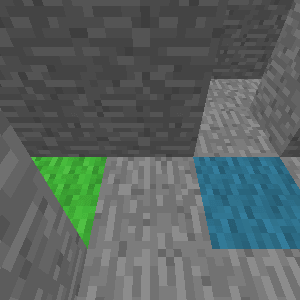}
\includegraphics[width=1\linewidth]{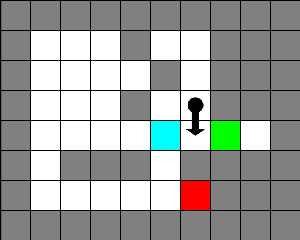}
\caption*{t = 16}
\end{subfigure}
&
\begin{subfigure}{0.08\linewidth}
\includegraphics[width=1\linewidth]{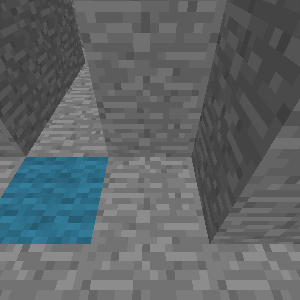}
\includegraphics[width=1\linewidth]{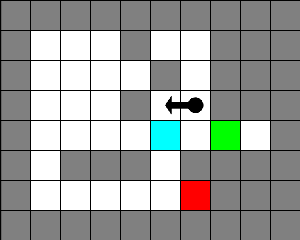}
\caption*{t = 17}
\end{subfigure}
&
\begin{subfigure}{0.08\linewidth}
\includegraphics[width=1\linewidth]{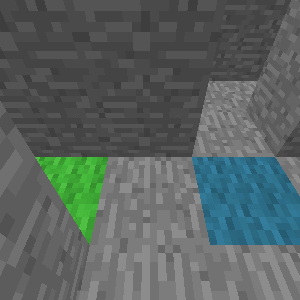}
\includegraphics[width=1\linewidth]{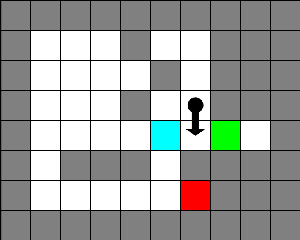}
\caption*{t = 18}
\end{subfigure}
&
\begin{subfigure}{0.08\linewidth}
\includegraphics[width=1\linewidth]{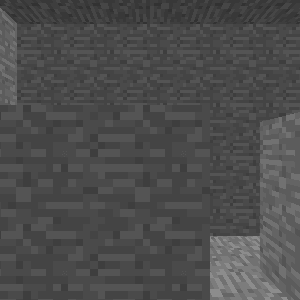}
\includegraphics[width=1\linewidth]{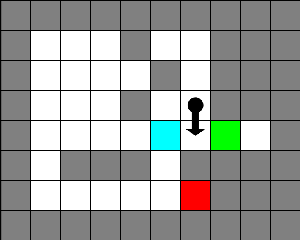}
\caption*{t = 19}
\end{subfigure}
&
\begin{subfigure}{0.08\linewidth}
\includegraphics[width=1\linewidth]{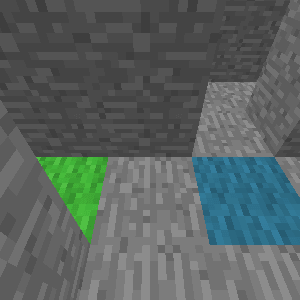}
\includegraphics[width=1\linewidth]{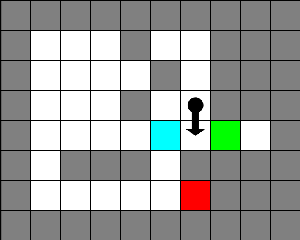}
\caption*{t = 20}
\end{subfigure}
&
\begin{subfigure}{0.08\linewidth}
\includegraphics[width=1\linewidth]{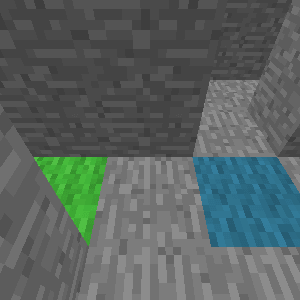}
\includegraphics[width=1\linewidth]{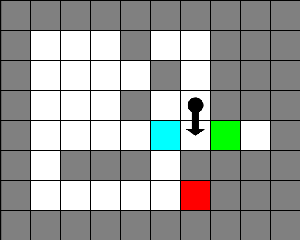}
\caption*{t = 21}
\end{subfigure}
&
\begin{subfigure}{0.08\linewidth}
\includegraphics[width=1\linewidth]{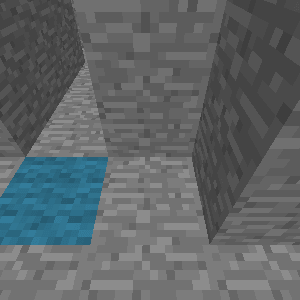}
\includegraphics[width=1\linewidth]{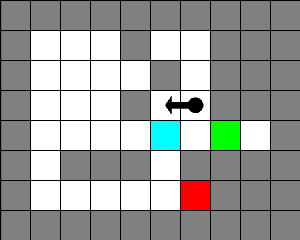}
\caption*{t = 22}
\end{subfigure}
&
\begin{subfigure}{0.08\linewidth}
\includegraphics[width=1\linewidth]{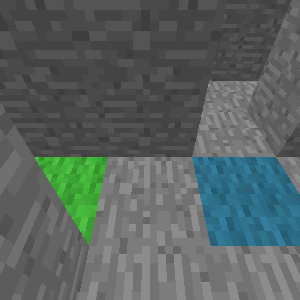}
\includegraphics[width=1\linewidth]{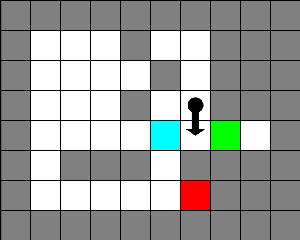}
\caption*{t = 23}
\end{subfigure}
&
\begin{subfigure}{0.08\linewidth}
\includegraphics[width=1\linewidth]{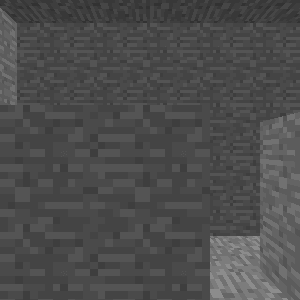}
\includegraphics[width=1\linewidth]{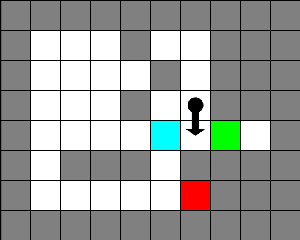}
\caption*{t = 24}
\end{subfigure}
\\
\\
\begin{subfigure}{0.08\linewidth}
\includegraphics[width=1\linewidth]{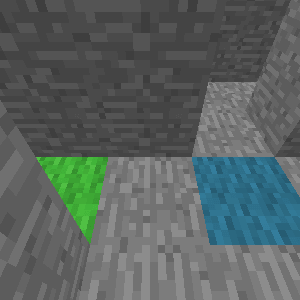}
\includegraphics[width=1\linewidth]{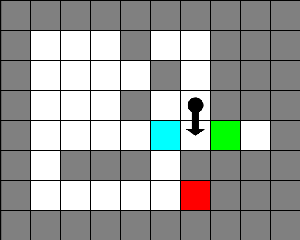}
\caption*{t = 25}
\end{subfigure}
&
\begin{subfigure}{0.08\linewidth}
\includegraphics[width=1\linewidth]{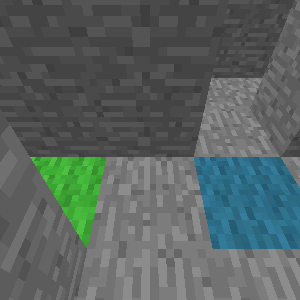}
\includegraphics[width=1\linewidth]{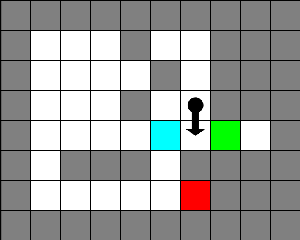}
\caption*{t = 26}
\end{subfigure}
&
\begin{subfigure}{0.08\linewidth}
\includegraphics[width=1\linewidth]{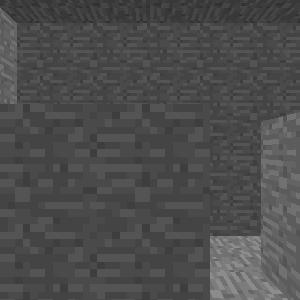}
\includegraphics[width=1\linewidth]{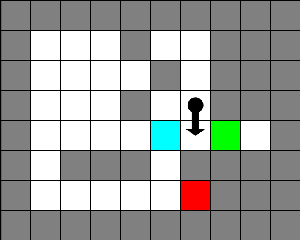}
\caption*{t = 27}
\end{subfigure}
&
\begin{subfigure}{0.08\linewidth}
\includegraphics[width=1\linewidth]{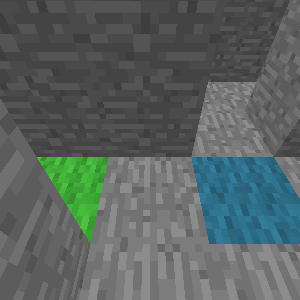}
\includegraphics[width=1\linewidth]{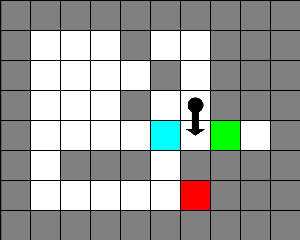}
\caption*{t = 28}
\end{subfigure}
&
\begin{subfigure}{0.08\linewidth}
\includegraphics[width=1\linewidth]{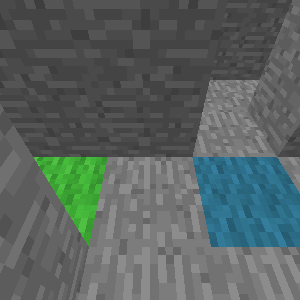}
\includegraphics[width=1\linewidth]{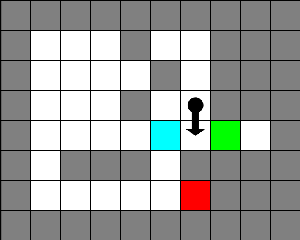}
\caption*{t = 29}
\end{subfigure}
&
\begin{subfigure}{0.08\linewidth}
\includegraphics[width=1\linewidth]{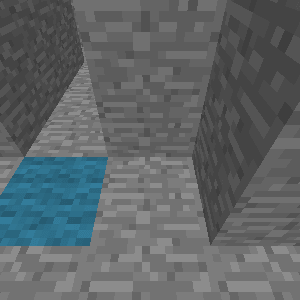}
\includegraphics[width=1\linewidth]{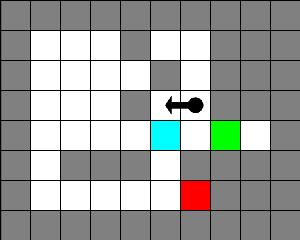}
\caption*{t = 30}
\end{subfigure}
&
\begin{subfigure}{0.08\linewidth}
\includegraphics[width=1\linewidth]{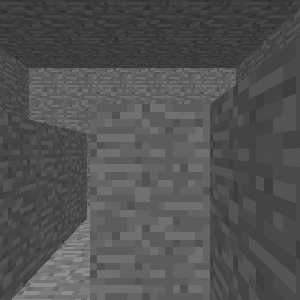}
\includegraphics[width=1\linewidth]{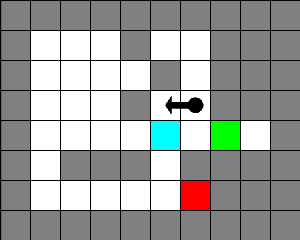}
\caption*{t = 31}
\end{subfigure}
&
\begin{subfigure}{0.08\linewidth}
\includegraphics[width=1\linewidth]{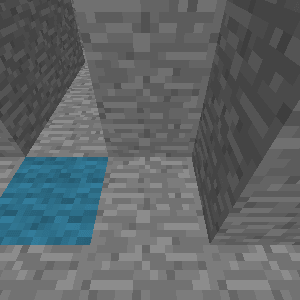}
\includegraphics[width=1\linewidth]{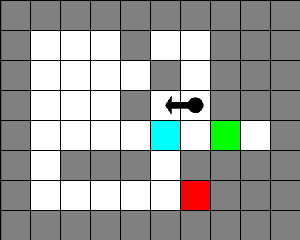}
\caption*{t = 32}
\end{subfigure}
&
\begin{subfigure}{0.08\linewidth}
\includegraphics[width=1\linewidth]{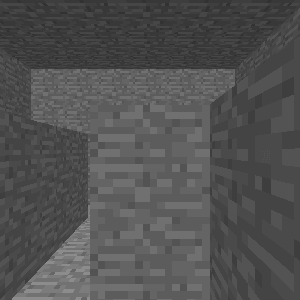}
\includegraphics[width=1\linewidth]{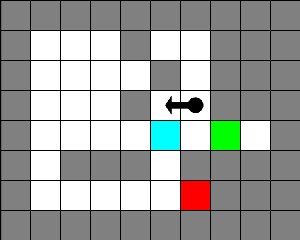}
\caption*{t = 33}
\end{subfigure}
&
\begin{subfigure}{0.08\linewidth}
\includegraphics[width=1\linewidth]{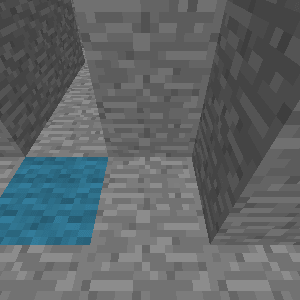}
\includegraphics[width=1\linewidth]{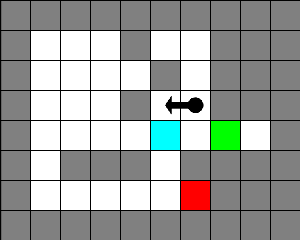}
\caption*{t = 34}
\end{subfigure}
&
\begin{subfigure}{0.08\linewidth}
\includegraphics[width=1\linewidth]{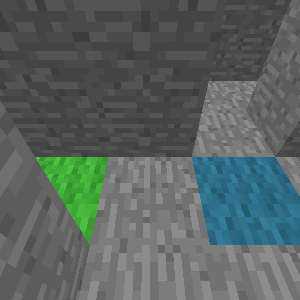}
\includegraphics[width=1\linewidth]{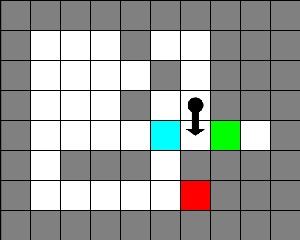}
\caption*{t = 35}
\end{subfigure}
&
\begin{subfigure}{0.08\linewidth}
\includegraphics[width=1\linewidth]{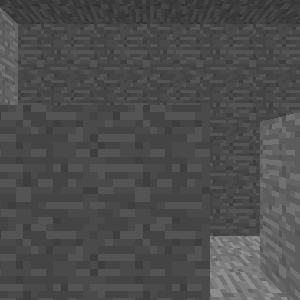}
\includegraphics[width=1\linewidth]{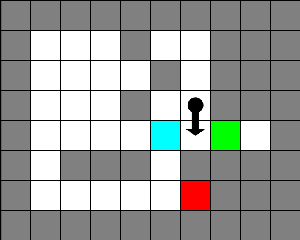}
\caption*{t = 36}
\end{subfigure}
\\
\\
\begin{subfigure}{0.08\linewidth}
\includegraphics[width=1\linewidth]{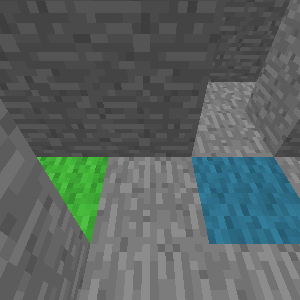}
\includegraphics[width=1\linewidth]{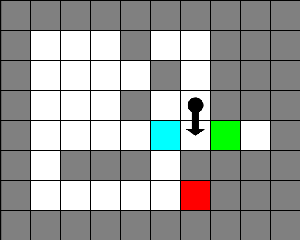}
\caption*{t = 37}
\end{subfigure}
&
\begin{subfigure}{0.08\linewidth}
\includegraphics[width=1\linewidth]{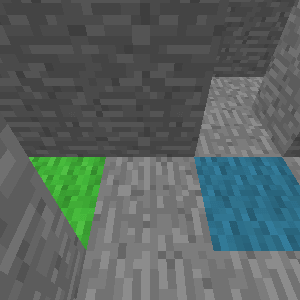}
\includegraphics[width=1\linewidth]{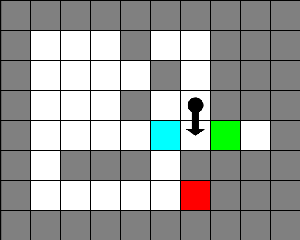}
\caption*{t = 38}
\end{subfigure}
&
\begin{subfigure}{0.08\linewidth}
\includegraphics[width=1\linewidth]{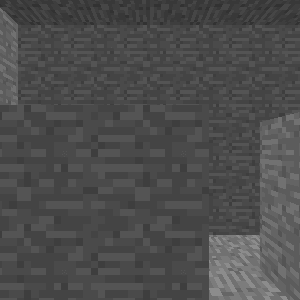}
\includegraphics[width=1\linewidth]{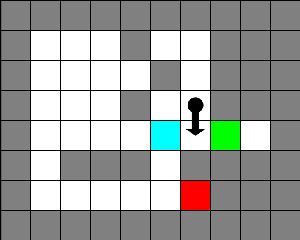}
\caption*{t = 39}
\end{subfigure}
&
\begin{subfigure}{0.08\linewidth}
\includegraphics[width=1\linewidth]{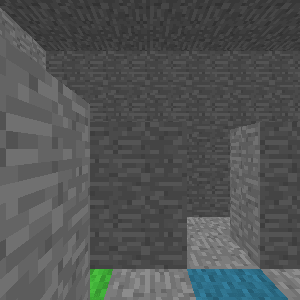}
\includegraphics[width=1\linewidth]{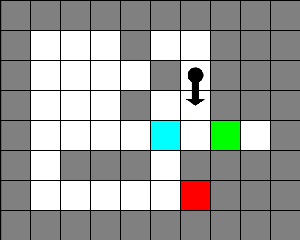}
\caption*{t = 40}
\end{subfigure}
&
\begin{subfigure}{0.08\linewidth}
\includegraphics[width=1\linewidth]{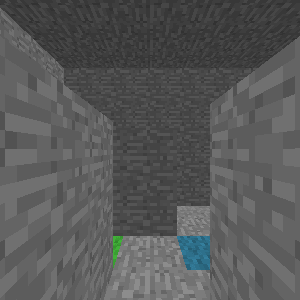}
\includegraphics[width=1\linewidth]{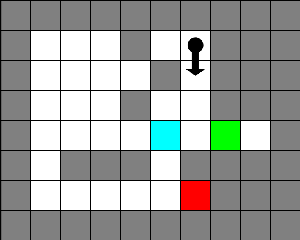}
\caption*{t = 41}
\end{subfigure}
&
\begin{subfigure}{0.08\linewidth}
\includegraphics[width=1\linewidth]{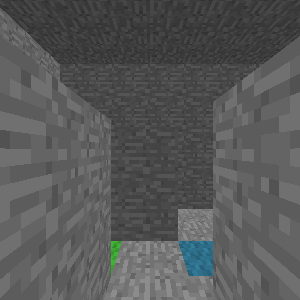}
\includegraphics[width=1\linewidth]{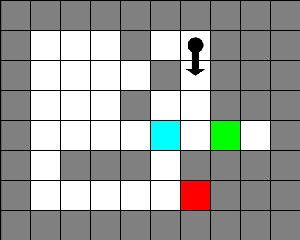}
\caption*{t = 42}
\end{subfigure}
&
\begin{subfigure}{0.08\linewidth}
\includegraphics[width=1\linewidth]{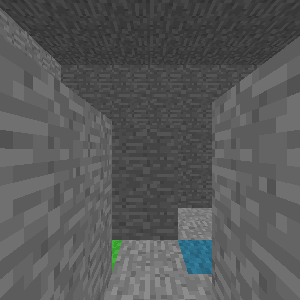}
\includegraphics[width=1\linewidth]{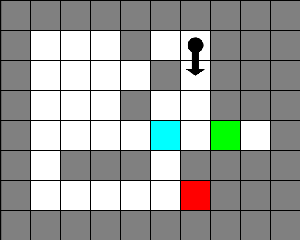}
\caption*{t = 43}
\end{subfigure}
&
\begin{subfigure}{0.08\linewidth}
\includegraphics[width=1\linewidth]{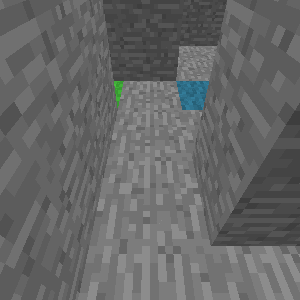}
\includegraphics[width=1\linewidth]{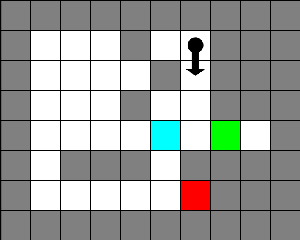}
\caption*{t = 44}
\end{subfigure}
&
\begin{subfigure}{0.08\linewidth}
\includegraphics[width=1\linewidth]{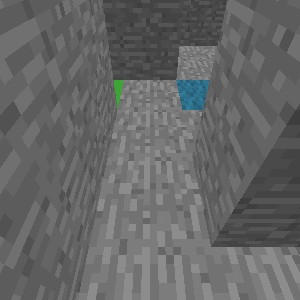}
\includegraphics[width=1\linewidth]{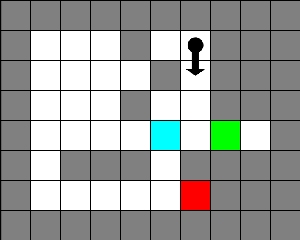}
\caption*{t = 45}
\end{subfigure}
&
\begin{subfigure}{0.08\linewidth}
\includegraphics[width=1\linewidth]{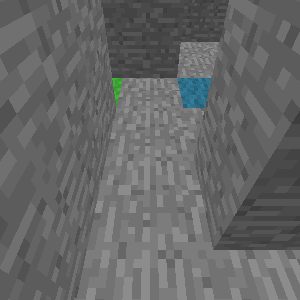}
\includegraphics[width=1\linewidth]{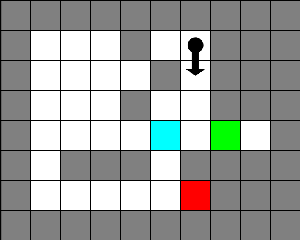}
\caption*{t = 46}
\end{subfigure}
&
\begin{subfigure}{0.08\linewidth}
\includegraphics[width=1\linewidth]{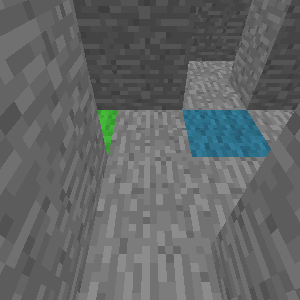}
\includegraphics[width=1\linewidth]{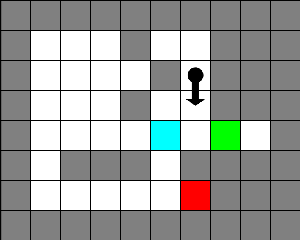}
\caption*{t = 47}
\end{subfigure}
&
\begin{subfigure}{0.08\linewidth}
\includegraphics[width=1\linewidth]{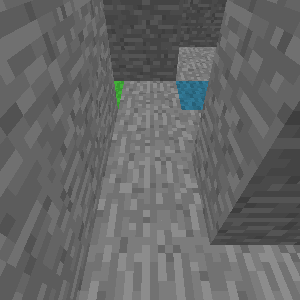}
\includegraphics[width=1\linewidth]{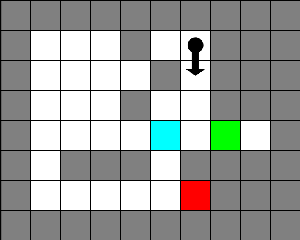}
\caption*{t = 48}
\end{subfigure}
\\
\\
\begin{subfigure}{0.08\linewidth}
\includegraphics[width=1\linewidth]{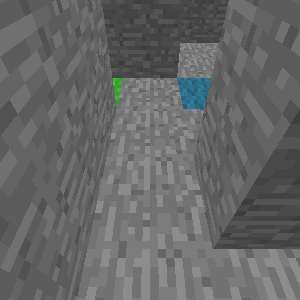}
\includegraphics[width=1\linewidth]{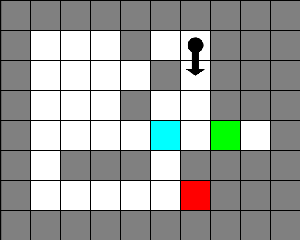}
\caption*{t = 49}
\end{subfigure}
&
\begin{subfigure}{0.08\linewidth}
\includegraphics[width=1\linewidth]{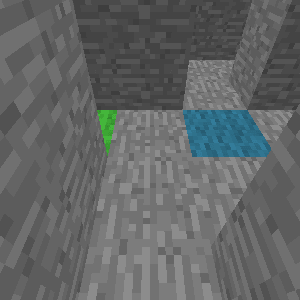}
\includegraphics[width=1\linewidth]{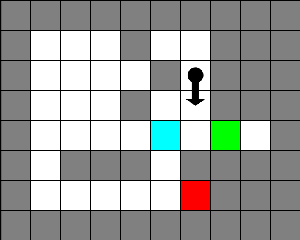}
\caption*{t = 50}
\end{subfigure}
\end{tabular}
\caption{FRMQN's play in a random maze with Sequential Goals with Indicator task. Given that the indicator is green, the agent has to visit the red block first and the blue block later. For this reason, the agent tries to avoid the blue block and search for another route to the red block (t=4-50). However, since there is no path to the red block (that avoids the blue block), the agent keeps searching until the episode terminates. }
\label{seq_i_fail}
\end{figure*}
\end{document}